\documentclass[twoside, nobind]{ociamthesis} 

\correctionstrue

\usepackage{pdfpages}

\usepackage[style=numeric-comp, sorting=none, backend=biber, doi=true, isbn=true, maxnames=99]{biblatex}

\usepackage[colorlinks]{hyperref}
\hypersetup{
hyperfootnotes=false,
colorlinks=true,
pdfpagelabels,
linktoc=page,
linkcolor=NavyBlue, 
citecolor=NavyBlue,
urlcolor=black 
}

\title{Towards holistic scene understanding:\\Semantic segmentation and beyond}
\author{Panagiotis Meletis}
\college{Department of Electrical Engineering}

\tuephdsubmissiontrue

\degree{Doctor of Philosophy}
\degreedate{ }

\usepackage{graphicx}
\graphicspath{ {images/} {data/} {figures/} }
\usepackage{tabularx}

\usepackage{floatflt}

\usepackage{caption}
\usepackage{subcaption}

\DeclareMathOperator*{\argmax}{arg\,max}

\usepackage{verbatim}

\usepackage{array}
\newcolumntype{C}[1]{>{\centering\arraybackslash}p{#1}}
\newcolumntype{Y}{>{\centering\arraybackslash}X}

\usepackage{bm}
\newcommand{\bsigma}{\bm{\sigma}}

\usepackage{mathtools}
\DeclarePairedDelimiter{\abs}{\lvert}{\rvert}

\usepackage{siunitx}

\usepackage{multirow}

\usepackage{makecell}

\usepackage{tablefootnote}

\usepackage{xspace}
\makeatletter
\DeclareRobustCommand\onedot{\futurelet\@let@token\@onedot}
\def\@onedot{\ifx\@let@token.\else.\null\fi\xspace}
\def\eg{e.g\onedot}

\def\ie{i.e\onedot}

\def\etc{etc\onedot}
\makeatother

\usepackage{pifont}
\newcommand*\rotnighty{\rotatebox{90}}
\newcommand*\rotseventy{\rotatebox{70}}
\newcommand*{\multirowrot}[2]{\multirow[b]{#1}{*}{\rotatebox[origin=lB]{90}{#2}}}

\makeatletter
\newsavebox\myboxA
\newsavebox\myboxB
\newlength\mylenA
\newcommand*\xoverline[2][0.75]{%
	\sbox{\myboxA}{$\m@th#2$}%
	\setbox\myboxB\null
	\ht\myboxB=\ht\myboxA%
	\dp\myboxB=\dp\myboxA%
	\wd\myboxB=#1\wd\myboxA
	\sbox\myboxB{$\m@th\overline{\copy\myboxB}$}
	\setlength\mylenA{\the\wd\myboxA}
	\addtolength\mylenA{-\the\wd\myboxB}%
	\ifdim\wd\myboxB<\wd\myboxA%
	\rlap{\hskip 0.5\mylenA\usebox\myboxB}{\usebox\myboxA}%
	\else
	\hskip -0.5\mylenA\rlap{\usebox\myboxA}{\hskip 0.5\mylenA\usebox\myboxB}%
	\fi}
\makeatother

\usepackage{floatrow}
\newfloatcommand{capbtabbox}{table}[][\FBwidth]

\usepackage{url}
\def\btex{B\kern-.05em{i}\kern-.025em{b}\kern-.08em\TeX}

\usepackage{stmaryrd}

\def\hlinefull{\noindent\rule{\textwidth}{1pt}}

\usepackage{booktabs}

\newcolumntype{H}{>{\setbox0=\hbox\bgroup}c<{\egroup}@{}}

\usepackage{setspace}

\usepackage{rotating}

\usepackage{enumitem}

\makeatletter
\newcommand\footnoteref[1]{\protected@xdef\@thefnmark{\ref{#1}}\@footnotemark}
\makeatother
\usepackage{cleveref}

\usepackage[export]{adjustbox}

\newcommand*{\statstabh}[2]{\rotatebox{0}{\makecell[c]{\textbf{#1}\\\textbf{#2}}}}

\usepackage{pdflscape}

\usepackage{xurl}

\usepackage{cleveref}

\let\svthefootnote\thefootnote
\newcommand\freefootnote[1]{%
	\let\thefootnote\relax%
	\footnotetext{#1}%
	\let\thefootnote\svthefootnote%
}

\begin{document}

\setlength{\textbaselineskip}{13pt plus2pt minus1pt}

\setlength{\frontmatterbaselineskip}{13pt plus1pt minus1pt}

\setlength{\baselineskip}{\textbaselineskip}


\setcounter{secnumdepth}{2}
\setcounter{tocdepth}{2}


\includepdf{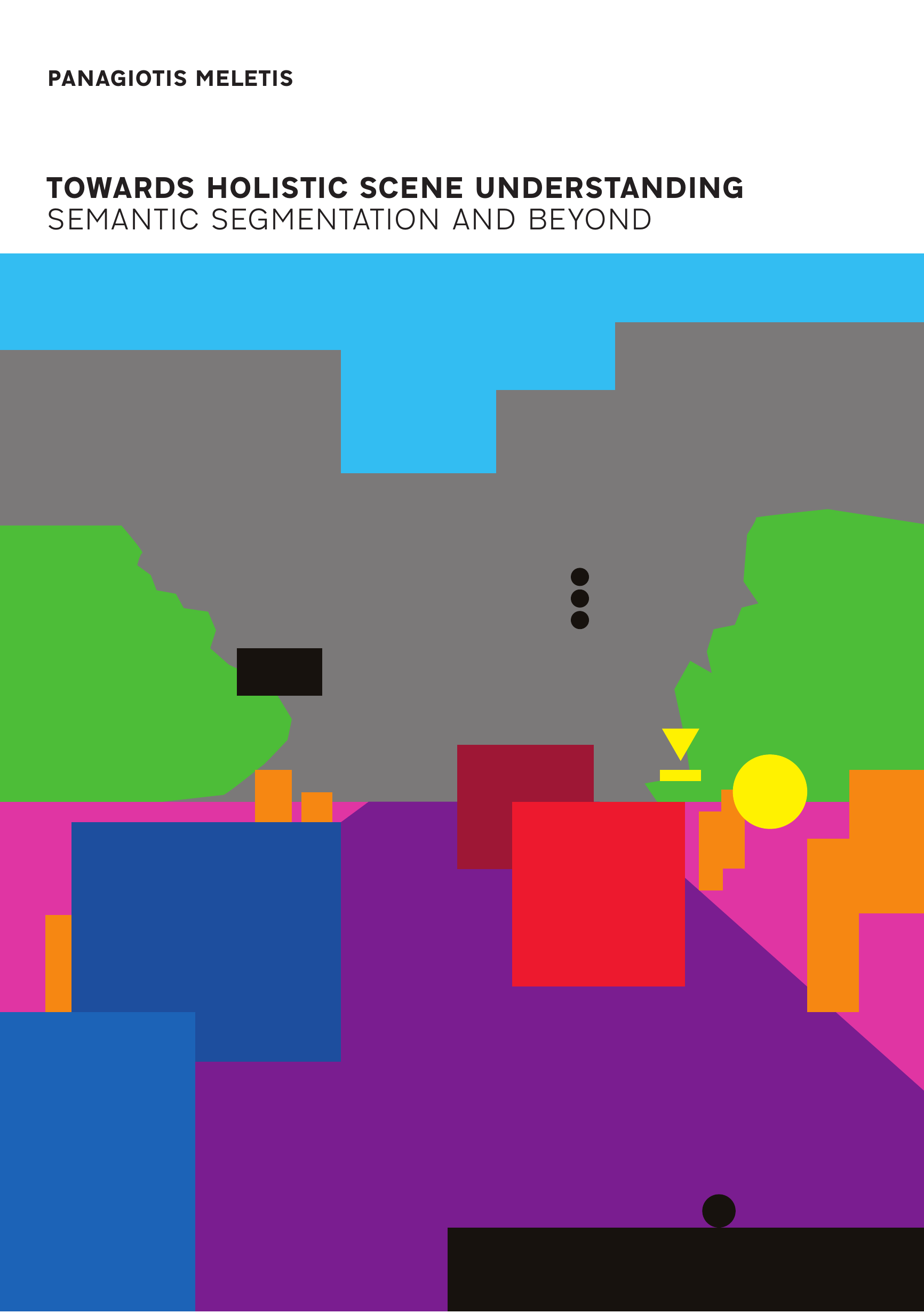}

\begin{romanpages}

\maketitle




\dominitoc 

\noindent Dit proefschrift is goedgekeurd door de promotor en de samenstelling van de promotiecommissie is als volgt:

\begin{table}[h]
	\begin{tabular}{ll}
		voorzitter: & prof. dr. J. K. Kok\\
		1e promotor: & prof. dr. ir. P. H. N. de With\\
		copromotor: & dr. G. Dubbelman\\
		leden: & prof. dr. W. Philips (University of Gent)\\
		           & dr. J. van Gemert (TU Delft)\\
		           & dr. ir. S. Stuijk\\
		           & prof. dr. S. Kollias (NTU Athens)\\
		           & dr. M. Nieto (Vicomtech, San Sebastián)
	\end{tabular}
\end{table}

\vspace*{\fill}

\noindent Het onderzoek of ontwerp dat in dit proefschrift wordt beschreven is uitge-voerd in overeenstemming met de TU/e Gedragscode \mbox{Wetenschapsbeoefening}.

\clearpage
\begin{flushright}
	\thispagestyle{empty}
	\vspace*{\fill}
	To my Family
\end{flushright}
\clearpage

\clearpage
\thispagestyle{empty}
\vspace*{\fill}

\hlinefull
\bigskip

\noindent Towards holistic scene understanding:\\Semantic segmentation and beyond
\bigskip

\noindent Panagiotis Meletis
\bigskip

\noindent Cover design: S. Grammenos, gramsdesign.gr
\bigskip

\noindent Printed by: Gildeprint drukkers
\bigskip

\noindent A catalogue record is available from the\\Eindhoven University of Technology Library\\
ISBN: 978-90-386-5381-5
\bigskip

\noindent NUR-code 959
\bigskip

\hlinefull
\bigskip

\noindent Copyright \textcopyright ~2021 by Panagiotis Meletis
\bigskip

\noindent All rights reserved. No part of this material may be reproduced or transmitted in any form or by any means, electronic, mechanical, including photocopying, recording or by any information storage and retrieval system, without the prior permission of the copyright owners.
\clearpage

\clearpage

\addcontentsline{toc}{chapter}{Summary}

\begin{flushright}
\huge
Summary
\bigskip
\bigskip
\end{flushright}


\begin{center}
\textbf{Towards holistic scene understanding:\\Semantic segmentation and beyond}
\end{center}

Scene understanding is an indispensable part of any autonomous system that needs to sense its surroundings to navigate and plan actions safely. The recent leaps in autonomous vehicles, robots, and augmented reality systems are strongly coupled with the corresponding platforms' capabilities to perceive their environments as humans do. Humans discover their world and create a very detailed model on different levels of abstraction. For example, persons can easily detect dynamic objects of a scene and distinguish them from their static environment. Humans can discern hundreds or even thousands of semantic concepts in their environment, for which they maintain a collection of useful attributes which are used for subsequent reasoning. Furthermore, humans decompose dynamic objects into their constituent parts, which assists them in anticipating their future trajectories. In the heart of all those functionalities lie the human capabilities for hierarchical and holistic scene understanding. Instilling these capabilities into an autonomous platform is a challenging and multi-perspective problem, but when solved it can offer numerous benefits. For example, a road accident involving an erroneous action of a pedestrian can be prevented if the pedestrian's intention is predicted using the position and orientation of his body parts. A robot or vehicle is able to reason better about the elements of a scene if it has detailed knowledge of semantics, properties, and locations of the objects surrounding it. Holistic scene understanding is the concept that will enable autonomous platforms to act independently and safely coexist in our environment.

A variety of sensory data can be exploited for scene understanding. Imagery is the most common modality due to the small size, low cost, and widespread availability of visual sensors. The recent advances of deep learning methods and especially convolutional networks for the analysis of this type of modality are another compelling factor. Convolutional network-based solutions have demonstrated immense capabilities in analyzing image data and form presently the state-of-the-art in semantic understanding and image recognition. This thesis focuses on visual scene understanding and investigates solutions based on contemporary convolutional networks.

Visual data contain rich information, but human-level recognition and scene understanding remains an open problem for deep learning systems. Scene understanding is traditionally partitioned into well-defined sub-tasks,~\eg image classification, object detection, semantic segmentation, or parts segmentation. Although this fragmentation is practical for concisely defining metrics and comparing systems, it leads to incoherences when different abstractions have to be consolidated. The common practice to solve a specific scene understanding sub-task is to collect a sufficiently rich dataset, annotate it according to task requirements, and train a deep network with the annotated data. However, when such a network is deployed in real-life scenarios or in environments that are not represented in the dataset, it often demonstrates poor performance, while its knowledge is limited to the semantics of the training dataset. There are at least two intertwined factors responsible for this undesired behavior,~\ie the richness of the dataset itself and the detailing level of the specific sub-task. On one hand, in a single, specialized dataset it is challenging to capture the tremendous diversity and the ever-changing conditions of our world. On the other hand, the more detailed predictions are required for a sub-task, the scarcer the available annotations and images become. For example, annotating datasets for semantic segmentation is up to 79 times more time-consuming than annotation for image classification, which is frequently causing a size difference of one to two orders of magnitude between available datasets. Instead of continuously generating larger and costlier datasets, an intuitive solution to this data shortage is to investigate how existing less-detailed datasets can be leveraged into more detailed tasks. This in turn demands to generalize the training procedure of deep networks, so it can admit a variety of datasets that are not specifically tailored for the task at hand.

The first part of this thesis explores how datasets with incompatible annotation types and conflicting semantics,~\ie heterogeneous datasets, can be combined for seamless training of semantic segmentation, in order to improve performance, generalizability and semantic knowledgeability of deep networks. The second part of the thesis investigates how the tasks of panoptic segmentation and part segmentation can be combined for solving them coherently and evaluating them consistently, which is a step towards holistic scene understanding.

Chapter~\ref{ch:3-iv2018} addresses semantic segmentation and explores benefits from training on multiple datasets in the context of street scene understanding. Existing datasets are limited in size and semantic diversity, and thus we propose to combine different semantic segmentation datasets, although they have conflicting semantics. To solve this issue, a framework is designed of hierarchical classifiers over a single convolutional backbone, which is then trained end-to-end on the combination of datasets. The framework enhances the mean IoU performance by up to +24.3\% for seen datasets, by +5.3\% in the cross-dataset (unseen) setting, while it expands the output semantic spaces. The results attained 3\textsuperscript{rd} place overall and 1\textsuperscript{st} place in WildDash, in the CVPR 2018 Robust Vision Challenge. Chapter~\ref{ch:4-iv2019} extends the framework of Chapter~\ref{ch:3-iv2018} by enriching semantic segmentation with weak supervision. Specifically, the described research proposes a mixed fully and weakly-supervised algorithm for training convolutional networks with bounding-box and image-tag supervision in conjunction with pixel supervision. Weak supervision for selected classes boosts the IoU performance by up to +13.2\%. However, the increasing number of employed datasets for simultaneous training augments the memory and computational load challenges. These aspects are addressed in Chapter~\ref{ch:5-itsc2019-wacv2019}, where we propose two methodologies for selecting informative and diverse image-label pairs from datasets with weak supervision, to reduce network training time and related ecological footprint without sacrificing performance. Specifically, the selection procedures reduce the training time by 20-90\%, while they require 100~times less pairs.

Building on the ideas from~\Cref{ch:3-iv2018,ch:4-iv2019,ch:5-itsc2019-wacv2019} and motivated by memory and computation efficiency requirements, Chapter~\ref{ch:6-journal}, reconsiders simultaneous training on heterogeneous datasets and improves the earlier described framework. The so-called Heterogeneous Training of Semantic Segmentation (HToSS) framework introduces the concept of semantic atoms for solving semantic conflicts between label spaces and incorporates the mixed fully and weakly-supervised training. HToSS exploiting various scene understanding datasets and achieves consistent gains in performance metrics that reach +20\% mIoU for seen (training) datasets, +16.6\% for unseen (generalization) datasets, and a relative increase of up to 250\% to the number of recognizable semantic classes. Although the trained networks for semantic segmentation demonstrate high performance and semantic knowledgeability, their predictions do not capture relevant scene abstractions,~\eg they neither distinguish individual objects nor parts that make up an object. To capture this information, Chapter~\ref{ch:7-panoptic} introduces the novel task of Part-aware Panoptic Segmentation (PPS), which consists of a coherent step towards holistic scene understanding. This task combines scene-level and part-level semantics together with instance-level enumeration, three abstractions that were not jointly investigated in the past. PPS is accompanied by two novel datasets to facilitate conflict-free training and evaluation, enabling possible future information interchange between the abstractions. Since the PPS task and the proposed datasets are novel, two networks are trained on them for setting the baselines.

In conclusion, the realized contributions span over convolutional network architectures, mixed fully and weakly-supervised learning, data selection, and part-aware panoptic segmentation. The promising results obtained for maximally exploiting heterogeneous training data for semantic segmentation, minimizing required computational resources and unifying scene and parts parsing, pave the way towards a holistic, knowledgeable, and sustainable visual scene understanding.


\cleardoublepage

\addcontentsline{toc}{chapter}{Samenvatting}

{
\selectlanguage{dutch}
\begin{flushright}
	\huge
	Samenvatting
	\bigskip
	\bigskip
\end{flushright}

Voor autonome systemen is het begrijpen van de omgeving een onmisbaar onderdeel om te kunnen navigeren en op een veilige manier acties te kunnen uitvoeren. De recente ontwikkelingen in autonome voertuigen, robots en augmented reality-systemen zijn sterk gerelateerd aan het begrijpen van de omgeving op een vergelijkbare manier als mensen dat doen. Mensen ontdekken en begrijpen hun omgeving op verschillende abstractieniveaus met een gedetailleerd model. Mensen kunnen bijv. bewegende objecten herkennen en deze onderscheiden van de statische omgeving. Daarnaast kunnen mensen een groot aantal semantische concepten in hun omgeving onderscheiden, herinneren en vervolgens toepassen in nieuwe situaties. Bovendien begrijpen mensen de verschillende onderdelen waaruit bewegende objecten zijn samengesteld, waardoor ze beter anticiperen op hoe die objecten zich gaan verplaatsen. Het menselijk vermogen om omgevingen te begrijpen op een hiërarchische en holistische manier staat hierin centraal. Het omzetten van deze vaardigheden in een autonoom platform is een uitdagend probleem, maar de mogelijke oplossing kan vele voordelen bieden. Zo kan bijv. een verkeersongeval waar een voetganger foutief handelt worden voorkomen als zijn intentie wordt voorspeld aan de hand van de positie en oriëntatie van zijn lichaamsdelen. Een voertuig kan de verschillende elementen in zijn omgeving beter begrijpen als het gedetailleerde kennis heeft van de betekenis, eigenschappen en locaties van de omliggende objecten. Het holistisch begrijpen van de omgeving is essentieel om autonome systemen onafhankelijk en veilig te laten handelen in hun omgeving.

Een scala aan sensordata kan worden gebruikt voor het begrijpen van scènes. Camerabeelden zijn de meest voorkomende modaliteit vanwege hun kleine formaat, de lage kosten en de brede beschikbaarheid van visuele sensoren. De recente ontwikkelingen van deep learning-methoden en vooral convolutionele netwerken voor de analyse van beeldgegevens zijn andere motiverende factoren. Oplossingen die gebaseerd zijn op convolutionele netwerken, hebben een enorme progressie doorgemaakt bij het analyseren van beelddata en vormen momenteel de state-of-the-art in de semantiek van beelden. Dit proefschrift focusseert op het begrijpen van visuele scènes en onderzoekt oplossingen op basis van moderne convolutionele netwerken.

Visuele gegevens bevatten zeer rijke informatie, maar herkenning op menselijk niveau en begrip van scènes blijft een probleem voor deep learning-systemen. Het begrijpen van scènes wordt traditioneel opgedeeld in goed gedefinieerde deeltaken, bijvoorbeeld beeldclassificatie, objectdetectie, semantische segmentatie of segmentatie van objectonderdelen. Hoewel deze fragmentatie het mogelijk maakt om metriekparameters  te definiëren en systemen te vergelijken, leidt het tot incoherenties wanneer verschillende abstracties tegelijk worden toegepast. Een populaire manier om een specifieke deeltaak voor het begrijpen van een scène te adresseren, is een omvangrijke dataset te verzamelen, deze te annoteren volgens de taakeisen, en een diep netwerk te trainen met de geannoteerde data. Wanneer een dergelijk netwerk echter wordt toegepast in realistische  scenario’s of in omgevingen die niet in de dataset aanwezig zijn, geeft het systeem vaak slechte prestaties, terwijl het begrip beperkt is tot de semantiek van de trainingsdataset. Dit wordt veroorzaakt door de omvangrijkheid van de dataset en het detailniveau van de specifieke deeltaak. Het annoteren van datasets voor bijv. semantische segmentatie is tot aan 79 keer tijdrovender dan voor beeldclassificatie. In plaats van voortdurend grotere en duurdere datasets te genereren, is een attractieve oplossing om bestaande, minder gedetailleerde datasets te gebruiken voor meer gedetailleerde taken. Dit heeft tot gevolg dat de trainingsprocedure van diepe netwerken gegeneraliseerd moet worden, zodat deze kan omgaan met een verscheidenheid aan datasets die niet specifiek bedoeld zijn voor de actuele taak.

Het eerste deel van dit proefschrift onderzoekt hoe datasets met incompatibele annotatietypes en conflicterende semantiek, m.a.w. heterogene datasets, kunnen worden gecombineerd voor een naadloze training van semantische segmentatie, met als doel om de prestaties, generaliseerbaarheid en semantische kennis van diepe netwerken te verbeteren. Het tweede deel van het proefschrift onderzoekt hoe de taken van panoptische segmentatie en objectdeelsegmentatie kunnen worden gecombineerd en samenhangend opgelost om ze daarna consistent te evalueren. Dit leidt tot een stap in de richting van holistisch begrip van scènes. 

Hoofdstuk~\ref{ch:3-iv2018} behandelt semantische segmentatie en exploreert de voordelen van training op meervoudige datasets in de begripscontext van straatbeelden. Bestaande datasets zijn beperkt in omvang en semantische diversiteit. Daarom wordt voorgesteld om verschillende semantische segmentatiedatasets te combineren, hoewel ze tegenstrijdige semantiek hebben. Het hiervoor ontworpen raamwerk bevat hiërarchische classificaties en is afgebeeld op een enkel convolutioneel netwerk, dat vervolgens end-to-end wordt getraind op de combinatie van datasets. Het raamwerk verbetert de gemiddelde IoU-prestaties met maximaal +24,3\% voor de trainingsdatasets, met +5,3\% in de cross-dataset (ongeziene) context, terwijl het de semantische ruimtes verbreed in de output. De resultaten hebben een derde plaats bereikt in algemene zin en de eerste plaats in WildDash, in de CVPR 2018 Robust Vision Challenge. Hoofdstuk~\ref{ch:4-iv2019} breidt het raamwerk van Hoofdstuk~\ref{ch:3-iv2018} uit door semantische segmentatie te verrijken met partiële supervisie. Specifiek stelt het beschreven onderzoek een gemengd algoritme voor met zowel volledige als partiële supervisie, voor het trainen van convolutionele netwerken met supervisie met gebruik van rechthoekige detectieventers en beeld-labels in combinatie met per-pixel supervisie. Partiële supervisie voor geselecteerde klassen verhoogt de IoU-prestaties tot aan +13,2\%. Het groeiende aantal gebruikte datasets voor gelijktijdige training vergroot echter het gebruik van geheugen en rekenkracht. Deze aspecten worden behandeld in Hoofdstuk~\ref{ch:5-itsc2019-wacv2019}, waar twee methodologieën worden beschreven voor het selecteren van informatieve en diverse beeld-labelparen uit datasets met partiële supervisie, om de trainingstijd van het netwerk en de gerelateerde kosten te verminderen zonder verlies in kwaliteit. De voorgestelde selectieprocedures verkorten de trainingstijd met 20-90\%, terwijl ze 100 keer minder dataparen nodig hebben.

Voortbouwend op de ideeën van Hoofdstuk~\ref{ch:3-iv2018} t/m Hoofdstuk~\ref{ch:5-itsc2019-wacv2019} en gemotiveerd door eisen voor geheugen- en rekenefficiëntie, adresseert  Hoofdstuk~\ref{ch:6-journal} opnieuw gelijktijdige training op heterogene datasets, waarbij het eerder beschreven raamwerk wordt verbeterd. Het zogenaamde Heterogeneous Training of Semantic Segmentation (HToSS) raamwerk introduceert het concept van semantische atomen voor het oplossen van semantische conflicten tussen labelruimtes en exploiteert ook de training met volledige en partiële supervisie. HToSS gebruikt verschillende datasets voor het begrijpen van scènes en behaalt consistente winsten in gemeten prestaties, die +20\% mIoU bereiken voor geziene (training) datasets, +16,6\% voor ongeziene (generalisatie) datasets, en een relatieve toename tot aan 250\% t.o.v. het aantal herkenbare semantische klassen. Hoewel de getrainde netwerken voor semantische segmentatie hoge prestaties en semantische kennis laten zien, realiseren hun voorspellingen geen relevante scène-abstracties, bijv. ze onderscheiden geen individuele objecten of delen daarvan. Om deze reden introduceert Hoofdstuk~\ref{ch:7-panoptic} de nieuwe taak van Part-aware Panoptic Segmentation (PPS), die semantiek op scène- en (object)deelniveau combineert met onderscheid op instantieniveau; drie abstracties die niet eerder gezamenlijk zijn onderzocht. Naast PPS zijn twee nieuwe datasets gebruikt om conflictvrije training en evaluatie te faciliteren, waardoor informatie-uitwisseling tussen de abstracties mogelijk wordt in de toekomst. Omdat de PPS-taak en de voorgestelde datasets nieuw zijn, worden twee individuele netwerken getraind om een uitgangswaarde te bepalen. 

Concluderend kan worden gesteld dat het proefschrift relevante bijdragen geeft aan convolutionele netwerkarchitecturen, het leren van volledige en partiële supervisie, dataselectie en objectdeel-gebaseerde panoptische segmentatie. Het maximaal benutten van heterogene trainingsgegevens voor semantische segmentatie, het minimaliseren van de benodigde rekenkracht en het verenigen van scène- en objectdeelanalyse, geven veelbelovende resultaten en bieden een perspectief naar een holistisch, goed onderbouwd en duurzaam begrip van visuele scènes.
}

\flushbottom

\tableofcontents




\end{romanpages}

\flushbottom

\chapter{\label{ch:1-intro}Introduction} 

This dissertation investigates scene understanding via image segmentation with an emphasis on autonomous vehicle applications and city-scene analysis. In this Chapter, the context of the dissertation is provided, the objectives are motivated and the contributions are presented in condensed form.

To this end, the opening Sections~\ref{ch1:sec:surw},~\ref{ch1:sec:cvdl} describe the context of the thesis and are dedicated to a historical perspective of computer vision and deep learning. Section~\ref{ch1:sec:scene-und} discusses an overview of scene understanding major tasks. The challenges and research questions are addressed in Section~\ref{ch1:sec:challenges-rq}. The contributions of the performed research are summarized in Section~\ref{ch1:sec:contributions}. Finally, the layout and organization of the thesis are presented in Section~\ref{ch1:sec:outline}.

\section{Scene understanding in the real world}
\label{ch1:sec:surw}
Scene understanding in the real world are of paramount importance and have numerous applications that ameliorate human lives, support society and provide efficiency in everyday tasks. The outcomes of scene understanding can improve safety in a variety of environments, including urban and rural regions, indoors or outdoors areas. For example and with focus on understanding, street scenes can be analyzed to extract information for traffic participants in order to optimize the traffic flow. Among other fields, scene and image understanding has tremendous potential in healthcare and sport events, and can also be applied for commercial purposes,~\eg to analyze client behavior. Visual surveillance and scene understanding are noting immense progress because they are developing in parallel with the growing market of visual sensors since the early years of this millennium. Visual surveillance at street level is one of the most mature areas of scene understanding with a dominant focus on person detection/recognition and human behavioral analysis.
The previous developments have fueled the continuous improvement of capturing devices such as visual sensors and affordable camera lenses. The low cost and small dimensions of these sensors have led to their omnipresent existence and integration into our everyday apparatus and devices, ranging from handheld smartphones to cameras assisting in car driving, and from harbor surveillance aiming at vessels to corridor monitoring in supermarkets.

The presence of image sensors is steadily contributing to an accumulation of large volumes of visual data. Moreover, an increasing number of companies and institutions are collecting visual data in a unprecedented scale and anticipate benefits from their exploitation. However, the abundance of raw visual data is not profitable, unless they are carefully organized into datasets, processed by automated means, and practical information is extracted from them. The growth of datasets also implies the increasing demands for computing.

The processing of the datasets can be achieved by computer vision and data analytics algorithms. The algorithms developed in the early ages of computer vision were aiming for lower-level processing and feature extraction and are nowadays less adequate for the high-level information extraction required by modern vision and surveillance systems. These systems should analyze visual information for static and dynamic objects, distinguish foreground and background, and detect moving and deformable objects. The analysis should also be applied at many levels of abstraction,~\eg at scene level and object parts level. Finally, an emerging analysis direction is the extraction of semantics, which could be used by other control or decision systems.

The semantics of a depicted scene,~\ie the human understandable concepts about the elements of the scene, are essential for identifying and categorizing a scene. For example, in street scenes identifying traffic participants (pedestrians, vehicles, traffic signs) is vital for automated driving algorithms or traffic flow optimization. As another example, in a port, vessels need to be categorized in cargo, commercial ships, river boats, cruising ships, etc. Humans have excellent capabilities in figuring the semantics of their environment. Similarly, the learning algorithms have the potential to grasp more features from the data that cannot be found by handcrafted feature extraction procedures. Thus, modern machine learning systems enable the deployed application of the understanding of objects and their actions from the previous examples to obtain a more general understanding of the scene and a higher recognition performance. In the following section, a brief historic perspective of these developments is presented.

\section{Computer Vision and Machine Learning}
\label{ch1:sec:cvdl}
Computer vision has a rich history. From the first efforts of introducing analog imaging in the early 19\textsuperscript{th} century until the commercialization of photography and the introduction of electronic cameras in the 2000s, people have used photos to capture moments and preserve visual information of any aspect of daily life. Two centuries later, image sensors are omnipresent in various devices that are daily used from smartphones to computers and from home use to professional applications. Besides this widespread use, the construction of these devices has been dramatically revolutionized by the transition to the digital data acquisition and processing. The easiness of image acquisition and the low-cost storage and data sharing, has created an exponential increase in captured images and information stored in them, showing an ever growing importance of imaging to the society. Indicatively, at present, every minute, 243,000 images are uploaded to Facebook\footnote{\url{https://www.brandwatch.com/blog/facebook-statistics}. Website accessed 28/06/21.}, and 500 hours of video to Youtube\footnote{\url{https://www.statista.com/statistics/259477/hours-of-video-uploaded-to-youtube-every-minute}. Website accessed 28/06/21.}. Human vision has been proven inadequate to analyze and process this immense amount of information, and automated visual processing is gradually taking over this task.

The benefits of computer vision and understanding of the content of images were foreseen and predicted from the late 1960s. Papert~\cite{papert1966summer} has introduced a summer project to mimic the human visual system, and many efforts to endow robots with intelligent vision have been taking place. Early studies in the 1970s have approached the problem from low-level feature extraction and image processing, finding edges, corners, or blobs, analyzing through color, and other pixel properties. Human-level perception of images has been proven very difficult, and in the next two decades (1980s - 2000s), computer vision was based on more rigorous mathematical analysis. The ideas of scale-space, contour models (snakes), active appearance models, and higher-level image content concepts such as objects, shapes, and textures have emerged in research. These methods were based on mathematical formulations, were verifiable and their limits could be identified using mathematical and explainable proofs.

Machine learning, a term coined by Arthur Samuel in the 1950s, was initially concerned with attempts to model the human brain. It started developing in parallel with computer vision, but their paths were not yet intersected. The invention of the perceptron (1957) was probably the first attempt to use neural networks for image recognition trained using supervised learning. The extension of the single-layer perceptron to networks with multiple layers in the mid-1960s and the development of backpropagation (1970) set the building blocks of contemporary machine learning and feedforward networks. Despite the progress at theoretical and practical levels (Universal Approximation theorem, Boosting) in the following years, machine learning and neural networks were not employed in large-scale computer vision problems.

The revolution for visual image data happened in 2012, and large-scale image classification using a convolutional neural network trained in a supervised learning regime is accomplished in the seminal work of Krizhevsky et al~\cite{krizhevsky2012alexnet}. The three enabling factors that led to that success were: i) Big data: the large-scale annotated ImageNet dataset, ii) Big compute: the large computational power of modern GPUs, and iii) Big networks: engineering tweaks for enabling training of multiple-layer (deep) networks. This approach, together with earlier successful applications of deep networks in natural language processing, gave birth to the Deep Learning trend in computer vision.

\section{Scene Understanding tasks}
\label{ch1:sec:scene-und}
Scene understanding is a vital component of vision systems, as explained in Section~\ref{ch1:sec:surw}. In the automated driving field, which is the main topic of this thesis, scene understanding can contribute to evade traffic accidents and thus loss of lives. Moreover, its results can be exploited from authorities for optimizing traffic or from individual vehicles for reducing trip times. Traffic accidents are a result of multiple interacting factors. The human error contributes to at least 90\% of the vehicle accidents that occur on roadways according to several studies~\cite{national20172016,pakgohar2011role}. These include long-term aspects (inexperience, age) or short-term causes (fatigue, limited view, distraction). Other influential factors for accidents are 70\% environmental and partly overlapping with 30\% vehicle-related issues, as indicated in the previously mentioned studies. These findings signify that by providing awareness to the mobility platforms for their surroundings (street conditions, traffic participants) and also their interior (driver detection, gaze tracking), the potential is created for reducing traffic incidents. This awareness can be provided by analyzing image data from low-cost, vehicle-mounted visual sensors through image scene understanding.

Scene understanding lies at the intersection of computer vision and machine learning. It is an umbrella concept that contains a variety of tasks related to the analysis of the scene structure and the identification of objects therein. This work concentrates on 2D image scene understanding, dominantly with using individual frames extracted from video sequences. Other perspectives of general scene understanding are not investigated, for example 3D perception or video-based temporal processing of the analyzed images. For the purpose of this thesis, we define four main aspects of image scene understanding related to a scene and its elements. These aspects will assist for the characterization and completeness of dataset annotations and prediction results throughout the thesis.
\begin{itemize}[noitemsep, topsep=0pt]
	\item \textit{Semantics}: the assignment of human-understandable semantic concepts to a scene or its regions. This is typically encoded by semantic classes,~\eg car, tree, pedestrian, etc.
	\item \textit{Localization}: the delineation of scene element positions in a topological or metric manner,~\eg with respect to their surroundings (relative), or positions based on image coordinates (absolute). The delineation is provided on a range of granularities, from coarse-to-fine localization,~\eg with bounding boxes or a region described at pixel level, and going from in-image localization to 3D mapping or world-based localization.
	\item \textit{Identity enumeration}: the identification of distinct instances of countable semantic classes in a scene. Some scene understanding tasks require the enumeration and separation of the localized semantic elements in the image and even within-class identity enumeration (similar objects at different locations).
	\item \textit{Coverage}: each task aims for understanding of different elements of a scene,~\eg some tasks concern only objects (things), other only parts of objects, and others only static scene components (stuff).
\end{itemize}

\begin{figure}
	\centering
	\begin{subfigure}{0.48\linewidth}
		\centering
		\includegraphics[width=\linewidth]{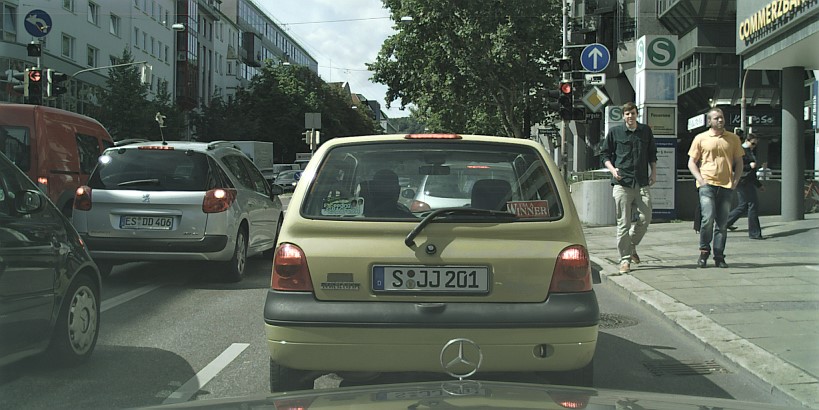}
		\caption{Input image}
	\end{subfigure}
	~~
	\begin{subfigure}{0.48\linewidth}
		\centering
		\includegraphics[width=\linewidth]{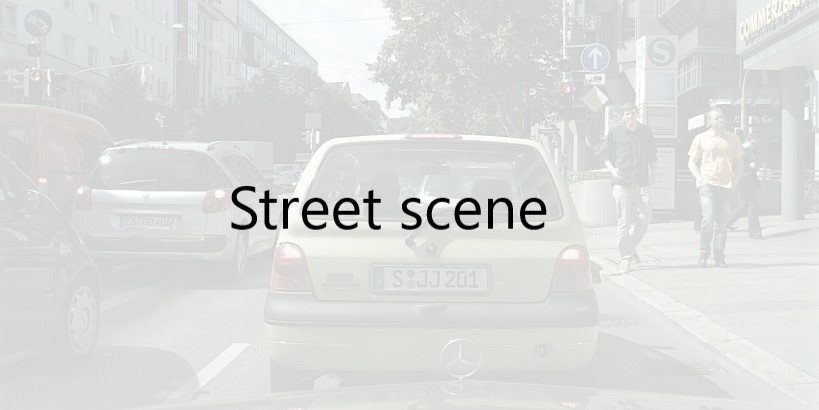}
		\caption{Scene recognition}
		\label{ch1:fig:task-def-screc}
	\end{subfigure}

	\vspace{8pt}

\begin{subfigure}{0.48\linewidth}
	\centering
	\includegraphics[width=\linewidth]{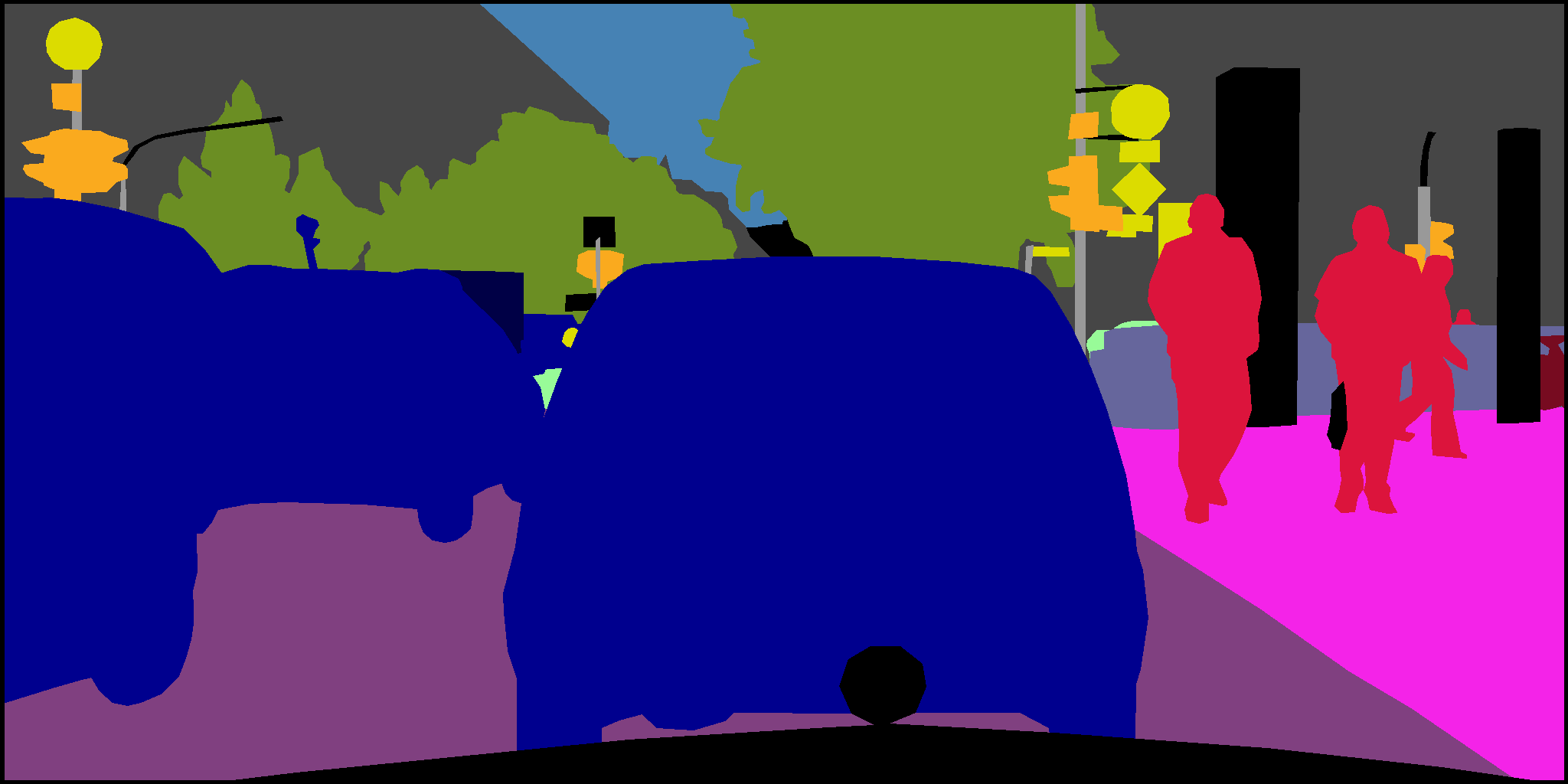}
	\caption{Semantic segmentation}
	\label{ch1:fig:task-def-semseg}
\end{subfigure}
~~
\begin{subfigure}{0.48\linewidth}
	\centering
	\includegraphics[width=\linewidth]{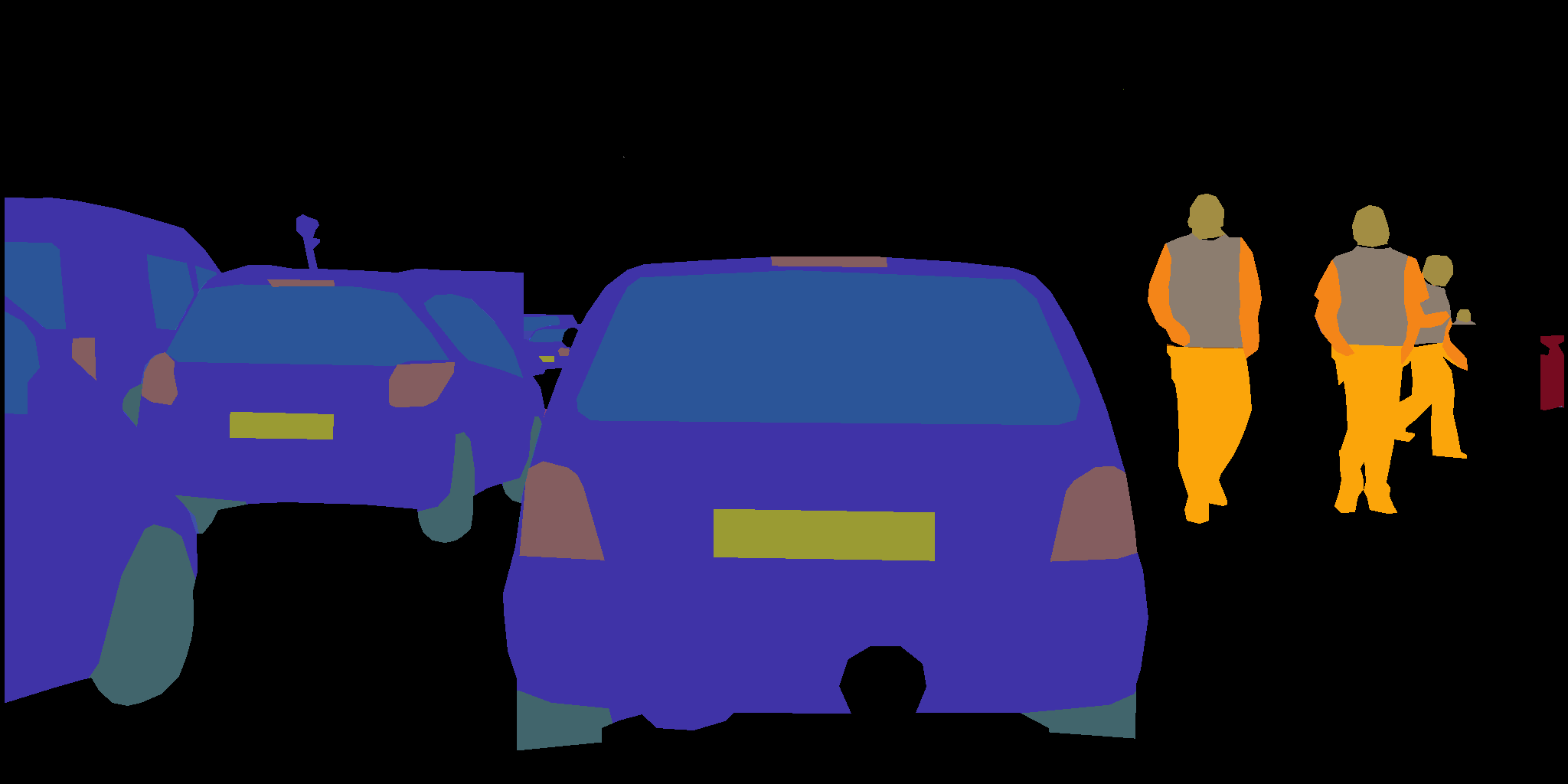}
	\caption{Part segmentation}
	\label{ch1:fig:task-def-partseg}
\end{subfigure}

\vspace{8pt}

	\begin{subfigure}{0.48\linewidth}
		\centering
		\includegraphics[width=\linewidth]{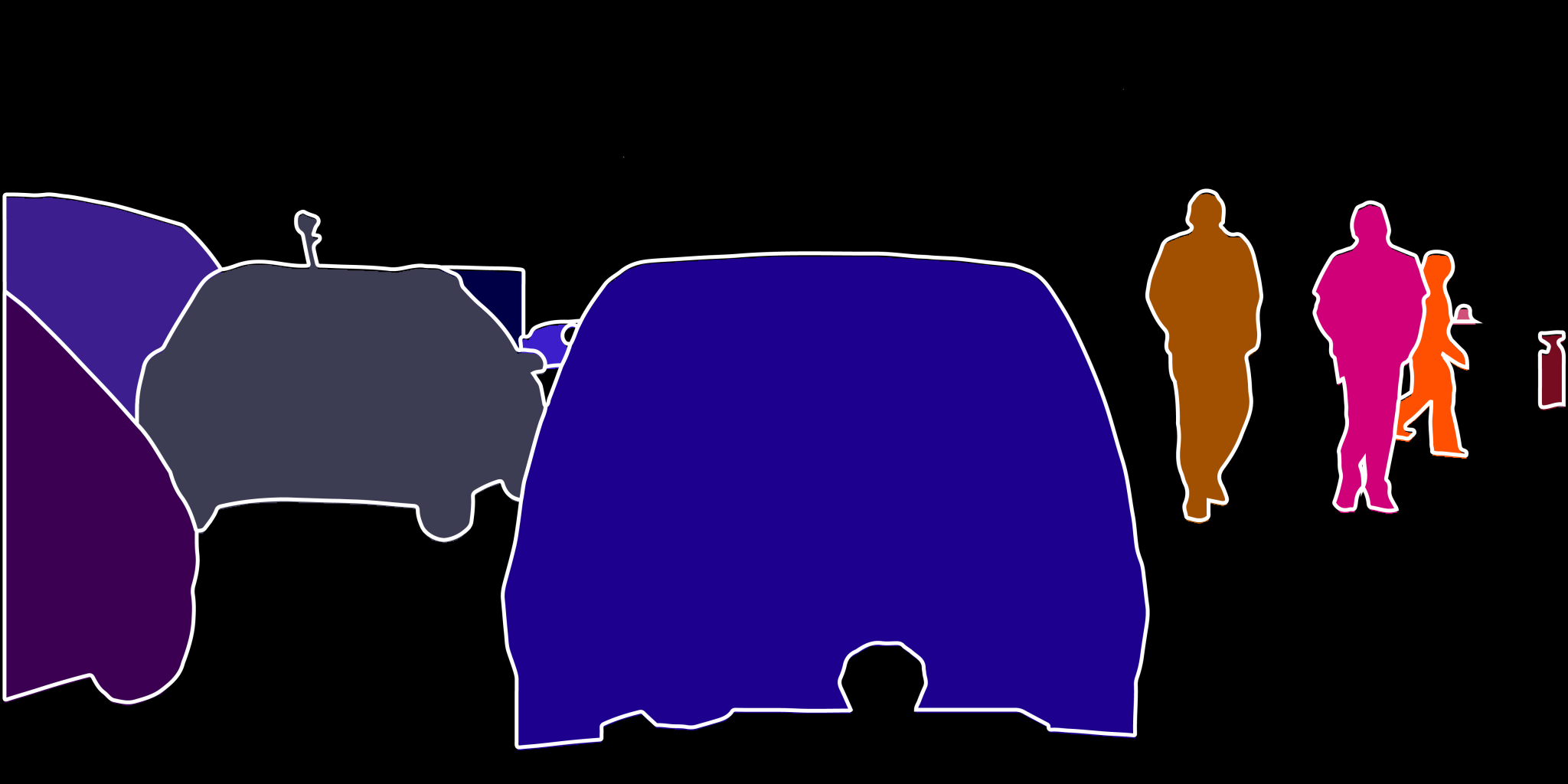}
		\caption{Instance segmentation}
		\label{ch1:fig:task-def-instseg}
	\end{subfigure}
	~~
	\begin{subfigure}{0.48\linewidth}
		\centering
		\includegraphics[width=\linewidth]{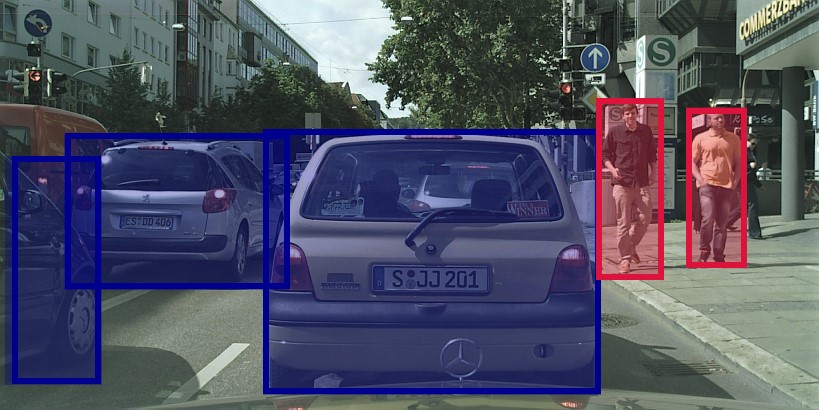}
		\caption{Object detection}
		\label{ch1:fig:task-def-objdet}
	\end{subfigure}

\vspace{8pt}

	\begin{subfigure}{0.48\linewidth}
		\centering
		\includegraphics[width=\linewidth]{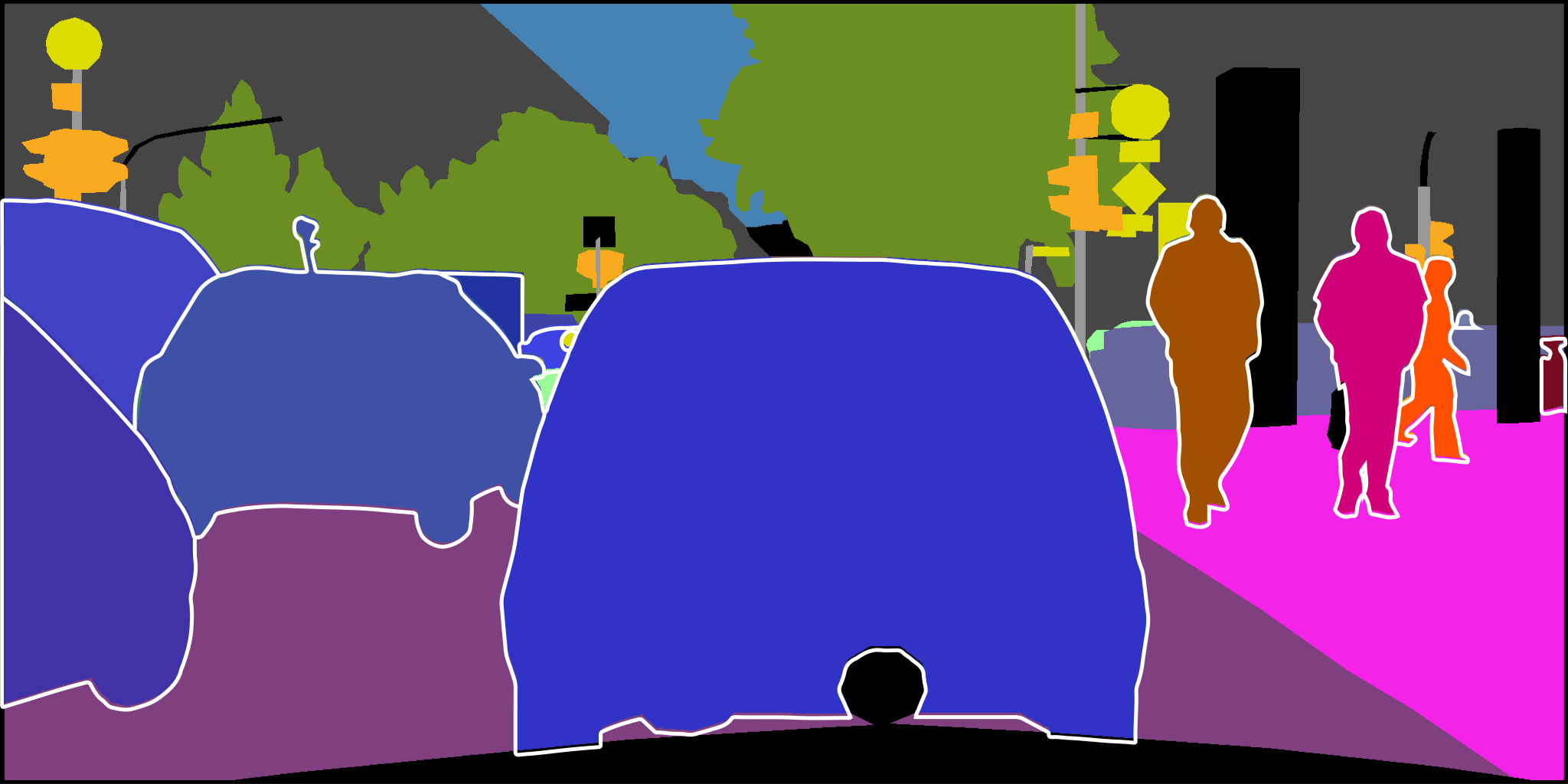}
		\caption{Panoptic segmentation}
	\end{subfigure}
	~~
	\begin{subfigure}{0.48\linewidth}
		\centering
		\includegraphics[width=\linewidth]{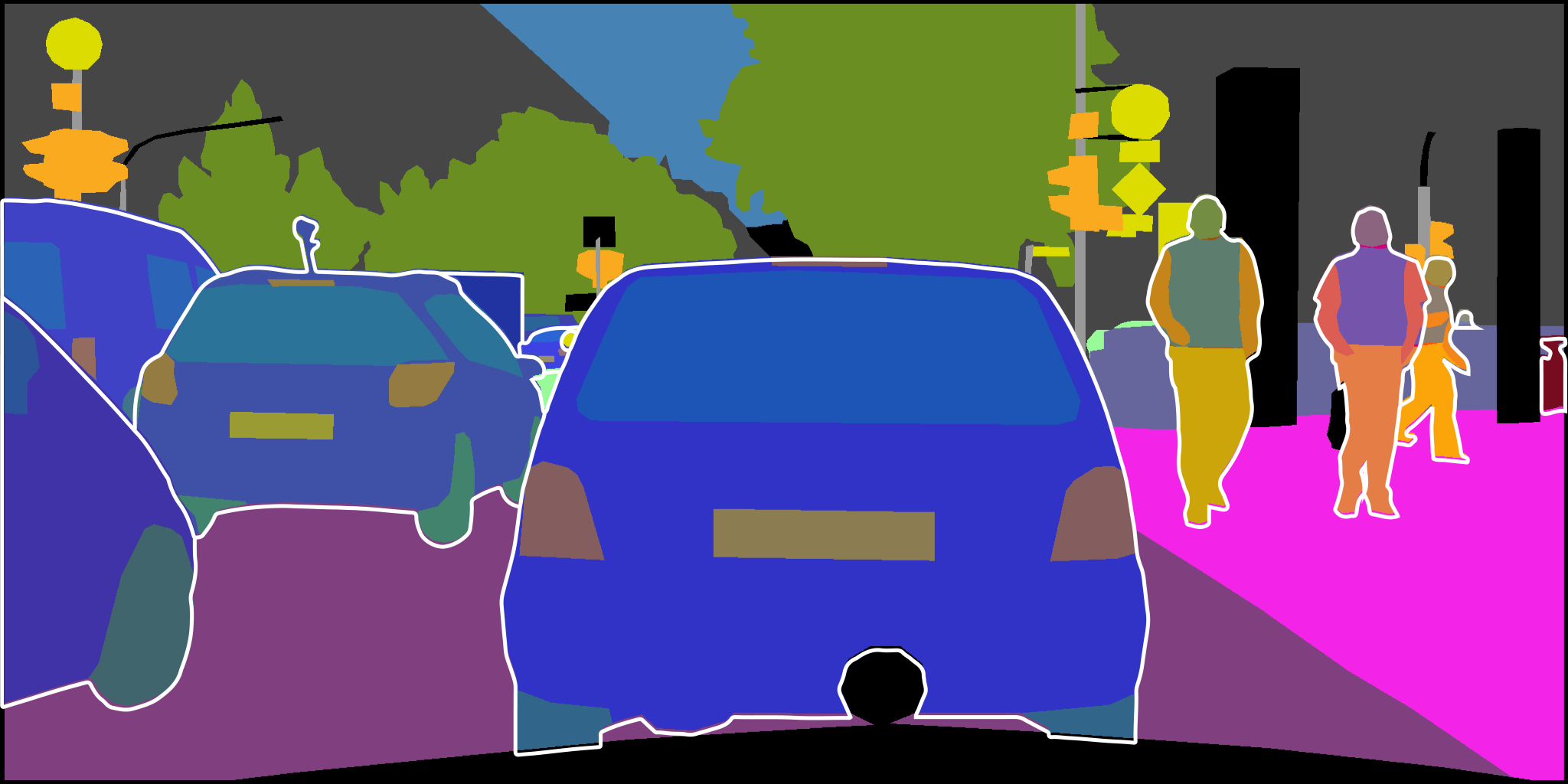}
		\caption{Part-aware panoptic segmentation}
	\end{subfigure}
	
	\caption{Various image scene-understanding tasks. \textit{Semantics} (scene or part-level) are indicated by colors. Annotations (\textit{localization}) are provided at image, bounding box, or pixel level. Object-instance boundaries (\textit{identity enumeration}) are emphasized with a white contour and color hues. The lack of annotations (\textit{coverage}) is indicated by pixels with black color or areas without bounding boxes.}
	\label{ch1:fig:task-def}
\end{figure}

Scene understanding represents an important stage of many downstream applications,~\eg autonomous driving, surveillance, augmented reality, and as such, it has received broad attention in recent years. A non-exhaustive list of major image scene-understanding tasks is provided below and depicted in Figure~\ref{ch1:fig:task-def}. All these tasks are accompanied by corresponding datasets for training and evaluation of algorithms, as well as specific metrics for quantifying performance for such tasks. The tasks and the involved aspects are described in the form of short paragraphs.

\paragraph{Image classification / Scene recognition} Classification aims to assign one or more semantic class labels/tags to an image, from a set of predefined labels, usually corresponding to the dominant object contained in it. The objective of scene recognition is to find the type of scene that the image depicts, and is an adjoint task to image classification, since the label corresponds to the direct surroundings or environment of dominant objects (Figure~\ref{ch1:fig:task-def-screc}).

\paragraph{Object detection / Instance segmentation} The detection of objects in an image can have many different levels of localization. Two of the most prominent tasks are shown in~\Cref{ch1:fig:task-def-objdet,ch1:fig:task-def-instseg}. The first one, object detection with bounding boxes, seeks for a set of, possibly overlapping, bounding boxes together with their semantic class labels that correspond to objects in the scene. The second one, instance segmentation, requires a stricter \textit{localization}, since the objects must be detected and delineated at a finer level using non-overlapping pixel-wise masks.

\paragraph{Semantic segmentation / Part segmentation} This task consists of assigning a semantic class label to every pixel in an image. Semantic segmentation is agnostic to enumerating objects and can be seen as classification at the pixel level (\Cref{ch1:fig:task-def-semseg,ch1:fig:task-def-partseg}). Alternatively, this task is also denoted as \textit{pixel-level semantic labeling} or \textit{scene parsing}. A specialized type of semantic segmentation is aiming at only segmenting parts of objects in the scene. This tasks operates at the part-level abstraction, while semantic segmentation concerns the scene-level abstraction. Pixels that do not belong to parts are ignored from the evaluation. 

\paragraph{Panoptic segmentation} The aim of panoptic segmentation is to find the most informative abstract representation of an image, compared to the other tasks, by investigating the first three aspects of a scene, as mentioned earlier, at their finest level. More specifically, it aims for pixel-level, non-overlapping masks of countable (cars, persons, traffic signs) or uncountable (sky, road, vegetation) semantic elements of an image. Compared to instance segmentation, it requires to assign semantic class labels also to non-object pixels in the image. Compared to semantic segmentation, countable objects (things) need also to be enumerated.

~

The aforementioned image scene-understanding tasks have different output formats, and span a variety of combinations from the list of aspects. For example, semantic segmentation requires predictions with very detailed \textit{localization},~\ie per-pixel granularity, but does not need instance \textit{enumeration}. Instance segmentation requires per-pixel segmentation and \textit{semantics} discovery, but covers only the countable objects of a scene (objects). As a consequence, within the fully-supervised setting explored in this thesis, training of networks require different types and granularity of supervision. This work has two tasks as starting points, which were selected to achieve a good trade-off between available supervision datasets, output granularity, contemporary research, and usability in the autonomous driving platform of our lab.

\Cref{ch:3-iv2018,ch:4-iv2019,ch:5-itsc2019-wacv2019,ch:6-journal} have the task of semantic segmentation as starting point. Semantic segmentation provides a concise and consistent description of scene semantics and has pixel-level granularity, making network predictions functional for a variety of systems. Moreover, the advantages of the task structure involve the uniform processing of images, easiness of handling of the input / output / training procedure, and fast processing, compared to other scene understanding tasks. Finally, the output predictions can be used by subsequent systems as is, or with simple post-processing or clustering.

\Cref{ch:7-panoptic} has the task of panoptic segmentation as starting point. Panoptic segmentation is the most general task of segmentation, because it combines the first three aspects of scene understanding at the most fine level of detail (pixel-level). Moreover, it can handle countable and uncountable semantic elements in a generalized way. Panoptic segmentation is adopted as a starting point because it can address the research issues of the chapters in the thesis. For example, to achieve accurate segmentation, pixel-wise predictions are preferred over coarse bounding-box localization, because they provide dense (pixel-wise) localization, semantics, and an unambiguous association between the image pixels of the depicted object and the actual real-life object. This accurate definition also supports a better \textit{localization} and \textit{enumeration} of same-class overlapping objects. These aspects will be addressed in the upcoming chapters of this thesis.

\section{Problem statement and research questions}
\label{ch1:sec:challenges-rq}

This section defines the main research themes of this thesis and the underlying research questions. Prior to each research theme, a short introduction is given on the essential aspects and their motivation, which then gradually leads to defining the theme and involved questions. The central research topic of this thesis is defined as follows.

\begin{quote}
	The objective of this thesis is to develop techniques for leveraging heterogeneous datasets to improve training and inference for semantic segmentation and panoptic segmentation, and extend these segmentation tasks towards holistic scene understanding with multiple abstractions in the perception of scenes or their enclosed elements.
\end{quote}

\begin{figure}
	\centering
	\begin{subfigure}{0.45\linewidth}
		\centering
		\includegraphics[width=\linewidth]{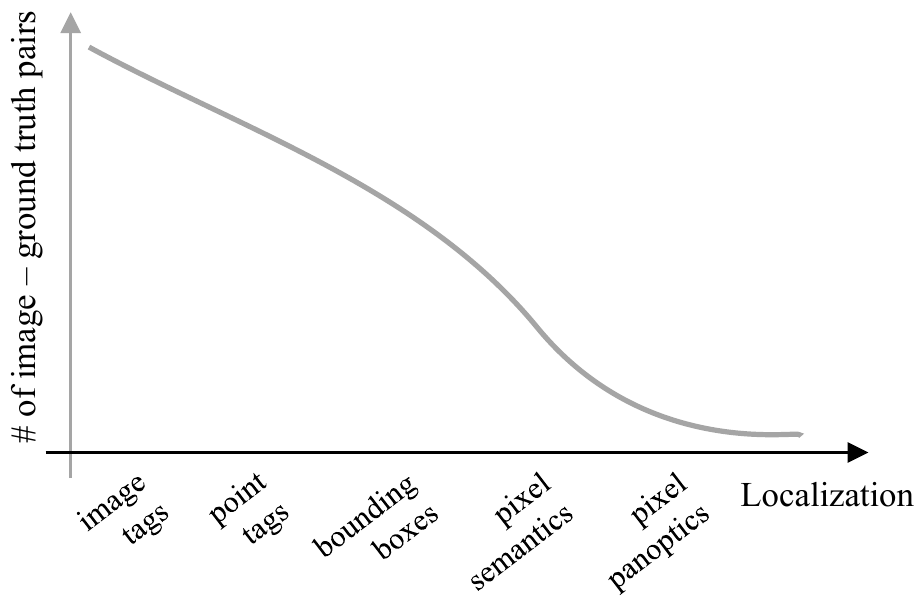} 
		\caption{Empirical trend of the number of labeled images with increasing annotation detail.}
		\label{ch1:fig:motivation1}
	\end{subfigure}
\qquad
	\begin{subfigure}{0.45\linewidth}
		\centering
		\includegraphics[width=\linewidth]{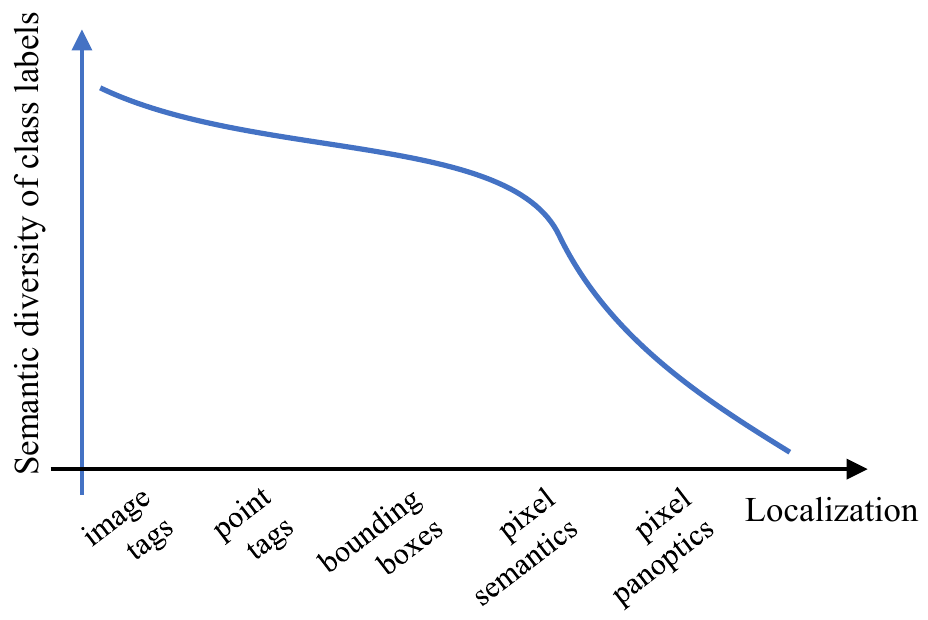} 
		\caption{Empirical trend of the number of semantic classes with increasing annotation detail.}
		\label{ch1:fig:motivation2}
	\end{subfigure}
	\caption{As the localization of manual annotations becomes more detailed and with increased coverage and counting,~\ie from coarse image-level tags (classification) to fine pixel-level labeling (panoptic segmentation), the variety of semantic class labels and the number of annotated images drop significantly.}
	\label{ch1:fig:motivation}
\end{figure}

Leveraging maximum results of the employed datasets is an important matter to consider when training convolutional networks for high accuracy. In scene understanding, supervised learning is the established paradigm of training, since it employs all the available annotation information from the datasets. Its success is based on image datasets that are collected and annotated by humans in order to conform to the task format at hand (refer to Section~\ref{ch1:sec:scene-und}). However, the strength of supervised learning,~\ie \mbox{(semi-)manual} labor, becomes at the same time its pitfall, because it involves costly procedures and manual effort.

Semantic segmentation is a fundamental part of scene understanding. It provides a comprehensive representation of images and their semantic content by encoding information with the finest localization,~\ie at the pixel level. It is the first stage of complex AI systems and an indispensable tool for reasoning and planning algorithms for tasks in~\eg autonomous driving. Semantic segmentation is driven by fully convolutional networks, trained on pixel-labeled datasets within the supervised learning paradigm. This trend imposes a weight shift of human endeavor from designing new networks and training procedures to collecting, handling, and meticulously annotating large-scale datasets. Overall, the landscape of available datasets for scene understanding tasks is growing rapidly, but their heterogeneity imposes constraints for combining them to improve specific processing and follow-up tasks. Moreover, as various applications have different semantic extents and requirements for the four aspects listed in Section~\ref{ch1:sec:scene-und}, the generated datasets are annotated over disjoint or conflicting label spaces, making combined training on these datasets not a trivial task if not complicated. These challenges motivate the following research themes.

\paragraph{Research theme 1}
Leverage heterogeneous datasets from a variety of scene-understanding tasks in order to improve semantic segmentation.

\begin{quote}
	RQ1a: Which challenges arise from using heterogeneous datasets to train convolutional networks for semantic segmentation?
\end{quote}

\begin{quote}
	RQ1b: How can we combine existing datasets for semantic segmentation that are annotated on disjoint label spaces?
\end{quote}

\begin{quote}
	RQ1c: Is it possible to use datasets with weak supervision to train convolutional networks for semantic segmentation?
\end{quote}

With the growing amount of datasets and their varying nature and quality, the training of multiple datasets for a joint purpose is becoming complicated. The training scenarios involving multiple datasets raise efficiency challenges and increases the needs for computational resources. As a consequence, the following research questions can be posed.

\paragraph{Research theme 2}
Efficient training and inference for semantic segmentation with a growing amount of datasets and increasing number of semantic classes.

\begin{quote}
	RQ2a: Does training for semantic segmentation with multiple datasets scale well with an increasing number of datasets? How is inference influenced when the output label space is large?
\end{quote}

\begin{quote}
	RQ2b: How can dataset imbalances be mitigated in a multi-dataset training scenario?
\end{quote}

\begin{quote}
	RQ2c: Given restricted memory resources, how can we maximize the number of employed datasets and reduce the throughput time for training?
\end{quote}

Datasets generated for semantic segmentation are biased towards providing very granular localization of labels, but they lack other aspects necessary for comprehensive scene understanding. Specifically, they exchange richness of semantics and the number of image-label pairs, with very granular annotation localization, to conform with the per-pixel nature of the semantic segmentation task. Moreover, semantic segmentation does not provide any information on instance counting and there is no separation of scene elements within the same semantic class. Consequently, networks for this task lack in scene understanding on many abstractions and include only a few labeled semantic concepts. The third research theme regards the scene understanding at multiple levels of abstraction.

\paragraph{Research theme 3}
Extend the scene-level panoptic segmentation task with part-level semantics towards holistic scene understanding.

\begin{quote}
	RQ3a: How can panoptic segmentation be combined with the concept of part segmentation to enrich the former with semantics of the latter? Can this be incorporated in an unambiguous and consistent manner?
\end{quote}

\begin{quote}
	RQ3b: Is it feasible to train a single network for scene-level and part-level semantics, together with instance-level separation?
\end{quote}

\begin{quote}
    RQ3c: Are the existing scene-understanding datasets adequate for training and evaluating systems for part-aware panoptic segmentation?
\end{quote} 


\section{Contributions}
\label{ch1:sec:contributions}

\subsection{Semantic segmentation}
A semantic segmentation system is characterized by its segmentation accuracy and its output \textit{semantics}, which is an important aspect of scene understanding (Section~\ref{ch1:sec:scene-und}). This work contributes to image-based semantic segmentation by increasing the recognized \textit{semantics}, and hence the predicted (output) \textit{coverage}, of a convolutional network, and its segmentation accuracy. This is achieved by exploiting multiple sources of information (datasets) simultaneously for training convolutional networks, as compared to single-dataset training. The proposed methodology delivers a data-oriented solution, provides training robustness, and focuses on street scenes and general outdoor scenes. The proposed methodology yields an Heterogeneous Training for Semantic Segmentation (HToSS) framework as the result of the last chapter on semantic segmentation and components of this framework are gradually elaborated in the first two chapters.

The contributions span three important directions in semantic segmentation. First, it is demonstrated that HToSS training improves performance (mean Intersection over Union) on the employed training (seen) datasets, compared to the accuracy of single-dataset conventional semantic segmentation. Second, it is shown that HToSS is advantageous for generalization,~\ie performance is increased also on unseen datasets that the networks have not been trained on. Third, semantic knowledgeability,~\ie the number of recognized semantic concepts, of single-network systems is also enhanced. 

Additionally, in a second stage, the HToSS framework is optimized to be efficient in memory during training and inference, enabling a larger number of datasets to be used under the same memory constraints. Moreover, special attention is paid to computational efficiency during training, in order to minimize costly floating-point operations, thereby minimizing power consumption.


\subsection{Deep learning with mixed supervision}
From the perspective of deep learning disciplines, this thesis contributes to fully-supervised and weakly-supervised learning. Specifically, the research explores and solves challenges arising from multiple-dataset training for semantic segmentation when, possibly conflicting, strong supervision (fully-supervised) or weak supervision (weakly supervised) data are available in separate or joint forms. This is achieved by solving semantic conflicts and relaxing \textit{localization} requirements for consistent training of semantic segmentation networks.

To this end, a Heterogeneous Training framework for Semantic Segmentation (HToSS) is developed, which enables training of fully convolutional networks with an arbitrary number of various scene-understanding datasets. The employed datasets comprise scenes of the same visual domain (\eg street scenes) or from a mixture of domains, demonstrating the effectiveness of HToSS. The framework aims to combine semantics from all employed datasets, while leveraging many types of supervision, either strong or weak. Networks trained with the proposed framework produce pixel-wise predictions, compatible with the semantic segmentation formulation and having a rich output semantic space. Moreover, the HToSS framework eradicates the need for any additional manual annotation effort for weak supervision, while it imposes minimal constraints to the label spaces of the employed datasets.

\subsection{Towards holistic scene understanding}
Scene understanding encapsulates a variety of tasks (Section~\ref{ch1:sec:scene-und}) focusing on different aspects of scene elements and having specifically defined goals, depending on the required analysis depth of the application at hand. This thesis explores three scene understanding tasks, which are traditionally solved separately,~\ie semantic segmentation, instance segmentation, and part segmentation. Our contribution involves the consistent unification of these tasks by formulating the novel task of part-aware panoptic segmentation, which paves the way towards holistic scene understanding. Addressing this task in an integral manner favors the minimization of resource utilization and is advantageous due to the interconnection between constituent tasks. The proposed baselines for part-aware panoptic segmentation provide the first single-network, all-encompassing solution that combines scene-level and part-level semantics together with object-instance enumeration.

In the context of this work and in order to support our hypotheses, we have created and made publicly available four datasets that aim at holistic scene understanding and increasing the \textit{semantics}, the \textit{localization}, and the \textit{coverage} of existing influential datasets. The first two, namely Cityscapes Traffic Signs and OpenScapes, aim at increasing the \textit{semantics} and \textit{coverage} of street scenes and are created by automated image selection and semi-manual annotation. The other two, namely Cityscapes Panoptic Parts and PASCAL Panoptic Parts, aim at the new part-aware panoptic segmentation task. These datasets increase the \textit{localization} and \textit{semantics} of the highly employed Cityscapes and PASCAL-VOC datasets. Finally, the fact that the extensions and new abstraction layers for all four datasets are provided on the same set of images, instead of providing them on a new set, facilitate exploitation of all datasets for multiple scene-understanding tasks and the related multiple levels of scene-understanding abstractions.

\section{Dissertation outline and scientific background}
\label{ch1:sec:outline}
\begin{figure}
	\centering
	\includegraphics[width=1.0\linewidth]{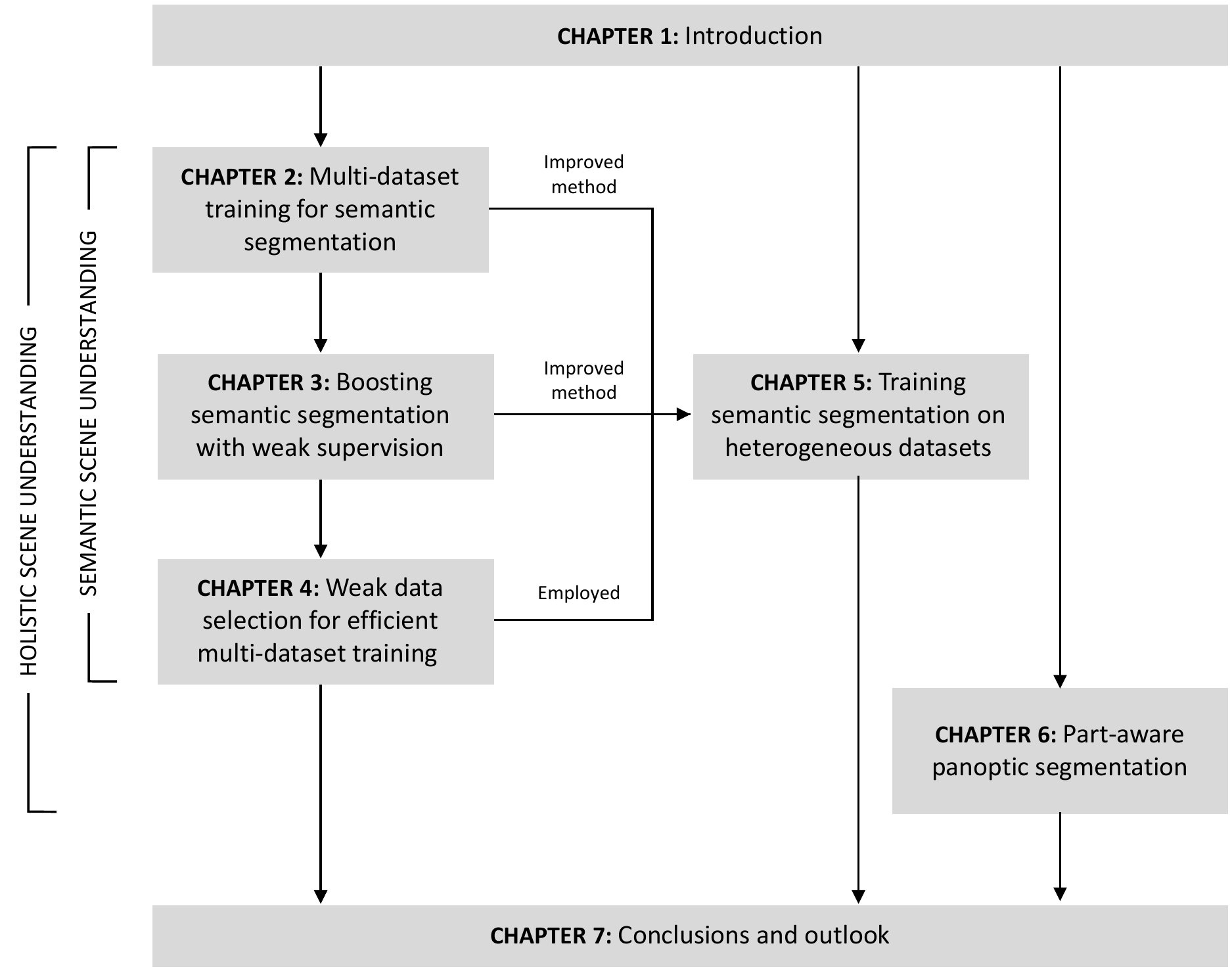}
	\caption{Organization of chapters and connections between them in this thesis.}
	\label{fig:ch1-outline}
\end{figure}

Figure~\ref{fig:ch1-outline} provides the layout of the thesis and the connection between chapters.

\paragraph{Chapter 2} This chapter focuses on semantic segmentation and identifies the limitations that current CNN training algorithms have, from the perspective of employed datasets. A preliminary formulation of the problem of training on multiple heterogeneous datasets is given, which partitions it into three challenges. Then, the solution to each of them is delineated, which extend the reach of CNNs to an extensive amount of unexploited information. The first challenge,~\ie the label granularity between datasets, is investigated and a solution is proposed that combines disjoint label sets into a single semantic tree that can effectively incorporate an arbitrary number of datasets. The contributions of this chapter were presented in the Proceedings of Int. Conf. IEEE IV 2018 and have been submitted (under review) in the IEEE Transactions on Neural Networks and Intelligent Systems in 2021. An extension of the developed system competed as part of the CVPR 2018 Robust Vision Challenge and obtained top-3 performance.

\paragraph{Chapter 3} This chapter investigates the benefits of training with multiple types of weak supervision, together with strong supervision for semantic segmentation, which is the second challenge in the formulation of the previous chapter. To this end, a methodology is developed for handling a variety of weak supervision types to seamlessly incorporate them into the established semantic segmentation FCN framework. From the semantic segmentation perspective, datasets with weak supervision are less costly to obtain are widely available. Thus the proposed method opens a new window to information that could not be exploited by previous networks. The contributions of this chapter were presented in the Proceedings of Int. Conf. IEEE IV 2019 and the paper was selected for an oral presentation.

\paragraph{Chapter 4}
Two critical complications that arise when training and testing CNN on multiple strongly and weakly-labeled datasets are examined in this chapter. The first one, namely dataset imbalances, influences the training procedure on the information assimilation capability of the networks derived from all datasets. The proposed solution amends the shortcoming on dataset imbalance by finding the most informative examples from datasets to enable balanced training. A second complication is related to the information richness of the weakly-labeled datasets and the poor localization of the annotations, which affects the training and discriminability of the networks. The proposed solution provides balance between strongly-localized and weakly-localized annotations in the batch content and reduces repeatability of examples with similar information richness. The contributions of this chapter were presented in the Proceedings of Int. Conf. IEEE ITSC 2019 as a joint work with Rob Romijnders and the paper was selected for an oral presentation.

\paragraph{Chapter 5}
This chapter continues and completes the research on heterogeneous training and proposes an integral framework (HToSS) for training CNNs on semantic segmentation using multiple datasets with strong and weak supervision. Ideas from the previous chapters are extended and integrated into a concise problem formulation for heterogeneous training. The proposed framework aims at enhancing CNNs for semantic segmentation in three directions: i) segmentation performance, yielding increased segmentation metrics on seen datasets, ii) generalization, giving improved segmentation metrics on unseen datasets, and iii) knowledgeability, providing an increased number of recognizable semantic concepts. The contributions of this chapter have been submitted (under review) as a journal paper at the IEEE Transactions on Neural Networks and Learning Systems in 2021.

\paragraph{Chapter 6} Whereas the previous chapters focus on exposing semantic segmentation to new sources of information, in this chapter the available information used for training networks is broadened at multiple levels of abstraction. Specifically, the novel task of part-aware panoptic segmentation is formulated. Apart from the existing panoptic segmentation components, this task requires segmenting objects into their constituent parts. A baseline solution is proposed, which integrates state-of-the-art panoptic segmentation with an extended network mechanism for exploiting part-level annotations. The contributions of this chapter were integrated in a joint paper with three other researchers, Daan de Geus, Chenyang Lu, and Xiaoxiao Wen, for publication in the Proceedings of Int. Conf. CVPR 2021.

\paragraph{Chapter 7}
In this chapter, conclusions are provided at per-chapter and thesis level, followed by a discussion on the research questions and the related contributions. Moreover, a general discussion and outlook to future work is given.

\begin{savequote}[8cm]
	
\end{savequote}

\chapter{Multi-dataset training for semantic segmentation}
\label{ch:3-iv2018}

\section{Introduction}
\label{sec:intro}
\freefootnote{\hspace*{-15pt}The contributions of this chapter were presented in the Proceedings of Int. Conf. IEEE IV 2018 and have been submitted (under review) in the IEEE Transactions on Neural Networks and Intelligent Systems in 2021.}
Visual semantic segmentation is a fundamental task in the perception sub-system of any highly automated vehicle or robot~\cite{janai17computer}. It provides awareness of the semantics of the environment in which the vehicle (agent) will move and perform actions. The understanding of semantics and spatial relationships between the scene components play an important role for higher-level reasoning and planning. Convolutional Neural Networks (CNNs) under a supervised learning setting have prevailed in semantic segmentation at the expense of manual human annotation effort~\cite{zhao2017survey}.

Semantic segmentation is part of a larger family of visual image understanding tasks, which include image classification and object detection. Compared to these tasks, semantic segmentation requires predictions at a highly detailed level,~\ie, pixel level, and thus requires even larger, per-pixel annotated datasets. This chapter focuses on the two following fundamental challenges that CNNs face in the context of semantic segmentation:
\begin{enumerate}
	\item \textit{Limited size of existing datasets}. Existing datasets~\cite{Cordts2016Cityscapes,neuhold2017mapillary} with per-pixel annotations contain typically in the order of 1k - 10k
	images, which is two to three orders of magnitude less than image classification datasets. This leads to poor generalization capabilities of CNNs trained on such datasets and consequently a lower performance in real-life settings.
	\item \textit{Low diversity of represented semantic concepts}. The complexity of manual, per-pixel labeling constrains the number of represented semantic classes into a few dozens, while other datasets for less-detailed tasks can reach up to thousands of semantic classes. As a result, CNNs trained on semantic segmentation datasets cannot recognize fine-grained semantic concepts and lose elements of the scene.
\end{enumerate}
Apart from these challenges, an important factor for robust training is the errors in labels induced my human annotators. In the context of this thesis, we hypothesize that the employed datasets have minimal annotation errors. The correctness hypothesis is also important for having confidence in comparisons between evaluation metrics on the validation/test splits of these datasets.

The natural way to address the first challenge, is to annotate more images from a single dataset with manual or semi-automatic means. Although this is a straightforward approach, manual labeling is costly and semi-automated procedures result in insufficient quality of annotations. These approaches emphasize the weight on costly manual labor, whereas the rich landscape of available semantic segmentation datasets allows to exploit semi-automated means for semantic segmentation. This is one of the keys to the approaches followed in this chapter.

The second challenge on increasing the number of recognizable semantic classes can be accomplished in two ways: (1) refine annotations of an existing dataset with extra (sub)classes, e.g.~\cite{petrovai2017semi}, or (2) use existing auxiliary datasets only for the new (sub)classes~\cite{zhou2016hierarchical,mao2016hierarchical}. The first approach can become very costly and laborious for big datasets and is actually unnecessary, as a plethora of datasets with fine-grained (sub)classes already exist for traffic scenes (\eg traffic sign types, vehicle types, pedestrians). We have adopted the second approach for working out this chapter and the aforementioned challenges.

The objective of this chapter is to exploit richer and larger datasets with the purpose to expand the number of classes and enhance the performance of automated classification. This implies we have to design a method for combining the label spaces of multiple datasets to result in a hierarchy of classes. This hierarchy should resolve any semantic conflicts or overlaps between dataset labels and should facilitate a successful automated classification across a multi-level hierarchy.

To achieve this objective and address the challenges, we propose to combine existing datasets in order to create a large, diverse training set of images and labels and use that for supervised learning of a CNN. The key challenge in this approach is the handling of semantic differences that exist between the datasets. To this end, we present a method that leverages multiple \textit{heterogeneous} datasets, to train a fully convolutional network for per-pixel semantic segmentation. This approach better exploits available datasets, thereby reducing annotation effort and increasing the number of classes that can be recognized.

This chapter is organized as follows. Section~\ref{ch3:sec:related-work} discusses the related work on multi-dataset training and Section~\ref{ch3:sec:problem-def} formulates the problem more formally. The fundamentals of our hierarchical approach are provided in Section~\ref{ch3:sec:proposed-method}. Section~\ref{ch3:ssec:three-level-hierarchy} presents the specifics of the chosen implementation. Section~\ref{ch3:sec:evaluation} demonstrates the performance gain of hierarchical classifiers with three \textit{heterogeneous} datasets, over flat, non-hierarchical classifiers. Furthermore, it is shown that multi-dataset training of a common feature representation using the proposed method, can improve performance across all datasets regardless of their structural differences.

\section{Related work}
\label{ch3:sec:related-work}
The majority of previous works focus on using multiple datasets with different label spaces but a single type of annotations, i.e., pixel-level labels. The authors of related work solve the challenges that arise from conflicts in semantics from multiple overlapping labels by either a dataset-based solutions~\cite{ros2016training,lambert2020mseg}, network architecture-based solutions~\cite{liang2018dynamic,kalluri2019universal,leonardi2019training,fang2020multi,sun2020real}, or loss-based solutions~\cite{kong2019training,fang2020multi}. Early works extend the conventional Fully Convolutional Network (FCN) architecture with multiple heads/decoders where one is used for each dataset~\cite{leonardi2019training} 
effectively approaching the problem from the multi-task learning perspective. Other works solve conflicts between datasets by merging all label spaces to a common space by merging or splitting classes and relabeling them. The authors of~\cite{ros2016training} unify 6 semantic segmentation datasets from multiple domains using manual relabeling or by ignoring classes. The authors of~\cite{lambert2020mseg} mix 13 datasets to create a large-scale training and testing platform. They solve semantic conflicts between label spaces by a handcrafted algorithm, which uses heuristics or requires manual relabeling. This work~\cite{lambert2020mseg} was published after performing and reporting the research of this chapter, hence it is referenced here only for completeness.

Contrary to existing work, we follow a novel approach addressing three aspects. First, we maintain the canonical FCN backbone architecture, irrespective of the number of employed datasets. Second, we keep the training datasets intact, without requiring any data relabeling. Third, we only modify the final classification layer, replacing the single classifier by multiple hierarchical classifiers and reformulating the cross-entropy loss accordingly.

\section{Problem formulation}
\label{ch3:sec:problem-def}
This section describes the challenges for simultaneous training of CNNs on multiple datasets aiming at semantic segmentation.

\subsubsection{A. Single-dataset training}
The task of Semantic Segmentation involves the per-pixel classification of images into a predetermined set of mutually exclusive semantic classes. In the conventional supervised learning setting, it requires a dataset $\mathcal{D} =  \{\left(\mathbf{x}, \mathbf{y}\right)_i,~i = 1,\dots,|\mathcal{D}|\}$ with images $\mathbf{x}$ and class-labeled image $\mathbf{y}$. The images are pixel-wise annotated over a predefined label space $\mathcal{L} = \{l_j,~j = 1,\dots,L\}$ of semantic classes (labels\footnote{We use the terms ``semantic labels'' and ``semantic classes'' interchangeably. However, the former appears in the context of a dataset description and the latter in the context of training networks.}). Each label image $\mathbf{y} \in \mathcal{L}^{H \times W}$ is a 2-D matrix with spatial size of $H \times W$ pixels and every position corresponds to an individual pixel (element) in the image $\mathbf{x}$.

Each label $l$ corresponds with a semantic concept or high-level abstraction that we recognize in a real-world scene,~\eg, car, tree, building, or person. In the context of a single dataset, it is important that the semantic classes $l_j$ have unambiguous and mutually exclusive semantic definitions. If this is not true, the features extracted by the CNN for the corresponding classes ``confuse'' the classifier during training and its performance drops proportionally to the extent of mislabeled pixels. In the following, we assume that datasets have annotations with negligible ambiguity.

The established Fully Convolutional Network (FCN) framework~\cite{long2015fully} for semantic segmentation consists of a CNN backbone and a softmax classifier, which outputs per-pixel probabilities over a predetermined set of mutually exclusive classes. In the single-dataset training scenario, these classes are distinct and their definitions do not overlap. We call this a ``flat'' classification over the set of classes $\mathcal{L}$, as opposed to our hierarchical approach. 

\subsubsection{B. Multi-dataset training}

\noindent In the multi-datasets training setting, the objective is to combine information from many sources in order to increase prediction accuracy of CNN outputs and the number of recognizable semantic concepts distinguished by the CNN. We assume that a collection of $D$ pixel-wise annotated datasets are available with their associated label spaces $\{\left(\mathcal{D}, \mathcal{L}\right)^i,~i = 1, \dots, D\}$. Each label space $\mathcal{L}^i = \{l_j^i,~j=1, \dots, L^i\}$ contains, as in the single-dataset case above, a predefined set of semantic labels $l_j^i$ that have unambiguous and concise definitions in the context of each dataset. Although this ensures that intra-dataset conflicts are absent, there is no limitation for label spaces \emph{across} datasets. 

Since there is no limitation across datasets, this usually leads to inter-dataset label space conflicts because they contain semantic concepts with different granularity (level of detail). The most typical case of conflicts arise when two labels from different datasets are combined. For example, the two datasets describe semantic concepts at different levels of detail,~\eg, \textit{rider} vs \textit{bicyclist}/\textit{motorcyclist}. However, another class may partially contain common concepts,~\eg, \textit{road}, then this class may contain \textit{lane markings} in one of the two datasets, which indicates the semantic concept of \textit{traffic sign} inside the \textit{road} class. It is readily clear that this generates confusion in combining of and training with multiple datasets.

In the multi-dataset training scenario, a naive stacking or merging of label spaces from all datasets, in order to train over the union of classes, cannot be directly performed due to the aforementioned conflicts. This chapter proposes a simple yet effective methodology for combining different datasets without the need of manual relabeling of existing datasets.

\begin{figure}
	\centering
	\includegraphics[height=0.5\textheight,angle=270]{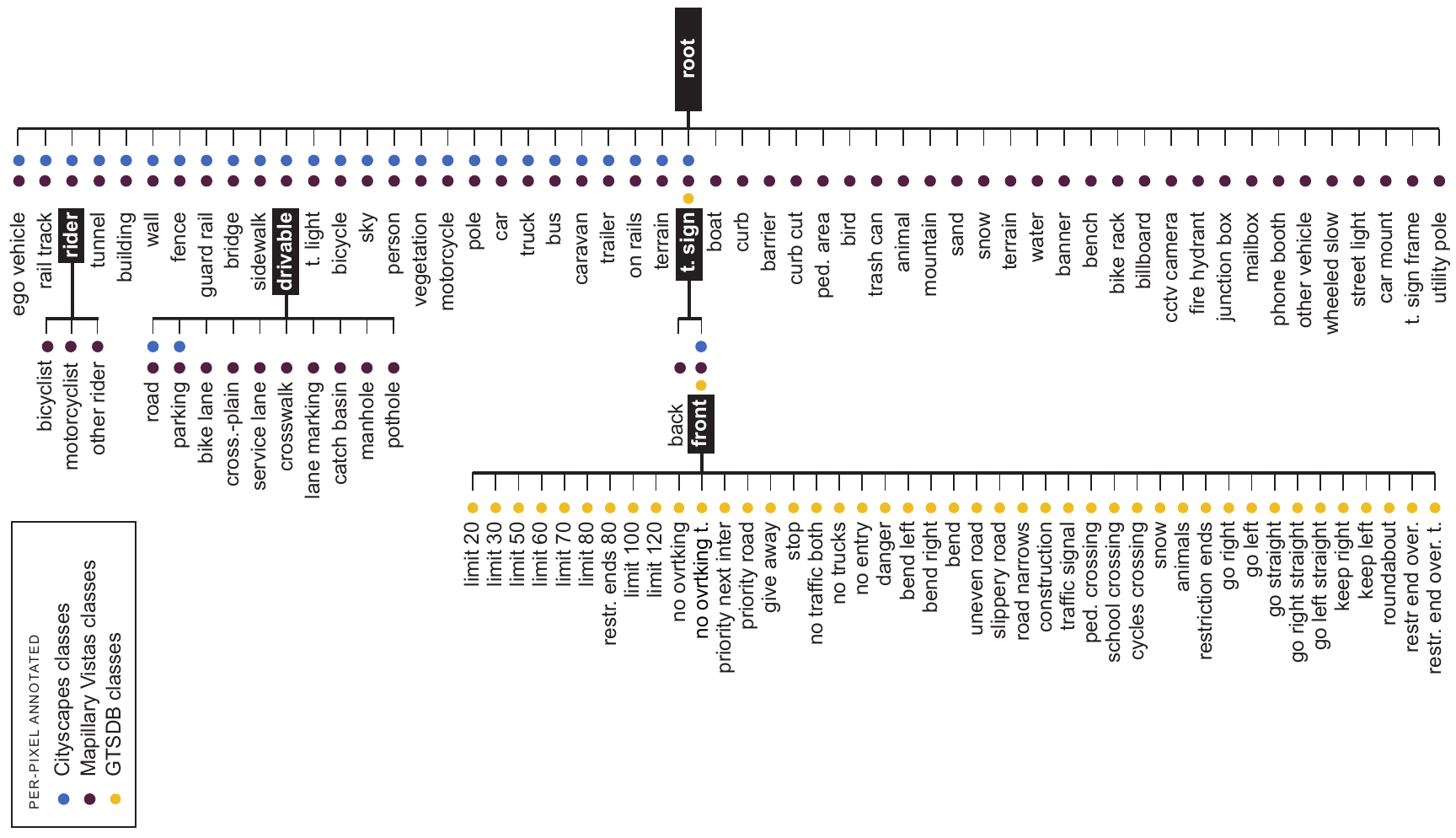}
	\caption{Three-level semantic label hierarchy combining 108 classes from Cityscapes, Mapillary Vistas and GTSDB dataset. Classes marked in black correspond to the L1, L2, and L3 classifiers of Fig.~\ref{ch3:fig:algorithm-overview}.}
	\label{fig:label-tree}
\end{figure}

\section{Method: CNN with hierarchical classifiers}
\label{ch3:sec:proposed-method}
This section describes the methodology for simultaneous training of CNNs with heterogeneous datasets for semantic segmentation. In the general case, heterogeneous datasets have conflicting label spaces, which makes stacking them impossible within a standard FCN~\cite{long2015fully} framework. Section~\ref{ch3:ssec:create-hierarchy} presents a solution to deal with conflicts by introducing a hierarchy of label spaces. Section~\ref{ch3:ssec:conv-netw-arch} analyzes the hierarchical classification components (\eg classifier, loss function, etc.) that are added to the FCN framework. Section~\ref{ch3:ssec:train-infer} discusses the training loss and the inference procedure. These components solve the challenges of Section~\ref{ch3:sec:problem-def} by introducing minimal assumptions on the datasets. Section~\ref{ch3:sec:evaluation} details experiments, which are based on an implementation with a triple-level hierarchy using three datasets.  The specifics of this implementation are provided in Section~\ref{ch3:ssec:three-level-hierarchy}.

\subsection{Hierarchy of label spaces}
\label{ch3:ssec:create-hierarchy}
The first step in the proposed approach is the construction of a semantic taxonomy, merging all label spaces from the available datasets. An example for three datasets is shown in Figure~\ref{fig:label-tree}. High-level classes from all datasets are placed at the top level (under the root) of the hierarchical tree. High-level classes are deemed the classes from all datasets that have common semantic definitions,~\ie, $\bigcap_j \text{def}(l_j)$ with $l_j \in \bigcup_i \mathcal{L}^i$. In case conflicting definitions exist,~\eg, \textit{rider} vs \textit{bicyclist} or \textit{traffic sign} vs \textit{stop sign}, then the more general class is selected. Below each high-level class, the remaining classes from all datasets are positioned in a hierarchical manner. Whenever two or more classes describe conflicting concepts, an intermediate tree node is introduced for removing ambiguity, and the classes are re-placed as its children. In Section~\ref{subsec:semantic-hierarchy-of-label-spaces} the process is described in details for our selection of 3 datasets. The only assumption during the procedure of generating the hierarchy of label spaces is that the selected datasets must contain either high-level classes or fine-grained sub-classes of existing high-level classes. For example, a traffic sign dataset can be incorporated if there exists a high-level \textit{traffic sign} class in the other datasets. Moreover, it should be noted that it is assumed that labels themselves do not have discrepancies,~\ie they contain negligible human miss-classification errors.

\subsection{CNN architecture}
\label{ch3:ssec:conv-netw-arch}
\begin{figure}
\centering
\includegraphics[width=0.9\linewidth]{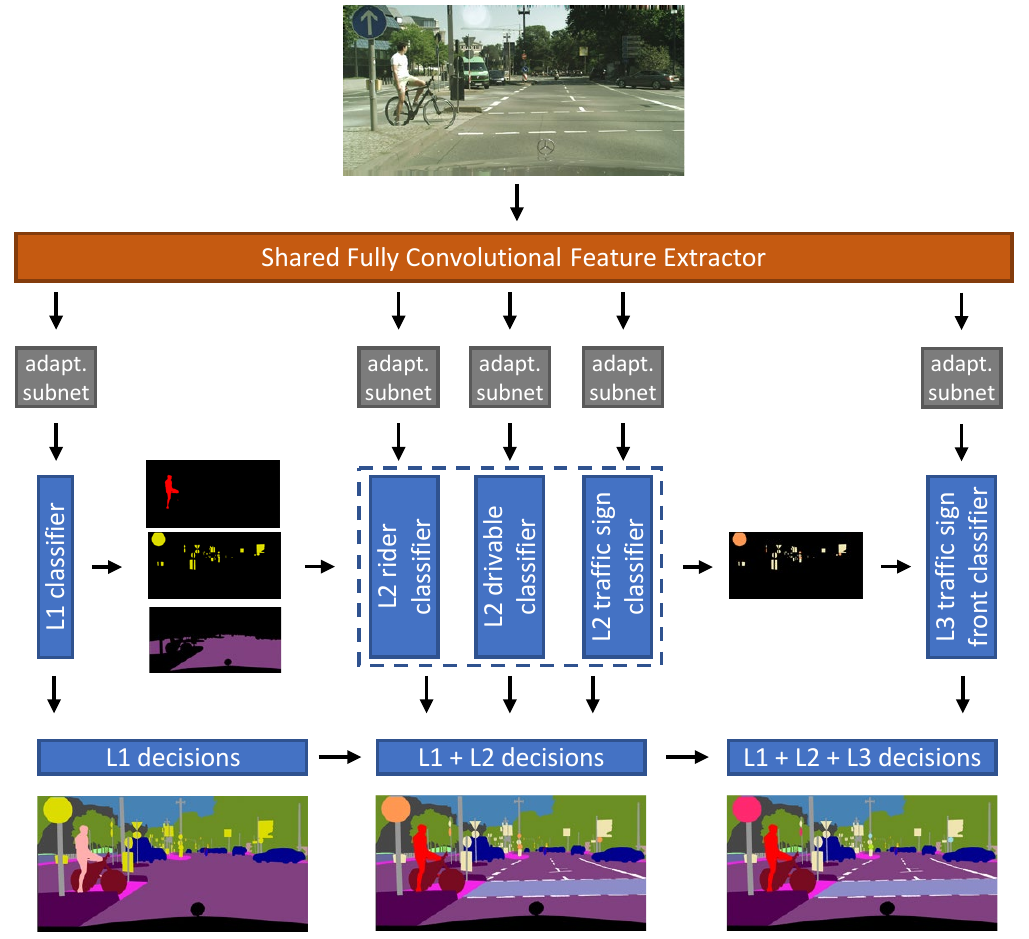}
\caption{Hierarchical classification network during inference. The input image is transformed by the backbone to a shared feature representation, which is passed to a hierarchy of classifiers through adaptation sub-networks. The Level-1 classifier outputs predictions for every pixel of the image, while each subsequent classifier infers only about its own set of classes. At the bottom-middle block, the L1 decisions are combined with L2 decisions. At the bottom-right, the output of all levels is combined to form the final fine-grained per-pixel segmentation.}
\label{ch3:fig:algorithm-overview}
\end{figure}

The proposed network architecture consists of a fully convolutional feature extractor for computing a dense, shared representation, and a set of classifiers, each corresponding to a node of the semantic hierarchy (see Fig.~\ref{ch3:fig:algorithm-overview}). Every classifier can be connected with classifiers one level down in the hierarchy, in order to pass its predictions for training and inference, as described in Section~\ref{ch3:ssec:train-infer}. Each classifier may be preceded by a shallow \textit{adaptation network}, which adapts the common representation, its depth, and receptive field to the needs of the classifier. This gives the network designer the opportunity to select different feature dimensions and receptive fields for each of the classifiers. For example, discriminating between,~\eg, \textit{traffic signs} is easier~\cite{ciregan2012multi}, as less features are needed, compared to high-level discrimination, like \textit{road} vs. \textit{sidewalk} and \textit{bushes} vs. \textit{trees}~\cite{c4}. The flexibility of applying different field-of-views to different classifiers, enables variable context aggregation, depending on the average object size of the classifier: \eg traffic signs appear generally in smaller scales than buildings or cars.

\subsection{Training and inference on hierarchical classifiers}
\label{ch3:ssec:train-infer}
This section describes the inference procedure and the training losses formulation per classifier during training of the CNN.

\subsubsection{A. Inference: hierarchical decision rule}
Inference is carried out per-pixel, in a hierarchical manner across the tree of softmax classifiers. Each classifier $j$ computes a per-pixel normalized vector $\bsigma^{j,p}$ of class probabilities for its own set of pixels $p \in \mathcal{P}^j$ and set of classes $\mathcal{C}^j$, and outputs per-pixel decisions $\hat{y}^{j,p} = \argmax_i \sigma_i^{j,p}$, where $\hat{y}^{j,p} \in \mathcal{C}^j$.\footnote{Notation: The symbol $\bsigma$ is a vector of probabilities (bold-face), which is the output of the softmax classifier (hence the choice of the letter, as softmax is the generalization of the sigmoid function). The symbol $\hat{y}$ is a scalar probability produced by the vector to scalar $\argmax$ function. The first superscript denotes the classifier, the second superscript denotes the pixel, and the subscript denotes the element of a vector.} The classifier outcomes are denoted as estimates $\hat{y}$, since later the ground truth $y$ will be described. The set of pixels $\mathcal{P}^j $, for which every classifier should produce decisions, is generated by its parent according to its own decisions. For example, the \textit{rider L2 classifier} has $\mathcal{C}^\text{rider} = \{\textit{bicyclist},~\textit{motorcyclist},~\textit{other rider}\}$ and $\mathcal{P}^\text{rider}$ are the pixels that are classified as \textit{rider} by the root classifier.

The hierarchical decision rule provides the freedom to assign to each pixel a predicted class with the desired level of detail, from the available set of labels $\{\hat{y}^{j,p}\}_{j \in \mathcal{J}}$, where $\mathcal{J}$ denotes the set of classifiers that produce decisions for this specific pixel. In conventional ``flat'' classification, the class granularity of predictions cannot be chosen. Multi-level predictions can be useful for applications that do not require high granularity,~\eg the specific type of a \textit{traffic sign} or the distinction between a \textit{bicyclist} and a \textit{motorcyclist}. Moreover, low-detail predictions have higher accuracy, since the network does not need to make the distinction between detailed semantics.

\subsubsection{B. Training: hierarchical classification loss}
We propose a hierarchical classification loss that consists of independent losses for the constituent classifiers of the hierarchy. The classifier $ j $ is trained on all labeled pixels $\mathcal{P}^j$ corresponding to its respective node. We use the standard cross-entropy loss for each classifier, specified by:
\begin{equation}
\label{eq:ss-losses}
L^j = - \frac{1}{\abs*{P^j}} \sum_{p \in P^j} \log \sigma^{j,p}_{y^{j,p}} ~,
\end{equation}
where $\abs*{\cdot}$ is the cardinality of the pixel set, and $y^{j,p} \in \mathcal{C}^j$ selects the element of $\bsigma$ that corresponds to the ground-truth class of pixel $p$ for classifier $j$. Equation~(\ref{eq:ss-losses}) specifies that for each pixel in the class the logarithm of the probability components from the vector are averaged. Finally, losses from all classifiers are collected and weighted with different hyper-parameters $\lambda^j$ to obtain the total objective loss for minimization, giving:
\begin{equation}
	\label{eq:total-loss}
	L^\text{total} = \sum_j \lambda^j \cdot L^j + \text{l2 regularizer} ~.
\end{equation}
We have chosen to experiment with a single, scalar weight per classifier and not to introduce a complex weight function depending on the classes or other factors, even though some classifiers might work better on some classes than others. This choice largely simplifies the hyper-parameter search.

\section{Three-level label hierarchy with street scenes datasets}
\label{ch3:ssec:three-level-hierarchy}
In this section, implementation details are outlined  to improve the repeatability of our experiments. The specifics of the three-level hierarchy with three datasets and the CNN architecture are described in the context of street scene understanding.

\subsection{Semantic hierarchies of datasets}
\label{subsec:semantic-hierarchy-of-label-spaces}
Figure~\ref{fig:label-tree} depicts the three-level hierarchy using all labels from Cityscapes, Vistas, and GTSDB datasets. The specific challenges that arise from combining their three label spaces are solved as follows. 1) A new high-level drivable class is introduced to solve the road-class semantic conflict when combining Cityscapes and Vistas. 2) A super-class of traffic signs is created and added as an intermediate node for differentiating Vistas \textit{backside} and \textit{frontside} traffic signs. 3) A \textit{rider} super-class is introduced to include the Cityscapes \textit{rider} class and the 3 Vistas \textit{rider} sub-classes. The semantic hierarchy of the labels induces a corresponding hierarchy of five classifiers, which is end-to-end trained, in a fully convolutional manner over a shared feature representation.

\subsection{Three-level CNN classifier architecture}
The network is depicted in Fig.~\ref{ch3:fig:algorithm-overview}. The feature extraction consists of the feature layers of the ResNet-50 architecture~\cite{he2016deep}, followed by a $1 \times 1$ convolutional layer (with ReLU and Batch Normalization), to decrease feature dimensions to 256. The stride on the input is reduced from 32 to 8, using dilated convolutions. The image representation is shared among 5 branches. Each branch has an extra bottleneck module~\cite{he2016deep} and ends at a softmax classifier, which includes a hybrid upsampling module. We choose the feature dimensions and the field-of-view of the per-classifier adaptation sub-networks to be identical for all branches, except for the \textit{L2 traffic sign classifier} where it is 64. The latter number is empirically determined. After experimenting with different upsampling techniques (fractional strided convolution, bilinear, convolutional), we concluded that the best performance and reduction of artifacts, is obtained by hybrid upsampling, which consists of one $2 \times 2$ learnable fractional strided convolutional layer, followed by bilinear upsampling to restore input dimensions.

\emph{Implementation details}:
We have used Tensorflow~\cite{abadi2016tensorflow} framework and a Titan X GPU (Pascal architecture) for training and inference. Due to limited memory, we set the batch size to 4 (Cityscapes:Vistas:GTSDB = 1:2:1) and the training dimensions to $512 \times 706$ (the average of Vistas images scaled to the smaller Cityscapes dimensions). During training, images are downscaled preserving their aspect ratio, and then randomly cropped. The network is trained for 17 Vistas epochs (early stopping) with Stochastic Gradient Descent and momentum of 0.9, L2 weight regularization with decay of 0.00017, initial learning rate of 0.01 that is halved three times, and batch normalization and exponential moving averages decay are both set to 0.9. The hyper-parameters $\lambda^j$ of Eq.~(\ref{eq:total-loss}) are chosen to be 1.0, 0.1 and 0.1 for the three levels of the hierarchy, respectively. For inference, we currently achieve a frame rate of 17~fps, \ie 58~ms per frame.

\section{Evaluation of the method}
\label{ch3:sec:evaluation}
To evaluate the proposed hierarchical classification approach the following experiments are conducted.
\begin{enumerate}[noitemsep,topsep=0pt]
	\item \emph{Baselines for flat classification}: The baseline performances of flat classifiers for single and multiple datasets training are derived.
	\item \emph{Hierarchical classification on three heterogeneous datasets}: The benefits of our complete method are indicated for combined training on three \textit{heterogeneous} datasets (Cityscapes, GTSDB, Vistas) with disjoint label spaces.
	\item \emph{Comparison of hierarchical and flat classification on cross-dataset setting}: This experiment validates the effectiveness of the proposed hierarchical approach against imbalanced classes, by isolating it on the per-pixel annotated Cityscapes-Traffic-Signs dataset with a two-level label space.
\end{enumerate}

\subsection{Datasets}
\label{subsec:datasets-generation}
Summaries of the statistics of the employed datasets that are used are listed in Table~\ref{tab:datasets}. Next, the extra annotations required for the experiments are described.

\begin{table}
\centering
\small
\begin{tabular}{@{}l|c|c|c|c@{}}
	\toprule
	& Cityscapes & Cityscapes T. & Mapillary & GTSDB\\
	& (fine) & Signs (prop.) & Vistas & (prop.)\\
	\midrule
	Resolution & $1024 \times 2048$ & $1024 \times 2048$ & 0.5 - 25 MP & $800 \times 1360$\\
	Images & 2,975 / 500 & 2,975 / 500 & 18,000 / 2,000 & 600 / 300\\
	Annotated pixels & $1.6 \cdot 10^9$ & $1.6 \cdot 10^9$ & {$156.2 \cdot 10^9$} & $0.003 \cdot 10^9$\\
	Traffic sign classes & n/a & 43 (18) & n/a & 43 (28)\\
	Other classes & 34 (27) & 34 (27) & 66 (65) & n/a\\
	Traffic signs & n/a & 3,158 & n/a & 900\\
	\bottomrule
\end{tabular}
\caption{Statistics of the three datasets. Images contain training and validation splits. The number of evaluated classes are shown between parentheses. The difference is caused by omitting void and boundary classes as explained below.}
\label{tab:datasets}
\end{table}

\subsubsection{A. Labeling Cityscapes with traffic sign classes}
We extend the label space of Cityscapes with 43 traffic sign classes corresponding to the GTSDB dataset. It should be considered that these annotations are only used for evaluation purposes and not for training the hierarchical classifiers. Cityscapes provides only per-pixel traffic sign annotations without differentiating between instances. To this end, we design an automated segmentation algorithm based on the 8-neighborhood distance, for separating connected traffic sign instances in the ground-truth traffic sign mask. Moreover, we develop a GUI annotation tool, which proposes image areas for labeling. Original and new annotations are captured under the name Cityscapes-Traffic-Signs. This dataset contains 2,778, and 380 traffic signs in the train and validation subsets, respectively.

\subsubsection{B. Converting GTSDB to per-pixel annotations}
\label{subsec:gtsdb-bboxes-to-perpixel}
The GTSDB bounding-box annotations are automatically converted to per-pixel annotations, using the traffic sign shapes (circle, triangle, hexagon) and inscribing them into their bounding box. This procedure can be problematic with an in-plane rotation of traffic signs, but after dataset inspection, we have observed that only a negligible amount of in-plane rotations occurs in practice.

\subsection{Metrics and evaluation conventions}
\label{subsec:metrics}

\subsubsection{A. Selected metrics}
Semantic segmentation is traditionally evaluated by multi-class mean Pixel Accuracy (mPA) and mean Intersection over Union (mIoU)~\cite{lin2014microsoft,Cordts2016Cityscapes}. mPA is defined as the per class ratio between the number of properly classified pixels and the total number of them, averaged over all classes. In the context of automated driving these metrics are relevant, since they measure geometric properties of objects in the image between predictions and ground truth. Moreover, these metrics follow well-known definitions in the field~\cite{garcia2017review,Cordts2016Cityscapes}.

\subsubsection{B. Evaluation conventions}
For Cityscapes, we report results on 27~classes (19 of the official benchmark + 8~common with Vistas). For the traffic sign classes, we evaluate on a subset of the 43~traffic signs that satisfy one major condition: their size is larger than 10\textsuperscript{3} pixels, for both the train and validation sets (GTSDB train set, GTSDB and Cityscapes-Traffic-Signs validation sets). We have adopted the 10\textsuperscript{3} pixels as a limit, since it is two orders of magnitude smaller than the least represented class in Cityscapes. For Vistas, we report results on the official 65~classes benchmark. Finally, the model performance is evaluated for each epoch to report the average over the last two checkpoints. The results for all experiments refer to the classes of Table~\ref{tab:datasets} unless otherwise stated. Furthermore, same-dataset evaluation means that training and testing are performed on splits of the same dataset, while cross-dataset evaluation is based on two different datasets.

\begin{table}
	\centering
	\footnotesize
	\begin{tabular}{@{}l|c|c|c|cc@{}}
		\toprule
		\multicolumn{1}{c}{} & \multicolumn{3}{c|}{Same dataset} & & Cross-dataset\\
		\midrule
		Tested on & Cityscapes & Vistas & GTSDB & & \makecell{Cityscapes-Traffic-Signs\\(traffic sign classes)}\\
		\midrule
		mPA (\%) & 53.6 & 36.5 & 25.4 & & 19.1\\
		mIoU (\%) & 46.2 & 29.6 & 17.2 & & 3.0\\
		\midrule
		Trained on & Cityscapes & Vistas & GTSDB & & Cityscapes + GTSDB\\
		\bottomrule
	\end{tabular}
	\caption{Flat classification performance baselines on per-pixel annotated datasets.}
	\label{tab:flat-baselines}
\end{table}
\begin{table}
	\centering
	\footnotesize
	\begin{tabular}{@{}l|c|c|c|cc@{}}
		\toprule
		\multicolumn{1}{c}{} & \multicolumn{3}{c|}{Same dataset} & & Cross-dataset\\
		\midrule
		Tested on & Cityscapes & Vistas & GTSDB & & \makecell{Cityscapes-Traffic-Signs\\(traffic sign classes)}\\
		\midrule
		mPA (\%) & 66.6 & 38.9 & 57.7 & & 29.7\\
		mIoU (\%) & 57.3 & 31.9 & 41.5 & & 8.3\\
		\midrule
		Trained on & \multicolumn{5}{c}{Cityscapes + Vistas + GTSDB}\\
		\bottomrule
	\end{tabular}
	\caption{Performance of our complete hierarchical classification approach on 4 datasets.}
	\label{tab:hierarchical}
\end{table}

\subsection{Exp. 1: Baselines for flat classification}
\label{sec:baselines}
In Table~\ref{tab:flat-baselines}, the same-dataset and cross-dataset baselines for the conventional flat classification approach are presented, in order to be able to fairly compare with the hierarchical results of Table~\ref{tab:hierarchical}. The results are based on using the same input dimensions and batch size as described in the implementation details of Section~\ref{ch3:ssec:three-level-hierarchy}. In Columns 1-3, the three models are trained on three datasets independently.

The right column provides cross-dataset results on Cityscapes-Traffic-Signs traffic sign classes for combined training on Cityscapes and GTSDB. It can be observed that simultaneous training on Cityscapes and GTSDB using a strong ``flat'' classification scheme, fails to achieve satisfactory cross-dataset results on the traffic sign classes of Cityscapes-Traffic-Signs. There are two factors responsible for the low performance: i) the training and the testing traffic signs may have different appearance, since they originate from different datasets, and ii) the number of pixels for a traffic sign class are not enough for training.

A new evaluation protocol is introduced to obtain the results of the last column. This protocol assures fairer comparisons for training a flat classifier on two datasets and consists of a strong baseline because it is more flexible towards misclassifications. When two datasets are combined, a semantic conflict appears between the high-level \textit{traffic sign} class of the first dataset and the traffic sign subclasses of the second dataset (Section~\ref{ch3:sec:problem-def}). This protocol resolves that conflict in the following way. The decision for a traffic sign pixel is deemed correct, if: 1) it is correctly labeled with any traffic sign subclass, or 2) it is labeled as a \textit{traffic sign} (which is also correct), and the second most probable choice is the correct traffic sign subclass. The models of the third and fourth columns are trained with the generated per-pixel annotations of the GTSDB dataset (see Section~\ref{subsec:gtsdb-bboxes-to-perpixel} for the details). All models are trained with the conventional cross-entropy loss.

This section sets the flat classification baselines for comparison with the experiment of the following section.

\subsection{Exp. 2: Hierarchical classification on 3 datasets}
\begin{figure}
	\centering
	\includegraphics[width=1.0\linewidth,trim={4.4cm 15.2cm 8.2cm 0.0cm},clip]{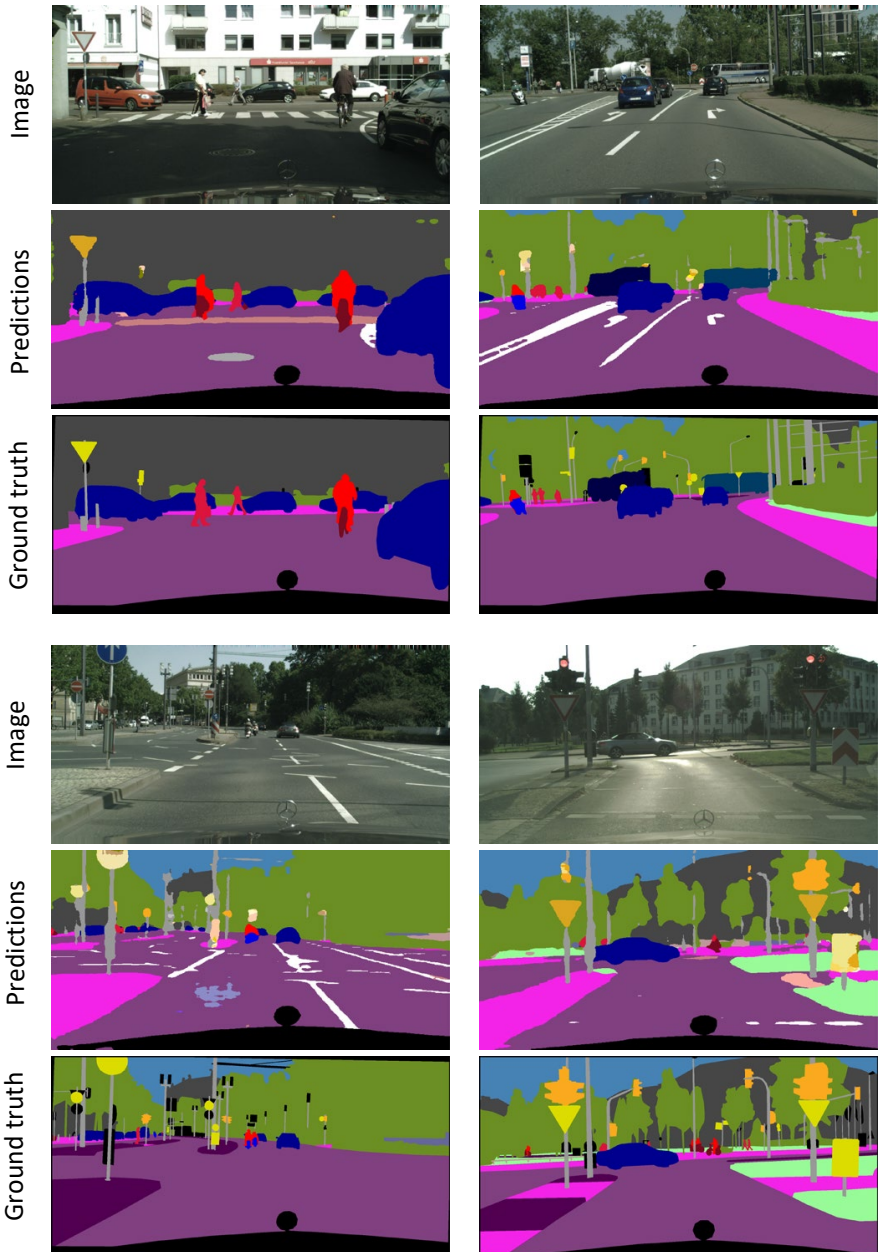}
	\caption{Examples from Cityscapes ``val'' split. The network predictions include decisions from L1-L3 levels of the hierarchy. The ground truth includes only one traffic sign superclass (yellow) without road attribute markings.}
	\label{fig:results-cityscapes}
\end{figure}

\begin{figure}
	\centering
	\includegraphics[width=0.75\linewidth,trim={4.4cm 9.0cm 8.0cm 0.0cm},clip]{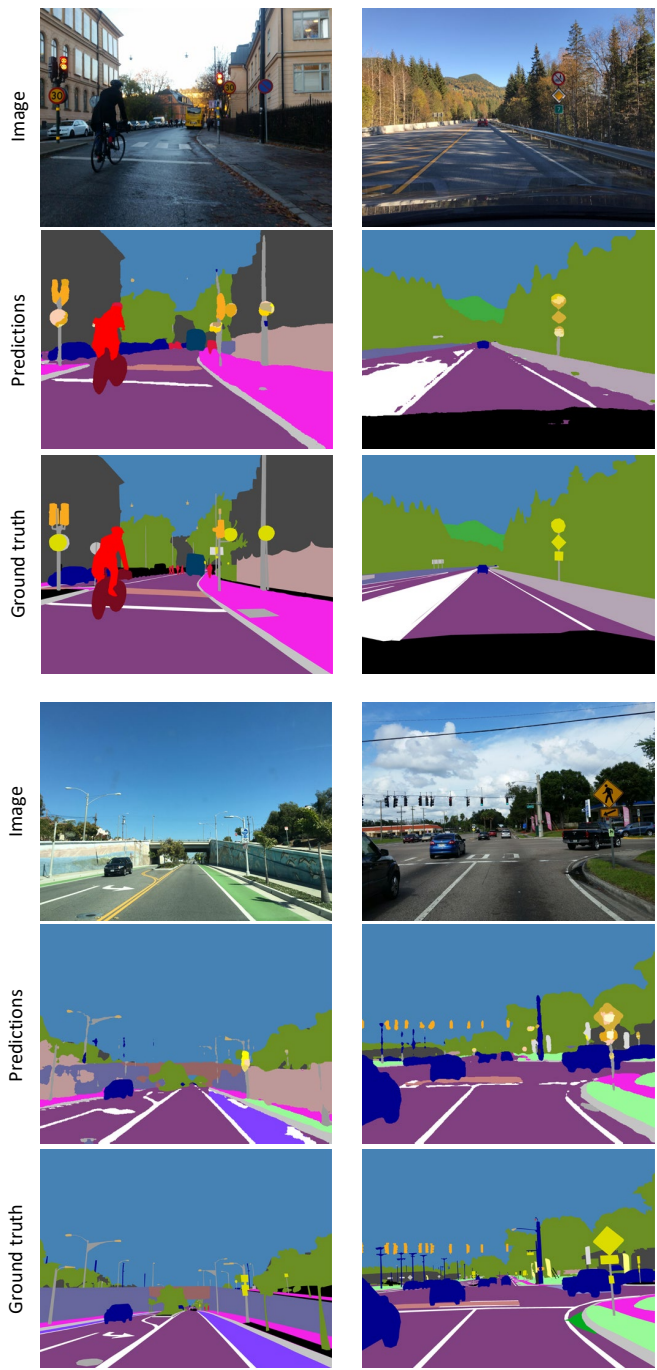}
	\caption{Examples from Mapillary Vistas validation split. The network predictions include decisions from L1-L3 levels of the hierarchy. The ground truth does not include traffic sign subclasses.}
	\label{fig:results-vistas}
\end{figure}

\begin{figure}
	\centering
	\includegraphics[width=0.95\linewidth,trim={4.4cm 13.0cm 8.2cm 0.0cm},clip]{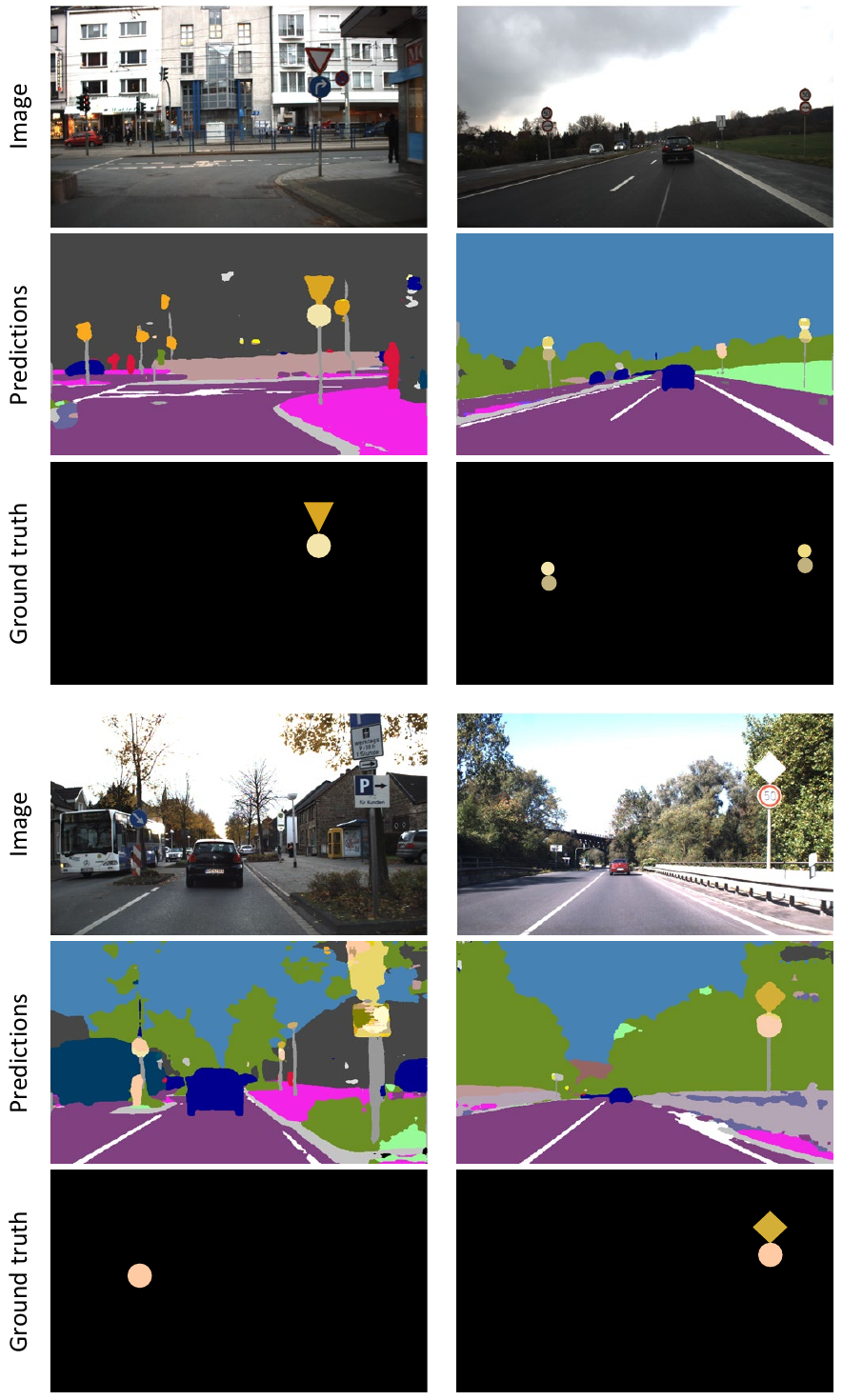}
	\caption{Examples from GTSDB test split. The network predictions include decisions from L1-L3 levels of the hierarchy. Note that the ground truth includes only traffic signs, since the rest of the pixels are unlabeled.}
	\label{fig:results-gtsdb}
\end{figure}

This experiment evaluates the complete hierarchical classification approach on three \textit{heterogeneous} datasets (Cityscapes, Mapillary Vistas, and GTSDB). In Figures~\ref{fig:results-cityscapes},~\ref{fig:results-vistas},~\ref{fig:results-gtsdb} qualitative results are depicted.

In Table~\ref{tab:hierarchical}, evaluation results are depicted on the validation splits of the three datasets that the model is trained with (left three columns) and results for traffic sign subclasses on Cityscapes-Traffic-Signs (right column). By comparing the left three columns of Tables~\ref{tab:flat-baselines} and \ref{tab:hierarchical}, performance is significantly increased in the mPA (in the range +2.4\% up to +32.3\%) and IoU (in the range +2.3\% up to +24.3\%) for all datasets. By comparing the right columns of Tables~\ref{tab:flat-baselines} and \ref{tab:hierarchical}, cross-dataset performance on traffic sign subclasses is also enhanced. Cityscapes-Traffic-Signs was not used by any means during the model training. The hierarchical multiple dataset training scheme is solely accounted for the +10.6\% increase in the mPA.

We conclude that hierarchical classification is highly advantageous for combined \textit{heterogeneous} dataset training, when datasets have conflicting label spaces and in- and between-dataset sample imbalances.

\subsection{Exp. 3: Comparison of hierarchical and flat classification on cross-dataset setting}
\label{subsec:hierarchical-vs-flat}
This experiment evaluates the hierarchical classification method on Cityscapes-Traffic-Signs with per-pixel annotations and a two-level label space. The goal is to isolate the proposed method on a single dataset, in order to show its effectiveness on highly imbalanced datasets against flat classification and rule out the visual domain differences as a source of performance degradation. The input dimensions are $512 \times 1024$ pixels and the batch size is 2.

From Table~\ref{tab:flat-vs-hierarchical}, we observe that hierarchical classification significantly increases the mPA (+26.0\%) and mIoU (+16.1\%) for L2 classes (i.e. GTSDB traffic sign subclasses) with respect to the flat classifier, while for L1 classes (i.e. Cityscapes classes), the increase is above +6\% for both mPA and IoU.

It is concluded that hierarchical classification is robust against class imbalances, even when applied on a single dataset with per-pixel annotations. Moreover, it assigns in each level, and thus classifier, the classes with the same order of examples.
\begin{table}
	\centering
	\footnotesize
	\begin{tabular}{@{}c|c|c|c|c@{}}
		\toprule
		& \multicolumn{2}{c|}{Flat classifier} & \multicolumn{2}{c}{Hierarchical classifiers}\\
		& L1 classes & L2 classes & L1 classes & L2 classes\\
		\midrule
		mPA (\%) & 69.4 (73.0) & \textbf{23.0} & 75.6 (74.8) & \textbf{49.0}\\
		mIoU (\%) & 60.4 (65.2) & \textbf{12.7} & 66.7 (65.7) & \textbf{28.8}\\
		\midrule
		Trained on & \multicolumn{4}{c}{Cityscapes-Traffic-Signs L1 and L2 classes}\\
		\bottomrule
	\end{tabular}
\caption{Flat versus proposed hierarchical classification performance on Cityscapes-Traffic-Signs. (In parenthesis performance for traffic sign L1 class.)}
	\label{tab:flat-vs-hierarchical}
\end{table}

\section{Extra experiments for Robust Vision Challenge}
During the research, we competed at the CVPR 2018 Robust Vision Challenge, using a modified version of the hierarchical semantic segmentation network called Augmented Hierarchical Semantic Segmentation (AHiSS). The challenge involves 4 datasets of diverse sizes,~\ie, Cityscapes, KITTI, ScanNet, WildDash, containing images from a variety of outdoor and indoor scenes. All datasets have large training and validation splits, except for WildDash, which has only a small validation split.

Our proposed solution attained 3\textsuperscript{rd} place overall\footnote{Official websites for results are accessed on June 1st 2018. The official table contains both anonymized and personalized submissions. We merged the anonymous IBN-PSP-SA and identified MapillaryAI submissions, as they appear to be using the same method. We attained 4\textsuperscript{th} place considering all submissions.} (Table~\ref{ch3:tab:rob-overall}) and 1\textsuperscript{st}~-~7\textsuperscript{th} places in the individual dataset benchmarks. The tables show the top-5 methods, while methods scoring lower or did not submit results at the accessing time are not listed in the tables. Per-dataset results are shown in Tables~\ref{ch3:tab:rob-cityscapes}, \ref{ch3:tab:rob-kitti}, \ref{ch3:tab:rob-scannet}, \ref{ch3:tab:rob-wilddash}, \ref{ch3:tab:rob-wilddash-impact} (ordered according to the datasets list above). The table lines are ordered according the column with the black triangle ($\blacktriangle$).

\begin{table}
\centering
\footnotesize
\begin{tabular}{lccccc}
\toprule
\thead[l]{Method name} & \thead{mIoU\textsubscript{cla}{\scriptsize $\blacktriangle$}} & \thead{miIoU\textsubscript{cla}} & \thead{mIoU\textsubscript{cat}} & \thead{miIoU\textsubscript{cat}} &\thead{ROB\\ranking}\\
\midrule
MapillaryAI  & \textbf{75.1} & \textbf{46.3} & \textbf{89.1} & \textbf{72.0} & 1\\
AHiSS (prop) & \textbf{70.6} & \textbf{39.8} & 84.2 & 62.9 & 4\\
VENUS        & 66.4 & 37.1 & \textbf{84.5} & \textbf{66.7} & 3\\
GoogLeNetV1  & 59.6 & 35.1 & 83.0 & 64.4 & 6\\
APMoE\_seg   & 56.5 & 30.6 & 83.5 & 66.1 & 6\\
\bottomrule
\end{tabular}
\caption{Cityscapes ROB results. The top-5 methods are shown ordered according to Cityscapes dataset ordering metric. Bold signifies the two best results per column. The {\scriptsize $\blacktriangle$} denotes the column values used for sorting the rows.}
\label{ch3:tab:rob-cityscapes}
\end{table}
\begin{table}
	\centering
	\footnotesize
	\begin{tabular}{lccccc}
		\toprule
		\thead[l]{Method name} & \thead{mIoU\textsubscript{cla}{\scriptsize $\blacktriangle$}} & \thead{miIoU\textsubscript{cla}} & \thead{mIoU\textsubscript{cat}} & \thead{miIoU\textsubscript{cat}} &\thead{ROB\\ranking}\\
		\midrule
		MapillaryAI  & \textbf{67.5} & \textbf{34.2} & \textbf{87.0} & \textbf{66.3} & 1\\
		AHiSS (prop) & \textbf{61.2} & 26.9 & \textbf{81.5} & 53.4 & 4\\
		VENUS        & 59.1 & \textbf{28.8} & 80.9 & \textbf{60.1} & 3\\
		VlocNet++    & 53.9 & 23.7 & 80.7 & 53.7 & 6\\
		APMoE\_seg   & 48.0 & 17.9 & 78.1 & 49.2 & 7\\
		\bottomrule
	\end{tabular}
	\caption{KITTI ROB results. The top-5 methods are shown ordered according to KITTI dataset ordering metric. Bold signifies the two best results per column. The {\scriptsize $\blacktriangle$} denotes the column values used for sorting the rows.}
	\label{ch3:tab:rob-kitti}
\end{table}

Each dataset defines a set of semantic classes (\eg \textit{person}, \textit{rider}) and groups them into a set of high-level categories (\eg \textit{human}) above the semantic classes. These class and category sets induce two IoU-based metrics that appear in all tables: mIoU\textsubscript{cla} and mIoU\textsubscript{cat}. Specifically, for Wild Dash (Tables~\ref{ch3:tab:rob-wilddash},~\ref{ch3:tab:rob-wilddash-impact}) another metric is defined. The metric \textit{Impact} evaluates the negative impact of common visual artifacts (Table~\ref{ch3:tab:hazards}) on algorithm output performance. This metric is calculated by:
\begin{equation}
\text{Impact} = \dfrac{\min(\text{mIoU}_\text{low}, \text{mIoU}_\text{high})}{\max(\text{mIoU}_\text{none}, \text{mIoU}_\text{low})} - 100.0~~[\%].
\end{equation}
The metrics mIoU\textsubscript{none, low, high} are evaluated on subsets of the dataset that correspond to the identified severity of the image artifacts,~\eg, the subset \textit{Blur\_high} contains images that have a lot of visible blur. Positive impacts are truncated to zero. For example, a \textit{Blur} impact of -10\% translates to an expected performance degradation for the algorithm of 10 percent when there is a considerable blur in the input image, as opposed to supplying the same algorithm a similar image without noticeable image blur. The following list of artifacts is considered.

\begin{table}
\centering
\scriptsize
\begin{tabular}{l|l|l}
\toprule
Artifact & Definition & Example\\
\midrule
Blur & Image is noticeably affected by blur. & motion, defocusing, compression\\
Coverage & Parts of the road are covered. & unusual lane markings, snow, leafs\\
Distortion & Visible lens distortion.\\
Hood & Ego-vehicle is visible, non-windscreen parts. & car hood, mirrors\\
Occlusion & Objects partially occluded or cut-off by border.\\
Overexposure & The scene is overexposed and too bright.\\
Particle & Particles in the air obstruct the view. & heavy rain, snow, fog\\
Screen & The windscreen is interfering. & reflections, wipers, rain\\
Underexposure & The image is underexposed and too dark.\\
Variation & Intra-class variations within the image.\\
\bottomrule
\end{tabular}
\caption{Visual artifacts of Wild Dash dataset covered by the \textit{Impact} metric.}
\label{ch3:tab:hazards}
\end{table}
\begin{table}
	\centering
	\footnotesize
	\setlength\tabcolsep{2.7pt}
	\begin{tabular}{@{}lccccccc@{}}
		\toprule
		\multirowthead{2}[-1ex][l]{Method name} & \thead{Meta AVG} & \multicolumn{4}{c}{Classic} & Negative & \multirowthead{2}[-1ex]{ROB\\ranking}\\
		\cmidrule(lr){2-2} \cmidrule(lr){3-6} \cmidrule(lr){7-7}
		& \thead{MmIoU\textsubscript{cla}{\scriptsize $\blacktriangle$}} & \thead{mIoU\textsubscript{cla}} & \thead{miIoU\textsubscript{cla}} & \thead{mIoU\textsubscript{cat}} & \thead{miIoU\textsubscript{cat}} & \thead{mIoU\textsubscript{cla}} &\\
		\midrule
		AHiSS (prop) & \textbf{39.0} & \textbf{41.0} & \textbf{32.2} & 53.9 & 39.3 & \textbf{43.6} & 2\\
		MapillaryAI  & \textbf{38.9} & \textbf{41.3} & \textbf{38.0} & \textbf{60.5} & \textbf{57.6} & 25.0 & 1\\
		LDN2         & 32.1 & 34.4 & 30.7 & \textbf{56.6} & \textbf{47.6} & 29.9 & 3\\
		BatMAN       & 31.7 & 31.4 & 17.4 & 51.9 & 37.3 & 36.3 & 5\\
		VENUS        & 28.2 & 29.8 & 22.7 & 51.5 & 35.0 & \textbf{50.6} & 8\\
		\bottomrule
	\end{tabular}
	\caption{Wild Dash ROB results. The top-5 methods are shown ordered according to Wild Dash dataset ordering metric. Bold signifies the two best results per column. The MmIoU metric at the left is an average of all other metrics for this dataset. The {\scriptsize $\blacktriangle$} denotes the column values used for sorting the rows.}
	\label{ch3:tab:rob-wilddash}
\end{table}
\begin{table}
	\centering
	\footnotesize
	\setlength\tabcolsep{3.0pt}
	\begin{tabular}{@{}lccccccccccc@{}}
		\toprule
		\multirowthead{2}[-6ex][l]{Method name} & \multirowthead{2}[-6ex]{Average\\impact} & \multicolumn{10}{c}{Impact [\%] (based on mIoU\textsubscript{cla})}\\
		\cmidrule(lr){3-12}
		& & \rotseventy{blur} & \rotseventy{coverage} & \rotseventy{distortion} & \rotseventy{hood} & \rotseventy{occlusion} & \rotseventy{overexposure} & \rotseventy{particle} & \rotseventy{screen} & \rotseventy{underexposure} & \rotseventy{variation}\\
		\midrule
		AHiSS (prop) & \textbf{-14.6} & -11 & -12 & \textbf{-2} & -24 & \textbf{0} & \textbf{-27} & -13 & \textbf{-13} & \textbf{-28} & \textbf{-16}\\
		MapillaryAI  & \textbf{-13.4} & -15 & -5 & -4 & \textbf{-23} & \textbf{0} & \textbf{-23} & \textbf{-12} & -21 & \textbf{-25} & \textbf{-6}\\
		LDN2         & -17.9 & \textbf{-7} & \textbf{0} & -11 & -36 & \textbf{0} & -37 & -16 & -24 & -42 & \textbf{-6}\\
		BatMAN       & \textbf{-14.6} & -9 & -8 & -11 & \textbf{-20} & -11 & -29 & \textbf{-5} & \textbf{-10} & -37 & \textbf{-6}\\
		VENUS        & -18.7 & \textbf{-3} & \textbf{0} & \textbf{0} & -32 & \textbf{0} & -42 & -15 & -31 & -43 & -21\\
		\bottomrule
	\end{tabular}
	\caption{Wild Dash ROB Impact results. The top-5 methods are shown ordered according to Wild Dash dataset ordering metric. Bold signifies the two best results per column.}
	\label{ch3:tab:rob-wilddash-impact}
\end{table}

\begin{table}
	\centering
	\footnotesize
	\begin{tabular}{@{}lcc@{}}
		\toprule
		\thead[l]{Method name} & \thead{mIoU\textsubscript{cla}{\scriptsize $\blacktriangle$}} &\thead{ROB\\ranking}\\
		\midrule
		MapillaryAI  & \textbf{43} & 1\\
		VENUS        & \textbf{37} & 3\\
		VlocNet++    & 36 & 4\\
		APMoE\_seg   & 34 & 6\\
		BatMAN       & 25 & 9\\
		AHiSS (prop) & 18 & 7\\
		\bottomrule
	\end{tabular}
	\caption{ScanNet ROB results. The top-5 methods are shown ordered according to ScanNet dataset ordering metric. Bold signifies the two best results per column. The {\scriptsize $\blacktriangle$} denotes the column values used for sorting the rows.}
	\label{ch3:tab:rob-scannet}
\end{table}

\begin{table}
	\centering
	\footnotesize
	\begin{tabular}{@{}lccccc@{}}
		\toprule
		\thead[l]{Method name} & \thead{Aggregate\\ranking {\scriptsize $\blacktriangle$}} &\thead{KITTI} &\thead{ScanNet} &\thead{Cityscapes} &\thead{WildDash}\\
		\midrule
		MapillaryAI  & 1 & 1 & 1 & 1 & 1\\
		LDN2         & 2 & 2 & 2 & 2 & 3\\
		AHiSS (prop) & 3 & 4 & 7 & 4 & 2\\
		VENUS        & 4 & 3 & 3 & 3 & 8\\
		AdaptNetv2   & 5 & 4 & 4 & 5 & 6\\
		\bottomrule
	\end{tabular}
	\caption{Final ROB consensus ranking based on multiple rankings of metrics on 4 datasets. The {\scriptsize $\blacktriangle$} denotes the column values used for sorting the rows.}
	\label{ch3:tab:rob-overall}
\end{table}

\section{Conclusions}
In this chapter, the problem of simultaneously training CNNs on multiple datasets for semantic segmentation is investigated. The objective of this research is threefold, to increase: i) same-dataset performance, ii) cross-dataset generalizability, and iii) semantic knowledgeability by a single network. We have presented a scheme of hierarchical classifiers as a replacement of the final classification layer of conventional CNNs. The proposed construction enables seamless training on multiple datasets, as shown in the experimentation, and resolves label-space conflicts between datasets.

This chapter contributes to the field of semantic segmentation in the following four ways.
\begin{itemize}[noitemsep,topsep=0pt]
	\item The problem is formulated for combining training with heterogeneously labeled datasets to improve semantic segmentation.
	\item The solution proposes a hierarchical classification methodology for dealing with disjoint, but semantically connected, label spaces.
	\item A modular architecture of hierarchical classifiers is developed that can replace the classification stage in modern convolutional networks.
	\item A dataset is generated by extending Cityscapes with per-pixel annotations over the GTSDB traffic sign subclasses. This dataset is referred to as Cityscapes-Traffic-Signs throughout the thesis.
\end{itemize}

Hierarchical classification maximally reuses dataset and computation resources and eliminates manual labeling effort. The proposal leverages the semantic relationships between datasets' labels to construct a hierarchy of classifiers, and introduces the respective hierarchical training and inference procedures. The final network segments an input image per-pixel into 108 diverse semantic classes from 8 high-level street-scene categories. The results clearly show the benefit of using the proposed hierarchical classification scheme for heterogeneous multi-dataset training.

In the context of the challenges presented in the introduction the results achieve:
\begin{itemize}[noitemsep,topsep=0pt]
	\item An up to sevenfold increase in the number of training examples for every training cycle. This results in an increase in the mPA and mIoU metrics for the same-dataset and cross-dataset settings up to +20\%.
	\item An up to a fivefold increase in the number of recognizable semantic classes by the final network trained on three datasets, compared to networks separately trained on each dataset.
\end{itemize}

This chapter has focused on performance and knowledgeability attained from training on multiple pixel-wise annotated datasets. Although the system of hierarchical classifiers is robust to dataset dissimilarities, it is not scalable to a larger number of datasets. This boundary results from memory limitations and is not fundamental. The discussed approach contains a part of a universal segmentation framework fully described in Chapter~\ref{ch:6-journal}, that addresses, among others, the memory scalability issue. Moreover, at the time of thesis publication, the proposed approach from this chapter was surpassed by other methods~\cite{lambert2020mseg}, including our complete solution presented in Chapter~\ref{ch:6-journal}.\\

The next chapter aims at increasing the performance of semantic segmentation by broadening the candidate datasets to be included during training, for a broad range of image recognition datasets. A final redesign of the complete framework, including memory scalability and inference latency constraints, is presented later in this thesis.

\begin{savequote}[8cm]
\end{savequote}

\chapter{Boosting semantic segmentation with weak supervision} 
\label{ch:4-iv2019}
\begin{spacing}{0.9}
\end{spacing}

\section{Introduction}
\freefootnote{\hspace*{-15pt}The contributions of this chapter were presented in the Proceedings of Int. Conf. IEEE IV 2019 and the paper was selected for an oral presentation.}
The previous chapter has explored multi-dataset training of CNNs for semantic segmentation. The utilization of multiple datasets enhances the generalization capabilities of the trained networks and enables them to segment a scene over an augmented label space, compared to single-dataset training. The proposed methodology is successfully combining datasets after resolving the semantic conflicts between their label spaces. However, the employed datasets belong to a relatively small group of datasets,~\ie those with pixel-wise annotations.

In this chapter, the improvement of CNN segmentation performance through multi-dataset training remains the objective, yet another extension is explored for this improvement purpose. The key to the extension is a specific inclusion from a broad and diverse family of image recognition datasets.

In the context of deep learning, semantic segmentation is traditionally formulated as a per-pixel (dense) classification task. Modeling this task with Fully Convolutional Networks (FCNs)~\cite{long2015fully} has become the \emph{de-facto} solution for images. The success of FCNs is based on the availability of sufficiently large, pixel-labeled datasets~\cite{Cordts2016Cityscapes, neuhold2017mapillary, huang2018apolloscape}. These datasets are costly to obtain and as a consequence, they have a limited size compared to other less accurate image recognition datasets. Moreover, some semantic concepts are not adequately represented in per-pixel datasets, since their frequency of occurrence in the real-world is highly varying,~\eg in traffic scenes, cars may be overrepresented, while riders can be heavily underrepresented. These two factors lead to either low overall performance or serious performance fluctuations among predicted classes. The high availability of less accurate datasets offers space for exploration to use them as additional sources of information for improving semantic segmentation. One of the concepts for exploration may be the use of weak supervision during the training when using these less accurate datasets from the perspective of semantic segmentation.
\begin{figure}
	\centering
	\includegraphics[width=0.7\linewidth]{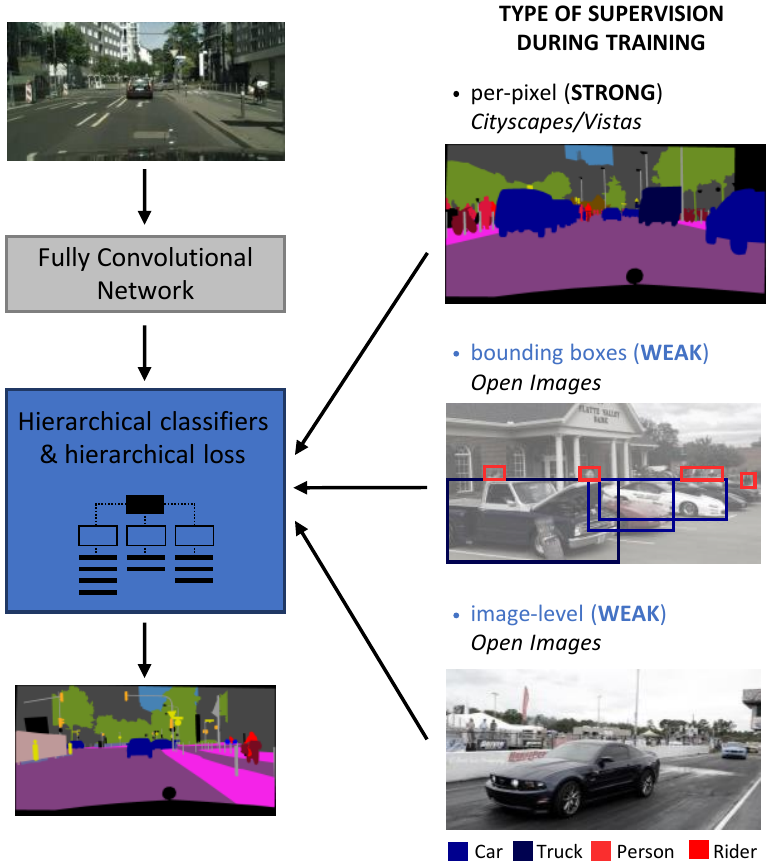}
	\caption{Key research area for this chapter is displayed in blue color. Using diverse types of weak supervision from the Open Images dataset, the aim is to increase the segmentation performance over datasets with strong supervision, by training an FCN with a hierarchy of classifiers and the corresponding loss functions.}
	\label{fig:eye-catcher}
\end{figure}

The objective of this chapter is to exploit weak supervision in conjunction with strong supervision to increase the performance of semantic segmentation using CNNs. The biggest challenge is the incompatibility of annotation formats between the two supervision modalities. FCNs require high localization of information (per-pixel) for ground truth during training. The coarse localization that the weakly labeled datasets provide is not adequate to train the networks.

Therefore, we design a method for per-pixel training of FCNs on multiple datasets simultaneously, containing images with strong (per-pixel) or weak (bounding boxes and image-level) labels. The proposed method fully solves, in a consistent and uniform manner the aforementioned challenge, while training on heterogeneous datasets. It extends the hierarchical classification scheme presented in the previous chapter by incorporating mixed supervision and adding specialized loss functions for training. When starting with such a concept, the following research questions for this chapter come to the foreground:
\begin{itemize}[noitemsep,topsep=0pt]
	\item What is a suitable CNN architecture for addressing a mixed supervision approach?
	\item How can a suitable supervision strategy be defined with a loss specification that matches with this problem?
	\item How to consolidate heterogeneous ground truth within a single training scheme?
\end{itemize}

The remainder of the chapter is organized as follows. The related work is discussed in Section~\ref{ch4:sec:related-work}. Section~\ref{ch4:sec:method} describes the mixed supervision training methodology. The platform for evaluation, including the collection of a new dataset with weak labels, namely \textit{OpenScapes}, together with the implementation details are presented in Section~\ref{ch4:sec:openscapes}. Section~\ref{ch4:sec:experiments} discussed the conducted experimentation, while Section~\ref{ch4:sec:conclusion} presents conclusions.

\section{Related work}
\label{ch4:sec:related-work}
Semantic segmentation is by definition a pixel-wise task and is solved following the established Fully Convolutional Networks (FCNs)~\cite{long2015fully} approach, which trains a CNN with per-pixel (strong) supervision. Previous works have used various types of less detailed (weak) supervision, either independently in a semi- or weakly supervised setting, or to accommodate strong supervision~\cite{zhu2014learning,dai2015boxsup,ibrahim2020semi}. Some methods generate candidate masks from annotated bounding boxes as ground truth  using external modules or pre-computed features. Others use heuristics to refine weak annotations and train networks with or without strong supervision. Weaker supervision has been employed for semantic segmentation with point-level \cite{bearman2016s} or even image-level \cite{wang2020weakly,pathak2015constrained,meng2019weakly} annotations, mainly under the Multiple Instance Learning (MIL) framework. Finally, some works~\cite{ye2018learning,papandreou2015weakly} exploit a combination of multiple types of weak supervision, such as bounding boxes and image-level tags.

Other datasets for related image recognition tasks, such as object detection and image classification, offer a much larger and diverse source of samples for all classes,~\eg ImageNet or Open Images~\cite{kuznetsova2018open}. The ability to train semantic segmentation networks with weakly labeled data from the perspective of semantic segmentation, is an important research trend in the field of image understanding~\cite{papandreou2015weakly, kolesnikov2016seed, pinheiro2015image, ye2018diverse}. The majority of previous works leverage weakly labeled datasets as the main source of information. The hypothesis being investigated here is that weakly labeled datasets can supplement the strongly labeled datasets during simultaneous training of networks for semantic segmentation.

Previous work is generally based on the fact that strong supervision is by definition limited due to involved cost and effort. Because of this limited availability we intend to use what is available as a good starting point and then considerably augment it with the broadly available weak supervision datasets. We aim at improving network performance for semantic segmentation by leveraging heterogeneous weak supervision aside strong supervision on a ``best-effort'' basis. This is achieved by generating pseudo per-pixel labels statically and refining them online using only existing network information (no external modules are needed).

\section{Method: Hierarchical classification with mixed supervision}
\label{ch4:sec:method}

In this section, the three components of the proposed methodology are explained, namely the convolutional architecture, the novel hierarchical loss, and the procedure for consolidating heterogeneous supervision. We have adopted a strong requirement on which the approach is based: all new components should be compatible with the standardized FCN pipeline without using any external modules. These modules may have extra memory or computational requirements, and make the training procedure complex. Although this prerequisite imposes strict limitations, it makes the proposed method potentially applicable to a plethora of existing and future CNN structures with minor modifications.

The key of the method consists of the hierarchical classification scheme developed in Chapter~\ref{ch:3-iv2018}. This is briefly summarized in Section~\ref{ssec:conv-net-arch} and extended to admit weak supervision. The process of adapting weak labels (bounding boxes and image tags) is illustrated in Section~\ref{ssec:pp-gt-gen}. Finally, the required loss functions for training are introduced in Section~\ref{ssec:hier-loss}.
\begin{figure}
\centering
\includegraphics[width=0.7\linewidth]{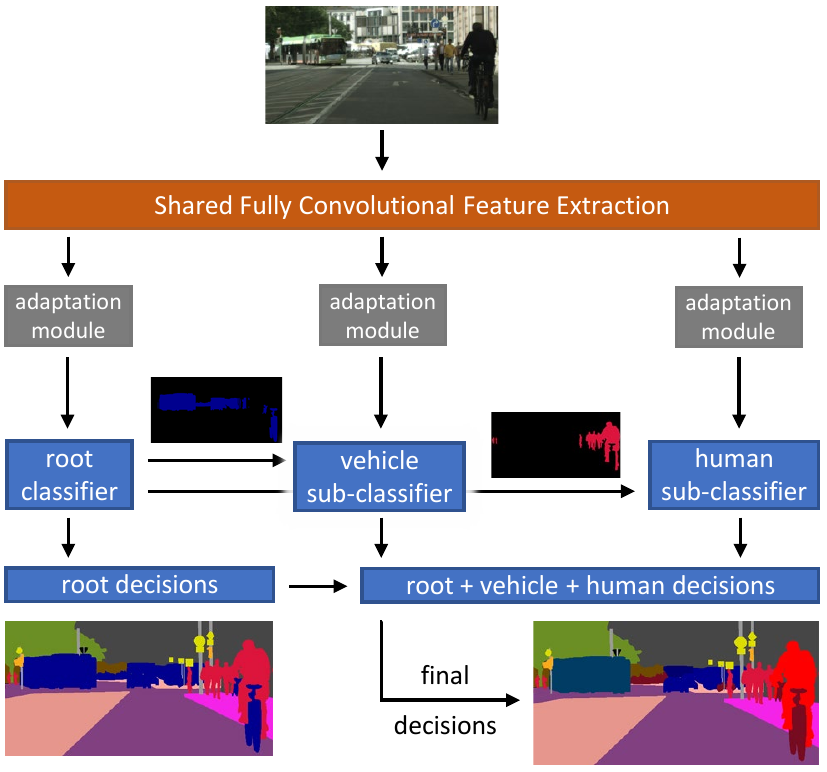}
\caption{Network architecture. The root classifier passes its decisions to the two sub-classifiers, which categorize only pixels that are assigned to them by the root classifier.}
\label{ch4:fig:architecture}
\end{figure}

\subsection{Convolutional Network Architecture}
\label{ssec:conv-net-arch}
The network architecture follows the design of CNN with hierarchical classifiers proposed in Chapter~\ref{ch:3-iv2018} and is depicted in Figure~\ref{ch4:fig:architecture}. Specifically, we opt for a CNN with a two-level hierarchical classification, which consists of a fully convolutional shared feature extraction and a set of, hierarchically arranged, classification stages. The shared feature representation passes through two shallow \textit{adaptation networks}.

The classification hierarchy is constructed according to the semantic taxonomy of classes and the availability of strong and weak supervision for each class. The root classification contains high-level classes with per-pixel labels. Each sub-classification corresponds to a high-level class and contains sub-classes with pixel, bounding-box, or image-tag labels. During training, weak labels are converted into per-pixel pseudo ground-truth as outlined in Section~\ref{ssec:pp-gt-gen}, while the corresponding losses are described in Section~\ref{ssec:hier-loss}.
\begin{figure}
	\centering
	\includegraphics[width=0.9\linewidth]{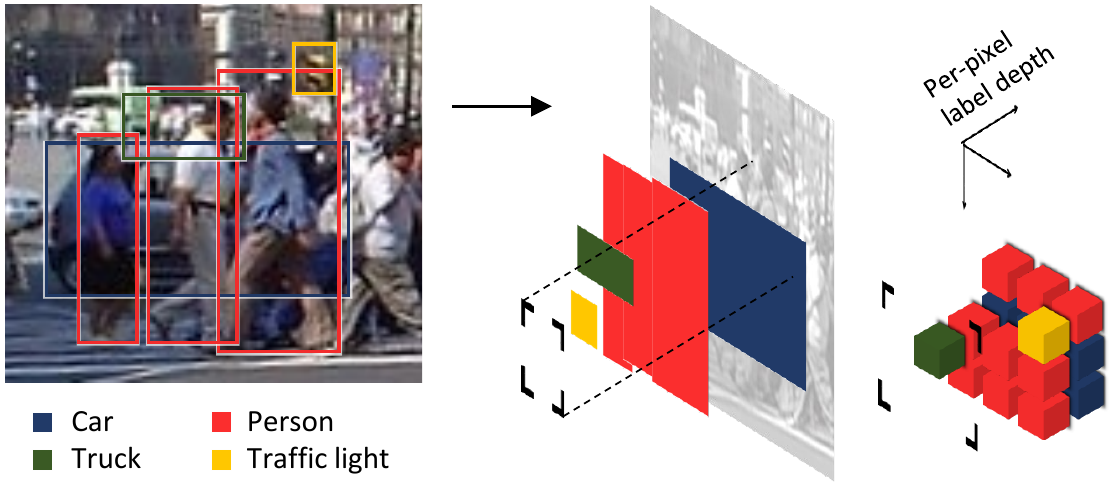}
	\caption{Generation of 3D per-pixel pseudo ground truth (GT). Left: image with a selected subset of bounding boxes, colored by class. Right: proposed 3D GT generation. The figure depicts the class voting process for all bounding boxes. The votes are later converted to a dense categorical probability vector. The same principle is used for generating GT from image-level tag labels: each image tag can be seen as a bounding box that extends into the whole image.}
	\label{fig:2d-vs-3d-gt-gen}
\end{figure}

The purpose of the hierarchical structure is twofold. First, it uniformly and simultaneously solves the training with mixed supervision, without using any external components. 
Second, the hierarchical structure also solves the label space incompatibilities between datasets, due to the unavailability of specific semantic classes in all datasets. This chapter concentrates on investigating cases for which the following assumption applies. The employed weak supervision should be provided for semantic classes that exist in strongly labeled datasets, which is defined and learned by the loss formulation. This formulation is then expanded for applying it in conjunction with the mixed supervision. The necessity of this assumption is discussed in Section~\ref{ch4:sec:discussion}.

\subsection{Generation of pseudo per-pixel labels from weak GT}
\label{ssec:pp-gt-gen}
As mentioned earlier, the goal is to maintain the FCN pipeline intact. This requires the conversion of bounding boxes and image-tags labels into per-pixel ground truth, which can be used to seamlessly train the network.

The regular per-pixel labels can be effectively modeled by a sparse (one-hot)\footnote{\label{ch4:ft:x-hot}The expressions sparse and one-hot throughout the thesis mean that only one element of a probability vector is unity and the others are zero. In dense probability vectors, more than one probability values can be non-zero and all elements should sum to unity. This is different from multi-hot vectors containing many unity elements as they do not represent probabilities. The terminology ``sparse'' and ``dense'' is used in related work and the Tensorflow software framework~\cite{abadi2016tensorflow}.} categorical probability distribution, since each pixel is annotated with a specific semantic class with probability unity. We extend this representation and convert the, possibly overlapping, weak labels into per-pixel labels, modeled with a dense\footnoteref{ch4:ft:x-hot} categorical probability distribution (see Figure~\ref{fig:2d-vs-3d-gt-gen}). This process assigns to each pixel a probability for every class, where the sum of class probabilities should amount to unity.

The conversion procedure follows a voting scheme. Each bounding box or image tag casts a vote at every pixel that they cover, by incrementing a per-pixel counter vector by unity. After all votes are collected, the counters are normalized (across all classes), which turns each counter vector for a pixel into a valid categorical distribution. These vectors represent the weak labels as coarse per-pixel labels and can be used as-is in the conventional cross-entropy loss, as will be described in Section~\ref{ssec:hier-loss}.

\subsection{Losses}
\label{ssec:hier-loss}
\begin{table}
\centering
\footnotesize
\begin{tabular}{@{}l|c|c@{}}
\toprule
\multirow{2}{*}{Classifier} & Per-pixel labeled data & Weakly labeled data\\
& (Cityscapes or Vistas) & (OpenScapes)\\
\midrule
Root & Sparse CCE & n/a\\
Vehicle sub-classifier & Dense CCE & Conditional dense CCE\\
Human sub-classifier & Dense CCE & Conditional dense CCE\\
\bottomrule
\end{tabular}
\caption{Loss components per classifier and per dataset. All losses are per-pixel Categorical Cross Entropy (CCE) losses between the dense or sparse categorical labels and the softmax probabilities of the associated classifier.}
\label{tab:loss-components}
\end{table}
The hierarchical classifiers are trained with a set of pixel-wise losses derived from the standardized Categorical Cross-Entropy (CCE) loss and are summarized in Table~\ref{tab:loss-components}. The losses are accumulated unconditionally for pixels belonging to the pixel-labeled datasets and conditionally for pixels belonging to the bounding box or image-level labeled datasets. We define five loss terms (Table~\ref{tab:loss-components}) that are added, together with the L2 regularization loss, to construct the total loss. The coefficients for the sub-classifiers are all $0.1$ and unity for the root classifier.

The taxonomy of classes assigns a set of possible outcomes to each classifier, where each outcome is simply given a sequence number $\{1, \dots, n\}$ here (\eg for the vehicle sub-classifier, each type gets a different number). The per-pixel labels are given in the form of a vector $\mathbf{y} = \left[y_1, \dots, y_n\right]$ with elements corresponding to each class. To use the labels in a CCE loss function, they are converted into a categorical probability vector $\mathbf{p}$, for which it holds that $\sum_i p_i = 1$. The general form of the per-pixel categorical cross-entropy loss for a softmax classifier with $n$ classes and softmax output $\mathbf{s} = [s_1, \dots, s_n]$ for all pixels $x \in \mathcal{X}$ is:
\begin{equation}
\label{eq:gen-loss}
\mathcal{L}^{\text{dense}} = -\frac{1}{\abs{\mathcal{X}}} \sum_{x \in \mathcal{X}} \sum_{i = 1}^n p^{(x)}_i \log s^{(x)}_i.
\end{equation}
In the root classifier, only the sparse categorical per-pixel labels are used, since this classifier receives supervision from the per-pixel labeled dataset (Cityscapes or Vistas). In this case, $p_{i^*} = 1$ for the correct class $i^*$ that the pixel is labeled with, and $p_{i \neq i^*} = 0$ for all other classes. Thus, Eq.~(\ref{eq:gen-loss}) can be reduced to the sparse CCE loss, which is specified by:
\begin{equation}
\label{eq:sparse-cce}
\mathcal{L}^{\text{sparse}} = -\frac{1}{\abs{\mathcal{X}}} \sum_{x \in \mathcal{X}} \log s^{(x)}_{i^*}.
\end{equation}
In the sub-classifiers, dense categorical per-pixel labels are used for both the per-pixel and the weakly labeled images (see Section~\ref{ssec:pp-gt-gen}). The sparse labels of the per-pixel labeled dataset can be converted into dense categorical labels, by defining a categorical distribution vector and assigning a probability of $p_{i^*} = 1$ to the ground-truth class $i^*$, and probability $p_{i \neq i^*} = 0$ to all other classes. For per-pixel labeled images, the loss of Eq.~(\ref{eq:gen-loss}) is accumulated unconditionally. For weakly labeled images, the loss of Eq.~(\ref{eq:gen-loss}) is accumulated from pixels $x \in \mathcal{X}$ that satisfy two conditions per pixel: 1) the per-pixel pseudo ground-truth probability is positive, and 2) the decision of the root classifier is in accordance with the per-pixel pseudo GT, yielding a non-zero probability in the per-pixel pseudo GT.
\begin{table}
	\centering
	\footnotesize
	\begin{tabular}{@{}l|c|c|c@{}}
		\toprule
		& Cityscapes & Vistas & \textit{OpenScapes} (prop.)\\
		\midrule
		Images & 2,975 & 18,000 & 200,000\\
		Classes & 27 & 65 & 14 \\
		Pixel labels & 1.6 $\cdot$ 10\textsuperscript{9} & 156.2 $\cdot$ 10\textsuperscript{9} & n/a \\
		Bound. boxes & n/a & n/a & 2,242,203\\
		Image tags & n/a & n/a & 1,199,582\\
		\bottomrule
	\end{tabular}
	\caption{\textit{OpenScapes} dataset overview and comparison with per-pixel labeled datasets. Training splits are shown.}
	\label{tab:openscapes-vs-others}
\end{table}

\section{OpenScapes Dataset and Implementation}
\label{ch4:sec:openscapes}
This section outlines the collection procedure of \textit{OpenScapes}, a new dataset containing street scenes, which consists of a realistic use case of a weakly labeled dataset. Moreover, statistics for the three datasets employed in this chapter (Cityscapes Vistas, \textit{OpenScapes}) are provided and the implementation details for our experiments are listed.
\begin{figure}
	\centering
		\includegraphics[width=1.0\linewidth]{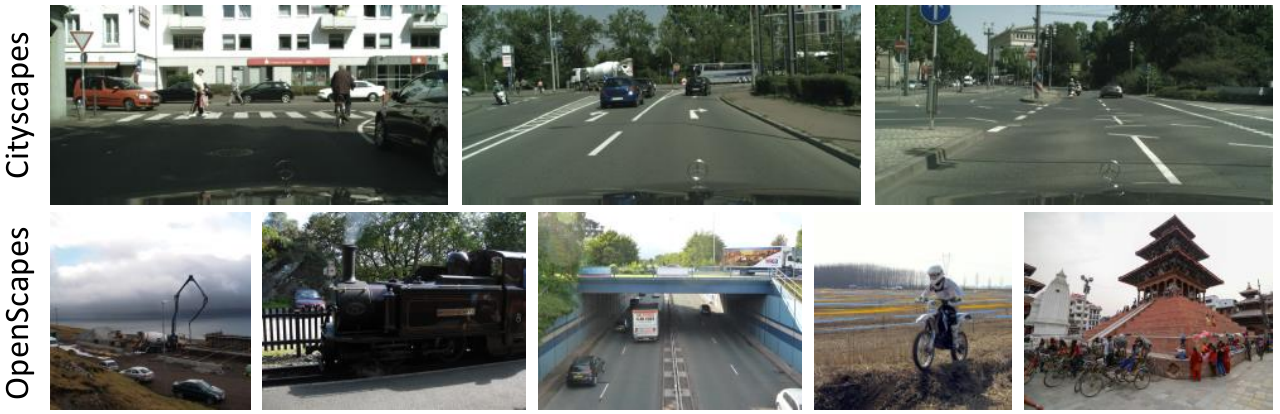}
	\caption{Example images from per-pixel labeled Cityscapes dataset and the weakly labeled \emph{OpenScapes} dataset that demonstrate the large \textit{domain gap}.}
	\label{fig:examples-domain-gap}
\end{figure}
\subsection{OpenScapes Street Scenes Dataset}
\label{ssec:openscapes}
We have collected images of street scenes from the very large-scale (Open-source) Open Images dataset~\cite{kuznetsova2018open} and created a subset named \textit{OpenScapes}. The Open Images dataset contains over 9,000,000 images, 14,600,000 bounding boxes for 600 object classes, and more than 27,900,000 human-verified image-level labels for 19,794 classes. The fully-automated collection procedure is described in Subsection~A of this section. The procedure selects 200,000 images, containing 2,242,203 bounding-box labels and 1,199,582 image-level labels from 14~classes, with ``best-effort'' possible street-scene content. After the selection, visual inspection of images demonstrated that a \textit{domain gap}~\cite{zhang2017transfer, csurka2017domain} occurs between the per-pixel datasets (Cityscapes, Vistas) and \textit{OpenScapes}. This can be seen by the image examples in Figures~\ref{fig:eye-catcher} and~\ref{fig:examples-domain-gap}, and is discussed in Section~\ref{ch4:sec:discussion}.

\subsubsection{A. Automated collection procedure}
\label{ssec:mining-proc}
The first step of the procedure is to rank images in descending order from the Open Images by the number of bounding boxes and image-level labels they contain for the 14 selected street-scene classes. Then, the procedure selects the top 100,000 images for the bounding-box labeled subset and then 100,000 images for the image-level labeled subset, while ensuring that there is no image overlap between the two subsets. For the ranking, a voting system is used, according to which classes in the weak labels of an image vote for an image to be a street-scene image or not. The more probable classes (\eg traffic light and license plate), assign more votes than other classes (\eg car, person) that may appear in non-street-scene contexts. 

\subsubsection{B. Statistics and comparison with pixel-labeled datasets}
\label{ssec:openscapes-vs-others}
In Table~\ref{tab:openscapes-vs-others}, the introduced \textit{OpenScapes} dataset is compared against two established per-pixel labeled datasets that are used in this chapter. Figure~\ref{fig:examples-domain-gap} displays example images from the Cityscapes and \textit{OpenScapes} datasets. As can be observed, the Cityscapes image domain is consistent: images are taken from the same point of view and in one country, contrary to the \textit{OpenScapes} set, which contains web-like images and does not represent a ``homogeneous visual domain''.

\subsection{Implementation details}
\label{ssec:impl-details}
The employed network architecture is depicted in Figure~\ref{ch4:fig:architecture}. The feature extraction consists of the ResNet-50 layers~\cite{he2016deep} (without the classifier), followed by a $1 \times 1$ convolutional layer, to decrease feature dimensions to 256, and a Pyramid Pooling Module~\cite{zhao2017pspnet}. The stride of the feature representation on the input is reduced from 32 to 8, using dilated convolutions. This choice achieves a good trade-off between output spatial detail and memory requirements and is aligned with related works. Each adaptation module consists of a bottleneck block~\cite{he2016deep}, a bilinear upsampling layer to recover the original resolution, and a softmax classifier.

Tensorflow~\cite{abadi2016tensorflow} and four Titan V GPUs (12-GB memory) are used for training. We have implemented synchronous, cross-GPU batch normalization, for accommodating to simultaneous training on multiple datasets. For all experiments, the batch size is set up to four images per GPU, depending on the experiment: one image from the per-pixel labeled dataset (Cityscapes or Vistas), two images from the bounding-box labeled dataset (\textit{OpenScapes} subset), and one image from the image-level labeled dataset (\textit{OpenScapes} subset).

For experiments involving Cityscapes images with dimensions of $512 \times 1024$ pixels and for Vistas $621 \times 855$ pixels are used. Since the \textit{OpenScapes} images have multiple aspect ratios, each image is upscaled to tightly fit the aspect ratio of the per-pixel labeled dataset and then a random patch of the same dimensions as the per-pixel labeled image is cropped. Networks with a batch size of three per GPU are trained for 26 epochs with an initial learning rate of 0.02 and four per GPU for 31 epochs with an initial learning rate of 0.03. All networks are trained with Stochastic Gradient Descent and a momentum of 0.9, L2 weight regularization with decay of 0.00017. The learning rate is halved three times, and the batch normalization moving averages decay is set to 0.9.

\section{Experiments}
\label{ch4:sec:experiments}
\begin{figure}
	\centering
	\includegraphics[height=0.95\textheight]{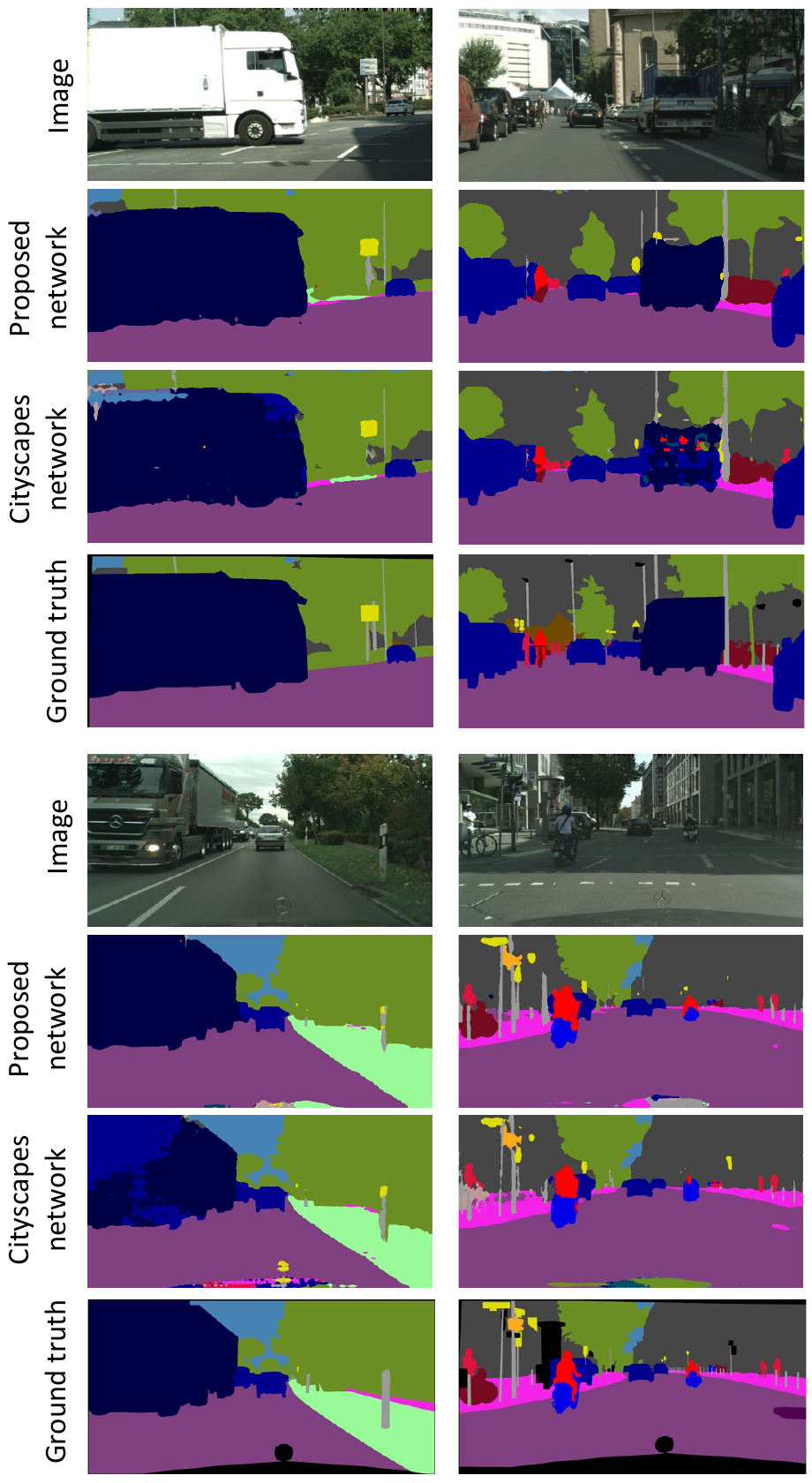}
	\caption{Comparison for Cityscapes validation split images for the hierarchical network trained on the \textit{OpenScapes} subset with bounding boxes and Cityscapes against the baseline network trained on Cityscapes only. The three classes with the largest improvement in IoU are Truck (+13.2\%), Rider (+3.5\%), and Person (+2.1\%), which is also visible.}
	\label{fig:citys-res}
\end{figure}

\begin{figure}
	\centering
	\includegraphics[height=0.95\textheight]{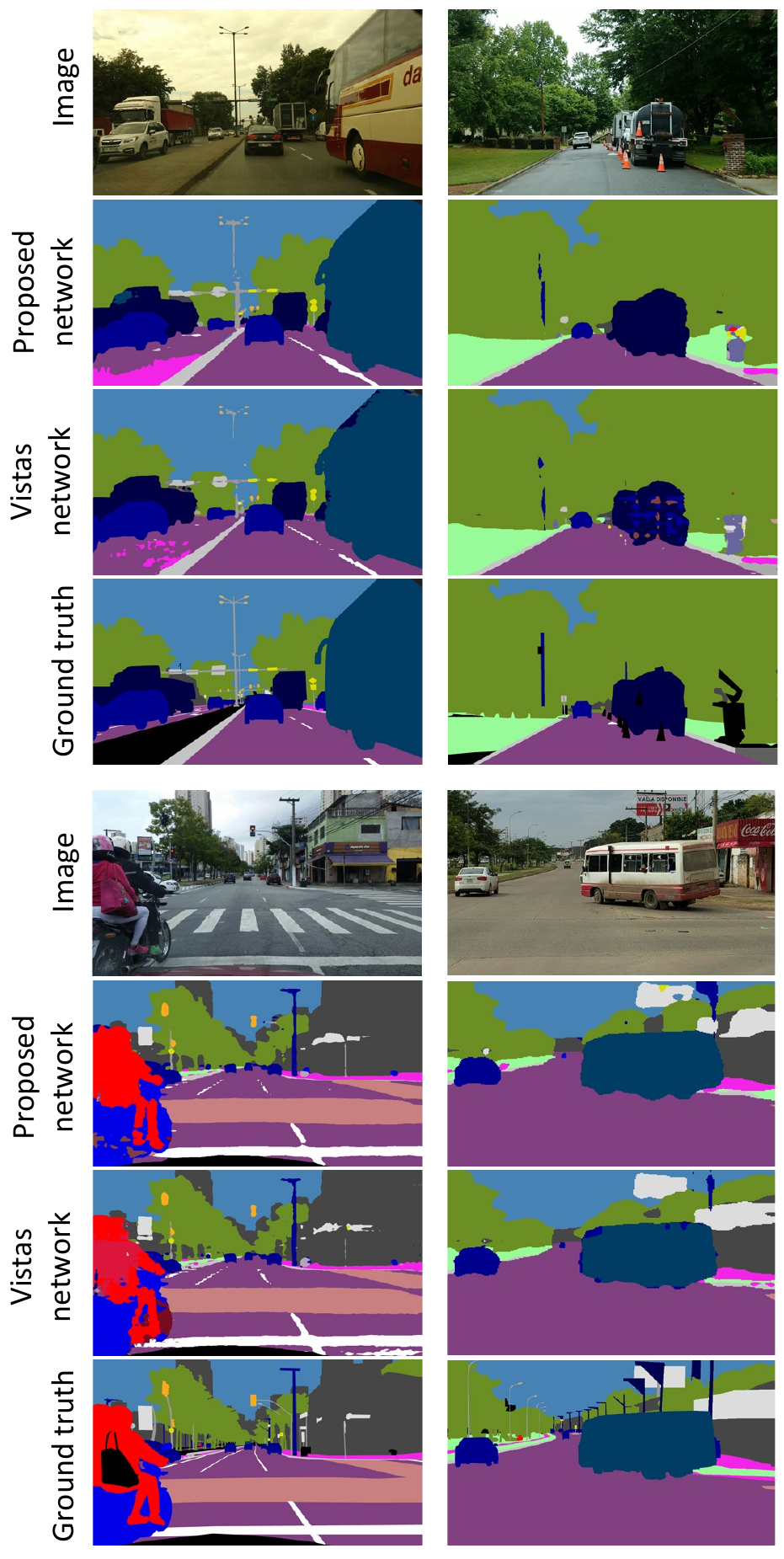}
	\caption{Comparison for Vistas validation split images for the hierarchical network trained on the \textit{OpenScapes} subset with bounding boxes and Vistas against the baseline network trained on Vistas only. The three classes with the largest improvement in IoU are Caravan (+11.3\%), On rails (+10.8\%), and Motorcyclist (+9.6\%), which is also visible.}
	\label{fig:vistas-res}
\end{figure}

As in the previous chapter, the performance is evaluated using two established multi-class metrics for semantic segmentation, namely mean pixel Accuracy (mAcc) and mean Intersection over Union (mIoU). Metrics for all experiments are evaluated on the validation splits of the per-pixel datasets (Cityscapes -- 500 images, Vistas -- 2000 images), and are averaged on the last three epochs to reduce variance. Subsection~\ref{ssec:overall} presents the overall results and Subsection~\ref{ssec:perf-per-class} the per-class results for the classes that receive extra weak supervision. In Subsection~\ref{ssec:perf-effect-of-size}, an ablation on the effect of varying the number of weak samples is provided. Example results from the validation datasets are shown in Figures~\ref{fig:citys-res} and~\ref{fig:vistas-res}.

\subsection{Overall results}
\label{ssec:overall}
\begin{table}
	\centering
	\footnotesize
	\begin{tabular}{@{}ccc|c|c|c|c@{}}
		\toprule
		\multicolumn{3}{c|}{GT origin (train splits)} & \multicolumn{4}{c}{Results (val splits)} \\
		Citys./Vistas & \multicolumn{2}{c|}{OpenScapes} &\multicolumn{2}{c|}{Cityscapes} & \multicolumn{2}{c}{Vistas} \\
		pixel & b. box & tag & mAcc & mIoU & mAcc & mIoU \\
		\midrule
		\ding{51} & & & 77.8 & 68.9 & 53.0 & 43.6 \\
		\ding{51} & \ding{51} & & 79.2 & 70.2 & 52.1 & 43.6 \\
		\ding{51} & \ding{51} & \ding{51} & 79.3 & 70.3 & 52.0 & 43.0\\
		\bottomrule
	\end{tabular}
	\caption{Overall performance of the proposed hierarchical segmentation system on validation splits of pixel-labeled datasets (last two rows). A combination of strong (Cityscapes or Vistas) and weak supervision (\textit{OpenScapes}) is used for the proposed system, compared to conventional segmentation baseline trained only on strong supervision of the respective dataset (first row).}
	\label{tab:perf-overall}
\end{table}
In Table~\ref{tab:perf-overall}, the overall results of the proposed hierarchical segmentation system for Cityscapes~\cite{Cordts2016Cityscapes} and Vistas~\cite{neuhold2017mapillary} are shown. All networks are trained with strong (per-pixel) supervision, from Cityscapes or Vistas, and a combination of weak (per bounding box or image-level or both) supervision from \textit{OpenScapes}. Two subsets of \textit{OpenImages} are used each with 100k images (Section~\ref{ssec:openscapes}) mixed in the batch with Cityscapes or Vistas images (see Section~\ref{ssec:impl-details} for implementation details). The pseudo per-pixel labels are generated as described in Section~\ref{ssec:pp-gt-gen}.

Using Cityscapes data, for the mAcc and mIoU metrics, a steady rise is observed after increasing the amount of weakly supervision included during training. However, for Vistas, training together with the \textit{OpenScapes} subsets slightly harms the performance. This trend corresponds to the average results for all classes and it does not hold for the specific classes, which receive extra supervision, as can be noticed from the following experiments. Overall, it is observed that by adding extra supervision for specific classes, the average performance over all classes is not harmed dramatically, and in most cases is even increased.

\subsection{Improve specific classes with weak supervision}
\label{ssec:perf-per-class}
\begin{table}
\centering
\footnotesize
\setlength\tabcolsep{4.0pt}
\begin{tabular}{@{}ccc|cccccc|c|cc|c@{}}
	\toprule
	& & & \multicolumn{7}{c|}{Vehicle sub-classifier} & \multicolumn{3}{c}{Human sub-class.} \\
	\cmidrule{4-13}
	& & & \multirowrot{4}{Bicycle} & \multirowrot{4}{Bus} & \multirowrot{4}{Car} & \multirowrot{4}{Motorc.} & \multirowrot{4}{Train} & \multirowrot{4}{Truck} & \multirowrot{4}{Average} & \multirowrot{4}{Person} & \multirowrot{4}{Rider} & \multirowrot{4}{Average}\\
	\multicolumn{3}{c|}{GT origin} & & & & & & & & & & \\
	Cityscapes & \multicolumn{2}{c|}{OpenScapes} & & & & & & & & & & \\
	pixel & b. box & tag & & & & & & & & & & \\
	\midrule
	\ding{51} & & & 67.0 & 79.7 & 91.9 & \textbf{52.2} & \textbf{69.3} & 62.3 & 70.4 & 70.2 & 47.9 & 59.0 \\
	\ding{51} & \ding{51} & & 67.8 & \textbf{81.8} & \textbf{92.5} & 50.3 & \textbf{69.3} & 71.4 & \textbf{72.2} & 71.9 & 50.7 & 61.3 \\
	\ding{51} & \ding{51} & \ding{51} & \textbf{67.9} & 79.1 & \textbf{92.5} & 48.7 & \textbf{69.3} & \textbf{75.5} & \textbf{72.2} & \textbf{72.3} & \textbf{51.4} & \textbf{61.9}\\
	\bottomrule
\end{tabular}
\caption{Proposed hierarchical segmentation system performance (last two rows) on Cityscapes compared to conventional segmentation baseline (first row) per-class; \textbf{IoU} (\textbf{\%}) values for the selected classes that receive extra supervision from the weakly labeled \textit{OpenScapes} dataset (100k subsets). Results are grouped per sub-classifier.}
\label{tab:perf-detail-citys}
\end{table}
\begin{table}
	\centering
	\scriptsize
	\setlength\tabcolsep{2.0pt}
	\begin{tabular}{@{}ccc|ccccccccccc|c|cccc|c@{}}
		\toprule
		& & & \multicolumn{12}{c|}{Vehicle sub-classifier} & \multicolumn{5}{c}{Human sub-classifier}\\
		\cmidrule{4-20}
		& & & \multirowrot{4}{Bicycle} & \multirowrot{4}{Boat} & \multirowrot{4}{Bus} & \multirowrot{4}{Car} & \multirowrot{4}{Caravan} & \multirowrot{4}{Motorc.} & \multirowrot{4}{On rails} & \multirowrot{4}{Oth. veh.} & \multirowrot{4}{Trailer} & \multirowrot{4}{Truck} & \multirowrot{4}{Wheeled} & \multirowrot{4}{Average} & \multirowrot{4}{Person} & \multirowrot{4}{Cyclist} & \multirowrot{4}{Motorc.} & \multirowrot{4}{O. rider} & \multirowrot{4}{Average}\\
		\multicolumn{3}{c|}{GT origin} & & & & & & & & & & & & & & & & & \\
		Vistas & \multicolumn{2}{c|}{OpenScapes} & & & & & & & & & & & & & & & & &\\
		pixel & b. box & tag & & & & & & & & & & & & & & & & &\\
		\midrule
		\ding{51} & & & 55.0 & \textbf{26.7} & \textbf{75.0} & \textbf{88.8} & 0.3 & \textbf{54.2} & 38.4 & 16.9 & 0.3 & 65.0 & 7.4 & 38.9 & \textbf{65.5} & \textbf{51.4} & 43.1 & 0.0 & 40.0\\
		\ding{51} & \ding{51} & & \textbf{56.1} & 21.2 & 73.8 & 88.6 & \textbf{11.6} & 53.9 & \textbf{49.2} & \textbf{18.4} & \textbf{0.9} & \textbf{66.9} & \textbf{10.7} & \textbf{41.0} & 64.7 & 47.1 & \textbf{52.7} & \textbf{0.4} & \textbf{41.2}\\
		\ding{51} & \ding{51} & \ding{51} & 54.5 & 21.2 & 74.0 & 88.4 & 11.4 & 52.8 & 49.0 & 18.1 & 0.8 & 66.0 & 10.6 & 40.6 & 64.6 & 47.1 & 49.9 & 0.3 & 40.5\\
		\bottomrule
	\end{tabular}
	\caption{Proposed hierarchical segmentation system (last two rows) performance on Vistas compared to conventional segmentation baseline (first row) per-class; \textbf{IoU} (\textbf{\%}) values, for the selected classes that receive extra supervision from the weakly labeled \textit{OpenScapes} dataset (100k subsets). Results are grouped per sub-classifier.}
	\label{tab:perfor-detail-vistas}
\end{table}

This section investigates the performance on specific important classes,~\eg vulnerable road users, which are the chosen classes to receive extra weak supervision. The results are provided in Tables~\ref{tab:perf-detail-citys} and~\ref{tab:perfor-detail-vistas}. In the case of using the \textit{OpenScapes} bounding-box-labeled subset, then the average IoUs of the chosen classes is improved for both datasets. In the case of using the \textit{OpenScapes} image-tag labeled subset results are less pronounced in the increase. For Cityscapes, the average IoUs continue to improve, but for Vistas they are decreased. This phenomenon is due to the large number of classes in the Vistas dataset, which constrains the available number of examples per class during each training step. This is further elaborated in the Section~\ref{ch4:sec:discussion}.

Overall, we conclude that compared to the baseline (without using any extra weak supervision) all average IoUs are increased with the proposed hierarchical segmentation system. The three largest gains in per-class IoUs are: +13.2\% for Cityscapes \textit{truck} class, +11.3\% for Vistas \textit{caravan} class, and +10.8\% for Vistas \textit{on rails} class.

\subsection{Ablation study: Effect of weakly labeled dataset size}
\label{ssec:perf-effect-of-size}
This experiment examines the effect of the size of the weakly labeled dataset on the performance. The hierarchical architecture is trained on Cityscapes with strong supervision, together with different portions of the \textit{OpenScapes} bounding-box labeled subset, where all other hyper-parameters fixed. The results of Table~\ref{tab:size-matters} indicate that performance gains are increasing proportionally to the amount of used weakly labeled images. Moreover, from row two of the same table, it is evident that there is a lower bound of employed weak supervision in order to attain gains, otherwise, performance may even drop.
\begin{table}
	\centering
	\footnotesize
		\begin{tabular}{r|c|c}
			\toprule
			Per-pixel + \#\textit{images} with b. box GT & mAcc & mIoU \\
			\midrule
			$0$ images ($0$ bboxes) & 77.8 & 68.9 \\
			$1k$ images ($17.3k$ bboxes) & 77.4 & 68.4 \\ 
			$10k$ images ($140.4k$ bboxes) & 78.2 & 69.2 \\ 
			$100k$ images ($1185.8k$ bboxes) & 79.3 & 70.3\\ 
			\bottomrule
		\end{tabular}
\caption{Performance (\%) of the proposed hierarchical system on Cityscapes with a different number of bounding boxes used to generate pseudo ground-truth labels for the weakly labeled dataset. The initial drop in performance is mainly due to the tuning of training hyper-parameters (number of epochs, batch size,~\etc) for the case of the third row (10k images).}
\label{tab:size-matters}
\end{table}

\section{Discussion}
\label{ch4:sec:discussion}

After having performed the experiments, we have observed that the performance of our method depends on three factors: 1) the amount of the employed weakly-labeled images, 2) the semantic connotation between classes from different datasets,~\ie the semantic definition of classes with same name among datasets is not identical, and 3) the \textit{domain gap}~\cite{zhang2017transfer, csurka2017domain, romijnders2019agnostic} between strongly and weakly-labeled datasets. The nature of this discussion is more conceptual, than actually discussing the results.

\subsubsection{A. Amount of weak supervision}
The ablation experiment of Table~\ref{tab:size-matters} provides clues on the performance gains as a function of the amount of weakly-labeled images. It is highly likely that the performance gain results from the growing amount of data with the labels. The table also shows with the numbers in the second that a sufficient amount of such data is needed in order to establish a performance gain.

However, it could be discussed that some part of the gain is also resulting from the method. This hypothesis could be valid when considering the combination of a weakly-labeled dataset with a disjoint image domain and highly conflicting \textit{semantic connotation extent} with a strongly-labeled dataset. In such a case, we expect a smaller increase of even a decrease in performance. This phenomenon occurs in one of our tables. This motivates our opinion on the performance gain being dependent on the method as well.

\subsubsection{B. Definition of image domain in literature}
\label{ch4:ssec:image-domain}
To broaden the discussion, we also bring in two aspects of a more theoretical nature.
In the first two chapters, we assumed that the images from all datasets that are trained simultaneously come from similar domains. In such a case, the network can extract consistent features from all images with a common feature extraction. This is important because, according to the proposed conditional losses, pixels that are not correctly predicted by the \textit{root classifier} in a training step are excluded from the sub-classifiers for that step. For example, in the extreme case where a car in the weakly-labeled dataset has completely different appearance and context from a car in the strongly-labeled dataset, the \textit{root classifier} will not detect it and all of its pixels will be concealed from the sub-classifiers.

In our experiments, this issue was minimally hindering the learning process, since we were monitoring the number of pixels corresponding to each sub-classifier during training. We hypothesize that the Cityscapes and Vistas image domains are rich enough, so the trained network can discover pixels with similar semantics from the Open Images domain,~\ie the trained network generalizes well, and thus the training of the sub-classifiers is unobstructed. However, this assumption may not hold in another combination of datasets,~\eg real-life and computer-rendered datasets. Moreover, the ``domain'' of a dataset in not concisely defined in literature and this field is still under research (see Figure~\ref{fig:examples-domain-gap} for examples). Finally, it should be noted that we have opted not to include any modules from the domain-adaptation field to account for this effect, since this is not the goal of this work and would violate our design rule to maintain the FCN pipeline identical. We have addressed this topic in recent work for agnostic inference~\cite{romijnders2019agnostic}, which is not described in this thesis.

\subsubsection{C. Definition of semantic classes among different datasets}
Another important topic is that the definition of semantic classes among different datasets can be different. This implies that the included semantic concepts for the same type of semantic object can be different, which we call the extent of \textit{class connotation}. 

In this chapter, we have assumed that classes with the same name across datasets are defined similarly and contain the same semantic concepts. For example, it is essential that small variations of a \textit{truck} in Vistas are still labeled as \textit{truck} in Cityscapes (and do not become \textit{caravan} for example). The effect of overlapping definitions between different classes and the lack of sufficient overlap for the same class is visible in the performance drop for the \textit{motorcycle} class in Table~\ref{tab:perf-detail-citys}, for which the semantic extent of the \textit{motorcycle} class definition diverges between Cityscapes and \textit{OpenScapes} datasets. In general, the more the fine-grained partitioning of semantic classes in a dataset, the easier it is for conflicts to appear with same-name classes in other datasets. This problem of \textit{semantic class connotations} is further explored in Chapter~\ref{ch:6-journal}.

\section{Conclusion}
\label{ch4:sec:conclusion}
We have presented a fully convolutional network with a hierarchy of classifiers for simultaneous training on strongly- and weakly-labeled datasets for semantic segmentation. The proposed architecture is coupled with suitable loss functions, enabling simultaneous training on multiple heterogeneous datasets. This chapter extends the developed hierarchical classification methodology of Chapter~\ref{ch:3-iv2018}, to include weak labels to be combined with strong labels, for extended training without requiring any external components or pre-training. We have evaluated the system on two established per-pixel labeled datasets, namely Cityscapes and Vistas. The final trained networks achieve improvements in IoU for the classes that receive extra weak supervision of up to +13.2\%, while the overall performance over all classes is improved in the majority of the cases.

To evaluate the efficacy of the approach in a realistic scenario, we have collected street-scene images from the Open Images dataset~\cite{kuznetsova2018open}, using an automated procedure and generating a weakly-labeled dataset called \textit{OpenScapes}. This new dataset contains 100,000 images with 2,242,203 bounding-box labels and 100,000 images with 1,199,582 image-tag labels, spanning 14 of the most important street-scene classes. Using \textit{OpenScapes}, we have shown that the performance for classes that receive extra weak supervision (up to 14) is increased, provided that a sufficient amount of weak labels are available. Moreover, we have examined the effect of the size of the weakly-labeled dataset and have demonstrated that the performance increase is proportional to the size of the employed weakly-labeled dataset. For our experiments, it is assumed that the definitions of the domains among the datasets have only small differences.

To summarize, the contributions of this chapter are as follows.
\begin{itemize}[noitemsep,topsep=0pt]
	\item An FCN architecture with a hierarchy of classifiers for simultaneous training on datasets with diverse supervision, including per-pixel, bounding-box, and image-tag labels.
	\item A novel cross-entropy loss that enables conditional training of sub-classifiers with weak supervision.
	\item The newly constructed \textit{OpenScapes} dataset exploiting also existing data, offers a large, weakly-labeled dataset with 200,000 images, 3,400,000 weak labels spanning over 14 semantic classes for street-scene recognition.
\end{itemize}

~

Weakly-labeled images contain, due to their nature, very coarse information from the perspective of semantic segmentation,~\eg per training step the useful labels belong on the average to approximately 90\% of per-pixel datasets and only up to 10\% of weakly-labeled datasets. As consequence, the training is slow and weak supervision is not used efficiently. In the following chapter, we will research how weakly-labeled images can be carefully selected, in order to maximize the information flow per training cycle and efficient use of training data.

\begin{savequote}[8cm]
	
\end{savequote}

\chapter{Weak data selection for efficient multi-dataset training}
\label{ch:5-itsc2019-wacv2019}
%
%
\section{Introduction}
\label{ch5:sec:intro}
\freefootnote{\hspace*{-15pt}The contributions of this chapter were presented in the Proceedings of Int. Conf. IEEE ITSC 2019 as a joint work with Rob Romijnders and the paper was selected for an oral presentation.}
In the previous chapter, a multi-dataset training scheme is developed to improve the segmentation accuracy of important (\eg vehicle, human), vulnerable (\eg pedestrian), or underrepresented (\eg train, rider) street-scene classes. According to the proposed method, an FCN with hierarchical classifiers is trained on various combinations of a pixel-labeled (strong) dataset and one or more weakly-labeled datasets. The results demonstrated that in the majority of the cases, information from bounding boxes and even image tags can be utilized, together with pixel labels, to contribute to increasing segmentation performance. However, as observed in that chapter, the coarser the spatial localization of a label, the less valuable it becomes for the fine, pixel-wise semantic segmentation task. The proposed system of the previous chapter has beneficially exploited the trade-off between the deficit in information density and the vast availability of labels, which appears in weakly-labeled datasets. Specifically, we have found that up to 30 times more weakly-labeled samples are required together with the strongly-labeled samples to account for this deficit. A crucial question for follow-up research is to consider whether it is possible to achieve similar gains by using fewer weakly-labeled samples to increase segmentation accuracy without sacrificing training efficiency,~\eg by selecting the most informative examples.

When using a limited amount of strongly-labeled data and at the same time reducing the amount of weakly-labeled samples, the information content of the data becomes lower. In any case, the computation time spent on low information samples is generally not well spent and should be minimized when possible. The previous research question also addresses the imbalance of datasets, because of the large ratio between weakly- and strongly- labeled data. Coupled to this issue is the problem of bias, because the amount of weakly-labeled data to be used as real information may favor the learning towards the poorly localized samples. As can be noticed, there are multiple issues connected to the posed research question, which are listed below. More specifically, the previous issues concern the following training procedure (TP) and information quality (IQ) challenges.
\begin{itemize}[noitemsep,topsep=0pt]
	\item \textit{IQ - Scarcity of information}: Images in weakly-labeled datasets are often annotated with only a few bounding boxes or image tags, which provide very little information for the pixel-level semantic segmentation task.
	\item \textit{IQ - Low richness of information}: Weakly-labeled datasets often have repetitive examples and thus low additional informative value. These examples have similar appearance, because for instance they come from the same sequence of frames, resulting in increased overfitting during training.
	\item \textit{TP - Inefficient training}: Computations are spent for images with little informative value and as a consequence, training time is increased.
	\item \textit{TP - Imbalanced training}: Different dataset sizes cause either a large portion of weak labels to remain unused or the repetitive processing of pixel-labeled images, thereby leading to overfitting of the network.
	\item \textit{TP - Batch contents}: The low informative value of weak labels mentioned for inefficient training, requires a large amount of weakly-labeled samples to be included in a batch, which can lead to a bias towards poorly localized samples, so that accuracy may even drop further.
\end{itemize}

The previous list of challenges is intended to show the involved research aspects and issues of the main research questions of this chapter, rather than a list that is going to be used for addressing each issue individually. In the sequel, we aim at solving the main research question, but at the same time attempt to address several of these issues simultaneously and achieve some benefits for them.
\begin{figure}
	\centering
	\includegraphics[width=0.85\linewidth]{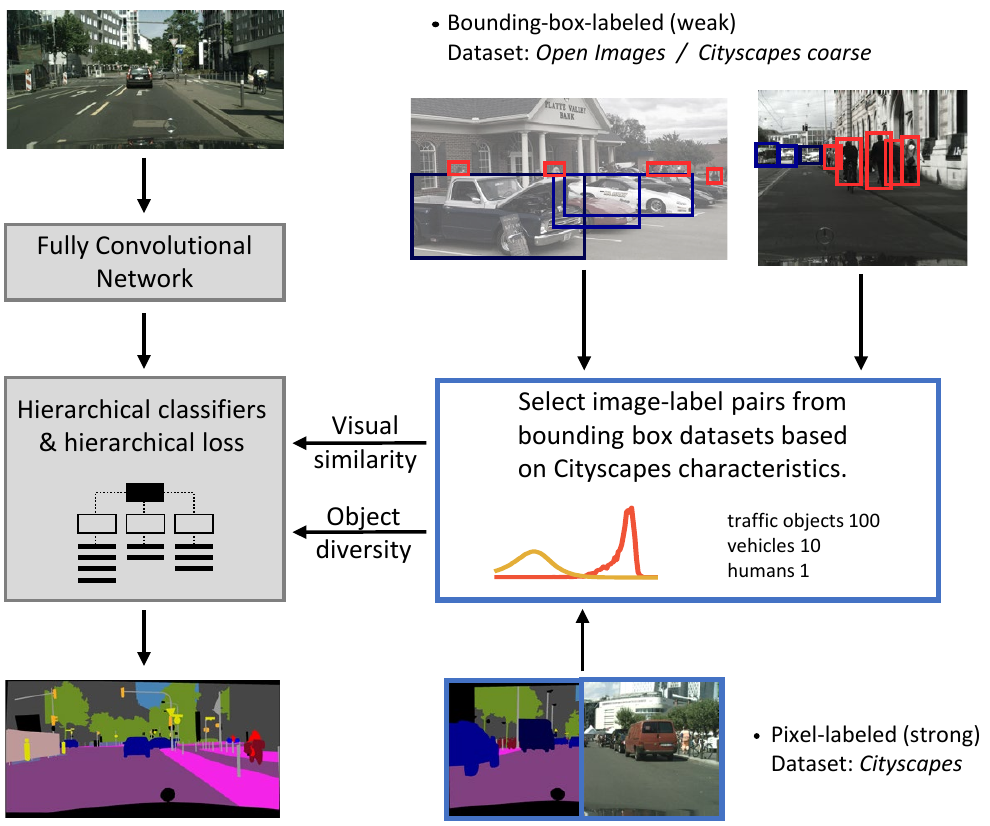}
	\caption{Method overview. The core of the method (center right) issues selected image-label pairs from the weakly-labeled datasets (top right) so they are visually similar to the strongly-labeled dataset (bottom right) and contain a high diversity of objects of interest. Subsequently, by training on both strongly- and selected weakly-labeled datasets, the benefits of data selection for multi-dataset semantic segmentation become visible.}
	\label{ch5:fig:eye-catcher}
\end{figure}

We hypothesize that a subset of a weakly-labeled dataset with informative image-label pairs can be selected to match some specific characteristics of the target strongly-labeled dataset. With the term specific we mean 1) the visual content of the subset belongs to the same domain and 2) the technical details like classes and resolution are matching between the datasets. When this match of characteristics is satisfied, it can lead to a balanced training involving less manual tuning of hyper-parameters, thereby improving segmentation accuracy. Hence, this chapter investigates data selection mechanisms, while preserving the hierarchical classification model for heterogeneous multi-dataset training.

The research questions imply an approach to mine diverse examples from weakly-labeled datasets and eliminate imbalances between strongly- and weakly-labeled datasets. To this end, we investigate two data selection methods, which indicate how image-label pairs can be chosen for maximizing \textit{object diversity} and \textit{visual similarity} among used datasets. The first method models image representations of a dataset with a Gaussian Mixture Model (GMM). This model can be used to find visually similar images between the strongly- and weakly- labeled datasets. The second method employs predefined scoring heuristics to rank image-label pairs according to the diversity of the semantic classes they contain.

The rankings induced by the two selection methods are used to choose the precise proportion of image-label pairs from the weakly-labeled dataset that matches the size of the strongly-labeled dataset and contains sufficient informative pairs. Subsequently, this selection is employed, together with pixel-labeled images, to efficiently train the multi-dataset FCN of the previous chapter and improve the resulting segmentation accuracy.

The remainder of chapter is organized as follows. Similar research fields are discussed in Section~\ref{ch5:sec:rel-work}. The core of the proposed method is discussed in Section~\ref{ch5:sec:method} and Section~\ref{ch5:sec:details} includes the specific details of the implementation and employed datasets. Section~\ref{ch5:sec:experiments} illustrates the usability and performance gains of the methodology through experimentation and ablations. The conclusions are presented in Section~\ref{ch5:sec:conclusion}.

\section{Position of the work in literature}
\label{ch5:sec:rel-work}
The proposed approach of this chapter can be categorized as residing on the intersection of image retrieval and semi-supervised learning. However, contrary to works specializing in each of these fields, the conducted research distincts itself in two ways:
\begin{enumerate}[noitemsep,topsep=0pt]
\item Model-related semi-supervised methods for improving performance are not employed, but instead the responsibility is given to a process of smart data selection;
\item Ground-truth data for relevant ``retrieved'' image-label pairs do not exist, thus the effectiveness of the approach cannot be quantified with retrieval metrics. Instead, the performance of the ranking is validated through the segmentation performance of the trained networks, which acts as a surrogate metric for quantifying the performance of the data-selection process.
\end{enumerate}
The above points and positioning of the work prohibits a normal discussion on related work. Nonetheless, in the following we briefly summarize some related work and survey papers about fields of solutions.
\begin{figure}
	\centering
	\includegraphics[width=1.0\linewidth, trim={0.0cm 27.5cm 0.0cm 0.0cm}, clip]{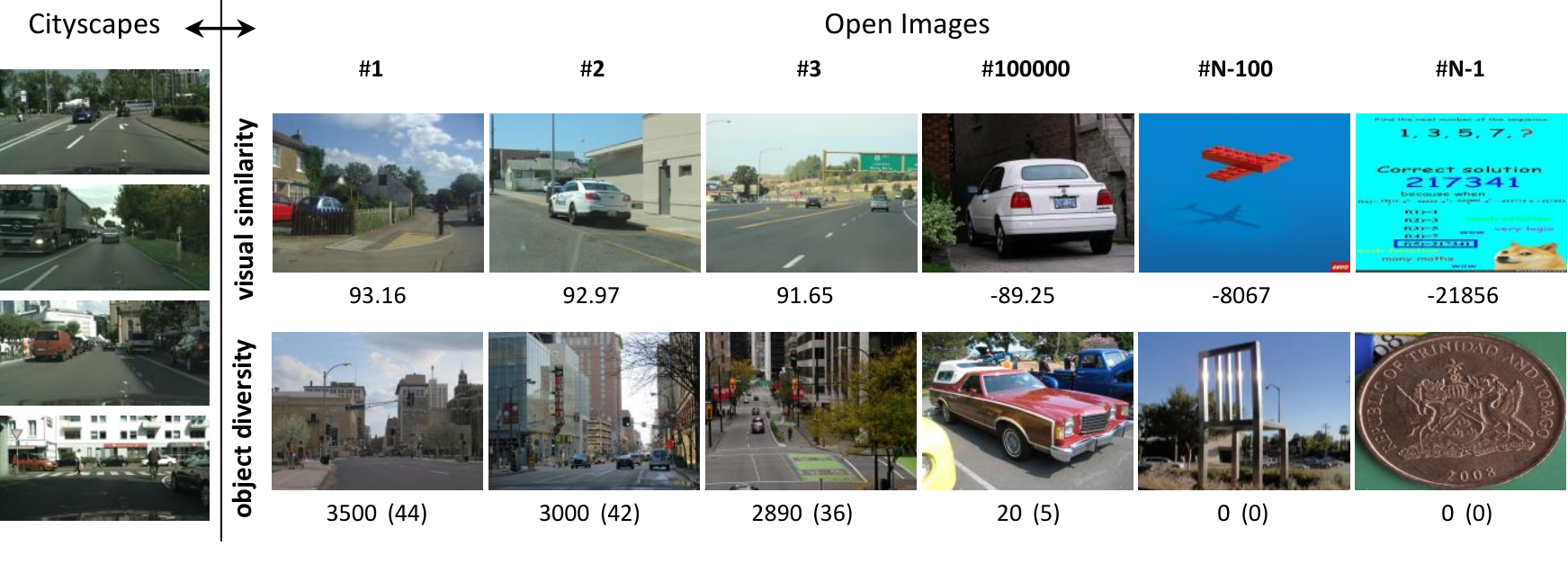}
	\caption{Examples of selected images from $N = 1.74$ million Open Images images using the propose data selection methods in descending order. First row: \textit{visual similarity} using GMM, the $sim_{citys}$ measure is shown. Second row: \textit{object diversity} using class scores, the heuristics scores and the number of objects of interest are shown.}
	\label{fig:sel-ims}
\end{figure}

The redundancy of training data has been studied in an intra-dataset context for classification~\cite{birodkar2019semantic, vodrahalli2018all, liu2018pixel}, but to the best of our knowledge, for semantic segmentation it remains largely unexplored. Training of convolutional networks for semantic segmentation with weak supervision is gaining traction in many different disciplines~\cite{Wang2018survey,Xu2019self,Wilson2020survey,Redko2020survey,Toldo2020review},~\eg \mbox{semi-/weakly-/self-} supervised learning, or domain adaptation. The majority of the works in these disciplines use weak labels as the single source for supervision, mainly in a single-dataset training setting. The main direction for addressing these problems is through model selection,~\ie design, train, and tune a convolutional network for robustness and performance. For example, in semi-supervised training, apart from the main segmentation loss, various extra losses have been proposed to handle weakly-labeled images. This chapter investigates a less studied branch that focuses on data selection~\cite{birodkar2019semantic, vodrahalli2018all, ghahremani2020vessels, liu2018pixel}, while not introducing any extra modules or losses to the model. Unlike domain adaptation networks, which receive no supervision for the target dataset, our employed multi-dataset training uses weak supervision as an additional source of information to enrich existing and strong supervision.

Content-based image retrieval~\cite{chen2021retrieval,rehman2012content,smeulders2000content,lew2006content} analyzes the content of images and retrieves images accordingly. Concept-based image indexing uses image metadata~\eg description, tags, or text, associated with the image and does not examine the image content. These two tasks are in principle symmetrical with our methods for finding \textit{visual similarity} and \textit{object diversity}. Other related branches of research include density estimation, clustering and data summarization~\cite{jin2005scalable}, and ranking~\cite{korba2017learning, korba2018learning}. Some examples of the second point are evolving for content-based retrieval experiments with special datasets,~\eg satellite imagery, aerial photography, or person re-identification. In the former case, the ground truth is not applicable to our field of interest, while in the latter the data is too specific for generalized scene understanding.

\section{Methodology}
\label{ch5:sec:method}

\subsection{Concept of the method}
\begin{figure}
	\centering
	\includegraphics[width=0.9\linewidth]{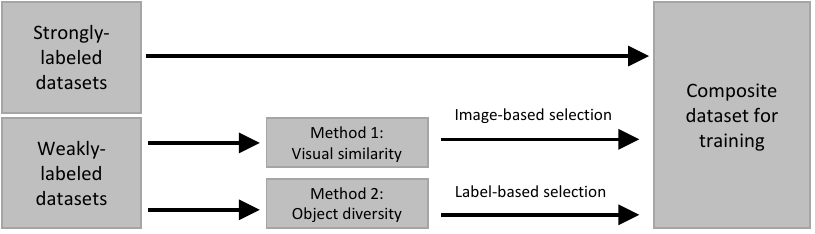}
	\caption{Concept of the proposed method.}
	\label{ch5:fig:concept}
\end{figure}
This section describes in detail the two proposed methods for selecting image-label pairs from bounding-box datasets and how they can be used separately and in combination. The first method aims at \textit{visual similarity} between two datasets and achieves that using only the images from the datasets (without requiring the associated labels). The second method aims at discovering \textit{object diversity} using only the labels from the bounding-box-labeled dataset (without exploiting the images).

The selected informative and diverse image-label pairs from the weakly-labeled datasets are then combined with the strongly-labeled dataset, in the multi-dataset training scheme developed in Chapter~\ref{ch:4-iv2019}. Ideally, \textit{visual similarity} will serve to find from the weakly-labeled datasets similar images to the Cityscapes image domain, while \textit{object diversity} will choose the most informative examples from these datasets. Both criteria indirectly aim at increasing segmentation accuracy through efficient and effective training.

For experimentation, a pixel-labeled dataset (Cityscapes~\cite{Cordts2016Cityscapes}) and two bounding-box-labeled datasets (Cityscapes Coarse~\cite{Cordts2016Cityscapes} and Open Images~\cite{kuznetsova2018open}) are considered (see Section~\ref{ssec:datasets} for more details). To provide an indication for the scale difference between the datasets, Cityscapes has 3,000 images, Cityscapes Coarse has 20,000 images, and Open Images has 1.74 million images. The large difference in up to three orders of magnitude in the number of images has to be mitigated in order to perform efficient combined training.

\subsection{Visual similarity by Gaussian Mixture Model}
\label{ch5:sec:method-gmm}
Inspired by~\cite{birodkar2019semantic, robotka2009image} this method consists of three distinct phases that are described in the following subsections. First, it uses a pretrained convolutional network to extract a low-dimensional representation for each Cityscapes Dense image, then a Gaussian Mixture Model (GMM) is fit to those representations, and finally, a model is used to rank the images of the weakly-labeled datasets. We hypothesize that: 1) images that are \textit{visually similar} to the Cityscapes Dense dataset depicting street scenes, will have a high probability density in the GMM, and 2) images from generic scenes,~\ie the majority of the images in the Open Images dataset not containing street scenes, will have low probability density.

\subsubsection{A. Extracting image representations}
\label{ssec:feat-extract}
In order to capture the distribution of the Cityscapes Dense image domain, image representations are extracted from a fully convolutional network trained for semantic segmentation on the Cityscapes Dense dataset. The first layers of the trained network facilitate the extraction of representations. It is known that initial layers of a neural network maintain information about the input images \cite{tishby2000information}, and thus their domain.

The backbone consists of a ResNet-50~\cite{he2016deep}, which we modify for semantic segmentation, similar to Chapters~\ref{ch:3-iv2018} and~\ref{ch:4-iv2019}. The features are extracted from the penultimate convolutional layer, which have tensor shape with dimensionality $H \times W \times C$ and we denote this sub-network as $\bm{f}$. If $\bm{X}_i$ is the input image, the convolutional representation can be denoted as the set of vectors:
\begin{equation}
	\varPhi_i = \left\{ \bm{f}_{h, w, :} \left( \bm{X}_i \right), ~ h = 1, 2, \dots, H, ~ \text{and} ~ w = 1, 2, \dots, W \right\} ~,
	\label{eq:repr-set}
\end{equation}
where $h, w$ index all the receptive fields of the penultimate layer corresponding to different regions on the input image $\bm{X}_i$. In other words, the output of $\bm{f}$ is sliced across the depth dimension and added to the set $\varPhi_i$ containing $H \times W$ vectors having $C$ elements (features) each.

\subsubsection{B. Modeling image representations}
The following step is to fit a probabilistic model to the low-dimensional representations $\varPhi_i$ for all images $\bm{X}_i$ of the Cityscapes Dense dataset. Such a model would assign a high probability density to the representations of the modeled domain (Cityscapes Dense), and low probability to images outside this domain.

We choose a GMM~\cite{mclachlan2019finite} for its simplicity and its explicitness in statistical modeling importance. Since assigning probability densities to entire Cityscapes images would be too costly, we assume that for every image the set $\varPhi_i$ contains independent and identically distributed representations and we average the set elements $\mathbf{e}$, so that
\begin{equation}
	\xoverline{\bm{\phi}}_i = \frac{1}{|\varPhi_i|} \sum_{\bm{\phi} \in \varPhi_i} \bm{\phi} ~,
	\label{eq:avg-repr-set}
\end{equation}
where the summation is performed element-wise.

Subsequently, the average vector representations $\xoverline{\bm{\phi}}_i$ for all images $i$ are modeled with a GMM. The employed GMM is a mixture of $K$ multi-variate Gaussian distributions, with variable mixture coefficients $\pi_k$, means $\bm{\mu}_k$, and covariance matrices $\bm{\Sigma}_k$ for the Gaussian distributions. These parameters are grouped into $\Psi = \left\{ \pi_1, \dots, \pi_K, \bm{\mu}_1, \dots, \bm{\mu}_K, \bm{\Sigma}_1, \dots, \bm{\Sigma}_K \right\} $. The log-likelihood function for $\Psi$, given the independent average representations $\xoverline{\bm{\phi}}_i$ for all images $N$ of Cityscapes Dense, can be expressed as
\begin{equation}
	\label{eq:log-likelihood}
	\log L\left(\Psi\right) = \sum_{i=1}^{N} \log \left( \sum_{k=1}^K \pi_k \mathcal{N} \left( \xoverline{\bm{\phi}}_i ~ ; ~ \bm{\mu}_k, \bm{\Sigma}_k \right) \right) ~.
\end{equation}
The maximum likelihood estimate $\Psi_\text{citys}$ is found using Eq.~\eqref{eq:log-likelihood} and the Expectation Maximization algorithm~\cite{mclachlan2019finite}.

\subsubsection{C. Image to dataset visual similarity}
From the previous subsection, the derived model from Eq.~\eqref{eq:log-likelihood} is used, which defines the probability density function of the GMM, where $\xoverline{\bm{\phi}}_i$ is modeled by the random variable $\xoverline{\bm{\phi}}$ leading to:
\begin{equation}
p \left(\xoverline{\bm{\phi}} ~; ~\Psi_\text{citys} \right) = \sum_{k=1}^K \pi_k \mathcal{N} \left( \xoverline{\bm{\phi}}; ~\bm{\mu}_k, \bm{\Sigma}_k \right) ~.
\end{equation}
We define a measure of similarity $\text{Sim}$ to the domain that is modeled by the GMM,~\ie Cityscapes Dense, as the maximum logarithm of the pdf of the model for all receptive fields of an image $\bm{X}_i$, which is specified by
\begin{equation}
\text{Sim}_\text{citys}(\bm{X}_i) = \max_{\bm{\phi} \in \varPhi_i} \left( \log p \left(\bm{\phi}(\bm{X}_i) ~; ~\Psi_\text{citys} \right) \right) ~.
\label{eq:sim-measure}
\end{equation}

According to our hypothesis, the larger the similarity $\text{Sim}$ value, the more visually similar the image is to the modeled Cityscapes Dense image domain. In this way, images from a weakly-labeled dataset can be ranked using $\text{Sim}_\text{citys}$ in descending order of similarity and various top portions can be selected for the experiments of Sections~\ref{ssec:perf-open},~\ref{ssec:perf-citys} to ensure the similarity in the data.

\subsection{Object diversity by scoring heuristics}
A training image has high \textit{object diversity} when it contains a large variety and number of objects of interest. Since the method is developed in the context of automated driving, we focus on three important categories of objects, namely traffic objects (traffic signs and traffic lights), vehicles (car, truck, bus, motorcycle, bicycle, train), and humans (pedestrian, rider). We assign to each category a score of importance. These scores are defined by empirical tests and manual inspection of the images, and they depend on each dataset. Because of the applied heuristic arguments for assigning scores, we will briefly give some rules on how we assigned the empirical scores below. Using these score heuristics and the bounding-box labels of an image, an aggregate score of importance per image-label pair is calculated by adding all scores. This approach deliberately favors images with many objects as they will contain more pixels with useful labels. Depending on the number of objects, the aggregate score will differ per image-label pair, which induces a ranking. It should be noted that the algorithm for object diversity can be applied independently of the GMM selection.

The following paragraph applies to the heuristics for scoring. In the Open Images dataset, the above categories are labeled in an instance-wise manner. We assign 100 points per traffic object, 10 points per vehicle, and 1 point per human to each instance. The general intuition behind assigning higher scores to traffic objects is that they are more likely to appear in street scenes only, while vehicles and humans can appear in various other scenes. For Cityscapes, traffic objects are not labeled instance-wise, so we assign scores for two categories only,~\ie 10 points per vehicle, and 1 point per human. The factor of 10 points difference between vehicle and human scores is explained by inspection of both datasets and the amount of occurrences for each of those classes,~\eg an image with 1 vehicle is preferred over an image with 9 people, as these people can be in indoor scenes. For each image, the total score from all labeled objects is accumulated. The images are ranked according to their score, and different top portions of the ranking are selected for the experiments of Sections~\ref{ssec:perf-open},~\ref{ssec:perf-citys}.

\subsection{Combination of the two selection methods}
\label{ch5:ssec:combine-sel-methods}
In the previous two sections, we have described the two selection schemes and how they result in two rankings of the images of a dataset. In general, the two rankings can have a different ordering, thus aggregating them into one collection is not a trivial task. Since \textit{visual similarity} and \textit{object diversity} have an equal preference, we opt for interleaving the rankings by interchangeably choosing images from the initial rankings to the final selection. In the process, if an image is already inserted in the final selection, it is not inserted twice.

\section{Datasets and implementation details}
\label{ch5:sec:details}
This section discusses the chosen convolutional model for simultaneous training on datasets with strong and weak supervision, and presents the employed datasets. All hyper-parameters for training are provided to enable reproducibility of the experiments.

\subsection{Datasets}
\label{ssec:datasets}
\textbf{Cityscapes Dense}:
The Cityscapes dataset \cite{Cordts2016Cityscapes} contains street-scene images from German cities, captured by a 2-Mpixel camera mounted on a car. We have used the training subset with 2,975 densely pixel-labeled images and the larger subset of 20,000 coarsely pixel-labeled images.

\noindent\textbf{Open Images v4}: This dataset \cite{kuznetsova2018open} contains 9 million images from everyday, rich scenes collected from the internet with multiple resolutions, shooting angles, containing several objects that are not relevant for automated driving. The official subset labeled with 14.6 million bounding boxes includes 1.74 million images.

\noindent\textbf{Cityscapes Coarse bboxes}:
This is a dataset with bounding boxes that was created for the purpose of this work from the coarse, per-pixel, instance labels of the Cityscapes Coarse subset. Specifically, for each labeled instance in an image, we define a bounding box using the minimum and maximum coordinates of per-pixel labels in each axis.

\subsection{FCN model for training on strong and weak supervision}
We use the model of Chapter~\ref{ch:4-iv2019} with minor hyper-parameter differences (batch size, weight decays), in order to make all experiments in this chapter consistent. Because of these differences in hyper-parameter settings, the results of this chapter with the same network of the previous chapter are not exactly equal anymore. The Cityscapes results in the final outcome tables of the previous chapter, therefore show 1-2\% difference in scores.

Here we provide a recapitulation of the network design from Chapter~\ref{ch:4-iv2019}. The network consists of a conventional ResNet-50 feature extraction that is modified to have semantic segmentation output with dilated convolutions and an upsampling module. Moreover, instead of one per-pixel classifier, it consists of a hierarchy of classifiers, arranged in a tree structure according to the class hierarchy. Each of the classifiers is fed with the same convolutional features of the feature extractor. During inference, the results from all classifiers are aggregated in a per-pixel manner to output the final decisions.

\subsection{Training details}
For fair comparisons, all networks in Section~\ref{ch5:sec:experiments} are trained with the same hyper-parameters and for the same number of epochs. For the image representation extraction of Section~\ref{ch5:sec:method-gmm}, the input image dimensions of $1024 \times 2048$ pixels are used, which are reduced to a grid of $256 \times 512$ receptive fields, each observing an area of approximately $200 \times 200$ pixels overlaying on the full image. The depth of feature representation is 256.

Using the above settings, Cityscapes images have in total $256 \cdot 512 \cdot 2975$ representations of 256 dimensions, leading to approximately $390 \cdot 10^6$ receptive fields. 
In training the GMM, we sample only 24 thousand representations from all the images in the Cityscapes Dense training set. For the GMM model, the parameters of the mixtures are found using Expectation Maximization. Updates are continued until the likelihood does not change from one E step to another by more than 0.001 nat.

\section{Experiments}
\label{ch5:sec:experiments}
First, an overall impression of the results are given in Section~\ref{ssec:perf-over} for the two selection methods applied on two diverse datasets. This overall presentation is given first because it creates an overview without being buried in large sets of numbers and enables cross-experiment comparisons. Afterwards, results are further analyzed in detail in Sections~\ref{ssec:perf-citys},~\ref{ssec:perf-open}. Section~\ref{ssec:analysis} contains ablation experiments for the parameters of the models.

All results refer to training a hierarchical segmentation network, as described in Section~\ref{ch5:sec:details}, with a combination of Cityscapes Dense (pixel labels) as the first set, and either Cityscapes Coarse or Open Images (bounding box labels) as the second set. The proposed data selection methods are evaluated on the pixel-labeled Cityscapes validation set, unless otherwise noted. We use the Intersection over Union metric~\cite{long2015fully}. The IoU results are averaged over the last 5 (Cityscapes) epochs when the model converges, since the variance is high. The mIoU results are the mean IoU over the classes that receive extra supervision from the weakly-labeled dataset.
\begin{figure}
	\centering
	\includegraphics[width=0.8\linewidth]{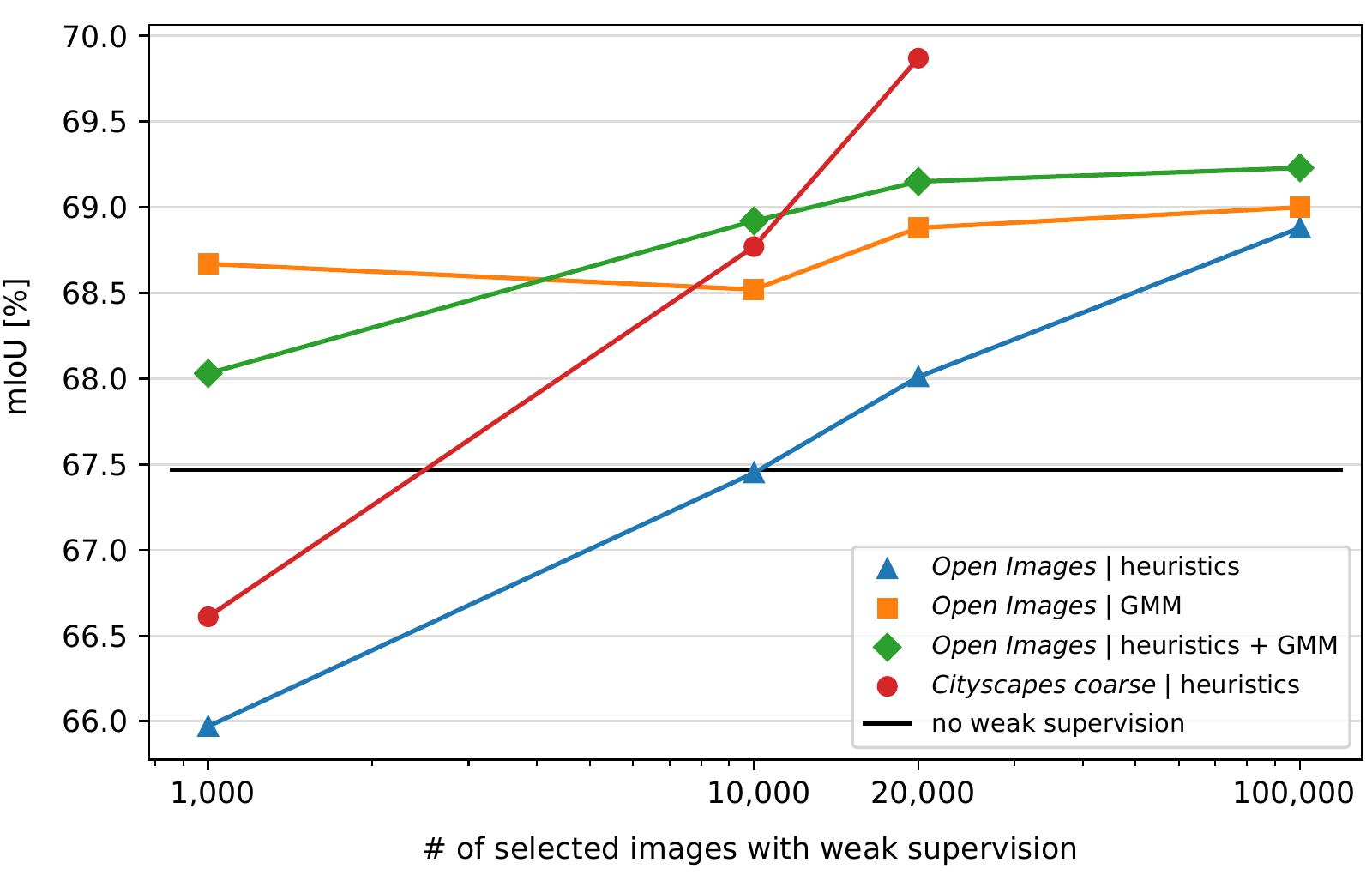}
	\caption{Performance (mIoU) on Cityscapes validation set. Each point from the same curve corresponds to a single training of the system for a different number of selected image-label pairs from the weakly-labeled datasets (Cityscapes Coarse and Open Images). Exception to this is the black horizontal line, which denotes the mIoU of training only on Cityscapes Dense.}
	\label{fig:perf-overall}
\end{figure}

\subsection{Overall presentation of results}
\label{ssec:perf-over}
In Figure~\ref{fig:perf-overall}, the mIoU performances on the Cityscapes validation set for various combinations of our selection methods and datasets are shown. We experiment with two different datasets, which contain data that are considered to be from different domains. Cityscapes Coarse is a subset of Cityscapes and as such contains images of street scenes all captured with the same system, from a specific point of view and in the same cities. Open Images is a generic scenes dataset collected from various image sources and points of view, where street scenes are rare. The experiments demonstrate that data selection from both datasets is beneficial for improving performance, while reducing the amount of required data.

From Figure~\ref{fig:perf-overall} we observe that class-scoring heuristics is advantageous for both datasets. As anticipated, in the case of Cityscapes Coarse, the mIoU for the same amount of selected data is always higher than with Open Images, since the Cityscapes Coarse dataset has \textit{visually similar} images to the Cityscapes validation set. The GMM selection demonstrates high performance using as little as 1,000 selected images, while the increase is only marginal when more images are selected.

Interestingly, when the selection quantity is limited to 1,000 images, the class-scoring heuristics perform even lower than the baseline (see red and blue curve at the left bottom in Figure~\ref{fig:perf-overall}). This outcome is not intuitive because theoretically speaking, by adding extra information to the system, the accuracy should not drop. A possible explanation relates this finding to overfitting, due to the small number of images that are trained with over multiple iterations. Moreover, the decision to maintain the same hyper-parameters (number of epochs, etc.) for all experiments is expected to give a negative impact on segmentation performance.

Overall, the selection methods appear to improve the quality of weak supervision to be used together with strong supervision in a simultaneous training scheme. The combination of the two methods yields better results in all cases where sufficient weakly-labeled images (\eg above 10,000 images) are available. The GMM selection is more effective when less data are available, while the class-scoring selection requires enough data in order to be successful. Finally, it is worth noting that the GMM selection attains almost the same performance as the class-scoring heuristics but requires up to 100 times fewer images.
\begin{table}
	\centering
	\small
	\begin{tabular}{l|cccc}
		\toprule
		& \multicolumn{4}{c}{\# of selected images}\\
		Method of selection & 1k (0.1\%) & 10k (1\%) & 20k (2\%) & 100k (10\%)\\
		\midrule
		Random (baseline) & 67.05 & 67.68 & 68.51 & 67.88\\
		\midrule
		Heuristics & 65.97 & 67.45 & 68.01 & 68.88\\
		GMM & 68.67 & 68.52 & 68.88 & 69.00\\
		Heuristics + GMM & 68.03 & 68.92 & 69.15 & 69.23\\
		\bottomrule
	\end{tabular}
	\caption{Performance (mIoU [\%]) on Cityscapes classes (8 traffic sign participants) that receive extra supervision from Open Images for various selection methods (no selection is equivalent to random). Each row corresponds to a network trained on all pixel-labeled Cityscapes images and the top n\% images from the bounding-box-labeled Open Images dataset.}
	\label{tab:perf-selec-from-open}
\end{table}
\begin{table}
	\centering
	\small
	\begin{tabular}{l|cccccccc}
		\toprule
		\# of images & \rotseventy{Car} & \rotseventy{Truck} & \rotseventy{Bus} & \rotseventy{Train} & \rotseventy{Motorcycle} & \rotseventy{Bicycle} & \rotseventy{Person} & \rotseventy{Rider}\\
		\midrule
		1k & 92.2 & 68.2 & 76.9 & \textbf{71.2} & 50.7 & 67.5 & 71.7 & 51.0\\
		10k & 92.2 & 69.7 & 79.3 & 65.6 & 48.2 & 67.7 & 71.6 & 51.2\\
		20k & \textbf{92.5} & \textbf{73.1} & \textbf{79.9} & 60.9 & \textbf{53.3} & \textbf{67.7} & 71.8 & 52.0\\
		100k & 92.4 & 69.6 & 78.8 & 67.8 & 51.1 & \textbf{68.0} & \textbf{71.9} & \textbf{52.6}\\
		\bottomrule
	\end{tabular}
	\caption{Detailed per class IoU [\%] for the combined Heuristics + GMM selection method using weakly-labeled data from Open Images. The 8 traffic participant classes out of 19 Cityscapes classes that receive extra supervision are listed here.}
	\label{tab:perf-detail}
\end{table}

\subsection{Detailed results for Cityscapes Dense and Open Images}
\label{ssec:perf-open}
The Open Images dataset~\cite{kuznetsova2018open} contains images from a variety of generic natural scenes, where the number of street-scene images is minimal, thus the \textit{domain gap} with Cityscapes is large. Open Images is labeled with 600 semantic classes, the majority of which are not relevant classes for urban street scenes.

Table~\ref{tab:perf-selec-from-open} shows the detailed mIoU performance on Cityscapes Dense for different number of selected image-label pairs. The hierarchical model is trained on a combination of pixel-labeled images from Cityscapes Dense and bounding-box-labeled images from Open Images. In the first row, the mIoU for random selection is shown, which represents a strong baseline. In the second and third row, the proposed techniques of Section~\ref{ch5:sec:method} are studied. It is observed that selection through GMM has a higher gain in a small amount of weakly-labeled images, while selection with scoring heuristics is better when using more than 20,000 weakly-labeled images.

The next step is to investigate the option of combining both selection methods, so high \textit{visual similarity} and \textit{object diversity} are simultaneously attained. The final collection of images is obtained by selecting the same amount from each of the two rankings, so that each method contributes half of the selected images after removing duplicate images. Interestingly, the two selection methods have dissimilar rankings, as can be seen from the analysis in the upcoming Section~\ref{ssec:common-images}.

As a conclusion, we conjecture that for a different number of available images with weak labels, a different selection method is more suitable. Knowing that Cityscapes Dense has 2,975 training images, if only 1,000 weakly-labeled images are available, then selecting similarity (GMM) over diversity (heuristics) works better, and the model does not overfit. In the case where weakly-labeled images are 100 times more available, then opting for \textit{object diversity} gives better results.

Table~\ref{tab:perf-detail} shows the IoU scores detailed per class for the GMM selection method. Three classes (car, bicycle, and person) have little gain in performance by exploiting the increased number of selected images. Four classes (truck, bus, motorcycle, and rider) have a significant gain in performance. A potential underlying reason for this behavior may be the different viewpoints used to depict these classes in the images of Cityscapes and Open Images.

An interesting case for analysis appears for the \textit{train} semantic class. It can be observed that the more images are included (up to 20k images), the more the IoU drops, and it rises back to a satisfactory level only when using 100k images. This clearly signifies that although the images including trains may appear \textit{visually similar} as a whole, the trains between Cityscapes and Open Images have different appearance, as discovered after manual inspection of the images. This enforces the need to investigate \textit{visual similarity} per class rather than per image. Further analysis is left for future research.

\subsection{Detailed results for Cityscapes Dense and Cityscapes Coarse}
\label{ssec:perf-citys}
\begin{table}
	\centering
	\small
	\begin{tabular}{l|ccc|c}
		\toprule
		& \multicolumn{4}{c}{\# of selected images}\\
		Method of selection & 1k (5\%) & 5k (25\%) & 10k (50\%) & 20k (100\%)\\
		\midrule
		Random & 66.68 & 66.82 & 69.38 & 69.87\\
		\midrule
		Heuristics & 66.61 & 67.58 & 68.77 & 69.87\\
		Heuristics + GMM & 68.37 & 68.29 & 67.41 & 69.87\\
		\bottomrule
	\end{tabular}
	\caption{Performance (mIoU [\%]) on Cityscapes classes (10 traffic participant classes) that receive extra supervision from Cityscapes Coarse for various selection methods. The network is trained on all pixel-labeled Cityscapes images and the top n\% images from the bounding-box-labeled Cityscapes Coarse.}
	\label{tab:perf-from-citys}
\end{table}
Cityscapes Coarse is a subset of Cityscapes, and thus is \textit{visually similar} to Cityscapes Dense by definition. Through this experiment, the selection method aiming for \textit{object diversity} is examined in isolation. However, for completeness, we present also results from combining both selection methods. Table~\ref{tab:perf-from-citys} illustrates that the proposed methods are useful when using few images from the weakly-labeled dataset, but there is no substantial benefit when using more than 10k images. The significant performance drop for the third column with 10k images is expected and is due to our chosen scheme of scoring heuristics.
Moreover, as can be seen from the last row of Table~\ref{tab:perf-from-citys}, the GMM selection does not add much, since \textit{visual similarity} is already attained by using the same dataset.

\subsection{Analysis and ablation experiments}
\label{ssec:analysis}
This section presents the results of the ablation experiments and the findings on using the GMM for characterizing the image domain of Cityscapes.

\subsubsection{A. Dataset characterization and visual similarity}
\label{ssec:dom-model}
\begin{figure}
	\centering
	\includegraphics[width=0.6\linewidth]{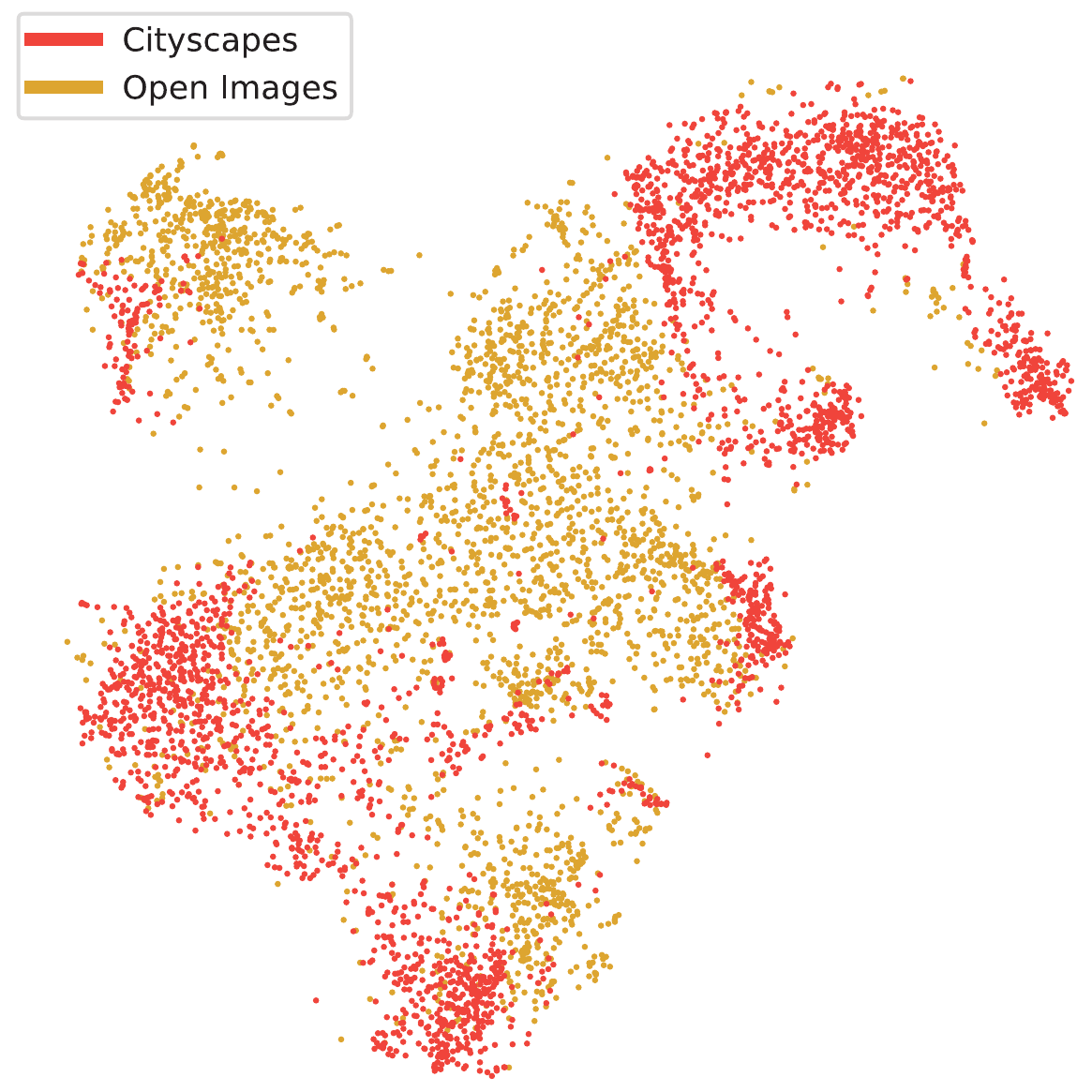}
	\caption{t-SNE plot of the image representations for a random sample from Cityscapes Dense and a random sample from Open Images.}
	\label{fig:tsne-citys-open}
\end{figure}
Figure \ref{fig:tsne-citys-open} visualizes the t-SNE embeddings~\cite{maaten2008visualizing} of the 256-dimensional image representations $\bar{\Phi}$ for image subsets from the two datasets. For both datasets, we randomly sampled 3,000 representations using our model on the respective training sets. It can be readily observed that the distributions of representations have minimal overlap. This separation explains why the GMM can fit the representations from Cityscapes so well and single out representations from Open Images that are dissimilar.

\begin{figure}
	\centering
	\includegraphics[width=0.7\linewidth, trim={0.5cm 12.5cm 0.5cm 4.0cm}, clip]{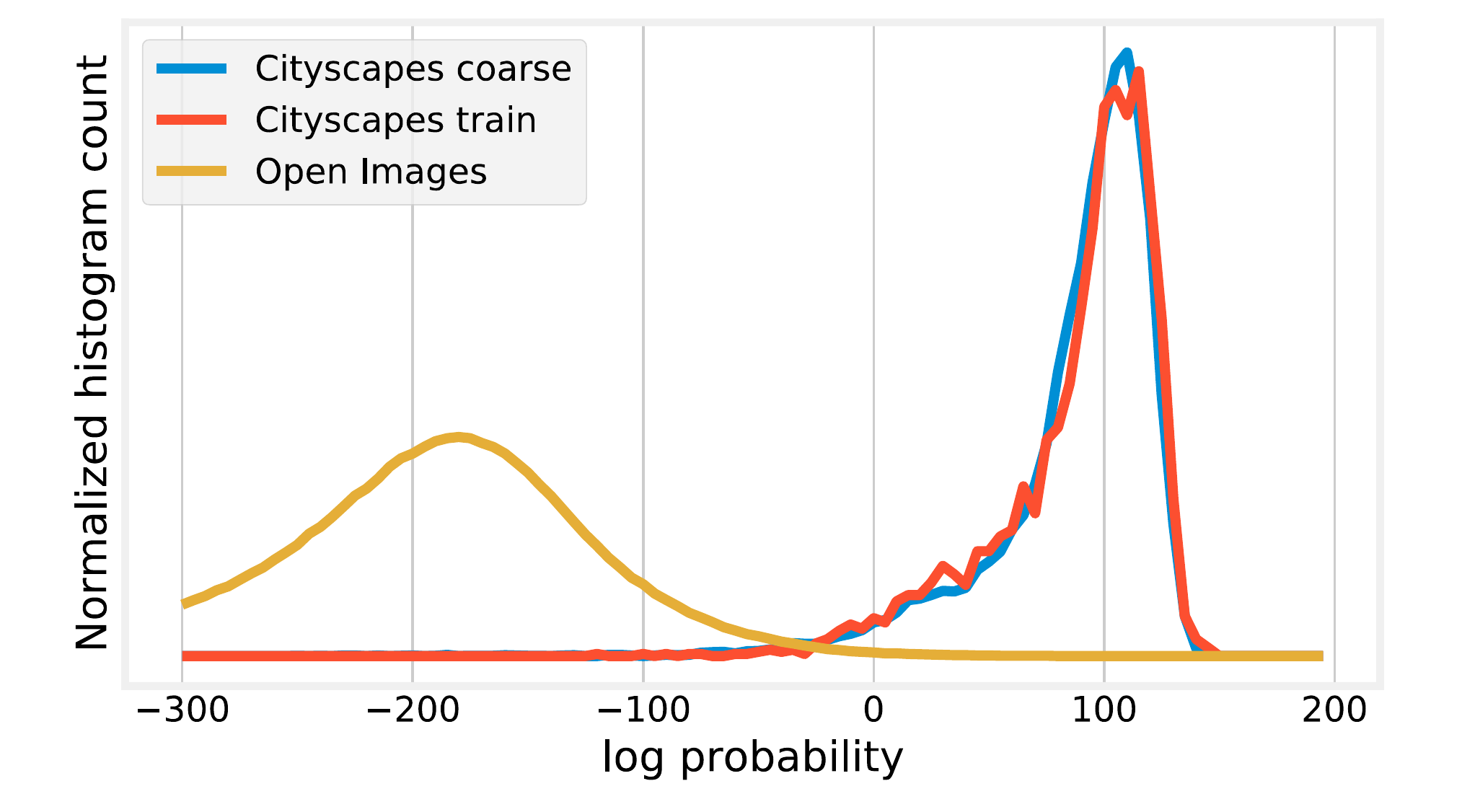}
	\caption{Empirical histogram of the log-probabilities for the three datasets. For each image in the training set, the max log-probability is calculated and incorporated in the histogram.}
	\label{fig:stats-logprobs}
\end{figure}

Figure~\ref{fig:stats-logprobs} illustrate the statistics of the \textit{visual similarity} measure defined in Section~\ref{ch5:sec:method},~\ie the max log-probability of the GMM, for all images of the three used datasets. From Figure~\ref{fig:stats-logprobs} it can be seen that the histogram of the max log-probability for the Cityscapes Dense and Coarse image subsets are very similar and confirms their common origin as subsets of Cityscapes. In contrast, the histogram of Open Images is more spread out and has a very small overlap with Cityscapes Dense. This spread of the distribution shows that the scene variety is high, and only a small subset is \textit{visually similar} to Cityscapes. The difference validates our hypothesis that the images from Open Images are generated with a different distribution.

\subsubsection{B. Number of GMM components}
In this ablation experiment, we investigate the optimal number $ K $ of GMM components for modeling the Open Images domain, such that the result of the GMM model outperforms the metric performance on Cityscapes validation. As can be seen from Table~\ref{tab:gmm-comp}, $ K = 5 $ gives higher mIoU results. The selection of this parameter follows the intuition that representations have a simple and compact structure, as indicated by the t-SNE plot of Figure~\ref{fig:tsne-citys-open}, and is guided by the Bayesian Information Criterion (BIC)\footnote{This criterion is computed by $\text{BIC} = \log(N_\text{samples}) \cdot N_\text{free\_parameters} - 2 \cdot \log L(\Psi) = \log(N_\text{samples}) \cdot K \cdot (256+256^2) - 2 \cdot \log L(\Psi)$}.
\begin{table}
	\centering
	\small
		\begin{tabular}{c|c|c|c}
			\toprule
			K components & 5 & 20 & 50\\
			\midrule
			BIC ($\cdot 10^6$) $\downarrow$ & 3.2 & 50.7 & 124.7\\
			mIoU $\uparrow$ & \textbf{68.67} & 65.86 & 64.18\\
			\bottomrule
		\end{tabular}
	\caption{Ablation on the number of components of GMM for the mIoU performance using the Open Images as the weak dataset.}
	\label{tab:gmm-comp}
\end{table}

\subsubsection{C. Common images in rankings}
\label{ssec:common-images}
When both rankings from different selections methods are used (experiment in Table~\ref{tab:perf-selec-from-open}), a conflict of ranking position arises (same image re-appearing in the same top-N list) as explained and solved in Section~\ref{ch5:ssec:combine-sel-methods}. Here we compute the common selections of the two ranking approaches. From Table~\ref{tab:common} it can be seen that the two selection methods have different preferences and also that \textit{visual similarity} does not induce \textit{object diversity} and vice versa. Specifically, it is interesting that in 1,000 selected images, only 0.3\% are present in the top 500 from both methods. This observation allows to use the selection methods and hence the data for different purposes.
\begin{table}
	\centering
	\small
		\begin{tabular}{l|cccccc}
			\toprule
			& \multicolumn{6}{c}{Top count of selected images}\\
			& 1k & 10k & 20k & 50k & 100k & 200k\\
			\midrule
			\# of common images & 3 & 117 & 385 & 1,492 & 4,303 & 10,704\\
			Percentage (common) & 0.3\% & 1.17\% & 1.93\% & 2.98\% & 4.3\% & 5.35\%\\
			\bottomrule
		\end{tabular}
\caption{Common images selected by two selection methods over different top-N options.}
\label{tab:common}
\end{table}

\section{Conclusion}
\label{ch5:sec:conclusion}
This chapter and Chapter~\ref{ch:4-iv2019} propose a single complete method for efficient training of the hierarchical FCN for semantic segmentation on unbalanced, mixed supervision datasets. The challenges of training efficiency and information scarcity (see Section~\ref{ch5:sec:intro}) that have emerged in the previous chapter are addressed in this chapter by developing two methodologies for data selection. The objective of the two methodologies is to select a reduced number of informative and diverse examples from large weakly-labeled datasets to accompany and enhance simultaneous training with smaller strongly-labeled datasets.

We have presented two methods for data selection aiming at \textit{visual similarity} and \textit{object diversity} for the problem of semantic segmentation. Both methods have been evaluated by employing an FCN with hierarchical classifiers developed in the previous chapter. On the one hand, visual similarity selection aims at finding image-label pairs from weakly-labeled datasets and functions well when only a small number of examples should be selected. On the other hand, object diversity selection aims at finding image-label pairs from weakly-labeled datasets and is preferable when a larger number of examples should be selected. The combination of both selection methods offers the best overall performance.

The selection methods have proven particularly useful for selecting images from the weakly-labeled datasets and dramatically decreased the number of required training images,~\ie 20 times for Cityscapes and 100 times for Open Images. As a bonus of this research, we have presented results for characterizing the visual domain of a dataset by Gaussian mixture modeling the representations of its images. To summarize the contributions, this chapter has:
\begin{itemize}[noitemsep,topsep=0pt]
	\item Proposed a selection method based on modeling image representations with a GMM, for finding \textit{visually similar} images to a given dataset;
	\item Proposed a selection method based on class-scoring heuristics, for finding rich-labeled images;
	\item Applied methods independently and jointly in weak supervision selection for semantic segmentation to reduce the number of required training examples while increasing performance;
	\item Offered a more generalized characterization of the image domain of a dataset through GMM modeling.
\end{itemize}

~

The following chapter reconsiders the problems that were studied in Chapters~\ref{ch:3-iv2018} and~\ref{ch:4-iv2019} and collectively re-addresses the involved challenges. This yields a result that re-formulates a generalized semantic segmentation task and proposes an all-encompassing solution, which also incorporates the developed tools of this chapter.

\begin{savequote}[8cm]
\end{savequote}

\chapter{Training semantic segmentation on heterogeneous datasets}
\label{ch:6-journal}

\section{Introduction}
\freefootnote{\hspace*{-15pt}The contributions of this chapter have been submitted (under review) as a journal paper at the IEEE Transactions on Neural Networks and Learning Systems in 2021.}
Chapter~\ref{ch:5-itsc2019-wacv2019} has investigated how multi-dataset semantic segmentation can be optimized with respect to the data resources required during training. The proposed combination of two selection methods dramatically reduces the number of examples needed during simultaneous multi-dataset training. The method is based on selecting diverse, but visually similar images between the datasets, while at the same time increasing the segmentation performance in the majority of cases, because the composition of the data into a joint dataset facilitates better learning.

This chapter reconsiders the problem of training networks with heterogeneous datasets for semantic segmentation and proposes a framework, namely Heterogeneous Training of Semantic Segmentation, which generalizes the methodologies introduced in Chapters~\ref{ch:3-iv2018} and~\ref{ch:4-iv2019}. Moreover, the findings of the previous chapter are also employed for improving training efficiency.

Semantic Segmentation~\cite{guo2017aro,garcia2018survey,minaee2020image} is a indispensable building block of visual analysis systems for various application domains, such as automated driving~\cite{zhu2017perception,janai2020computer}, biomedical image analysis~\cite{taghanaki2020deep}, virtual/augmented reality, and surveillance~\cite{minaee2020image}. In street-scene understanding, semantic segmentation is the first step of scene analysis and provides the necessary platform towards higher level reasoning and planning. The segmentation task is part of the bigger family of image recognition tasks, which include, among others, image classification and object detection. The early success of supervised Convolutional Networks (CNNs) in image classification~\cite{krizhevsky2012alexnet,szegedy2015going,he2016deep} defines such networks as the \textit{de-facto} solution for related image recognition tasks, which is in part attributed to the successful exploitation of very large datasets. Unlike CNNs trained for classification or detection, where data collection is easier, CNNs for semantic segmentation face the following two fundamental challenges.

\begin{figure}
	\centering
	\includegraphics[width=\textwidth]{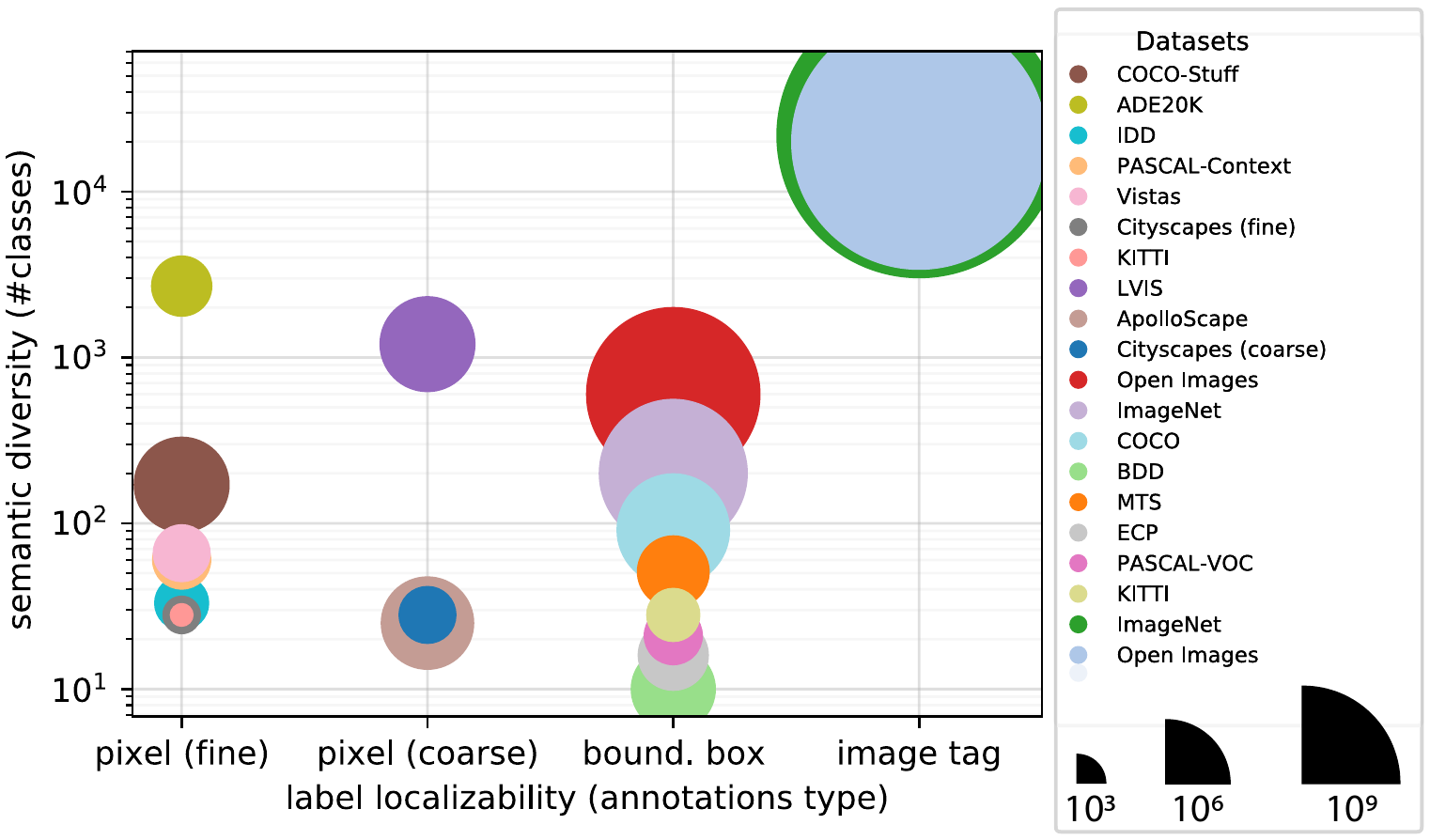}
	\caption{Comparison of various image understanding datasets with respect to their: i) annotation type, ii) number of semantic classes, and iii) number of images (visualized by the squared radius of the circles). Networks with segmentation output typically use a single dataset, in a fully-supervised (pixel-labeled) fashion or with a weakly-supervised (\mbox{bbox-/tag-labeled}) method.}
	\label{fig:datasets-statistics}
\end{figure}

\textit{Limited size of existing datasets}. The lack of rich annotated datasets causes CNNs for semantic segmentation to exhibit limited performance when evaluated on seen datasets for training and poor generalization capabilities on unseen datasets for testing. This challenge becomes particularly acute in data-scarce areas, such as street-scene understanding. Compared to the field of street surveillance for safety, where the focus has been on the behavior of moving objects like persons and cars, the area of street-scene semantic segmentation is still at an early stage of development, considering the fact that the first datasets are only 5 years old. Existing datasets for semantic segmentation~\cite{cordts2016Cityscapes,neuhold2017mapillary,varma2019idd} contain typically 100 to 1000 times less images than datasets for image classification and object detection. The main reason for the differences is the level of fine-grained detail in annotations. For example, COCO creators~\cite{lin2014microsoft} report that annotating pixels for semantic segmentation is 15 times slower than drawing bounding boxes, while according to~\cite{bearman2016s} annotating pixels is 78 times slower than choosing image-level tags. This difference in dataset sizes is illustrated in Figure~\ref{fig:datasets-statistics}, where some popular datasets for computer vision tasks are included and compared with respect to size and detail.

\textit{Low diversity of represented semantic concepts}. CNNs trained on semantic segmentation datasets can recognize typically a few dozens of scene concepts, while fine-grained semantic classes are rare in these datasets. The complexity of (manual) per-pixel labeling practically constrains the annotated semantic classes to represent an order of 100 different scene concepts, while datasets with less detailed spatial annotations (bounding boxes or tags)~\cite{gupta2019lvis,deng2009imagenet,kuznetsova2020open} can reach up to 1,000 - 10,000 unique semantic classes (see vertical axis of Figure~\ref{fig:datasets-statistics}).

The natural way to address the aforementioned two challenges is to annotate more images at the pixel level, or refine existing annotations with finer-grained \mbox{(sub-)}classes through manual or semi-automated means. This is a straightforward yet costly approach, since manual labeling is laborious and semi-automated procedures result in insufficient quality of annotations when not complemented with human quality control. An alternative approach is to merge existing datasets and train a CNN with the combined dataset, in order to amend the aforementioned challenges. However, combining datasets is not an easy task, due to their structural differences.

In this chapter, we investigate the involved problems when combining multiple datasets and present advancements in three directions: \textit{performance} on training of seen datasets, \textit{generalizability} on testing of unseen datasets, and \textit{knowledgeability},~\ie, the number of recognizable semantic concepts. Instead of combining only datasets for semantic segmentation,~\ie pixel-labeled datasets, a larger candidate pool of heterogeneous datasets is admitted to be combined from various image understanding tasks (Figure~\ref{fig:datasets-statistics}). The aim for this chapter leads to the following detailed problems.
\begin{itemize}[noitemsep,topsep=0pt]
\item \textit{Heterogeneous annotation formats}: Image understanding datasets contain a variety of label formats and the majority of them are not compatible with fully convolutional training losses, thus using information from these annotations is not directly possible.
\item \textit{Heterogeneous label spaces}: Datasets often contain disjoint or conflicting semantic label spaces, thus simultaneous training on more than one space is not feasible without \textit{a-priori} proper handling of the conflicts.
\item \textit{Training resources}: Simultaneous training with a large pool of heterogeneous datasets can dramatically increase the memory requirements and the computation needs up to a level where execution becomes impractical.
\end{itemize}

Datasets created for different tasks or domains are not homogeneous,~\eg can contain conflicting label spaces or incompatible annotation types, as referred to in the first two problems. As a consequence, a collection of these datasets is then infeasible and cannot be incorporated in a Fully Convolutional Network (FCN)~\cite{long2015fully} training pipeline, which is the established approach for semantic segmentation using CNNs. To advance the state of the art in multi-dataset training, we generalize semantic segmentation over heterogeneous datasets and analyze the challenges. Subsequently, we propose a unified methodology that de-couples dataset specifics (structure, annotation types, label spaces) from the task formulation of semantic segmentation. In this way, a plethora of existing image understanding datasets can be leveraged within the same consistent and robust FCN-based framework. From now on, this framework will be called Heterogeneous Training of Semantic Segmentation (HToSS).

This chapter is organized as follows. Section~\ref{sec:rel-work} reviews related approaches and similar tasks. In Section~\ref{sec:challenges}, a definition of the exact problem statement is given. The details of our approach are described in Section~\ref{sec:methodology} and extensive experimentation is provided in Section~\ref{sec:experimentation}. Finally, Section~\ref{ch5:sec:conclusion} summarizes the findings for handling heterogeneous datasets and concludes the chapter.

\section{Related work}
\label{sec:rel-work}
Multi-dataset training is gaining traction in various areas,~\eg in object detection~\cite{zhou2021simple,zhao2020object}, depth estimation~\cite{ranftl2020depth}, and domain adaptation~\cite{zhao2020multisource,sun2015survey}, since it improves network robustness and generalization capabilities. This work focuses on semantic segmentation and relaxation of the requirements that a dataset has to comply to, in order to be suited for multi-dataset training. The proposed work generalizes related literature in semantic segmentation~\cite{lambert2020mseg,jain2020scaling}  and complements emerging recent work~\cite{bevandic2020multi}.

\subsection{Multi-dataset semantic segmentation}
\label{subsec:cross-dataset-ss}
The majority of previous works focus on using multiple datasets with possibly different label spaces, but a single type of supervision,~\ie pixel-level labels. Most of the works solve the challenges that arise from conflicts in label semantics through dataset-based solutions~\cite{ros2016training,lambert2020mseg}, architecture-based solutions~\cite{liang2018dynamic,meletis2018heterogeneous,kalluri2019universal,leonardi2019training,fang2020multi,sun2020real}, or loss-based solutions~\cite{kong2019training,meletis2018heterogeneous,fang2020multi}. In these works, all label spaces of the employed datasets are combined into a common taxonomy, by merging, splitting, or ignoring semantic concepts or by manual re-labeling when needed. Early works extend the conventional FCN architecture with multiple heads/decoders up to one for each dataset~\cite{leonardi2019training}, or multiple (hierarchical) classifiers~\cite{meletis2018heterogeneous}, thereby effectively approaching the problem from the multi-task learning perspective. The authors of \cite{ros2016training} combine 6 datasets, while in \cite{lambert2020mseg} 13 datasets are combined to create a large-scale training and testing platform.

Contrary to existing works, the proposed HToSS framework does not require any image relabeling or ignoring classes to simultaneously train an FCN with multiple datasets, and solves any label space conflict at the stage of loss calculation.

\subsection{Semantic Segmentation with weak supervision}
Semantic segmentation is by definition a pixel-based task and it is conventionally realized by training a CNN with per-pixel homogeneous supervision. Previous works have used a diverse set of less detailed (weak) heterogeneous supervision, either to accommodate strong supervision, or independently in a semi/weakly-supervised setting~\cite{zou2020pseudoseg,ibrahim2020semi,meletis2019boosting,papandreou2015weakly,kalluri2019universal}. Several methods generate candidate masks from bounding-box supervision~\cite{zhu2014learning,meletis2019boosting,dai2015boxsup,ibrahim2020semi,meletis2018heterogeneous} using external modules, internal network predictions, or heuristics to refine weak annotations. These masks are used to train networks alone or together with strong supervision. Even weaker forms of supervision has been employed and examples of this include point-level \cite{bearman2016s} and image-level \cite{wang2020weakly,pathak2015constrained,meng2019weakly} annotations, mainly within a multiple instance learning formulation. Finally, methods that use a combination of multiple weaker types of supervision have been proposed, such as bounding boxes and image-level tags \cite{ye2018learning,meletis2019data,papandreou2015weakly,li2019weaklier,li2018weakly}.

Inspired by earlier works, the proposed framework achieves pixel-accurate training using weak supervision by a pre-processing step that generates pseudo-labels and a refinement process during training. Moreover, unlike previous methods, the HToSS framework treats all types of weakly-labeled datasets uniformly and uses them in combination with strongly-labeled datasets.

\subsection{Other related tasks}

Two related semantic segmentation tasks that encapsulate multiple datasets in their formulation are transfer learning~\cite{weiss2016survey} and domain adaptation~\cite{sun2015survey,zhao2020multisource}. These tasks aim at transferring rich knowledge from a source dataset/domain to a target dataset/domain, where knowledge is scarce or even non-existing. They mainly concentrate on the performance in the target domain, which may be available during training, in some limited form. Recently, variations of these tasks also track performance in the source domain and investigate multiple-source versions of the problems~\cite{he2021multi,piva2021exploiting}. The HToSS formulation considers performance on all employed datasets and during training it does not depend on information from the testing datasets explicitly, as in domain adaptation.

The following four tasks are briefly addressed for their relevance with aspects of HToSS. First, multi-dataset semantic segmentation has been addressed in literature using \textit{multi-task learning} ~\cite{ranftl2020depth,nekrasov2019real,kokkinos2017ubernet,vandenhende2021multi,crawshaw2020multitask}, where a network head/branch is devoted to each dataset independently,~\ie segmentation for each dataset is modeled as a separate ``task''. Multi-task learning is briefly addressed here because it can be defined as a baseline, where simply a network is devoted to each independent dataset. Second, the knowledgeability perspective of networks has been studied in a \textit{continual learning} setting, where new classes are discovered or added during training or inference~\cite{nakajima2019incremental,klingner2020class} and old data may not be available. Third, the absence of labels during training for some datasets/domains has been addressed by \textit{self-training} or \textit{pseudo-label} approaches~\cite{saporta2020esl,sanberg2017free,zoph2020rethinking,zou2020pseudoseg,zhan2018mix} have addressed the absence of labels during training for some datasets/domains. Finally, the problem of conflicting label spaces has been solved also through the prism of \textit{learning with partial labels}~\cite{zhang2017disambiguation,cid2012proper}. The partial-label formulation associates a training sample with a set of candidate labels among which at most one is the correct label.

This plethora of research shows that training with multiple and heterogeneous datasets as done in the HToSS approach is a desired capability of modern training pipelines.

\begin{figure}
	\centering
	\includegraphics[width=1.0\linewidth]{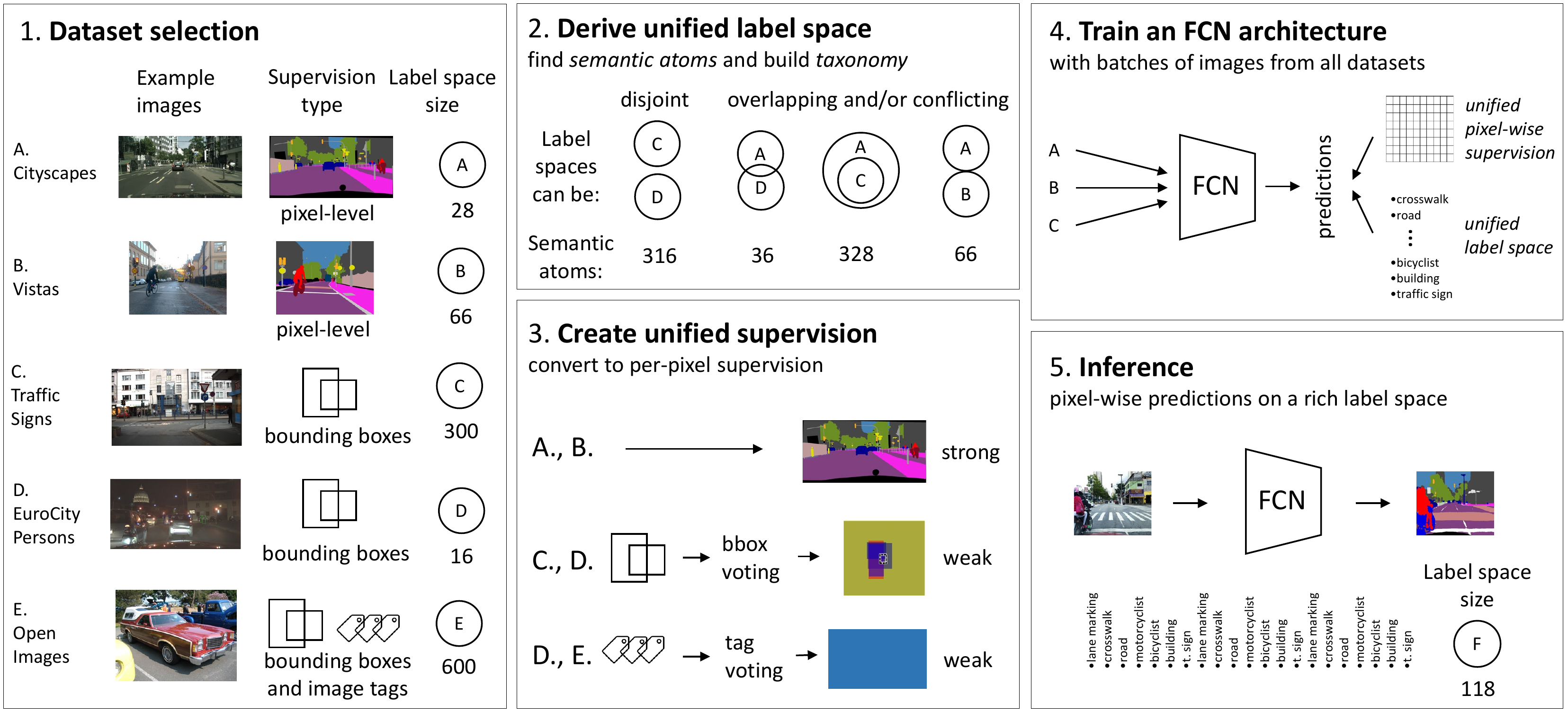}
	\caption{Motivation and overview of the proposed framework. Universal Semantic Segmentation aims at using a wide range of heterogeneous image-understanding datasets with incompatible annotation formats and conflicting label spaces (Block 1). Our methodology derives a unified label space (2) and consolidates supervision (3), so they can be used to simultaneously train an FCN (4) on all datasets. The trained network (5) generalizes better to unseen images and recognizes fine-grained semantic classes.}
	\label{fig:motivation-b}
\end{figure}

\section{Problem formulation and challenges of HToSS}
\label{sec:challenges}
The purpose of this section is to first more accurately define the problem of generalized semantic segmentation, prior to presenting our method on Heterogeneous Training of Semantic Segmentation in the next section. Here, the text starts with a review of the conventional semantic segmentation task and proceeds with aspects for a more generalized view of this task. Then, in a set of subsections, the main issues are addressed. Section~\ref{ssec:prob-form} commences with a detailed mathematical formulation of the semantic segmentation problem. Afterwards, Section~\ref{ssec:challenges} presents the challenges of the discussed generalized view.

\subsection{Preliminaries}
\label{ch6:ssec:prelim}

\subsubsection{A. Semantic Segmentation}
The task of Semantic Segmentation~\cite{guo2017aro,long2015fully,garcia2018survey,cordts2016Cityscapes,minaee2020image} involves the per-pixel classification of an image into a predetermined set of mutually-exclusive semantic classes. A semantic segmentation system has a 2-D image $\mathbf{x}$ as input and uses a given label space $\mathcal{L}^\text{pred}$ of semantic classes. The aim of the system is to predict a 2-D matrix $\mathbf{y}^\text{pred}$, where each element corresponds to an image pixel with a semantic class assigned to it.

In the conventional supervised learning setting, the segmentation task entails a dataset $\mathcal{S} =  (\mathcal{D}, \mathcal{L})$, which consists of $N$ image-label pairs ~$\mathcal{D} =  \{\left(\mathbf{x}_i, \mathbf{y}_i\right),~i = 1, \dots, N\}$ and a label space $\mathcal{L} = \left\{l_j,~j = 0, \dots, L\right\}$ of $L$ semantic classes and one special class $l_0$ representing semantics not included in the $L$ classes (unlabeled or void pixels). Every label $\mathbf{y} \in \mathcal{L}^{H \times W}$ is a 2-D matrix with spatial size $H \times W$ and every position corresponds to a single pixel in the image $\mathbf{x}$.

It is essential that the semantic classes $l_j$ have unambiguous and mutually-exclusive semantic definitions $\text{def}(l_j)$, which represent semantic entities of a scene,~\eg vehicle, person, tree, sky. If this is not true,~\ie annotations are noisy or concepts overlap between classes, then the trained classifier may be confused and evaluation is inaccurate. In literature, the classes assigned to pixel regions are considered to be unambiguous and non-overlapping (\eg a red car driving on a black road) with each other inside each label space. The ambiguity begins from the concepts included in the single word that describes a label (\eg a caravan starting to look like a truck and vice versa). When this happens, we call the labels to have noise. As reported and common practice in literature, in the sequel of this chapter, we assume that such noise in labels is negligible.

\subsubsection{B. Generalized Heterogeneous Semantic Segmentation}
In the conventional setting described above, the formats of the task definition and the given dataset are in full agreement with each other. In this case, the output label space can be set to be the dataset label space $\mathcal{L}^{\text{pred}} \equiv \mathcal{L}$ and the predictions are congruent to the per-pixel annotations $\mathbf{y}^{\text{pred}} \cong \mathbf{y}$. The symmetry between the training data and the task goal is advantageous, however, this constrains the available information that a system can be exposed to,~\eg only one dataset from the first column of Figure~\ref{fig:datasets-statistics}. These datasets are limited in size and semantics, which renders them inadequate for large-scale semantic segmentation training or even in-the-wild segmentation deployment.

We consider two extensions to the traditional semantic segmentation problem and encapsulate them in a unified formulation. The first aims at enriching the output semantic space of the task using heterogeneous label spaces. The second intends to increase the amount of available supervision by generalizing segmentation to more types of supervision. We incorporate these generalizations under a common formulation by maintaining the task format identical, while relaxing the requirements for the given datasets. This enables potential inclusion of datasets, which are originally destined for other image recognition problems,~\eg multiple datasets from all columns of Figure~\ref{fig:datasets-statistics}, when training a network for semantic segmentation.

Heterogeneous semantic segmentation enables trained networks to aggregate information from diverse datasets under a consistent formulation and it could demonstrate potential improvements in the following three aspects.
\begin{enumerate}[noitemsep,topsep=0pt]
	\item \textit{Accuracy on seen datasets}: The inclusion of multiple datasets during training increases samples for underrepresented classes and provides diversity in recognizable semantics. This should result in an increase of the standard segmentation accuracy metrics (mIoU, mPA (mean Pixel Accuracy)) on the testing splits of the datasets used for training.
	\item \textit{Generalizability on unseen datasets}: The segmentation performance on datasets that were not used during training is generally expected to be lower than on seen datasets. Multi-source training is expected to contribute to long-sought generalizability, which we evaluate using segmentation metrics on unseen datasets (\textit{cross-dataset zero-shot} setting~\cite{lambert2020mseg}).
	\item \textit{Semantic knowledgeability}: The semantic richness of network predictions can be enhanced by incorporating semantics from multiple datasets. As can be observed from Figure~\ref{fig:datasets-statistics}, label spaces for finely annotated datasets are smaller than coarsely annotated datasets,~\ie $\mathcal{L}^\text{pixel} \ll \mathcal{L}^\text{bbox} \ll \mathcal{L}^\text{tag}$. To the best of our knowledge, the variety of output classes in $\mathcal{L}^\text{pred}$ has not been quantified in this context and system performance cannot be compared across datasets w.r.t. the number of recognizable semantic concepts. To this end, we propose a single metric (refer to Section~\ref{ssec:knowledgeability}) to quantify the semantic richness of the output classes and define the trade-off against the prediction accuracy for these classes.
\end{enumerate}

\subsection{Problem formulation of HToSS}
\label{ssec:prob-form}
Similar to the conventional segmentation task definition, generalized semantic segmentation aims at predicting a \mbox{2-D} matrix $\mathbf{y}^\text{pred}$ with semantic classes, given a 2-D image $\mathbf{x}$ and a label space $\mathcal{L}^\text{pred}$. Contrary to the traditional single-dataset formulation, we assume that a set $\mathbb{S}$ of $D$ heterogeneous datasets is available, where each dataset $\mathcal{S}^{(i)} = ( \mathcal{D}^{(i)}, \mathcal{L}^{(i)} )$ includes $N$ image-label pairs $\mathcal{D}^{(i)}$ and the corresponding label space $\mathcal{L}^{(i)}$ with $L^{(i)}$ semantic classes:
\begin{flalign}
	& \mathbb{S} = \left\{ \mathcal{S}^{(i)}, ~i = 1, \dots, D \right\} ~,\\
	& \mathcal{S}^{(i)} = \left( \mathcal{D}^{(i)}, \mathcal{L}^{(i)} \right) ~,\\
	& \mathcal{D}^{(i)} =  \left\{ \left( \mathbf{x}^{(i)}_j, \mathbf{y}^{(i)}_j \right),~j = 1, \dots, N^{(i)}, ~i = 1, \dots, D \right\} ~, \label{eq:dataset-def}\\
	& \mathcal{L}^{(i)} = \left\{ l^{(i)}_m, ~m = 0, \dots, L^{(i)}, ~i = 1, \dots, D \right\} ~.
\end{flalign}
The goal is to train a system for semantic segmentation, such that it utilizes information from all heterogeneous datasets in $\mathbb{S}$. The system should have a consistent label space and recognize the semantic concepts from all available label spaces $\mathcal{L}^\text{pred} = \cup_{i=1}^D \mathcal{L}^{(i)}$.

Within the investigated problem formulation, the following conditions should hold for the employed datasets.
\begin{enumerate}
\item \textit{Intra-dataset label-space consistency}. Each label space $\mathcal{L}^{(i)}$ should include consistent and mutually-exclusive semantic classes, as explained in Subsection~\ref{ch6:ssec:prelim}. However, there is no constraint between label spaces across datasets such that:
\begin{equation}
\text{def}\left( l^{(i)}_m \right) \cap \text{def}\left( l^{(i)}_n \right) = \emptyset, ~\forall ~i = 1, \dots, D ~,
\label{eq:hypothesis-1}
\end{equation}
where $\text{def}(l)$ denotes the proper definition of the semantic class $l$ as a set of all semantic concepts that $l$ contains. 

\item \textit{Condition for weakly-labeled classes}. Any semantic class from a weakly-labeled dataset $\mathcal{S}^{(W)}$ should either correspond identically to, or contain partially semantics from, a class in a strongly-labeled dataset $\mathcal{S}^{(S)}$, which specifies that:
\begin{equation}
	\exists ~l^{(S)} ~\text{so that} ~def\left( l^{(W)} \right) \subseteq def\left( l^{(S)} \right) ~.
	\label{ch6:eq:cond-2}
\end{equation}
\end{enumerate}

It is noted that: i) cond. 2 implies that there must be at least one pixel-labeled dataset available for training, and ii) cond. 1 does not imply inter-dataset label space consistency, which is one of the challenges addressed by our HToSS framework.

\subsection{Challenges}
\label{ssec:challenges}
The previously presented extensions in the problem formulation are accompanied by various challenges, due to the structural and intrinsic differences among datasets. Weak labels introduce spatial localization uncertainties during pixel-wise training. Moreover, the inter-dataset sample and semantic imbalances become more apparent in a multi-dataset training setting compared to single-dataset training. The most prominent challenges reside in the annotation formats and the conflicting label spaces of the employed datasets. The following paragraphs analyze these aspects.
\begin{table}
	\small
	\centering
	\begin{tabular}{@{}ll@{}}
		\toprule
		Label type & Definition\\
		\midrule
		Pixel (dense) & $\mathbf{y} \in \mathcal{L}^{H \times W}, ~\text{count}(y_k = l_0) \ll \text{count}(y_k \neq l_0), ~y_k \in \mathbf{y}$\\
		Pixel (coarse) & $\mathbf{y} \in \mathcal{L}^{H \times W}, ~\text{count}(y_k = l_0) \gg \text{count}(y_k \neq l_0), ~y_k \in \mathbf{y}$\\
		Bound. boxes & $\mathbf{y} = \left\{ \left( l_k, ~\text{bbox-coords}_k \right), ~k = 1, \dots, B_j \right\}$\\
		Image tags & $\mathbf{y} = \left\{ l_k, ~k = 1, \dots, T_j \right\}$\\
		\bottomrule
	\end{tabular}
	\caption{Variety of annotation formats considered in HToSS. The dataset indexing superscript $(i)$ and the dataset sample subscript $j$ are omitted for clarity. Class label $l_0$ is the \textit{void} class. The function $\text{count}(\cdot)$ counts the number of elements for which the condition argument holds.}
	\label{tab:annot-formats}
\end{table}

\subsubsection{A. Label-space conflicts}
\label{sec:semantic-level-of-detail}
Datasets are annotated over different label spaces on a vast spectrum of semantic detail, as they are collected to serve different purposes, which leads to conflicting or overlapping definitions of classes between datasets. If the class definitions for all labels is matching between datasets then a simple union of the label spaces is feasible. However, this is usually not doable because of potential conflicts. The main source of conflicts stems from partial overlapping semantic class definitions between two arbitrary datasets $\mathcal{S}^{(X)}$ and $\mathcal{S}^{(Y)}$, which is specified by:
\begin{equation}
	def\left(l^{(X)}\right) \cap def\left(l^{(Y)}\right) \neq \emptyset ~.
\end{equation}

Since, the class definitions can overlap only partially, merging them or including them both in the combined label space will introduce ambiguity to the output label space of a trained network. A special common case occurs when conflicts arise from differences in the semantic level-of-detail between classes. For example, a class $l^{(X)}$ from dataset $\mathcal{S}^{(X)}$ describes a high-level concept which contains many more fine-grained classes for dataset $\mathcal{S}^{(Y)}$, giving:
\begin{equation}
	def\left(l^{(X)}\right) = \bigcup_m def\left(l_m^{(Y)}\right) ~.
\end{equation}
The inclusion of all classes $l^{(X)}$, $l^{(Y)}_m, \forall m$ in a single label space would also imply introducing conflicts.

These challenges are addressed in literature by i) keeping only a reduced subset of common, non-overlapping classes,~\ie the intersection of label spaces, or ii) by training multiple networks, or iii) by re-annotating datasets to the finest semantic concepts. These solutions either reduce the available semantic diversity of the final system, or require increased costs and computations during training and inference, compared to devising a single, consistent label space and use that to train a single-backbone network, as we do in HToSS. 

\subsubsection{B. Annotation format incompatibilities}
\label{sec:annotation-types}
A plethora of diverse datasets for image recognition are not used in semantic segmentation due to their incompatible annotation formats. Semantic segmentation is by definition a pixel-wise task, thus it is convenient that training datasets provide annotations at the same pixel-level format. The spatial localization of labels from other datasets of Figure~\ref{fig:datasets-statistics} is not adequate to train a network for pixel-accurate segmentation. The incompatibilities in annotation formats are even more pronounced in a multi-dataset training scenario, where a variety of incompatible annotation formats can exist. Generalized heterogeneous semantic segmentation requires to extract useful supervision at the pixel level from a much coarser source of information.

\section{Methodology}
\label{sec:methodology}
\begin{table}
	\small
	\centering
	\begin{tabular}{cc|ll}
		\toprule
		\makecell[c]{Label\\spaces} & \makecell[c]{Label\\type} & \makecell[l]{Preparation\\(once)} & \makecell[l]{Supervision during\\training (each step)}\\
		\midrule
		\multirow{2}{*}{\rotnighty{\makecell{no\\conflicts}}} & strong & - & standard cross-entropy (CE)\\
		\cmidrule{2-4}
		& \makecell{weak\\(Sec.~\ref{sec:weak-superv})} & \makecell[l]{create pseudo-labels\\(Sec.~\ref{ssec:unify}, Fig.~\ref{fig:temp-convert-supervision})} & \makecell[l]{conditional CE, refine pseu-\\do-labels (Sec.~\ref{ssec:hier-loss}, Fig.~\ref{fig:temp-loss-comps})}\\
		\midrule
		\multirow{2}{*}{\rotnighty{\makecell{with\\conflicts}}} & \makecell{strong\\(Sec.~\ref{sec:label-taxonomy})} & \makecell[l]{build taxonomy\\(Sec.~\ref{ssec:gen-taxonomy}, Fig.~\ref{fig:hierarchy})} & \makecell[l]{supervise semantic atoms\\(Sec.~\ref{ssec:converters}, Fig.~\ref{fig:temp-hierarchical-classifier})}\\
		\cmidrule{2-4}
		& \makecell{weak\\(Sec.~\ref{ssec:general-case})} & \makecell[l]{\textbf{all above}} & \makecell[l]{\textbf{all above}}\\
		\bottomrule
	\end{tabular}
	\caption{Overview of HToSS methodologies developed in this chapter. Each row includes methods for combining a strongly-labeled dataset and any other dataset(s) with the type of supervision denoted by the second column.}
	\label{ch6:tab:overview-methods}
\end{table}
The development of our methodology for heterogeneous multi-dataset training abides to the design principle of maintaining the established FCN architecture for semantic segmentation~\cite{long2015fully}. Although this may seem as a hidden assumption, we conjecture it is not, because that established architecture can handle uniformly pixel-accurate annotations. The objective of this chapter is to extend the framework of semantic segmentation into a generalized heterogeneous concept, which is also pixel-accurate and accepts simply for heterogeneity in label spaces and annotation types. Therefore, an FCN can just be extended to handle this broader formulation.

As a starting point, the network is a single-backbone, convolutional-based structure for image feature extraction, and a single-head classifier for per-pixel classification of the extracted features. This strong desideratum enables straightforward applicability of the proposed methods to current or future FCN-based architectures, and scalability to an arbitrary number of datasets. Other approaches,~\eg multi-classifier or multi-backbone architectures that depend on the number of datasets, or detection networks that have also segmentation output, specialize the FCN design principle into a specific instantiation. In order to satisfy the FCN design requirement, we make an assumption regarding the label spaces of weak datasets, which was already formulated as the second condition of the problem formulation (Section~\ref{ssec:prob-form}). 

The Heterogeneous Training of Semantic Segmentation framework proposes solutions to the challenges of Section~\ref{ssec:challenges} by introducing a methodology for combing disjoint or conflicting label spaces (Section~\ref{sec:label-taxonomy}), and training a single-backbone network with strong and weak supervision simultaneously (Section~\ref{sec:weak-superv}). These components of the solution are summarized in Table~\ref{ch6:tab:overview-methods}.
Our objective, as described in detail in the problem statement~\ref{ssec:prob-form}, is threefold: 1) improve segmentation performance on seen datasets, 2) increase generalizability to unseen datasets, 3) enhance the semantic knowledgeability,~\ie the number of distinct classes that a network can predict.

\subsection{Combine datasets with different label spaces}
\label{sec:label-taxonomy}
This section describes the proposed approach for training a single-backbone, single-classifier FCN on multiple pixel-labeled datasets with disjoint or conflicting semantic label spaces. As explained in Section~\ref{ssec:challenges}, the naive approach of training with all datasets and output predictions over the union of classes is not applicable in the general case. Specific classes may not be present in all datasets, or they can describe semantics with different granularity, leading to semantic confusion and ambiguities during training and inference.

%
\begin{figure}
	\centering
	\includegraphics[width=0.55\linewidth]{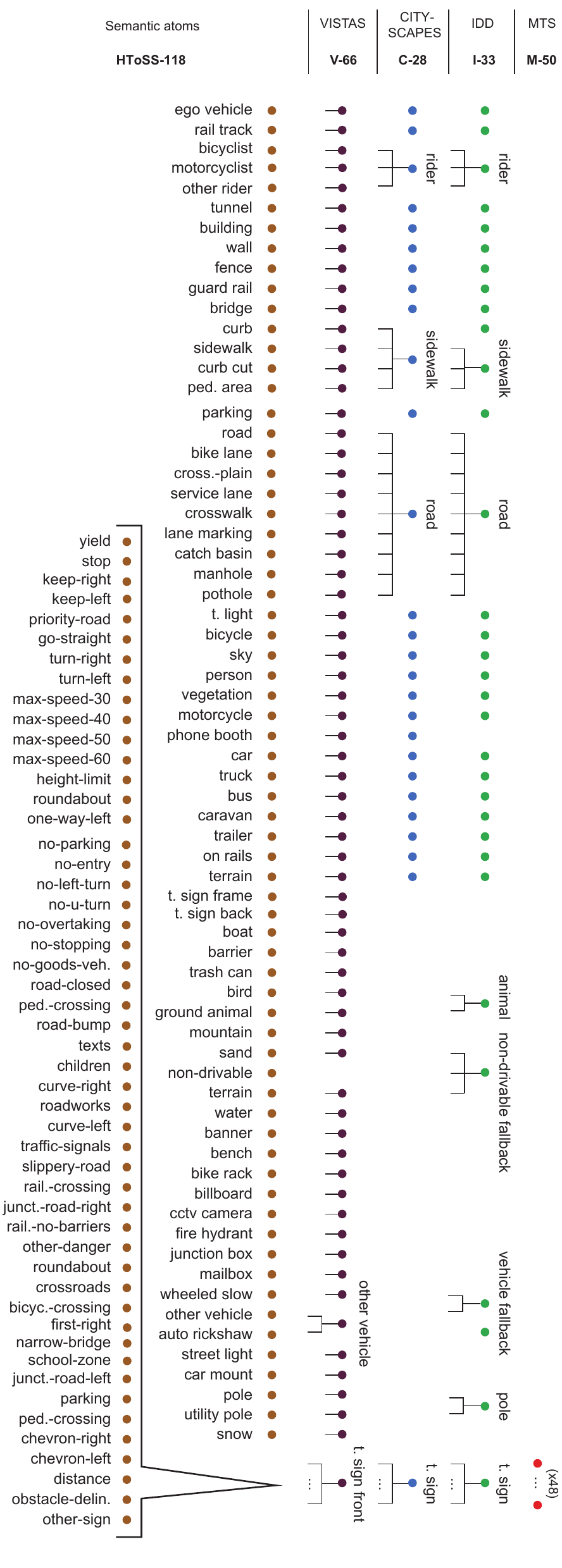}
	\caption{Combined taxonomy of 118 \textit{semantic atoms} (\textit{void} atom is not shown) merging a total of 174 semantic classes from Cityscapes (27), Vistas (65), IDD (33), and MTS (50) datasets. Each dataset section corresponds to the grouping of the combined taxonomy labels in order to apply strong supervision from the respective dataset.}
	\label{fig:hierarchy}
\end{figure}

\begin{figure}
	\centering
	\includegraphics[width=1.0\linewidth]{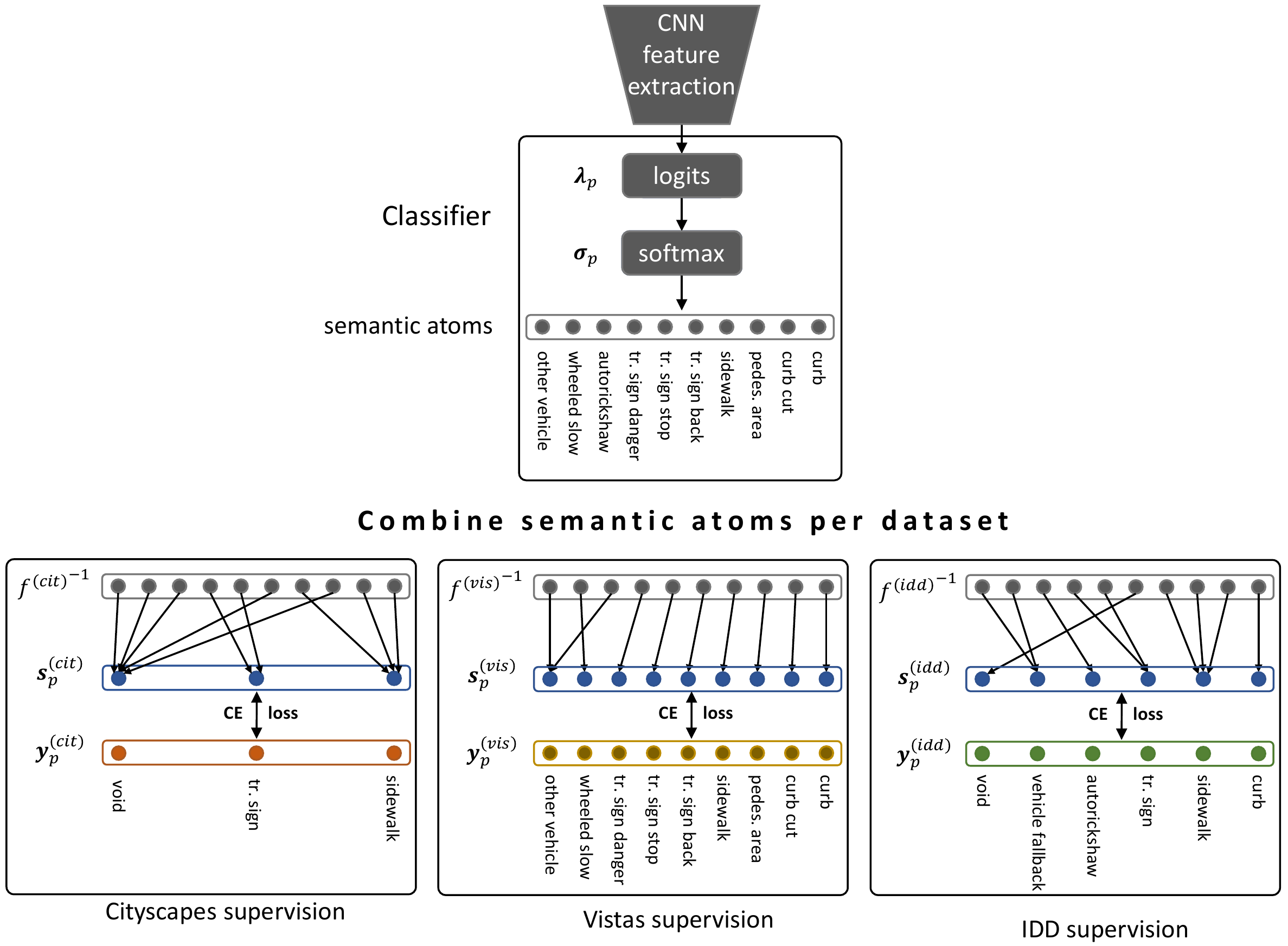}
	\caption{
		The HToSS classifier is supervised using the standard cross-entropy (CE) loss between the \textit{semantic atoms} vector and
		a different ground-truth vector per dataset (in boxes). This is achieved by combining the CNN output onto a new multi-nomial
		distribution vector that matches the distribution of the labels of each dataset. A selection of classes from Figure~\ref{fig:hierarchy} is depicted in order to explain the procedure of probabilities accumulation.}
	\label{fig:temp-hierarchical-classifier}
\end{figure}

Our method employs a regular FCN and enables training on an arbitrary number of datasets by operating exclusively on the supervision stage, by manipulating predictions of network. The procedure breaks down in two steps. First, ahead of FCN training, the semantics of the training datasets are consolidated by generating a unified taxonomy of their label spaces (Section~\ref{ssec:gen-taxonomy}). Then, the classifier makes predictions on this unified taxonomy. Second, during training, specific per-dataset converters (Section~\ref{ssec:converters}) transform the classifier output label space to match the label space of each dataset, and subsequently the segmentation loss function is applied for the optimization algorithm.

\subsubsection{A. Generation of unified taxonomy of label spaces}
\label{ssec:gen-taxonomy}
As outlined in the problem statement (Section~\ref{ssec:prob-form}), there are no constraints imposed on the label spaces of strongly-labeled datasets. Hence, it is highly probable that classes between datasets have disjoint or conflicting definitions, as described in Section~\ref{ssec:challenges}. In order to solve conflicts among different label spaces, we introduce the concept of the \textit{semantic atoms}. With the use of the \textit{semantic atoms}, we aim at deriving a unified label space from all datasets $\mathbb{S}$.

A \textit{semantic atom} $\alpha$ is a fine-level semantic primitive/class, of which the definition corresponds either fully or partially to a definition of a semantic class from a dataset in $\mathbb{S}$. A set of properly defined \textit{semantic atoms} $A = \left\{ \alpha_m, ~m = 1, \dots, N_A \right\}$ can fully cover the semantics of all employed datasets. The following three properties hold for all \textit{semantic atoms}. First, each \textit{semantic atom} should have a concise and unique semantic definition that does not overlap with any of the other \textit{semantic atoms}:
\begin{equation}
	\text{def}\left(\alpha_k\right) \cap \text{def}\left(\alpha_m\right) = \emptyset, ~\forall ~k \neq m ~.
\end{equation}
Second, its definition matches fully or partially to a definition of a semantic class from a dataset and thus every \textit{semantic atom} corresponds to (\textit{is-a}) at most one semantic class:
%
\begin{equation}
	\text{def}(\alpha_m) \subseteq \text{def}(l_n), ~\forall \alpha_m \in \mathcal{A}, ~l_n \in \mathfrak{L} \label{eq:sem-atom-def} ~,
\end{equation}
where $\mathfrak{L} = \cup_i \mathcal{L}^{(i)}, ~i = 1, \dots, D$ is the set of labels from all label spaces of the datasets to be combined.
Third, the set of all \textit{semantic atoms} should completely describe the semantics of all datasets, which yields that every semantic class $l_n$ consists of (\textit{has-a}) at least one \textit{semantic atom}:
%
\begin{equation}
	\text{def}(l_n) = \bigcup_{m \in M_n} \text{def}(\alpha_m) , ~\forall l_n \in \mathfrak{L} ~,
\end{equation}
where $M_n$ is a set of indices corresponding to class $l_n$. The second equation states that a set all labels in the label space contains the union of the \textit{atoms} that are inside the label space found by the preceding equation.

The universal taxonomy,~\ie the set of all \textit{semantic atoms} $\mathcal{A}$ can be easily generated by inspecting the definitions of semantic classes from the multi-set of label spaces $\mathfrak{L}$, and either splitting them into, or including them as, primitive semantic classes, such that $\mathcal{A}$ has mutually exclusive \textit{semantic atoms}. This step cannot be currently automated and requires human involvement, since in existing datasets the definition format of semantic classes and the ambiguity in their (natural language) descriptions does not intrinsically allow automated processing. However, if datasets provide semantics in a more formalized manner,~\eg with ontology/attribute-based semantic definitions~\cite{kulmanov2020semantic}, this step may also be automated. Alternatively, the automation can be achieved by statistical analysis of annotations and semantics, in order to uncover correlation between labels, which can lead to label merging or splitting.

As long as the manual process for the extraction of \textit{semantic atoms} is completed, the taxonomy of semantic classes from all datasets can be generated (see Figure~\ref{fig:hierarchy}). Then, we can train a single-classifier FCN using the \textit{semantic atoms} as output classes and supervise this CNN using the original label spaces from each dataset, by combing the \textit{atoms} using the generated unified taxonomy as described in the next section.

The \textit{semantic atom} and the semantic classes of all datasets should adhere to the following requirements.
\begin{itemize}
	\item Every semantic class contains (\textit{has-a}) at least one \textit{semantic atom}.
	\item Every \textit{semantic atom} corresponds (\textit{is-a}) to at most one semantic class ($N_A \ge L^{(i)}$).
	\item Each \textit{semantic atom} has a concise and unique semantic definition that does not overlap with any of the other \textit{semantic atoms}.
\end{itemize}

Summarizing, the unified taxonomy of the label space from all datasets is generated using the extracted \textit{semantic atoms}. This consists of combining all \textit{atoms} from all datasets into the same taxonomy (see Figure~\ref{fig:hierarchy}). We can train a single-classifier CNN using the \textit{semantic atoms} as output classes and supervise this CNN using the original label spaces from each dataset as input, as described in the next section.

\subsubsection{B. Supervision of \textit{semantic atoms} with original label spaces}
\label{ssec:converters}
Having extracted the set of \textit{semantic atoms} $\mathcal{A}$ that fully covers the semantics of the employed datasets $\mathbb{S}$, a single-backbone, single-classifier FCN with output label space $\mathcal{A}$ can be trained. This procedure is shown in Figure~\ref{fig:temp-hierarchical-classifier} for a selection of \textit{semantic atoms} from Figure~\ref{fig:hierarchy}. The output of the classifier for spatial position (pixel) $p$ and dataset $i$ is the categorical probability vector $\bsigma^{(i)}_p \in \left[0, 1\right]^A$, where each element corresponds to the probability of a \textit{semantic atom} in $\mathcal{A}$. Since $\bsigma^{(i)}_p$ represents a categorical probability it holds that $\sum_m \sigma^{(i)}_{p, m}=1$. In the following, it is described how $\bsigma^{(i)}_p$ is transformed to be compatible with the original label space $\mathcal{L}^{(i)}$ of each dataset of the taxonomy, in order to train the classifier using the conventional cross-entropy loss.

Conceptually, for each supervising dataset $i$, the categorical output $\bsigma_p^{(i)}$ is mapped to the categorical labels. Via this mapping the labels of the original dataset can supervise (in)directly the training of the semantic atoms. The extraction of \textit{semantic atoms} induces a collection of sets $\{G^{(i)}_m, ~i = 1, \dots, D, ~m = 0, \dots, L^{(i)}\}$. Each $G^{(i)}_m$ contains the \textit{semantic atoms} that correspond to class $l^{(i)}_m$ from dataset $i$. According to the taxonomy construction process (Section~\ref{ssec:gen-taxonomy}), the extracted \textit{semantic atoms} fully describe the semantics of all classes from all selected datasets. As a consequence, an arbitrary dataset class is represented by either a single or a combination of \textit{semantic atom}(\textit{s}). Using this property, $\bsigma^{(i)}_p$ is partitioned into groups according to sets $G^{(i)}_m$ and accumulate their probabilities into a reduced vector $\bm{s}_p^{(i)} \in {[0, 1]}^{L^{(i)}}$ for each dataset $i$. This process can be written concisely as:
\begin{equation}
	s^{(i)}_{p, m}(\bsigma_p) = \sum_{\alpha \in G^{(i)}_m} \sigma^{(i)}_{p, \alpha} ~,
\end{equation}
where the \textit{atoms} $\alpha$ are used for indexing, which is possible by assigning an integer number to each of them,~\ie $\mathcal{A} = \{1, 2, \dots, A\}$. Since $\bsigma^{(i)}_p$ is a categorical distribution, then $\bm{s}^{(i)}_p$ is also a categorical distribution\footnote{Each collection of sets $\{G^{(i)}_m\}$ is a partition of the set $\mathcal{A}$, thus all elements of $\bm{\sigma}_p$ are used exactly once in a summation.}. Moreover, it contains classes that correspond one-by-one to the ground truth $\bm{y}^{(i)}_p$. Thus, they can be used in the standard cross-entropy loss formulation.

During batch-wise training, a batch can contain images from many datasets. Without loss of generality, the cross-entropy loss for a single image $j$ from dataset $i$ in the batch is formulated. The label $\bm{y}^{(i)}_j$ (Eq.~\eqref{eq:dataset-def}) has shape $H \times W \times L^{(i)}$ (one-hot encoding), and the output $\bsigma^{(i)}_j$ has shape $H \times W \times A$. By using a single index $p \in P$ to enumerate spatial positions ($H, W$), the cross-entropy loss for each image $j$ can be expressed as:
\begin{equation}
	\text{Loss} \left(\bm{y}^{(i)}, \bsigma^{(i)} \right) = -\dfrac{1}{\left|P\right|} \sum_{p \in \mathcal{P}} \sum_m y^{(i)}_{p, m} \log s^{(i)}_{p, m} ~, \label{eq:loss-per-image}
\end{equation}
where $j$ has been omitted from all symbols for clarity.

In the following, the gradients of the loss wrt. the logits of the network is derived and it is shown that the proposed method is a generalization of the standard formulation. As this is independent of the dataset and the position indices, they are dropped for minimizing notation clutter. The logits $\bm{\lambda} \in \mathbb{R}^A$ are the input of the softmax $\bsigma$, where $\sigma_i (\bm{\lambda}) = e^{\lambda_i} / \sum_j e^{\lambda_j}$ and the converted outputs of the network can be expressed as $ \bm{s}\left(\bsigma\left(\bm{\lambda}\right)\right)$. Using the backpropagation rule the gradient of the loss wrt. the logits is:
\begin{equation}
	\frac{\partial \text{Loss}}{\partial \bm{\lambda}} = \frac{\partial \text{Loss}}{\partial \bm{s}} \cdot \frac{\partial \bm{s}}{\partial \bsigma} \cdot \frac{\partial \bsigma}{\partial \bm{\lambda}} ~. \label{eq:part-der}
\end{equation}
Since each pixel in the annotations has a single class (one-hot),~\ie $y_m = 1$ for class $m = m^*$ and $y_m = 0,~m \ne m^* $, the loss of Eq.~\eqref{eq:loss-per-image} (omitting the summation over positions $p$) reduces to $-\sum_{m} \llbracket m = m^* \rrbracket \log s_m = -\log s_{m^*}$, where $\llbracket \cdot \rrbracket$ is the Iverson bracket. It is easy to show that the partial derivatives of the factors in Eq.~\eqref{eq:part-der} are:
$\partial \text{Loss}/\partial s_i = -\llbracket i = i^* \rrbracket/\sigma_{i} $,
$\partial s_i / \partial \sigma_j = \llbracket j \in G_i \rrbracket$, and 
$\partial \sigma_i / \partial \lambda_j = \sigma_i \left( \llbracket i = j \rrbracket - \sigma_j \right)$. Substituting these into Eq.~\eqref{eq:part-der} it yields:
\begin{equation}
	\frac{\partial \text{Loss}}{\partial \lambda_m} = \sigma_m - \llbracket m \in G_{m^*} \rrbracket ~,
\end{equation}
which is a mere generalization of the loss derivative in the original FCN framework, which is $\partial \text{Loss} / \partial \lambda_m = \sigma_m - \llbracket m = m^* \rrbracket $. This property ensures comparable gradient flows between FCN and our framework, and thus no architectural changes or loss modifications are needed.

\subsection{Combine datasets with different annotation types}
\label{sec:weak-superv}
This section describes how a single-backbone, single-classifier FCN is trained on multiple weakly-labeled or strongly-labeled datasets. For now, we assume that the label spaces of all datasets are identical up to the size and the labels. This limitation is lifted in Section~\ref{ssec:general-case}, where combined training with different annotation types and conflicting labels spaces is investigated.

The spatial localization of annotations in weakly-labeled datasets,~\eg bounding boxes and image tags, is inadequate for providing useful pixel-level supervision, as commented in Section~\ref{ssec:challenges}. However, if properly conditioned or refined, these spatial localizations have the potential to provide helpful cues for increasing segmentation performance. A two-step approach is followed with the design principle of conforming to the FCN framework without adding extra modules to the network, as explained at the start of Section~\ref{sec:methodology}. First, weak annotations from all datasets are converted to per-pixel pseudo labels, so they can be seamlessly used together with pixel labels from strongly-labeled datasets for pixel-wise training. Second, during each training step, the pseudo labels are refined, using only information from the network at this step, without requiring any external knowledge. By applying these guidelines, we adopt the design basis of the previous chapter, which however will be extended for solving the issues in this chapter.
\begin{figure}
	\centering
	\includegraphics[width=0.8\linewidth]{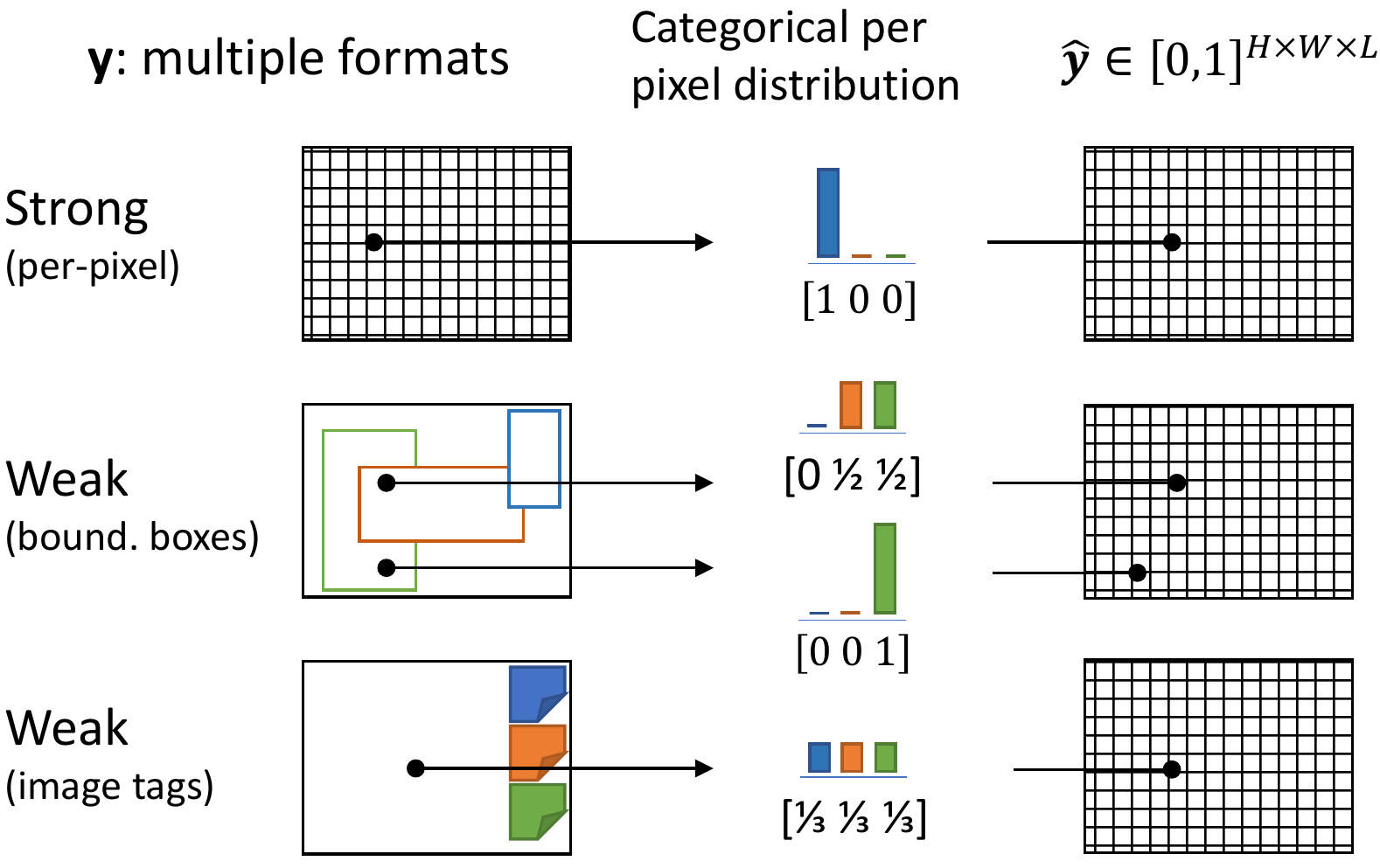}
	\caption{Creation of pseudo ground truth from weakly annotated datasets before training. For each pixel in the image a categorical probability vector is created with elements corresponding to the set of annotated classes ($L=3$ in this example). Each element is assigned a probability according to how many bounding boxes cover the underline pixel.}
	\label{fig:temp-convert-supervision}
\end{figure}

\subsubsection{A. Unifying weak and strong annotations}
\label{ssec:unify}
The objective is to transform the weak annotations into per-pixel pseudo labels, so they can be integrated in the pixel-wise training loss. The pseudo labels are then refined during training, to provide a best-effort approximation of the ideal fine labels.

As listed in Table~\ref{tab:annot-formats}, weak supervision from bounding boxes and image-level tags are considered. Bounding-box annotations have orthogonal boundaries, which rarely match the smooth object boundaries, of~\eg poles, humans, bicycles. Image tags have even a coarser localization. We treat image tags as bounding boxes that cover the whole image with their image tags. This forms a basis to handle both annotation formats under a common method.

The core of the method involves representing the per-pixel label as a categorical probability vector $\hat{\bm{y}}_p \in \left[0, 1\right]^L$ over the set of all classes $\mathcal{L}$ of the dataset it belongs to. This choice enables including information from all bounding boxes, even if they heavily overlap, and does not require hard choices to assign a single class to each pixel,~\eg assigning randomly or by heuristics. The algorithm is described in the following paragraph and visualized in Figure~\ref{fig:temp-convert-supervision}.

\paragraph{Algorithm for conversion of weak labels to per-pixel labels}
A 3-D label canvas is initiated with two spatial dimensions, being equal to the image size, and a depth dimension with size $L^{(i)}$ for the semantic classes. Each bounding box in $\mathcal{B}$ casts a unity vote to all pixels (spatial locations) being covered within its spatial limits and at a single position $l_j$ in the depth dimension that corresponds to its semantic class. After the voting is completed from all boxes, the 3-D label canvas is normalized along the depth dimension (semantic classes) to unity, by dividing the votes by the sum of votes for that position. Then, another 2-D slice is concatenated stretched along the same spatial dimensions, corresponding to the \textit{unlabeled} semantic class. Finally, for the pixels that are not covered by any bounding box, the \textit{unlabeled} class probability is set to unity. At this point, a valid categorical probability distribution vector is obtained ($\sum_m \hat{y}_{p, m} = 1$) for each image pixel $p$, which can be directly used as-is in the conventional per-pixel cross-entropy loss.

\subsubsection{B. Supervision of \textit{semantic atoms} with unified annotations}
\label{ch6:ssec:hier-loss}
In the previous subsection, it was sketched how weak annotations are transformed into per-pixel pseudo-labels that are directly usable in the cross-entropy loss formulation. Here, the refinement of these coarse labels is described. The refinement is performed in an online fashion during training and generates more accurate labels for supervision. It is achieved by applying two conditions to pseudo-labels that omit supervision for uncertain or ambiguous pixels,~\eg pixels that may reside outside the borders of an object.
\begin{figure}
	\centering
	\includegraphics[width=1.0\linewidth]{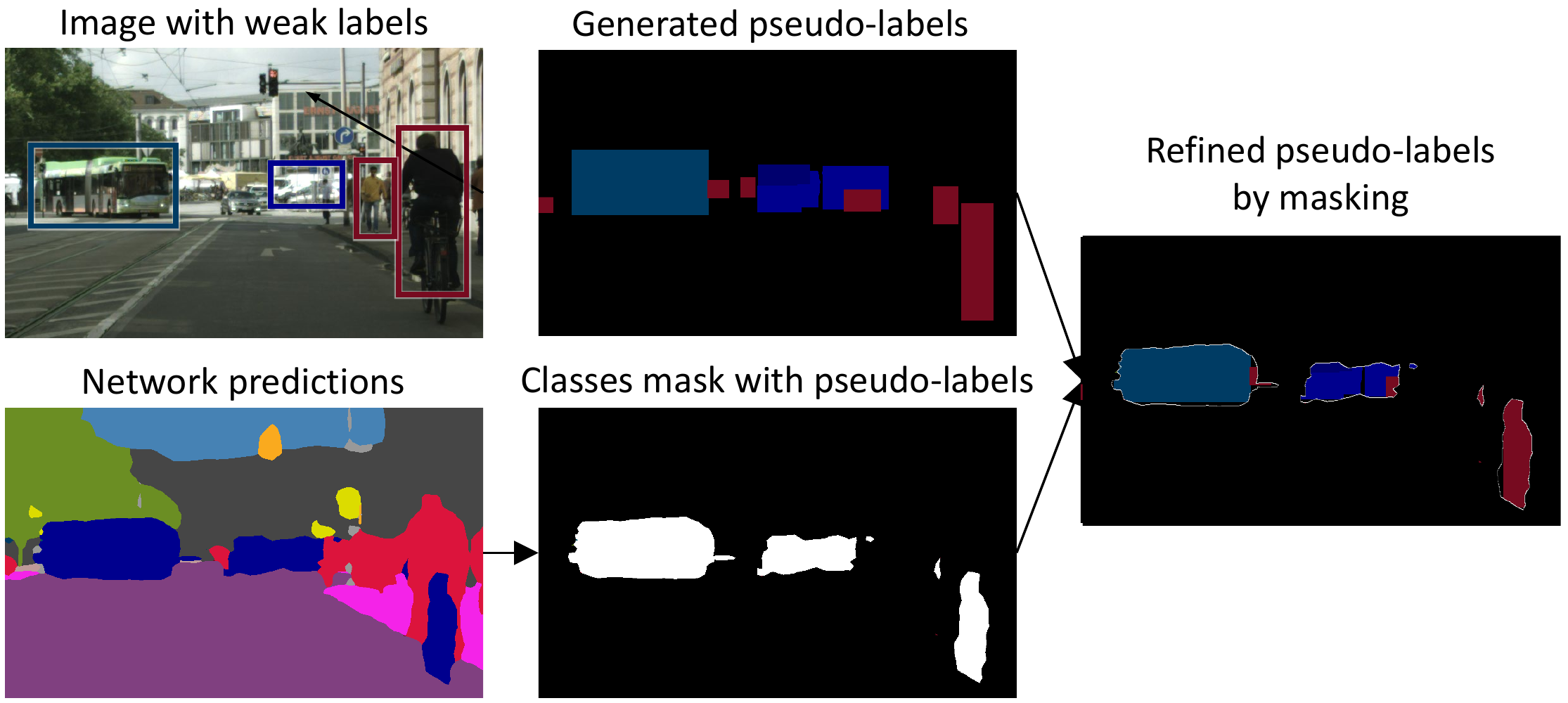}
	\caption{Refinement of pseudo labels during each training step using the predictions of the network in that step.}
	\label{fig:temp-loss-comps}
\end{figure}

\paragraph{Algorithm for refinement of generated per-pixel labels}
For the refinement, we assume that a collection of $\mathbb{S}^{(S)}$ datasets with strong labels and $\mathbb{S}^{(W)}$ datasets with weak labels are given, where all datasets have an identical label space. First, the weak labels $\bm{y}$ of $\mathbb{S}^{(W)}$ are converted to pseudo-labels $\hat{\bm{y}}$, using the procedure of Section~\ref{ssec:unify} A. During a training step, the best estimate of the segmentation result for an image with pseudo-labels is the prediction of the network for this image. We rely on the network's online predictions to improve the pseudo-labels. The following applies to every dataset in $\mathbb{S}^{(W)}$, hence the dataset super-script ($i$) is omitted for simplified notation. If $\bsigma_p \in [0, 1]^L$ is the softmax output of the classifier for position (pixel) $p$ and label space size $L$, then the prediction is the class having the maximum probability, expressed as $\pi_p = \argmax_m \sigma_{p, m}$.

The refined pseudo-labels $\tilde{\mathbf{y}}_p$ are obtained for pixel $p$ by keeping the pseudo-label if the prediction agrees and the corresponding probability is higher than a threshold $T$, as follows:
\begin{equation}
	\tilde{\mathbf{y}}_p = 
	\begin{cases}
		\hat{\mathbf{y}}_p, & \text{if} ~\pi_p = \argmax_m \hat{y}_{p, m} ~\text{and} ~\sigma_{p, \pi_p} \ge T,\\
		\text{unlabeled}, & \text{otherwise} ~.
	\end{cases}
	\label{ch6:eq:refine-ps-labels}
\end{equation}
This is the first condition and the process is illustrated in Figure~\ref{fig:temp-loss-comps}.

The second condition refers to the magnitude of the probability of the predictions, which can be view as a measure of confidence. Specifically, a heuristically chosen threshold $T$ is used, which should be exceeded by the probability of the highest predicted class, in order to be deemed reliable. This threshold provides a good trade-off between utilizing enough weak labels, while maintaining their confidence high. For the experiments, we have empirically chosen  $T = 0.9$. The final loss for the batch with images from both weakly-labeled  $(W)$ and strongly-labeled $(S)$ datasets is computed by:
\begin{equation}
	\text{Loss} = - \sum_{p \in \mathcal{P}} \sum_j z_{p,j} \log \sigma_{p,j} ~,
	\label{eq:loss-strong-weak}
\end{equation}
where $\mathcal{P} = \mathcal{P}^{(S)} \cup \mathcal{P}^{(W)}$ is the set of all pixels and $z_{p, j}$ is defined as:
\begin{equation}
	\bm{z}_p = 
	\begin{cases}
		\frac{1}{\abs{\mathcal{P}^{(S)}}} \bm{y}^{(S)}_p, & p \in \mathcal{P}^{(S)} \\
		\frac{1}{\abs{\mathcal{P}^{(W)}}} \tilde{\bm{y}}^{(W)}_p, & p \in \mathcal{P}^{(W)} ~.
	\end{cases}
\end{equation}

Note that $\bm{y}_p$, $\hat{\bm{y}}_p$ in the last equation are vectors based on the elements $y_{p, j}$ as used in the last function. The specification of the vectors and the loss function are similar to the construct in Chapter~\ref{ch:4-iv2019}.

\subsection{Combining datasets with conflicting label spaces and annotation types}
\label{ssec:general-case}
Section~\ref{sec:label-taxonomy} proposed a solution for label-space conflicts considering only strongly-labeled datasets. Section~\ref{sec:weak-superv} proposed a solution for training networks with multiple annotation types considering only datasets with identical label spaces. The combination of the two approaches that is able to simultaneously train networks with any datasets (case of last row of Table~\ref{ch6:tab:overview-methods}) is described in this section.

The extraction of \textit{semantic atoms} (Section~\ref{ssec:gen-taxonomy}) and the conversion of weak to per-pixel annotations (Section~\ref{ssec:unify}) can be directly applied to the selected datasets. However, for supervising the \textit{semantic atoms}, the formulas of Section~\ref{ssec:converters}, and ~\ref{ssec:hier-loss} cannot be directly applied for any \textit{semantic atom} and a small set of them requires a different handling. According to this distinction, the \textit{semantic atoms} are split into two sets $\mathcal{A}^a$ and  $\mathcal{A}^s$ with classes that need special care. Each \textit{atom} in $\mathcal{A}^a$ is either a class with only strong labels, or with weak and strong labels from different datasets. Each \textit{atom} in $\mathcal{A}^s$ is strictly a class with weak labels for which the condition of Eq.~\eqref{ch6:eq:cond-2} holds. The localization cues of the \textit{atoms} in $\mathcal{A}^s$ are extremely sparse, due to their weak annotations and the fact that they do not appear in strongly-labeled datasets. Consequently, the refinement step (Eq.~\eqref{ch6:eq:refine-ps-labels}) is ineffective for pixel-accurate segmentation. As a solution, for these \textit{atoms} (\eg the traffic sign sub-classes in the taxonomy of Figure~\ref{fig:hierarchy}), we use the parent (strongly-labeled) classes $\mathcal{A}^p$ (\eg \textit{traffic sign front}) as cues for pixel-accurate segmentation. Then, fine-grained semantics can be attained using classification over $\mathcal{A}^s$.

\begin{figure}
	\centering
	\includegraphics[width=0.7\linewidth]{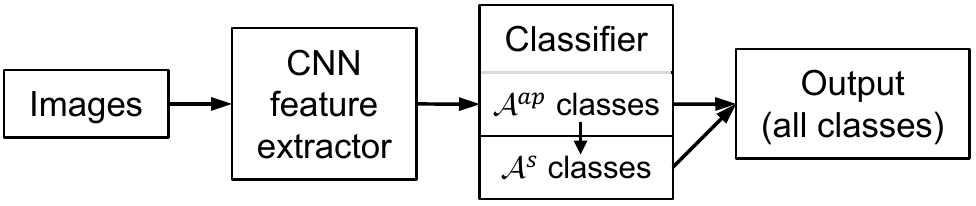}
	\caption{Classifier structure in case of conflicting label spaces and mixed (strong and weak) supervision. The final predictions (right block) are the predictions of classifier part $\mathcal{A}^{ap}$, where the pixels that are assigned classes belonging to classifier part $\mathcal{A}^p$ are replaced by sub-class predictions of classifier part $\mathcal{A}^s$.}
	\label{fig:classifiers-structure}
\end{figure}

The predictions of the classifier, as illustrated in Figure~\ref{fig:classifiers-structure}, are over two sets of classes: $\mathcal{A}^{ap} = \mathcal{A}^{a} \cup \mathcal{A}^p$ and $\mathcal{A}^s$. For the sub-classes in $\mathcal{A}^s$ the relationship between them and the parent classes $\mathcal{A}^p$ is leveraged. Specifically, the predictions of the corresponding parent class in $\mathcal{A}^p$ are used to provide cues for the refinement process (for example the predicted segmentation masks of the \textit{traffic sign front} class are used to refine the bounding boxes of the traffic sign sub-classes of Figure~\ref{fig:hierarchy}). The classifier is trained using the losses from Eq.~\eqref{eq:loss-per-image} and~\eqref{eq:loss-strong-weak}. During inference, the sub-classes of $\mathcal{A}^s$ simply replace their parent classes from $\mathcal{A}^p$ in the final predictions.

\section{Experimental evaluation}
\label{sec:experimentation}
Extensive experimentation is conducted with various combinations of datasets to validate the proposed methodology for multi-dataset simultaneous training with mixed supervision and conflicting label spaces. The experiments and results are assessed on three directions: segmentation performance on seen (training) and unseen (testing) datasets, and semantic knowledgeability, as mentioned in the problem formulation at the start of this chapter. Section~\ref{ssec:metrics} describes the evaluation metrics and Section~\ref{ssec:setup} discusses the technical details of our experiments. The following three Sections~\ref{ssec:exps-combine-label-spaces},~\ref{ssec:exps-strong-weak-nonconfl},~\ref{ssec:exps-combine-supervision} investigate the three scenarios in multi-dataset training appearing in Table~\ref{ch6:tab:overview-methods}. Finally, Section~\ref{ssec:exps-ablations} contains ablation studies for a selection of experiments in this work. A collection of diverse datasets for street scene understanding with strong and weak supervision are used. An overview of the employed datasets is shown in Table~\ref{tab:datasets-overview}, where for each dataset, the respective label spaces are defined.

~

A detailed dataset description is omitted here, since the experiments are based on the same datasets of the previous chapters. However, the labeling and abbreviation and label space indications are provided in the table that is discussed in the actual experiments later in this section. Therefore, the dataset table is presented close to the experiments.

\subsection{Evaluation metrics}
\label{ssec:metrics}
We use two metric families to quantify the performance of the models. The first family consists of the standard Intersection over Union (IoU) metric~\cite{lin2014microsoft,cordts2016Cityscapes} and various averages of it (arithmetic -- mIoU) to summarize performance over multiple classes and across different datasets. The second metric family is based on a new metric, namely \textit{Knowledgeability} that we define in the following subsection. This metric evaluates how many semantic concepts a model can recognize with sufficient segmentation accuracy that is measured with the IoU as the underlying metric.

\subsubsection{Knowledgeability metric}
\label{ssec:knowledgeability}
We introduce a new metric called \textit{Knowledgeability} that quantifies in a single value the semantic richness of the output of a semantic segmentation system, by evaluating how many semantic concepts (atoms) it can recognize with sufficient segmentation accuracy. 
Using existing metrics, one way to achieve this is to report the size of the system's output label space and separately the IoU performance per class. However, this approach has some pitfalls. First, merely reporting the label space size is not a reliable metric for semantic richness of predictions, since the IoU performance for some classes can be very low or even zero. Second, IoU-based average aggregates,~\eg mIoU, do not reflect the number of recognizable classes, because they assess a system purely at segmentation level. Finally, these aggregates are intrinsically dependent on the size of the evaluated label space: as the size increases, the difficulty of assigning the correct class increases, eventually leading to mIoU reduction (due to the smoothing properties of averaging). The new metric is designed to explicitly consider the size of the label spaces of both the system output and the evaluated dataset together with the segmentation performance for the output classes.

The core of the metric is based on counting the number of classes that achieve an IoU higher than a threshold $t$ wrt. the total number of classes $c$ that are considered for computing the metric. To make the metric independent of the need for proper selection of $t$, the counting is averaged over a set of $N_T$ thresholds, which in this work are chosen to be equidistant,~\ie $T = \{0.0, ~1/N_T, ~\dots, ~1.0-1/N_T\}$.
Other values for T can be chosen depending on the application and datasets specifics. Assuming an output label space of a model that contains $L$ discreet semantic classes, the set of all per-class IoUs $\mathcal{E} = \{\text{IoU}_i\}_{i = 1}^L$ can be constructed by evaluating the model output against the ground truth. Subsequently, the set $\mathcal{E}$ is used to generate all the subsets $\tilde{\mathcal{E}}_t=\{\text{IoU}~|~\text{IoU} > t,~\text{IoU} \in \mathcal{E}\}$ containing the IoUs above the threshold $t$ from $T$. To this end, \textit{Knowledgeability} is defined as the $c$-normalized number of classes averaged over $T$:
\begin{equation}
	\mathcal{K}^c_T = \frac{1}{N_T} \sum_{t \in T} \frac{\min( | \tilde{\mathcal{E}}_t |, c )}{c} ~.
	\label{eq:knowledgeability-definition}
\end{equation}

This definition guarantees $\mathcal{K}^c_T$ to be between 0 and 1,~\ie $0 \le \mathcal{K} \le \mathcal{K}_\text{max} = \min\left(L, c\right) / c \le 1$, which is achieved by creating the sets $\tilde{\mathcal{E}}_t$ using the strictly greater condition ($\text{IoU} > t$) and by employing the $\min$ function. The bounds enable the use of the metric for comparison across datasets with different number of classes and semantic segmentation systems with different number of output classes. The new metric \textit{Knowledgeability} allows us to express the increase in the number of recognizable classes and at the same time consider the performance on these classes, without the need to choose a specific single threshold.

\begin{table}
	\small
	\centering
	\begin{tabular}{@{}lHcrc@{}H@{}}
		\toprule
		Dataset name & Abv. & Labels & \multicolumn{1}{c}{\# imgs} & \makecell{Train./Eval.\\Lab. space} & \makecell{L. space\\Subsets}\\ 
		\midrule
		\textbf{Training datasets}\\
		~~Cityscapes~\cite{cordts2016Cityscapes} & C & pixel & 2,975 & C-28 & C-20 \\
		~~Cityscapes Coarse~\cite{cordts2016Cityscapes} & CC & \makecell{pixel, bbox} & 19,997 & C-28 & C-20 \\
		~~Cityscapes T. Signs~\cite{meletis2018heterogeneous} & CT & pixel & 2,975 & CT-62 & C-20, C-20\\
		~~Mapillary Vistas~\cite{neuhold2017mapillary} & V & pixel & 18,000 & V-66\\
		~~Indian Driving D.~\cite{varma2019idd} & I & pixel & 20,000 & I-33\\
		~~EuroCity-Persons~\cite{braun2019eurocity} & E & bbox & 40,000 & E-2\\
		~~Map. Traffic Signs~\cite{ertler2020mapillary} & T & bbox & 36,589 & T-51 \\
		~~Open Images~\cite{kuznetsova2020open} & O & \makecell{bbox, im. tag} & 200,000 & O-51\\
		\midrule
		\textbf{Testing datasets}\\
		~~Cityscapes (val)~\cite{cordts2016Cityscapes} & Cv & pixel & 500 & C-20 & CC-10, E-2\\
		~~Cityscapes T. Signs~\cite{meletis2018heterogeneous} & Cv & pixel & 500 & CT-34 & CT-20\\
		~~Mapil. Vistas (val)~\cite{neuhold2017mapillary} & Vv & pixel & 2,000 & V-66 & CC-10, E-2\\
		~~IDD (val)~\cite{varma2019idd} & Iv & pixel & 2,036 & I-33 &  CC-10, E-2 \\
		\midrule
		\multicolumn{3}{l}{\textbf{Generalization (unseen) datasets}}\\
		~~Wild Dash (val)~\cite{zendel2018wilddash} & W & pixel & 70 & W-19 \\
		~~KITTI (train)~\cite{geiger2013vision} & K & pixel & 200 & K-20 \\
		\bottomrule
	\end{tabular}
	\caption{Overview of employed datasets for experimentation. The type of annotations, the number of images and the label spaces are shown. The networks are trained with the label spaces of the training datasets. The evaluation label spaces may be smaller than the training counterparts for the same dataset, due to missing classes in testing/generalization datasets or smaller official splits.} 
	\label{tab:datasets-overview}
\end{table}

\subsection{Network and implementation details}
\label{ssec:setup}

\subsubsection{A. Convolutional network architecture}
ResNet-50~\cite{he2016deep} is used for feature extraction, since it provides a good trade-off between the segmentation performance, training time, and memory requirements. Although potentially better backbones have been proposed, the ResNet-50 network is sufficient for the purpose of our experiments. The original network is designed for ImageNet classification, thus adaptation for semantic segmentation is needed. To this end, the following changes are applied to its four architecture blocks: 1) the strides of the first convolutional block are changed from 1 to 2 and the last convolutional block from 2 to 1, since the images under consideration are larger than the ImageNet images and reduction of spatial resolution is preferred as early as possible, 2) a dilation rate~\cite{yu2016dilated,zhao2017pyramid} of 2 and 4 is applied to the bottleneck's middle convolution of the last two blocks respectively, in order to maintain spatial details of the representation, 3) the output 2,048-dimension features are projected with an 1x1 convolution layer to 256-dimension features to reduce memory requirements, and 4) a pyramid pooling module~\cite{zhao2017pyramid} is appended, which enriches the output features with contextual information by enlarging the receptive field and including multi-scale features.

The training procedure is similar to Chapter~\ref{ch4:sec:experiments}. The most important parameters are summarized here. The network is trained with Stochastic Gradient Descent with a momentum of 0.9 and initial learning rate 0.02 that is halved three times over equidistant periods. The L2 weight regularization has a decay of 0.00017 and the batch normalization moving averages decay is set to 0.9. The batch size and input image size are discussed in the following subsection.

\subsubsection{B. Hyper-parameter tuning and training implementation details}
Two factors that emerge in multi-dataset training and significantly affect the accuracy of CNNs for segmentation are the batch size and the spatial input image size for extracting features. The former is connected with the optimization algorithm's (SGD) robustness and the latter with the representation's detail and receptive field. Both of them are highly dependent on the amount of available computational resources and are an important aspect to consider in multi-dataset training. From the perspective of training a single network with multiple datasets, we have a limited amount of resources (4 Titan V GPUs), so that we tune the above factors separately for each experiment in order to have a good performance trade-off and training feasibility per experiment. This leads to different baseline performance for each experiment and therefore, the hyper-parameters are also set per experiment. 

\noindent\textit{Batch size}: The global (per-step) and per-dataset batch size are crucial hyper-parameters that determine balanced training across all classes and all datasets. The need for larger batch sizes increases proportionally to the size of the output label space. Moreover, including more datasets in the training requires a sufficiently large batch size. The experiments of this chapter involve a large variety of datasets with various sizes of label spaces. Henceforth, we tune the batch sizes per experiment, so all results are comparable per table, while at the same time, we are monitoring the performance to be satisfactory.

\noindent\textit{Image size}: The input image size of the feature extraction determines the scale and detail of the extracted features throughout the network. Larger input dimensions yield more detailed representations and better segmentation results, but increase the GPU memory requirements. Since different experiments contain various datasets, we tune this important hyper-parameter separately per table. Increasing the number of simultaneously employed datasets requires to reduce input image dimensions per-dataset, in order to fit images from all datasets in one batch.

In all experiment,  we use a three-stage procedure to reach the desired training image size: 1) Scale: images from all datasets are resized to have similar scales (height dimensions are close to each other), 2) Random crop: image patches are cropped from images if the image size from Step 1 is not the predefined input size of the feature extractor, 3) Scale: the patches are scaled again to the correct input size of the feature extractor.

\subsection{Strong supervision, conflicting label spaces}
\label{ssec:exps-combine-label-spaces}

\textit{Structure of the experiments}:
The experiments are organized according to the structure of the Methodology Section~\ref{sec:methodology}. The first three experiment subsections correspond one-by-one to the three proposed components of the methodology. The first one investigates the combination of datasets with strong supervision and conflicting label spaces. The second one treats datasets with non-conflicting label spaces and a combination of strong and weak supervision. The third subsection combines conflicting label spaces and weak supervision all together. The corresponding methodology elements are summarized in Table~\ref{ch6:tab:overview-methods}.

~

\begin{table*}
	\centering
	\scriptsize
	\setlength\tabcolsep{1.6pt}
	\begin{tabular}{@{}ccc|c||cccccccccccc@{}}
		\toprule
		\multicolumn{3}{c}{} & \multirow{3}{*}{\makecell{Output\\label\\space}} & \multicolumn{6}{c}{Generalization on unseen datasets} & \multicolumn{6}{c}{Performance on seen datasets (val)}\\
		\cmidrule(l{3pt}r{3pt}){5-10} \cmidrule(l{3pt}r{3pt}){11-16}
		\multicolumn{3}{c|}{Training datasets} &  & \multicolumn{2}{c}{WildDash} & \multicolumn{2}{c}{KITTI} & \multicolumn{2}{c}{W.Dash+KITTI} & \multicolumn{2}{c}{Citys} & \multicolumn{2}{c}{IDD} & \multicolumn{2}{c}{Vistas}\\
		Citys & IDD & Vistas & & mIoU & $\mathcal{K}^{19} $ & mIoU & $\mathcal{K}^{20}$ & mmIoU & m$\mathcal{K}$ & mIoU & $\mathcal{K}^{20}$ & mIoU & $\mathcal{K}^{33}$ & mIoU & $\mathcal{K}^{66}$\\ 
		\midrule
		\ding{51} & & & C-20 & 27.8 & 39.2 & 48.0 & 43.1 & 37.0 & 43.2 & 63.0 & 50.3 & 29.8 & 30.4 & 22.5 & 26.0\\ 
		& \ding{51} & & I-33 & 36.8 & 48.6 & 40.2 & 50.9 & 38.3 & 49.1 & 47.3 & 55.3 & \underline{63.4} & \underline{65.0} & 23.7 & 49.1\\ 
		& & \ding{51} & V-66 & 42.4 & 53.2 & 49.5 & 57.5 & 46.0 & 55.2 & 67.6 & 60.3 & 41.0 & 53.0 & 40.9 & 63.5\\ 
		\midrule
		\ding{51} & \ding{51} & & HToSS-34 & 36.3 & 49.9 & 55.3 & 50.5 & 46.4 & 50.2 & 73.0 & 56.8 & 61.3 & 58.5 & 26.4 & 49.8\\
		\ding{51} & & \ding{51} & HToSS-66 & \underline{44.4} & 60.0 & 51.6 & 62.4 & 48.0 & 48.0 & 61.2 & 62.3 & 43.7 & 61.0 & \underline{43.1} & 62.9\\
		& \ding{51} & \ding{51} & HToSS-68 & 44.0 & 59.5 & 53.8 & 63.0 & 48.9 & 61.3 & 69.2 & 60.1 & 58.0 & 62.4 & 42.6 & 63.0\\
		\ding{51} & \ding{51} & \ding{51} & HToSS-68 & \underline{44.4} & \underline{64.2} & \underline{56.5} & \underline{64.4} & \underline{50.5} & \underline{64.3} & \underline{74.9} & \underline{66.7} & 57.9 & \underline{65.0} & \underline{43.1} & \underline{68.3}\\ 
		\bottomrule
	\end{tabular}
	\caption{Overall results for the HToSS on combinations of pixel-labeled datasets with conflicting label spaces (bottom rows) compared to single-dataset counterparts (top rows). All models are trained with the same hyper-parameters per dataset. Segmentation performance and knowledgeability are assessed on seen and unseen datasets from Table~\ref{tab:datasets-overview}.}
	\label{ch6:tab:pp-confl}
\end{table*}

In the first set of experiments, we focus on combining pixel-labeled datasets with conflicting label spaces (case of third row in Table~\ref{ch6:tab:overview-methods}). This scenario is commonly occurring, where different pixel-labeled datasets (for semantic segmentation) are annotated at various levels of semantic granularity. In the experiments, the label spaces of three datasets (Cityscapes, Vistas, IDD) are combined, as described in Section~\ref{sec:label-taxonomy}, resulting in the taxonomy of Fig.~\ref{fig:hierarchy} (without the MTS dataset). The direct solution of training with the union of datasets and their label spaces is not applicable, since the semantic conflicts among the label spaces introduce ambiguities in the concatenated output label space. For example, the \textit{rider} Cityscapes class conflicts with the Vistas \textit{motorcyclist}/\textit{bicyclist} classes, since they describe the same semantic concepts with different granularity. These conflicts are resolved by our method to deduce a universal taxonomy. To compare all possible dataset combinations, four HToSS networks are trained and compared among each other and with separate single-dataset trainings. The experimental results are shown in Table~\ref{ch6:tab:pp-confl}. For these experiments, the input image size is $799 \times 799$ and batch size formation is 1 image from Cityscapes, 2 from Vistas and 1 from IDD, for each GPU.

The HToSS networks are able to segment, in a single pass, an image over more semantic concepts and with higher attainable mIoU compared to single-dataset counterparts, as indicated by the IoU and \textit{knowledgeability} $\mathcal{K}$ metrics. HToSS networks generalize well, demonstrating mIoU gains of up to +16.6\% for unseen datasets and up to +11.9\% for seen datasets. Moreover, the \textit{knowledgeability} for the HToSS networks increases proportionally to the size of the output label spaces (\eg Columns 2, 4, 12) in the majority of the cases. An exception to these observations is the case of the IDD testing dataset, where the single-dataset training reaches or outperforms HToSS-based networks (Columns 9, 10). After careful visual examination of the dataset, we have observed that the semantic annotations of IDD have a high degree of overlapping concepts. For example, the \textit{road} class and the \textit{drivable fallback} class have partially overlapping definitions,~\ie they both contain semantic atoms like \textit{pothole} or \textit{crosswalk}. This contradicts our hypothesis on assuming non-conflicting semantic class definitions (refer to Eq.~\eqref{eq:hypothesis-1}) and possibly explains the discrepancy in the results. Finally, from the last row of the table, it can be seen that, as the size of the evaluated label space increases, the \textit{knowledgeability} metric remains in equal levels or slightly increases, while the mIoU decreases significantly. This demonstrates the ability of the \textit{knowledgeability} metric to measure the ability of the HToSS-68 network to recognize more classes without being affected by the average deficiency of the mIoU metric. The reader should bear in mind that \textit{c-Knowledgeability} is a comparable metric across datasets, while mIoU is not suited for that purpose.

\subsection{Strong \& weak supervision, non-conflicting label spaces}
\label{ssec:exps-strong-weak-nonconfl}

\begin{table*}
	\centering
	\footnotesize
	\setlength\tabcolsep{1.7pt}
	\begin{tabular}{cc|cc|c||ccccccc}
		\toprule
		\multicolumn{4}{c}{Training datasets} & \multirow{3}{*}{\makecell{Output\\label\\space}} & \multicolumn{3}{c}{WildDash (unseen)} & \multicolumn{4}{c}{Cityscapes (seen)}\\
		\cmidrule(l{3pt}r{3pt}){6-8} \cmidrule(l{3pt}r{3pt}){9-12}
		\multicolumn{2}{c|}{pixel-labeled} & \multicolumn{2}{c|}{bbox-labeled} & & ECP-2 & CitysC-10 & W-19 & ECP-2 & CitysC-10 & \multicolumn{2}{c}{CitysC-20}\\
		Citys & CitysC & ECP & CitysC & & mIoU & mIoU & mIoU & mIoU & mIoU & mIoU & $\mathcal{K}_{20}$\\
		\midrule
		\ding{51} & & & & C-20 & 12.9 & 16.7 & 23.0 & 65.0 & 69.6 & 61.3 & 50.1\\ 
		\ding{51} & \ding{51} & & & C-20 & 13.8 & 17.4 & 23.1 & 65.2 & 70.1 & 61.5 & 51.4\\ 
		\midrule
		\ding{51} & & \ding{51} &  & HToSS-20 & \underline{31.7} & 20.7 & 22.7 & \underline{66.7} & 70.5 & 61.3 & 50.9\\ 
		\ding{51} & & & \ding{51} & HToSS-20 & \underline{31.4} & 20.8 & \underline{24.4} & 65.9 & 70.5 & \underline{63.4} & 53.5\\ 
		\ding{51} & & \ding{51} & \ding{51} & HToSS-20 & 31.1 & \underline{26.2} & 22.4 & 64.0 & \underline{71.6} & 61.7 & \underline{53.7}\\ 
		\bottomrule
	\end{tabular}
	\caption{Overall results on segmentation performance and knowledgeability for the HToSS with pixel-labeled (Cityscapes) and bounding-box-labeled (Cityscapes Coarse, ECP) datasets with non-conflicting label spaces (bottom rows) compared to training with only per-pixel supervision (top rows). The first 2 columns under each evaluated dataset refer to class subsets (ECP-2, CitysC-10) that receive the extra weak supervision from ECP and CitysC. The last column(s) contain results over all classes of the denoted datasets. The pixel-labeled CitysC is used in this experiment only to set the oracle for the experiments involving the weakly-labeled CitysC.}
	\label{ch6:tab:pp-pb-noconfl}
\end{table*}

\begin{table}
	\centering
	\footnotesize
	\setlength\tabcolsep{3.2pt}
	\begin{tabular}{@{}ccc|cccccc|c||cc|c@{}}
		\toprule
		&&& \multicolumn{10}{c}{Cityscapes (seen)} \\
		\multicolumn{3}{c|}{\multirow{2}{*}{Train datasets}} & {\multirow{5}{*}{\rotnighty{Bicycle}}} & \multirow{5}{*}{\rotnighty{Bus}} & \multirow{5}{*}{\rotnighty{Car}} & \multirow{5}{*}{\rotnighty{Motorc.}} & \multirow{5}{*}{\rotnighty{Train}} & \multirow{5}{*}{ \rotnighty{Truck}} & \multirow{5}{*}{\rotnighty{\makecell[l]{Vehicle\\mIoU}}} & \multirow{5}{*}{\rotnighty{Person}} & \multirow{5}{*}{\rotnighty{Rider}} & \multirow{5}{*}{\rotnighty{\makecell[l]{Human\\mIoU}}}\\
		& & & & & & & & & & & & \\
		Citys & \multicolumn{2}{c|}{Open Im.} & & & & & & & & & & \\
		pixel & bbox & tag & & & & & & & & & & \\
		\midrule
		\ding{51} & & & 67.8 & 80.1 & 92.3 & \underline{51.9} & \underline{69.6} & 63.2 & 70.8 & 70.9 & 48.5 & 59.7 \\
		\ding{51} & \ding{51} & & 68.7 & \underline{82.1} & \underline{92.9} & 50.2 & \underline{69.8} & 71.9 & \underline{72.6} & 72.5 & 51.2 & 61.9 \\
		\ding{51} & \ding{51} & \ding{51} & \underline{69.1} & 79.7 & \underline{92.8} & 48.9 & \underline{69.6} & \underline{76.3} & \underline{72.7} & \underline{72.9} & \underline{52.5} & \underline{62.7}\\
		\bottomrule
	\end{tabular}
	\caption{Cityscapes per class mIoU (\%) improvements, for the classes that receive extra supervision from the weakly labeled \textit{OpenScapes} dataset (100k subsets).}
	\label{tab:perf-detail-citys}
\end{table}

This section investigates HToSS on a mix of strongly-labeled and weakly-labeled datasets and non-conflicting label spaces. Strongly-labeled datasets often have classes with few annotated pixels, mostly due to their size/scarcity in the real world. The hypothesis is that weakly-labeled datasets consist a source of examples for underrepresented classes in stronly-labeled datasets, and hence, combined training will increase segmentation performance for these classes. Using the approach developed in Section~\ref{sec:weak-superv} we train HToSS networks on pixel- and bounding-box-labeled Cityscapes, Cityscapes Coarse and ECP datasets (see Table~\ref{tab:datasets-overview}). The results are provided in Table~\ref{ch6:tab:pp-pb-noconfl}. For these experiments, the input image size is $699 \times 699$ and batch size formation is 1 image from Cityscapes, 2 from Cityscapes Coarse, and 2 from ECP, for each GPU.

The first two rows in Table~\ref{ch6:tab:pp-pb-noconfl} show baseline results on pixel-labeled Cityscapes and Cityscapes Coarse datasets, which have 2,000 and 20,000 images, respectively. The last three rows use the HToSS methodology for mixed-supervision training with different combinations of Cityscapes together with either the same-domain Cityscapes Coarse or the different-domain ECP datasets. Performance is evaluated on different subsets of the unseen Wild Dash dataset or the seen Cityscapes dataset. The subsets correspond to the classes that receive extra weak supervision from the bounding-box-labeled datasets. As can be seen, HToSS improves segmentation performance and generalization of the networks without requiring pixel-labeled data. Especially, in the case of Cityscapes + ECP (third row), segmentation performance on the two underrepresented classes of Cityscapes (\textit{person} and \textit{rider}) is increased by +18.8\% for Wild Dash and +1.7\% for Cityscapes. As a conclusion, we can state that HToSS framework can successfully leverage weak aside strong supervision to increase segmentation performance on selected classes (\eg vulnerable road users), without reducing overall performance.


A second experiment, of which the results are provided in Table~\ref{tab:perf-detail-citys}), examines the segmentation accuracy of specific classes when adding weak supervision from bounding boxes and image tags. The weak supervision from the Open Images dataset is increased in steps (second and third row), over strong supervision from Cityscapes. The results demonstrate that on average mIoU performance for classes that receive weak supervision is slightly improved (up to +2.9\%), but specific classes are substantially benefited with an increase of up to +13.2\% IoU. The inclusion of image-tag supervision, improves or maintains IoUs for 6/8 classes, however the improvement is less significant compared to the including only bounding-box supervision. This shows that the weaker and less localized forms of supervision have smaller impact in performance within the HToSS.

To conclude, the performed experiments have shown that weak supervision in conjunction with strong supervision can increase the segmentation performance for the aimed classes, while this approach maintains or even slightly increases the overall performance for all classes. Moreover, it is observed that weak supervision on the same image domain (Cityscapes + Cityscapes Coarse) yields higher segmentation performance gains.

\begin{table*}
	\centering
	\footnotesize
	\setlength\tabcolsep{4.0pt}
	\begin{tabular}{cc|cc|c||ccccc}
		\toprule
		\multicolumn{4}{c|}{Training datasets} & \multirow{3}{*}{\makecell{Output\\label\\space}} & \multicolumn{1}{c}{W.Dash*} & \multicolumn{4}{c}{Cityscapes T. Signs (seen)}\\
		\cmidrule(l{3pt}r{3pt}){6-6} \cmidrule(l{3pt}r{3pt}){7-10}
		\multicolumn{2}{c|}{pixel-labeled} & \multicolumn{2}{c|}{bbox-labeled} &  & W-19 &C-20 & CitysT-14 & \multicolumn{2}{c}{CitysT-34}\\
		Citys & CitysT & MTS & CitysT &  & mIoU & mIoU & mIoU & mIoU & $\mathcal{K}_{34}$\\
		\midrule
		\ding{51} & & & & C-20 & 27.1 & 70.3 & n/a & n/a & 39.3\\
		\ding{51} & \ding{51} & & & CT-34 & 27.8 & 69.5 & \underline{17.7} & 47.5 & 46.2\\
		\midrule
		\ding{51} & & \ding{51} & & HToSS-70 & 28.9 & \underline{70.7} & 11.6 & 45.6 & 44.3\\
		\ding{51} & & & \ding{51} & HToSS-34 & \underline{30.2} & 69.8 & 17.0 & \underline{46.9} & \underline{44.5}\\
		\bottomrule
	\end{tabular}
	\caption{Overall results on segmentation performance and \textit{knowledgeability} for the HToSS on pixel-labeled (Cityscapes) and bounding-box-labeled (Cityscapes traffic signs, Mapillary traffic signs) datasets with conflicting label spaces (bottom rows) compared to training with only per-pixel supervision (top rows). Cityscapes Traffic Signs (CitysT) is originally a pixel-labeled dataset. The version appearing under the bbox-labeled column contains only the traffic sign classes and the converted pixel labels to bounding boxes~\cite{meletis2019boosting}. * (unseen dataset)}
	\label{tab:combine-annotations-conflicting-1}
\end{table*}

\subsection{Strong \& weak supervision, conflicting label spaces}
\label{ssec:exps-combine-supervision}
The most challenging scenario involves simultaneously training with strong and weak supervision and conflicting label spaces. At the same time, this is the most rewarding scenario, since label spaces of weakly-labeled datasets can increase the output label space with numerous semantic concepts that do not exist in pixel-labeled datasets (see Figure~\ref{fig:datasets-statistics}). To examine this scenario, we augment the label space of Cityscapes with traffic sign classes from the bounding-box-labeled datasets: MTS (50) and Cityscapes Traffic Signs (14). In this case, the methodology developed in Section~\ref{ssec:general-case} is utilized, and the network is trained with input image size of $599 \times 599$ and batch size formation is 1 image from Cityscapes, 3 from Cityscapes Traffic Signs, and 3 from Mapillary Traffic Signs, for each GPU.

Table~\ref{tab:combine-annotations-conflicting-1} shows the results of models trained on Cityscapes and on combinations of Cityscapes Traffic Signs and MTS.  The combined label space is part of HToSS-68 (Figure~\ref{fig:hierarchy}). We evaluate all networks on the original Cityscapes classes (C-20), the traffic sign classes subset (CitysT-14), all CitysT-34 classes (CitysT-14 traffic signs + C-20), as well as, on the unseen Wild Dash, whose classes are a subset of Cityscapes (W-19 $\subset$ \mbox{C-20}). Comparing the last two rows and the first row, demonstrates that the \textit{knowledgeability} increases by +5\% for the HToSS label spaces, while generalization performance (first column) is also increased by at least +1.8\%.

\begin{figure*}
	\begin{subfigure}{0.47\linewidth}
		\includegraphics[width=\linewidth]{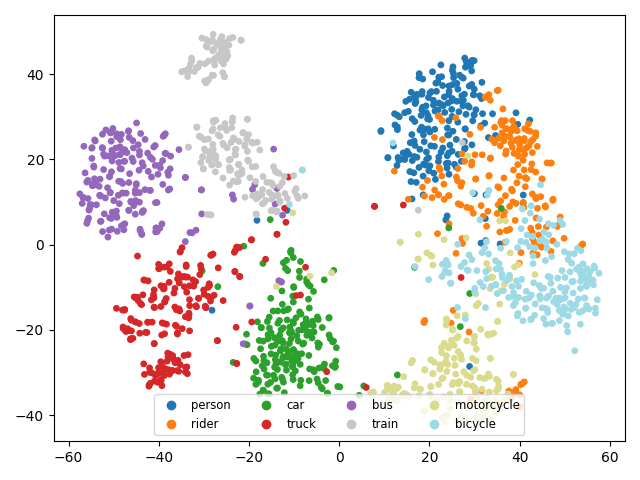} 
		\caption{Single dataset training on Cityscapes.\\~}
		\label{fig:subim1}
	\end{subfigure} ~
	\begin{subfigure}{0.47\linewidth}
		\includegraphics[width=\linewidth]{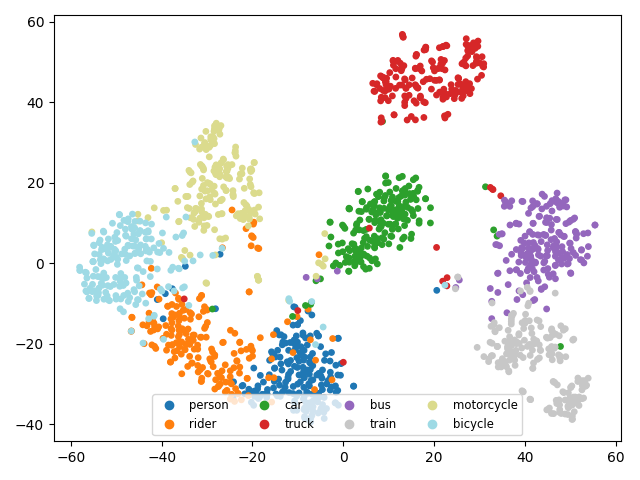}
		\caption{HToSS on Cityscapes + Vistas + IDD.\\~}
		\label{fig:subim2}
	\end{subfigure} ~
	\begin{subfigure}{0.45\linewidth}
		\includegraphics[width=\linewidth]{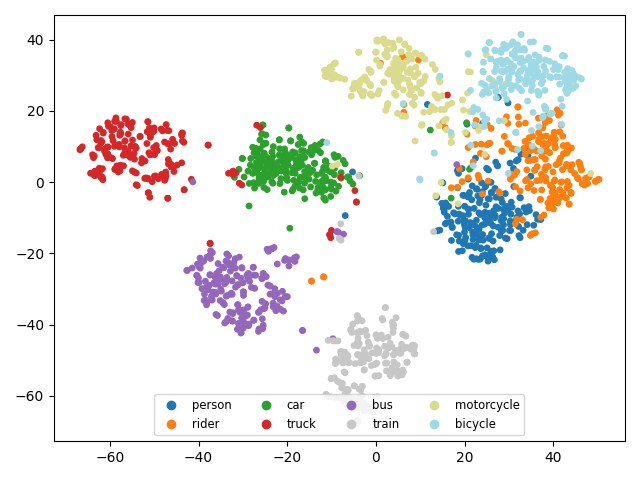}
		\caption{HToSS on Cityscapes + EuroCity Persons + Cityscapes Coarse.}
		\label{fig:subim3}
	\end{subfigure}
	\caption{Progress of features, while adding more datasets, as a 2-D t-SNE visualization (using same t-SNE hyper-parameters). Clusters become less scattered (intra-class distance) and better separated (inter-class distance).}
	\label{fig:tsne}
\end{figure*}

\begin{table}
	\centering
	\footnotesize
	\begin{tabular}{ll|cccc}
		\toprule
		Model & \makecell{Conflicts\\resolution} & \makecell{Memory\\$\Delta$ Params} & \makecell{Inference\\$\Delta$ ms} & \makecell{Unseen dts.\\mmIoU} & $\mathcal{K}^{66}$\\
		\midrule
		\makecell[l]{one per\\dataset} & \makecell[l]{post-proc.\\merging} & $+ 5.4 \cdot 10^7$ & $+ 127.1$ & 46.2 & 62.6\\ 
		\midrule
		\multirow{2}{*}[2pt]{\makecell[l]{shared\\backbone,\\head per\\dataset}} & \makecell[l]{common\\classes} & $+ 2.1 \cdot 10^5$ & $+ 1.7$ & 34.2 & 45.6\\
		\cmidrule{2-6}
		& \makecell[l]{post-proc.\\merging} & $+ 2.1 \cdot 10^5$ & $+ 1.8$ & 45.4 & 62.8\\ 
		\midrule
		\multirow{2}{*}[2pt]{\makecell[l]{shared\\backbone,\\single\\head}} & \makecell[l]{common\\classes} & \textit{reference} & \textit{reference} & 30.2 & 42.1\\ 
		\cmidrule{2-6}
		& \makecell[l]{\textbf{semantic}\\\textbf{atoms (ours)}} & $+ 5.1 \cdot 10^3$ & $+ 0.0$ & 50.5 & 68.3 \\ 
		\bottomrule
	\end{tabular}
	\caption{Common baselines methodologies for combining three pixel-labeled datasets (Cityscapes, Vistas, IDD) with conflicting label spaces. All methods use ResNet-50 backbones and softmax classifiers. The $\Delta$'s for the total number of parameters (Params) and the single-image inference time (ms) are w.r.t. the \textit{reference} row 4,~\ie keeping only the common, non-conflicting classes from all datasets.}
	\label{tab:comb-labels-2}
\end{table}


\subsection{Ablations and Insights}
\label{ssec:exps-ablations}
An analysis is provided for the conducted experiments. First, an ablation on how the amount of weak supervision affects performance is presented in Table~\ref{tab:size-matters} for the experiment of Section~\ref{ssec:exps-combine-supervision}. An increasing number of images and bounding boxes from the weakly-labeled dataset are added per step. We observe that as weakly-labeled images are included, the segmentation performance increases accordingly.

Second, we provide t-SNE plots in Fig.~\ref{fig:tsne} for experiments from Table~\ref{tab:comb-labels-1}.
The plots capture the 2-D projections of the output of feature extractor before and after adding multiple datasets. It can be observed that the representations when adding more datasets have better properties from the perspective of classificability/separability. This may be explained by the increase in the variety of examples for the majority of the \textit{semantic atoms} used from auxiliary datasets. The desirable properties of the t-SNE plots are reflected in metric results as well.

Finally, we provide comparisons of the HToSS methodology against various baselines in Table~\ref{tab:comb-labels-2} examining memory and time factors wrt. the attained performance. The first three rows describe direct solutions using existing trained networks and post processing for solving conflicts. The single-network approach (fourth row) is the closest to HToSS, but resolves conflicts by maintaining only the common classes. This leads to a significant loss in \textit{knowledgeability}, as the number of recognizable classes reduce. Overall, the HToSS approach uses a reduced number of parameters and performs fast inference, since it uses a common backbone and a single classifier.

\begin{table}
	\centering
	\small
	\begin{tabular}{cc||cc}
		\toprule
		pixel & bbox-labeled & \multicolumn{2}{c}{Cityscapes (C-20)}\\
		Citys & Open Images & mAcc & mIoU \\
		\midrule
		\ding{51} & - & 81.2 & 70.2\\
		\midrule
		\ding{51} & $1k$ images ($17.3k$ bboxes) & 80.7 & 69.8\\ 
		\ding{51} & $10k$ images ($140.4k$ bboxes) & 81.6 & 70.6\\ 
		\ding{51} & $100k$ images ($1185.8k$ bboxes) & 83.7 & 72.3\\ 
		\bottomrule
	\end{tabular}
	\caption{Segmentation accuracy with different number of bounding boxes used to generate pseudo labels from the weakly-labeled Open Images.}
	\label{tab:size-matters}
\end{table}

\section{Conclusions}
\label{ch5:sec:conclusions}
This chapter concludes the line of research on addressing generalized semantic segmentation and complements the proposed methods from Chapters~\ref{ch:3-iv2018} and~\ref{ch:4-iv2019}. This chapter has proposed a complete methodology, namely Heterogeneous Training of Semantic Segmentation (HToSS), for combining an arbitrary number of strongly-labeled and weakly-labeled datasets, including datasets with pixel-level, bounding-box-level, and image-level annotations. Simultaneous training of HToSS networks with these datasets is achieved, irrespective of any conflicts in their label spaces, which are resolved with the concept of \textit{semantic atoms}. This concept together with weak supervision handling facilitates the applicability of the proposed algorithms in a large number of existing image recognition datasets. The methodology achieves universal segmentation by making the single hypothesis that the semantic classes of weakly-labeled datasets contain semantic concepts of one or more classes from the strongly-labeled datasets.

The experimentation on multiple combinations of 8 training datasets have demonstrated that the segmentation performance (IoU) is increased in accordance with the number of datasets added in the combination. Moreover, that increase is larger when the networks are evaluated in the same or similar domains to those of the training datasets. Furthermore, an interesting finding with respect to generalization, concerns the large IoU improvements in specific datasets (WildDash) when weak supervision is used. We hypothesize that this may be explained by the robustness that noisy examples enforce in the learning process. The new metric \textit{Knowledgeability} behaves in agreement with the desire to distinguish more semantic concepts in the scenes, having as a requirement to segment these concepts sufficiently.

The contributions of this chapter can be summarized as follows. 
\begin{enumerate}[noitemsep,topsep=0pt]
\item Formulation and characterization of challenges of the generalized semantic segmentation problem for simultaneous heterogeneous multi-dataset training.
\item A methodology for combining label spaces with variable semantic level of detail and with different classes, thereby enabling simultaneous training on datasets with disjoint or conflicting label spaces.
\item A methodology for consolidating strong (pixel) and weak (bounding-box or image-tag) supervision, which facilitates simultaneous training on datasets with mixed supervision.
\item A novel metric that quantifies \textit{Knowledgeability} of a network predictions,~\ie the number of recognizable semantic classes, while maintaining achievable performance for these classes, which can be used to compare the performance of a network across datasets.
\end{enumerate}

~

This chapter has achieved training on heterogeneous datasets which are intended for the task of semantic segmentation. HToSS allows supplementing the pixel-labeled training data with other relevant datasets that otherwise would not be compatible. These properties make the HToSS approach useful to many applications in which training data for semantic segmentation are too scarce to achieve required performance. The following chapter takes one step ahead and attempts to generalize the task of Panoptic Segmentation to enrich it with object-part-aware semantics. Both chapters pave the wave to generalized scene understanding in computer vision.

\begin{savequote}[8cm]
And Hera set a watcher upon her, great and strong Argus Panoptes, who with four eyes looks every way. And the goddess stirred in him unwearying strength: sleep never fell upon his eyes; but he kept sure watch always.

\textit{Hesiod, Aegimius}
\end{savequote}

\chapter{Part-aware panoptic segmentation} 
\label{ch:7-panoptic}

\section{Introduction}
\freefootnote{\hspace*{-15pt}The contributions of this chapter were integrated in a joint paper with three other researchers, Daan de Geus, Chenyang Lu, and Xiaoxiao Wen, for publication in the Proceedings of Int. Conf. CVPR 2021.}
The presented methodologies in the previous chapters aimed at enabling the training of fully convolutional networks with a variety of heterogeneous datasets for semantic segmentation. Specifically, these methodologies expanded the admissible annotation formats and label spaces of datasets and allowed the inclusion of an arbitrary number of datasets in a single training round. This final chapter generalizes panoptic segmentation, which is a super-task of regular semantic segmentation. This generalization of the segmentation introduces part semantics, which paves the way towards achieving holistic scene understanding.

The task of semantic segmentation, which has been studied in the previous chapters, considers the semantics of a scene with pixel-level precision. Although scene semantics are crucial for scene understanding and subsequently performing higher-level tasks,~\eg control or navigation, they do not provide a rich multi-level representation of the scene. For example, knowing which pixels of an image belong to the semantic class \textit{car} does not include any information about how many cars are in the scene, or which pixels belong to which specific car instance. The task that extends semantic segmentation and satisfies these requirements is manifested in panoptic segmentation~\cite{Kirillov2019PS}. Panoptic segmentation achieves scene-level parsing by requiring semantic and instance information to be predicted simultaneously in a pixel-wise manner. However, the question is whether this prediction is sufficient for decision-making systems, like assessing the intention of a pedestrian or the trajectory of a vehicle.

Humans distinguish easily the semantic parts of the objects in their surroundings and create a description of a scene at multiple levels of abstraction. This in turn helps them to reason about intention and trajectory of other dynamic objects, leading to a better anticipation of the future. For example, a pedestrian will see parts of a car (turning lights, wheels) and consequently he will not cross a street if he anticipates a car to turn. As another example, a driver will stop his car if he expects a dog or a person to cross the street by observing their legs, torso, or arms. It becomes clear from these examples that semantic parts of scene participants play an essential role for any system that claims to achieve holistic scene understanding.

\textit{Panoptic segmentation} captures semantic and instance information, but operates only at the scene-level abstraction. \textit{Parts parsing} in the original formulation captures only semantic information on the part-level abstraction for a single object class (\eg person parts segmentation~\cite{chen2014detect}). Recently, \textit{parts parsing} has been extended for multiple object classes in an instance-agnostic manner~\cite{michieli2020gmnet, Zhao2019BSANet, chen2014detect} or has become instance-aware~\cite{gong2018instance, li2017holistic, zhao2018understanding}. 

\begin{figure}
	\centering
		\begin{tabular}{c@{\extracolsep{4pt}}c}
			\centering
			
			\includegraphics[width=0.48\linewidth]{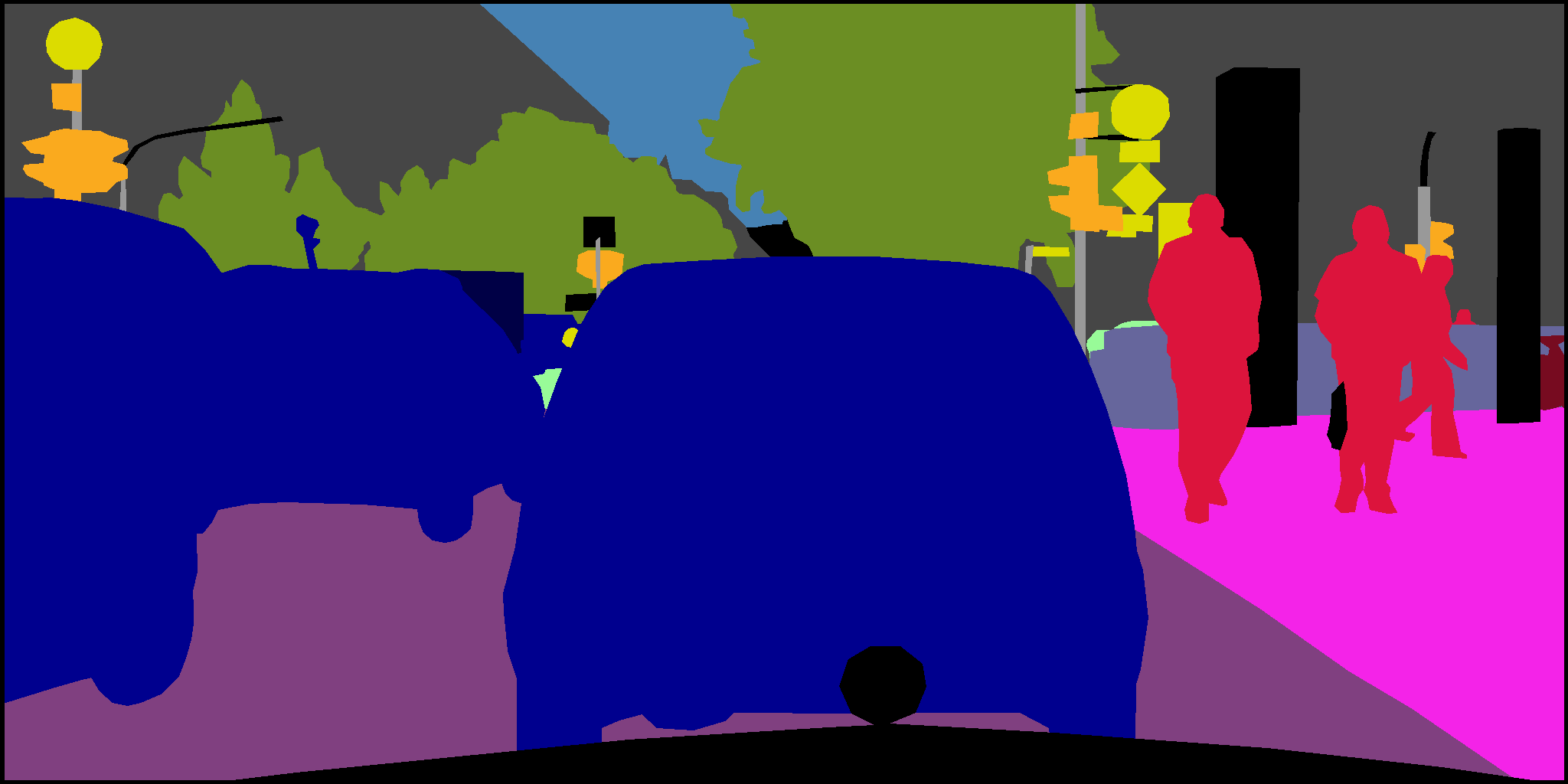} &
			\includegraphics[width=0.48\linewidth]{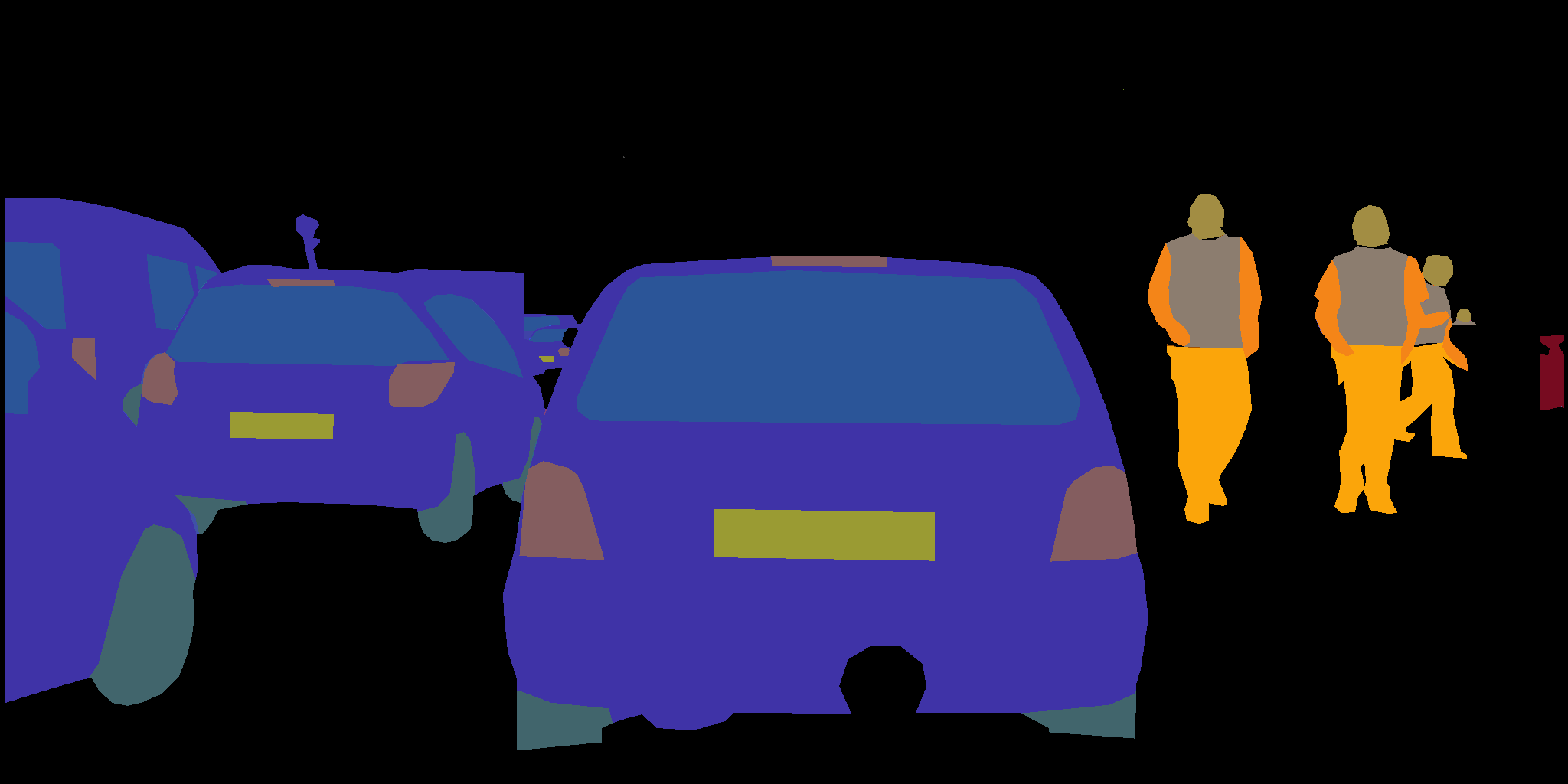}\\
			
			{\footnotesize Semantic Segmentation} & {\footnotesize Part Segmentation} \\

			\includegraphics[width=0.48\linewidth]{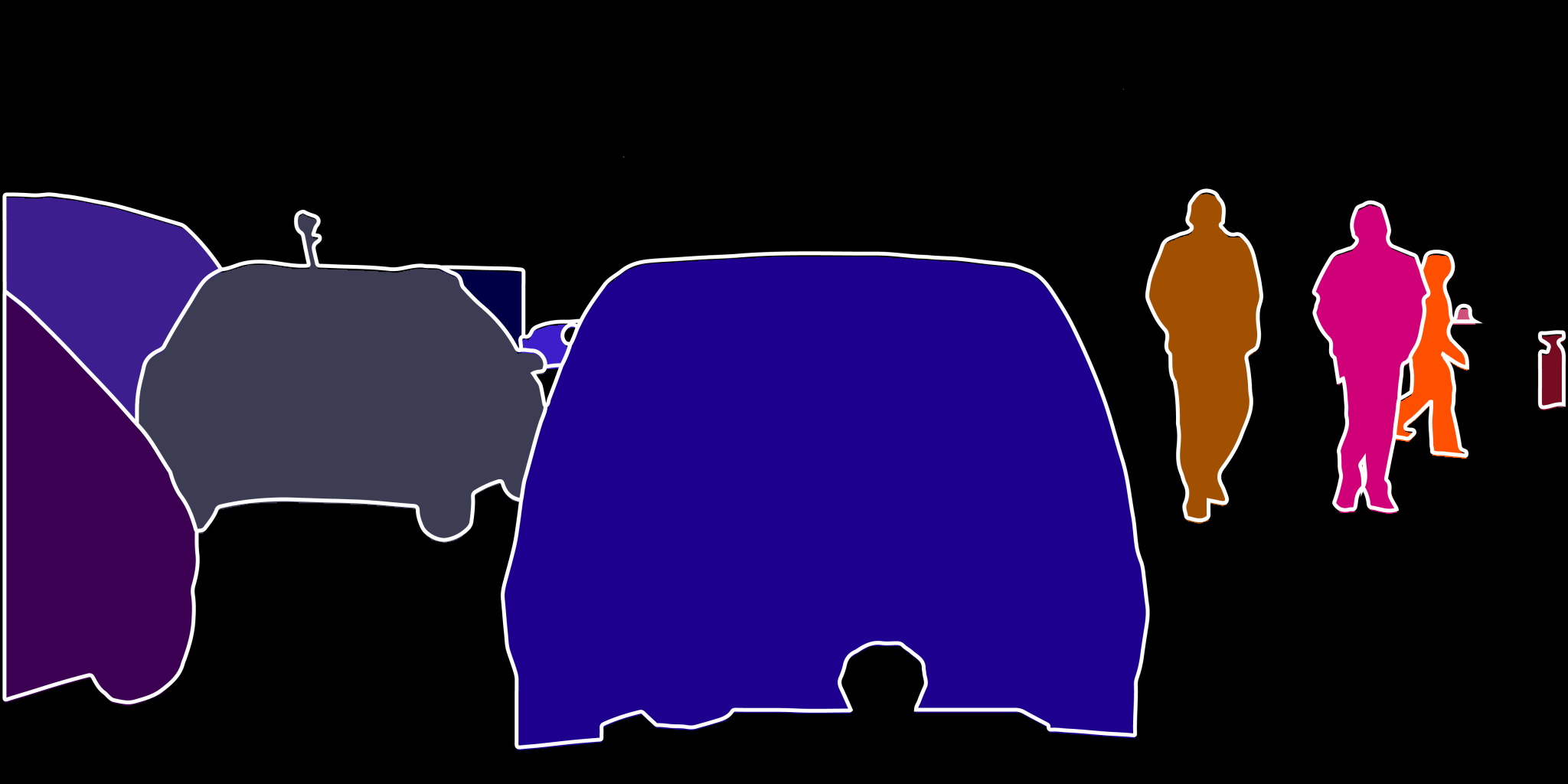} &
			\includegraphics[width=0.48\linewidth]{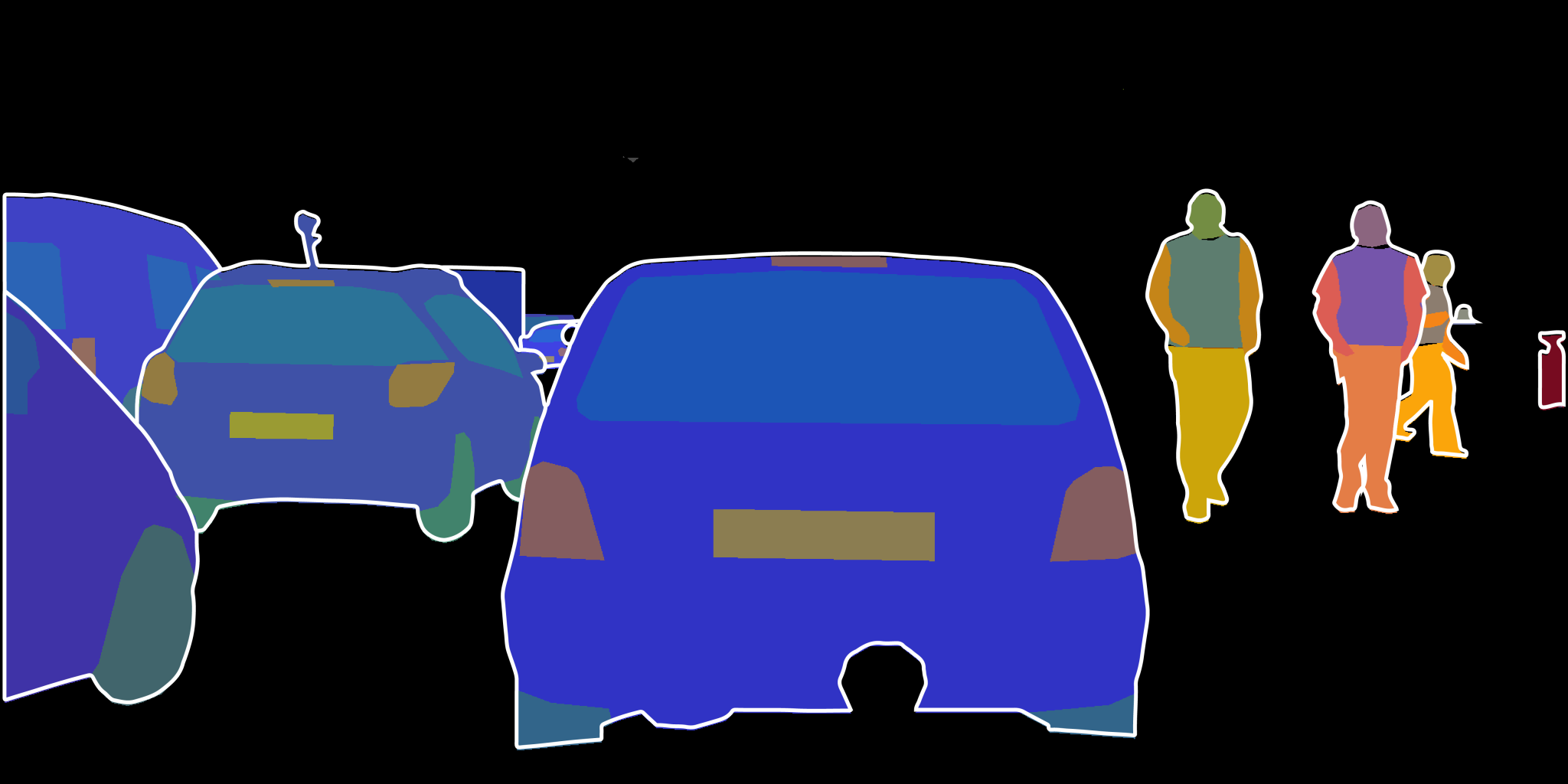}\\
			
			{\footnotesize Instance Segmentation} & {\footnotesize Instance-aware Part Segmentation} \\
			
			\includegraphics[width=0.48\linewidth]{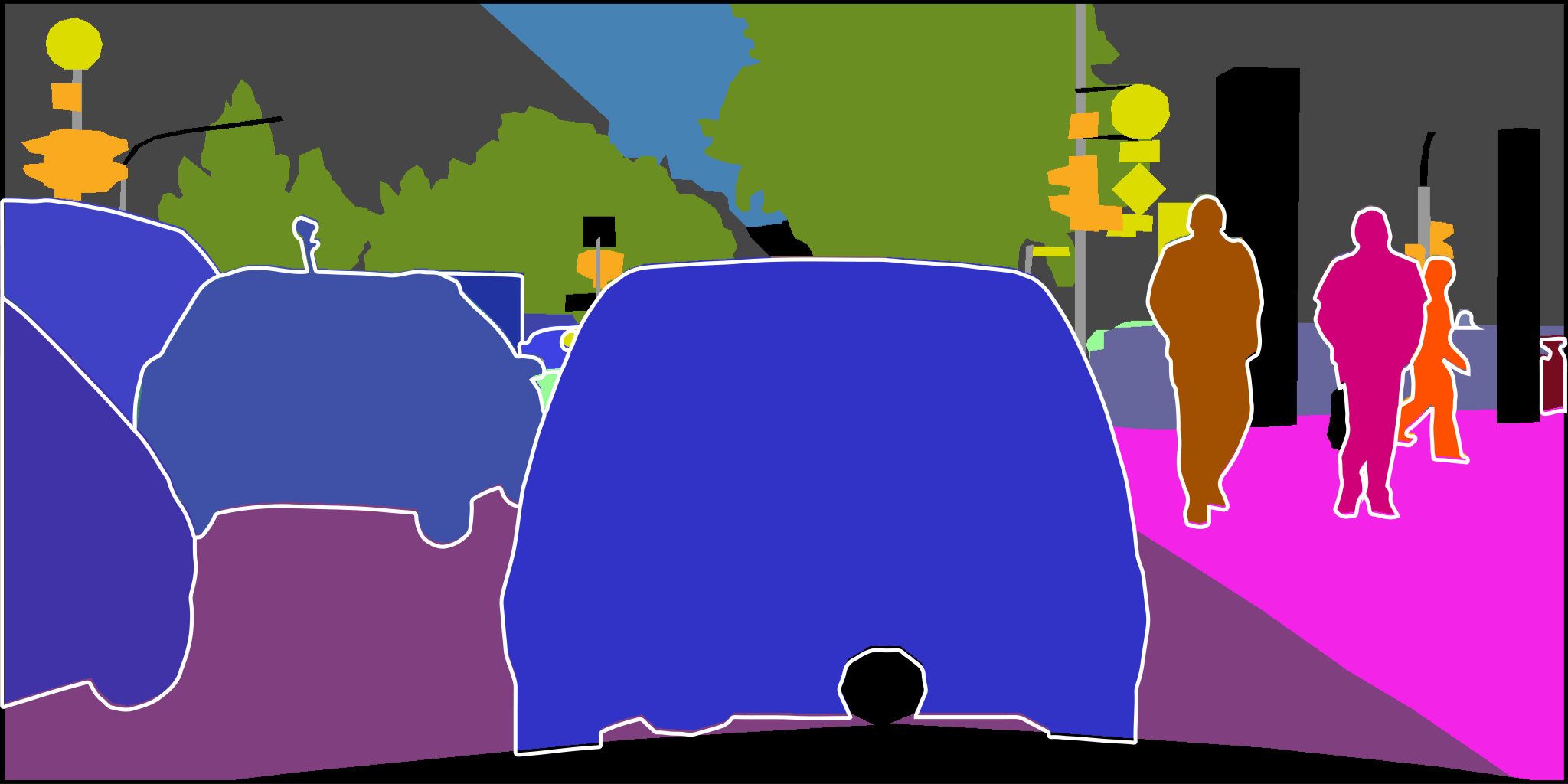} &
			\includegraphics[width=0.48\linewidth]{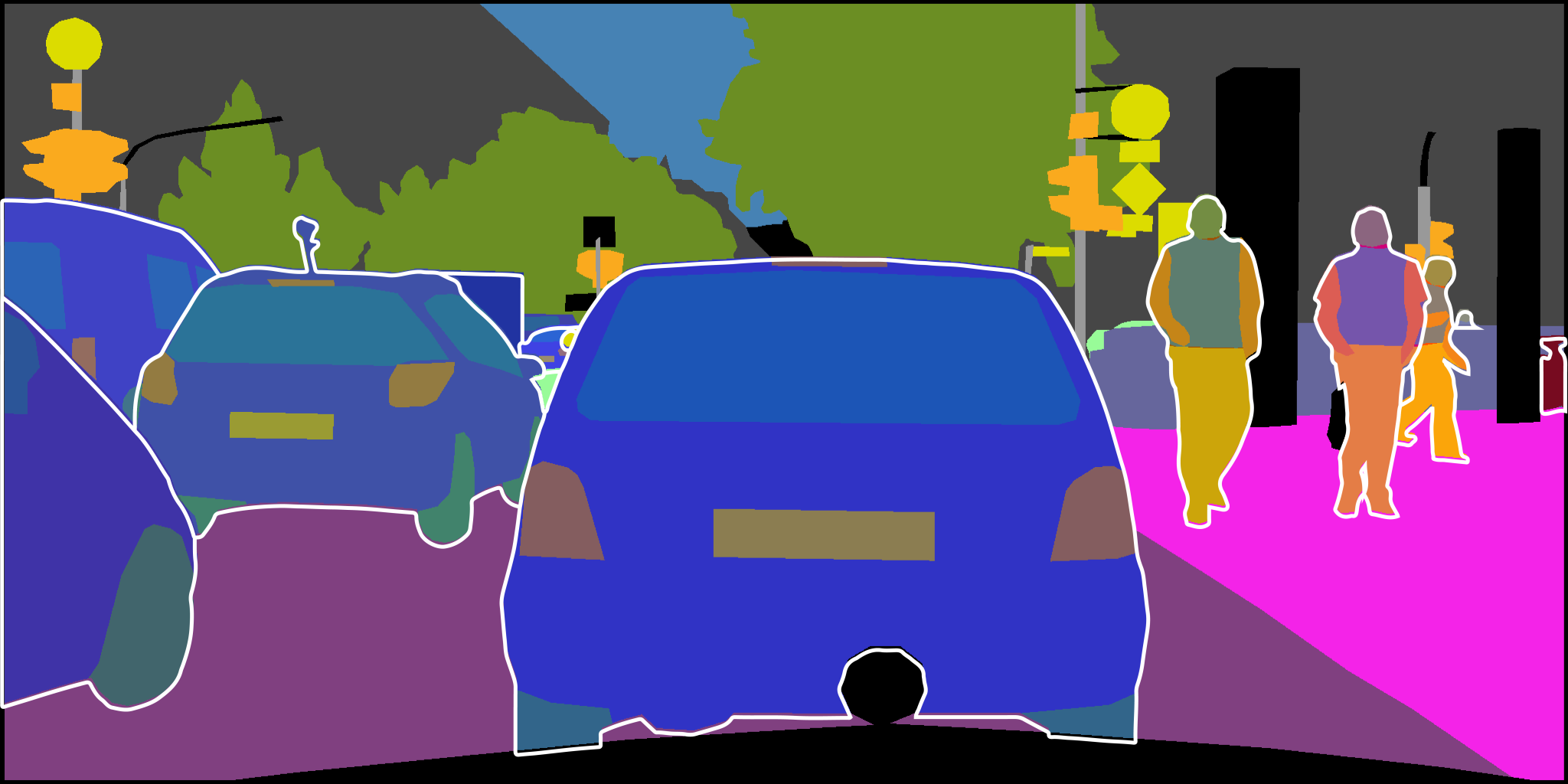}\\
			{\footnotesize{Panoptic Segmentation}} & {\footnotesize{\textbf{Part-aware Panoptic Segmentation}}} \\
		\end{tabular}
	\caption{\textit{Evolution of scene understanding tasks}: from semantic to panoptic (top to bottom) and from part-agnostic to part-aware (left to right). Colors indicate scene-level and part-level semantics. Instance-level boundaries are emphasized with a white contour.}
	\label{ch7:fig:task-def}
\end{figure}

The current chapter proposes the novel task of \textit{Part-aware Panoptic Segmentation} (PPS), which combines \textit{scene parsing} and \textit{parts parsing}. PPS encompasses the 1) scene-level classification of each image pixel, 2) scene-level clustering of \textit{things} pixels into individual instances, and 3) part-level classification of pixels belonging to scene-level classes with parts. The previous three steps are important and will be addressed later in this chapter. On one hand, PPS generalizes panoptic segmentation with part-level semantics. On the other hand, PPS generalizes (instance-aware) parts segmentation with scene-level \textit{stuff} classes semantics and multiple \textit{things} classes. Figure~\ref{ch7:fig:task-def} visualizes the conceptual differences between PPS and related scene understanding tasks.

To allow for research on the new task of PPS, we introduce consistent part-aware panoptic annotations for two commonly used datasets. For Cityscapes \cite{Cordts2016Cityscapes}, which contains urban street scenes, we have labeled part classes for all 3,475 images of the training and validation set. These annotations are a superset of the existing panoptic annotations. For Pascal VOC 2010~\cite{Everingham2010Pascal}, which contains generic/every-day scenes, we have combined the existing datasets for semantic segmentation~\cite{mottaghi14pascalcontext} and instance-aware part segmentation~\cite{chen2014detect} to generate unambiguous and consistent annotations for PPS. Section~\ref{ch7:sec:datasets} provides further details and statistics on these datasets. Moreover, this chapter establishes several benchmarks and baselines based on state-of-the-art panoptic segmentation or instance-aware part segmentation networks.

This leads to the follow problem statements for this chapter.
\begin{itemize}[noitemsep,topsep=0pt]
\item How can \textit{Part-aware Panoptic Segmentation} be formulated such that it consists of a superset of related scene understanding tasks and does not conflict with them while requiring richer predictions?
\item Are existing datasets adequate for providing sufficient information to networks for training or validating for \textit{Part-aware Panoptic Segmentation}?
\item When creating baselines for this new task, what are the design choices that have to be made for generating predictions that satisfy complete task requirements?
\end{itemize}

This chapter is organized as follows. Section~\ref{ch7:sec:rel-work} reviews related scene understanding tasks and datasets. In Section~\ref{ch7:sec:prob-def}, a precise mathematical formulation of PPS is given. The details of combining and annotating the datasets, together with statistics and comparisons are described in Section~\ref{ch7:sec:datasets}. The PPS baseline architectures and results are provided in Section~\ref{ch7:sec:experiments}. Finally, Section~\ref{ch7:sec:conclusions} summarizes the findings from introducing the PPS task and concludes the chapter.

\section{Related work}
\label{ch7:sec:rel-work}
Visual scene understanding aims to extract rich and all-encompassing information from images with the long-term goal to reach human-level holistic understanding. Scene understanding has been researched at various levels of semantics, either scene or parts, and on different abstractions (refer to Section~\ref{ch1:sec:scene-und}). However, different abstractions are not frequently addressed at the same time. In this chapter, a single coherent task encompassing multiple levels of abstraction is proposed, which unifies scene parsing and part parsing. 

\subsection{Scene parsing}
Scene parsing refers to tasks that address scene understanding at the scene-level classes and distinguish between individual instances of \textit{things}. The recently proposed task of panoptic segmentation~\cite{Kirillov2019PS}, a unification of the typically distinct tasks of semantic segmentation and instance segmentation, is a scene parsing task. In earlier forms, this task has been investigated in~\cite{tu2005image, yao2012describing}.

Early works on panoptic segmentation applied a multi-task network that trains and outputs instance segmentation and semantic segmentation in parallel, followed by a merging operation to generate panoptic segmentation results~\cite{degeus2019single, Kirillov2019PanopticFPN, li2019attention, mohan2020efficientps, porzi2019seamless}. These works were followed by optimizing the process of merging to panoptic segmentation~\cite{Lazarow2020OCFusion, liu2019end, xiong2019upsnet, Yang2020SOGNet}, or try to solve the task more efficiently~\cite{Cheng2020PanopticDeepLab, degeus2020fpsnet, Hou2020Real, Li2020Unifying, yang2019deeperlab}. Although, panoptic segmentation covers many aspects of scene understanding, it does not analyze a scene with part-level semantics, which have a vital role in planning and reasoning,~\eg future anticipation of pedestrians in street scenes.

\subsection{Part parsing}
Part parsing refers to tasks that address scene understanding based on part-level semantics. Two representative tasks are part segmentation and pose estimation. Part segmentation requires a pixel-level prediction for all identified parts, whereas pose estimation aims at the detection of connected keypoints between the parts for each object. Pose estimation is inherently instance-aware and is surveyed in~\cite{dang2019deep, liu2015survey}. 

Part segmentation is usually treated as a semantic segmentation problem~\cite{gong2019graphonomy, jiang2018cnn, jiang2019cnn, li2017holistic, liang2015human, liu2018cross, luo2013pedestrian, luo2018trusted, michieli2020gmnet, Zhao2019BSANet}, and as such it remained for a long time instance-agnostic. The related dense pose task was introduced in~\cite{alp2018densepose} and a unification of pose estimation and part segmentation was provided in~\cite{dong2014towards}. Only recently, to the best of our knowledge, an instance-aware human-part segmentation task was introduced and studied in~\cite{gong2018instance, li2017holistic, zhao2018understanding}. Most research has focused on part segmentation for humans~\cite{dong2013deformable, gong2018instance, ladicky2013human, li2018multi, li2019self, liang2016semantic, liang2018look, lin2019cross, ruan2019devil, yang2019parsing, zhao2018understanding}, but other parts have also received attention, \eg facial parts~\cite{lin2019face}, and animal parts~\cite{chen2014detect, wang2015joint}. A limited amount of papers have addressed multi-class part segmentation~\cite{michieli2020gmnet, Zhao2019BSANet}, but up to this point these methods are not instance-aware. As a result, instance-aware part segmentation on a more general dataset, consisting of a wider range of classes and parts, remains an open issue.

\subsection{Datasets}
Part-aware panoptic segmentation requires scene-level and part-level semantics, as well as instance-level object enumeration. Training and evaluation on this new task cannot be straightaway accomplished with existing datasets. Although a plethora of datasets exist for object detection and semantic segmentation, only few have labels compatible with the panoptic segmentation task,~\eg~\cite{Cordts2016Cityscapes, Lin2014COCO}. For part-level segmentation, the datasets are even more scarce. LIP~\cite{liang2018look}, MHP~\cite{zhao2018understanding} and CIHP~\cite{gong2018instance} provide instance-aware, part-level annotations, but only for human parts. To the best of our knowledge, Pascal-Parts is the only dataset that has part-level annotations for a more general set of classes~\cite{chen2014detect}. However, these annotations do not contain any information on classes without parts.

From the aforementioned observations, it can be deduced that there is no dataset that covers all the requirements for the PPS task. To enable single-dataset consistent evaluation and training, we present part-aware panoptic annotations on two datasets. First, Cityscapes~\cite{Cordts2016Cityscapes}, a commonly used dataset for panoptic segmentation, is extended by manually annotating parts for five different \textit{things} classes. Second, Pascal VOC~\cite{Everingham2010Pascal} is extended by collecting and merging different annotation sub-sets to generate a complete and consistent annotation set its 10,103 images. An overview of the proposed datasets and comparisons with related datasets can be found in Table~\ref{ch7:tab:related-datasets}.

\section{Task definition and metrics}
\label{ch7:sec:prob-def}

\subsection{Task definition of Part-aware Panoptic Segmentation}
The task of \textit{Part-aware Panoptic Segmentation} (PPS) is a scene-understanding task that is designed to encapsulate information at two visual abstraction levels, namely the scene and the part levels. Specifically, it captures the following information layers: 1) scene-level semantics, 2) scene-level instance clustering, and 3) part-level semantics. To achieve this, the PPS task is based on panoptic segmentation~\cite{Kirillov2019PS}, which is enriched with part-level semantics. 

A part-aware panoptic segmentation algorithm describes every pixel in an image with a set jointly containing semantic and instance information. This can be expressed for pixel $i$ in the form $(l, p, z)_i$, where label $l$ represents the scene-level semantic class, label $p$ the part-level semantic class, and $z \in \mathbb{N}$ the \textit{identity} of the instance (\textit{ID}). The scene-level and part-level semantic classes are predefined and usually correspond to the available semantic granularity of the labels of the considered dataset, while the instance \textit{ID} is an unbounded integer separating, per image, distinct instances of the same scene-level semantic class.

The scene-level semantic class $l$ is chosen from a predetermined set of $\mathcal{L} := \{l_1,\dots,l_{L}\}$ classes. For any of these classes, a set of part-level semantic classes $\mathcal{P}_l := \{p_{l,1},\dots,p_{l,P_l}\}$ containing $P_l$ semantic parts may be defined. The superset of all parts is denoted as $\mathfrak{P} = \cup_l \mathcal{P}_l$, with $l \in \mathcal{L}$. The set $\mathcal{L}$ can be separated into disjoint subsets in two different ways. First, $\mathcal{L} = \mathcal{L}^\text{St} \cup \mathcal{L}^\text{Th}$, where $\mathcal{L}^\text{St}$ consists of the stuff classes, \ie uncountable entities with amorphous shape (\eg sky, sea), while $\mathcal{L}^\text{Th}$ contains the things classes, which are classes for countable objects with a well-defined shape (\eg car, person). Second, $\mathcal{L}$ can be also separated in a subset of scene-level classes that have parts (\eg limbs, car parts), $\mathcal{L}^\text{parts}$, and scene-level classes that do not have parts, $\mathcal{L}^\text{no-parts}$. Consequently, it should hold that $\mathcal{L} = \mathcal{L}^\text{parts} \cup \mathcal{L}^\text{no-parts}$. We require that both $\mathcal{L}^\text{St} \cap \mathcal{L}^\text{Th} = \emptyset$ and $\mathcal{L}^\text{parts} \cap \mathcal{L}^\text{no-parts} = \emptyset$ hold (non-overlapping sets). The selection of classes belonging to the four subsets $\mathcal{L}^\text{St}$, $\mathcal{L}^\text{Th}$, $\mathcal{L}^\text{parts}$, $\mathcal{L}^\text{no-parts}$ is a design choice that is typically determined based on the requirements of the application, or the purpose of a dataset, as for~\cite{Kirillov2019PS}. 

A PPS algorithm makes predictions for an image, which adheres to the following requirements: 1) a scene-level semantic class from $\mathcal{L}$ must be assigned to all image pixels, 2) a scene-level instance \textit{ID} is provided only for pixels that are assigned a scene-level class from $\mathcal{L}^\text{Th}$, and 3) a part-level semantic class must be assigned only to pixels that are assigned a scene-level class from $\mathcal{L}^\text{parts}$. In summary, a pixel can be labeled with one of the following combinations:
\begin{itemize}[noitemsep,topsep=0pt]
	\item Stuff class: $(l, -, -)$, where $l \in \mathcal{L}^{St}$;
	\item Stuff class with parts: $(l, p, -)$, where $l \in \mathcal{L}^{St} \cap \mathcal{L}^\text{parts}$ with $p \in \mathcal{P}_l$;
	\item Things class: $(l, -, z)$, where $l \in \mathcal{L}^{Th}$ with $z \in \mathbb{N}$;
	\item Things class with parts: $(l, p, z)$, where $l \in \mathcal{L}^{Th} \cap \mathcal{L}^\text{parts}$ with $p \in \mathcal{P}_l$,
\end{itemize}
where ``$-$'' denotes that the involved specific information is irrelevant. Finally, the PPS format accepts a special \textit{void} label for scene-level and part-level semantics, which represents ambiguous pixels or concepts not included in any subset $\mathcal{L}$.

\subsection{Relationship to other scene understanding tasks}
\textit{Part-aware panoptic segmentation} (PPS) is related to and generalizes various per-pixel segmentation tasks. \textit{Part segmentation} is specialized semantic segmentation focusing on segmenting object parts, but it does not require separating parts according to the object instance they belong to. In the PPS format, this task can be described as $(l, p, -)_i$, $l \in \mathcal{L}^\text{parts}$, $p \in \mathfrak{P}$. \textit{Instance-aware part segmentation}, can be described as $(l, p, z)_i$, $l \in \mathcal{L}^\text{Th} \cap \mathcal{L}^\text{parts}$, $p \in \mathfrak{P}$, and pivots part parsing at an instance level, but treats any non-things pixel as background, losing environmental context. Finally, \textit{panoptic segmentation}, $(l, -, z)_i$, $l \in \mathcal{L}$, includes no notion of part semantics.

\subsection{Metrics}
\label{ch7:ssec:metrics}
Two metric families are used to assess the accuracy of a system prediction for PPS. The first, Part-aware Panoptic Quality, is proposed in~\cite{us2021part} and is briefly discussed in this section. The second is the conventional IoU-based metric for semantic segmentation, which is adapted to the multi-level abstraction setting.

\subsubsection{A. Part-aware Panoptic Quality}
Part-aware Panoptic Quality (PartPQ)~\cite{us2021part} is inspired by the Panoptic Quality (PQ) metric~\cite{Kirillov2019PS} and additionally extends it considering the part-level abstraction. The PartPQ per scene-level class $l$ is defined by:
\begin{equation}
	\textrm{PartPQ} = \frac{\sum_{(p,g) \in \textit{TP}}\textrm{IOU\textsubscript{p}}(p,g)}{|\textit{TP}| + \frac{1}{2}|\textit{FP}|+ \frac{1}{2}|\textit{FN}|} ~.
	\label{eq:partpq}
\end{equation}
The metric is based on the overlap between a predicted segment $p$ and a ground-truth segment $g$ for a class $l$. The formula counts the number of true positive (TP), false positive (FP), and false negative (FN) segments, using the Intersection Over Union (IOU) as the underlying metric. A prediction is a TP if it has an overlap with a ground-truth segment with an $\textrm{IOU} > 0.5$. An FP is a predicted segment that is not matched with the ground truth, and an FN is a ground-truth segment not matched with a prediction. The part-level abstraction is captured by the $\textrm{IOU\textsubscript{p}}(p,g)$ term in Equation~\eqref{eq:partpq}. The IOU\textsubscript{p} definition depends on whether the scene-level class $l$ has parts or not, hence:
\begin{equation}
	\textrm{IOU\textsubscript{p}}(p,g) =
	\begin{cases}
		\textrm{mean IOU\textsubscript{part}}(p,g), & \textrm{$l \in \mathcal{L}^\text{parts}$} ~;\\
		\textrm{IOU\textsubscript{inst}}(p,g), & \textrm{$l \in \mathcal{L}^\text{no-parts}$} ~.
	\end{cases}   
\end{equation}
For the classes in $\mathcal{L}^\text{parts}$, the mean IoU for all part classes in the two matched segments is calculated. For the classes in $\mathcal{L}^\text{no-parts}$, the instance-wise IoU is computed as in the original PQ metric. The PartPQ for a dataset with multiple scene-level classes is calculated by averaging over all per-class $\textrm{PartPQ}$ scores for all $l \in \mathcal{L}$.

\subsubsection{B. Part-aware Intersection over Union}
Part-aware Intersection Over Union (PartIOU) is equal to the conventional IoU computed for part-level classes. As such, a different PartIOU is calculated for each scene-level class and the average of all per-class PartIOUs is the one reported.

\begin{landscape}
\begin{table}
	\footnotesize
	\renewcommand{\arraystretch}{1.3}
	\setlength\tabcolsep{3.7pt}
	\centering
	\begin{tabular}{lccccccccccc}
		\toprule
		\textbf{Dataset} & \statstabh{Instance}{-aware} & \statstabh{Panoptic}{-aware} & \statstabh{Parts}{-aware} & \statstabh{Stuff}{classes} & \statstabh{Things}{classes} & \statstabh{Parts}{classes} & \statstabh{Human}{parts} & \statstabh{Vehicle}{parts} & \statstabh{\#Images}{train / val.} & \statstabh{Average}{image size} & \statstabh{Average}{\#inst./img}\\
		\midrule
		PASCAL-Context~\cite{mottaghi14pascalcontext} & - & - & - & 459 (59) & - & - & - & - & 4998 / 5105 & 387 $\times$ 470 & -\\
		LIP~\cite{liang2018look} & - & - & \ding{51} & - & 1 & 20 & 20 & - & 30.5k / 10k & 325 $\times$ 240 & -\\
		CIHP~\cite{gong2018instance} & \ding{51} & - & \ding{51} & - & 1 & 20 & 20 & - & 28.3k / 5k & 484 $\times$ 578 & 3.4\\
		MHP v2.0~\cite{zhao2018understanding} & \ding{51} & - & \ding{51} & - & 1 & 59 & 59 & - & 15.4k / 5k & 644 $\times$ 718 & 3\\
		PASCAL-Person-Parts~\cite{chen2014detect} & \ding{51} & - & \ding{51} & - & 1 & 6 & 6 & - & 1716 / 1817 & 387 $\times$ 470 & 2.2\\
		PASCAL-Parts~\cite{chen2014detect} & \ding{51} & - & \ding{51} & - & 20 & 194 & 24 & 57 & 4998 / 5105 & 387 $\times$ 470 & 2.5\\
		Cityscapes~\cite{Cordts2016Cityscapes} & \ding{51} & \ding{51} & - & 23 & 8 & - & - & - & 2975 / 500 & 1024 $\times$ 2048 & 17.9\\
		\midrule
		\textbf{\textit{This work}} & \multicolumn{3}{c}{} & \multicolumn{5}{c}{}\\
		~~\textbf{PASCAL Panoptic Parts} & \ding{51} & \ding{51} & \ding{51} & 80 & 20 & 194 & 24 & 57 & 4998 / 5105 & 387 $\times$ 470 & 2.5\\
		~~\textbf{Cityscapes Panoptic Parts} & \ding{51} & \ding{51} & \ding{51} & 23 & 8 & 23 & 4 & 5 & 2975 / 500 & 1024 $\times$ 2048 & 17.9\\
		\bottomrule
	\end{tabular}
	\caption{\textit{Dataset statistics} for related (part-) segmentation datasets and the proposed datasets. \textit{PASCAL-Context} has 459 semantic classes but only 59 of them are included in the official split. It should be noted that the amount of classes does not decrease despite the combination of datasets. Cityscapes Panoptic Parts has a very rich instance information layer, since it contains almost 9 times more instances with parts per image than the other datasets.}
	\label{ch7:tab:related-datasets}
\end{table}
\end{landscape}

\section{Panoptic Parts datasets}
\label{ch7:sec:datasets}
\begin{figure}
	\centering
	\captionsetup[subfigure]{labelformat=empty,font=footnotesize}
	\begin{subfigure}[b]{0.4\linewidth}
		\centering
		\includegraphics[width=\linewidth]{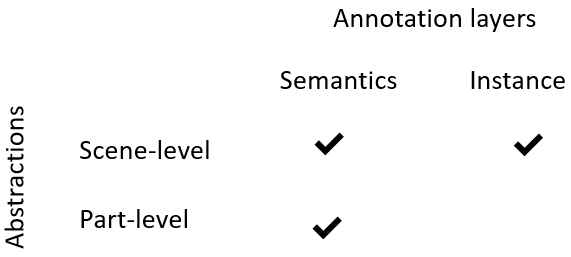}
		\caption{Cityscapes Panoptic Parts.}
	\end{subfigure}
	\qquad \qquad
	\begin{subfigure}[b]{0.4\linewidth}
		\centering
		\includegraphics[width=\linewidth]{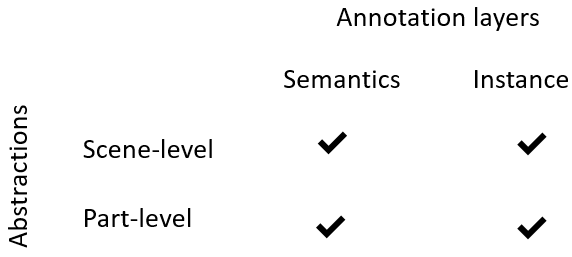}
		\caption{PASCAL Panoptic Parts.}
	\end{subfigure}
	\caption{Information included in the annotations for the two abstraction levels.}
	\label{ch7:fig:abstr-info}
\end{figure}

We accompany the PPS task with two new datasets, Cityscapes Panoptic Parts (CPP) and PASCAL Panoptic Parts (PPP), which are based on the established scene understanding datasets Cityscapes~\cite{Cordts2016Cityscapes} and PASCAL VOC~\cite{Everingham2010Pascal}, respectively. The introduced datasets include per-pixel annotations at two levels of visual abstraction, namely at scene level and part level, and up to two information layers, namely semantics and instance-wise annotations. As can be observed from Table~\ref{ch7:tab:related-datasets}, the existing datasets landscape is inadequate for PPS, since no dataset entails all of the required information. If any arbitrary combination of the existing datasets is used to achieve multi-level abstraction, conflicts would arise at the pixel level caused by overlapping labels. The proposed datasets comprise a consistent set of annotations, which are free of such conflicts. The information provided by the datasets is summarized in Figure~\ref{ch7:fig:abstr-info}.
\begin{figure}
	\centering
	\includegraphics[width=0.8\linewidth]{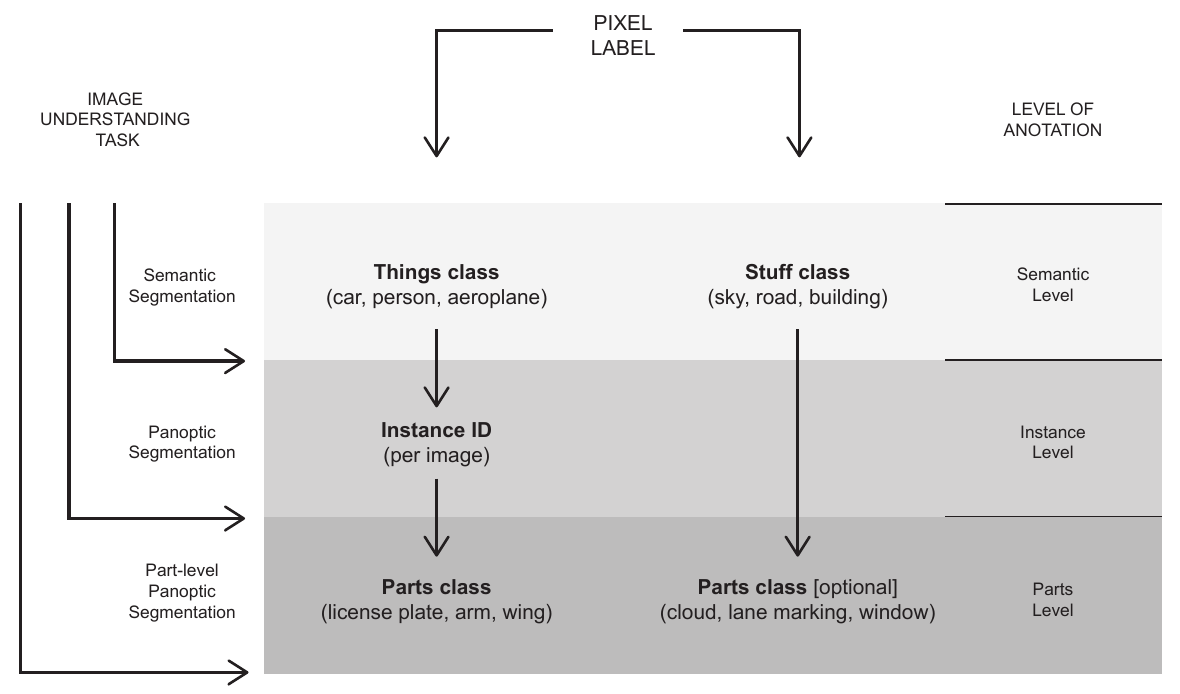}
	\caption{Annotations hierarchy according to the provided annotations and the related image understanding tasks that can be used.}
	\label{fig:tasks}
\end{figure}

\subsection{Dataset: Cityscapes Panoptic Parts (CPP)}
\label{ch6:ssec:cpp}
Cityscapes Panoptic Parts (CPP) extends the popular Cityscapes dataset~\cite{Cordts2016Cityscapes} with part-level semantics. This dataset contains urban scenes recorded in Germany and neighboring countries. A group of technical people manually annotated the original publicly available 2,975 training and 500 validation images with 23 part-level semantic classes. We generated a pipeline tool that takes advantage of original annotations to guide and hint the annotators actively. The proposed CPP dataset is fully compatible with the original Cityscapes panoptic annotations and is, to the best of our knowledge, the first urban scenes dataset with scene-level and part-level semantic annotations, enhanced with instance-wise separation on the same set of images.

\begin{figure}
	\begin{center}
		\includegraphics[width=0.65\linewidth, trim={0.0cm 0.1cm 0.0cm 0.0cm}, clip]{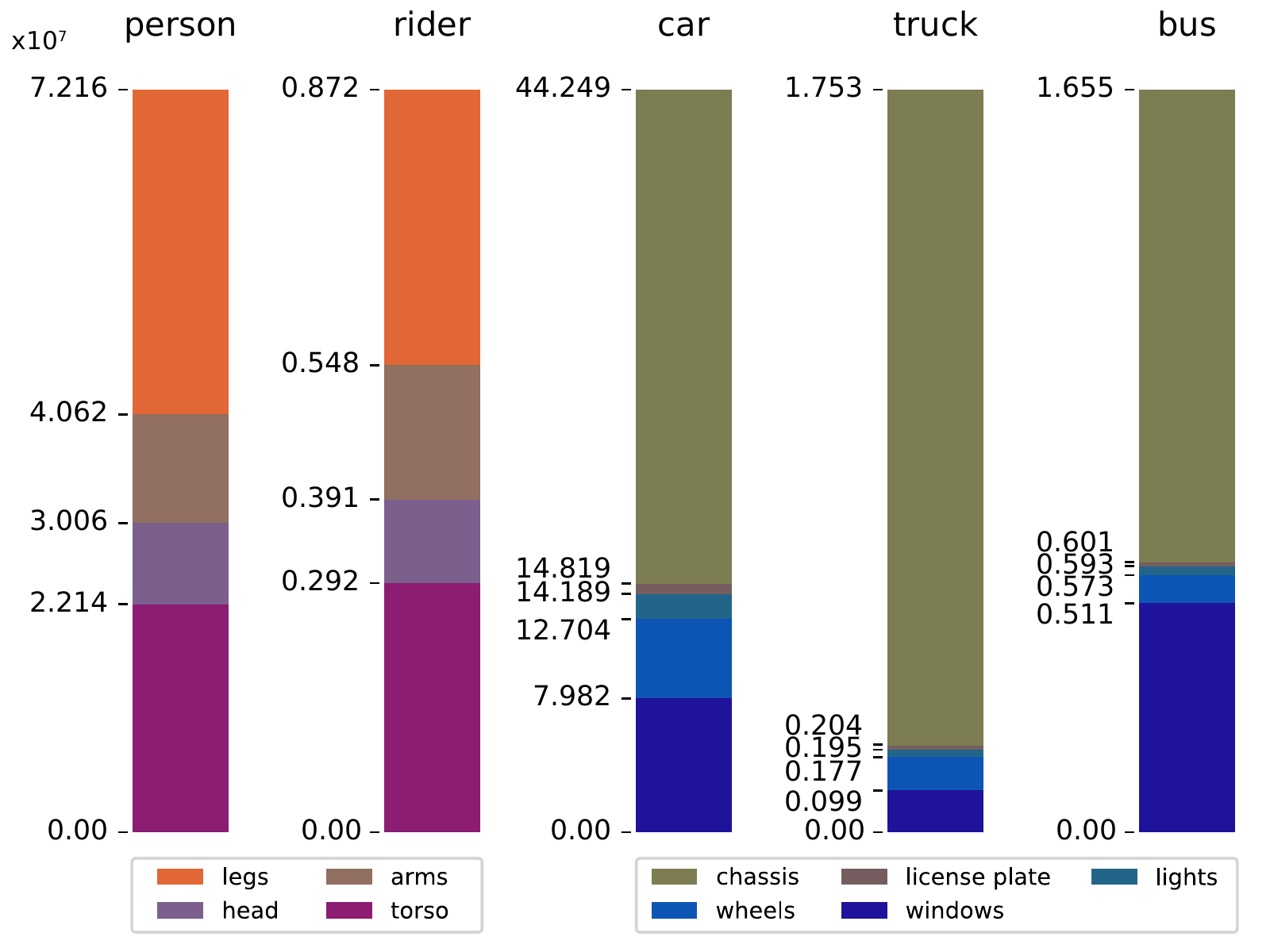}\\
	\end{center}
	\caption{Cumulative absolute number of the proposed CPP dataset pixels ($\times 10^7$) that are annotated by the technical team per semantic class and per human/vehicle part.}
	\label{ch7:fig:num-pixels-cityscapes}
\end{figure}

Taking into consideration the complexity of scenes and the variety in number and pose of traffic participants, we have selected 5 scene-level semantic classes from the \textit{human} and \textit{vehicle} high-level categories to be annotated with parts,~\ie, $\mathcal{L^\text{parts} = \{\textit{person}, \textit{rider}, \textit{car}, \textit{truck}, \textit{bus}\}}$. The \textit{human} categories are annotated with $\mathcal{P}^\textit{human} = \{\textit{torso}, \textit{head}, \textit{arm}, \textit{leg}\}$ and the \textit{vehicle} categories with $\mathcal{P}^\textit{vehicle} = \{\textit{chassis}, \textit{window}, \textit{wheel}, \textit{light}, \textit{license plate}\}$. Statistics for CPP are presented in Table~\ref{ch7:tab:related-datasets} and in Figure~\ref{ch7:fig:num-pixels-cityscapes}.

\subsection{Dataset: PASCAL Panoptic Parts (PPP)}
\label{ch6:ssec:ppp}
The PASCAL Panoptic Parts (PPP) dataset extends the PASCAL VOC 2010 benchmark~\cite{Everingham2010Pascal} with part-level and scene-level semantics. The original PASCAL VOC dataset is labeled with scene-level semantics, and only partly at instance-level. A large number of subsequent extensions have been proposed with annotations over different levels of abstraction, leading to various inconsistencies between them at the pixel level. We have created PPP by carefully merging PASCAL-Context~\cite{mottaghi14pascalcontext} and PASCAL-Parts~\cite{chen2014detect} to maintain the high quality of annotations and solve any conflicts. As the PPP dataset solves conflicts between PASCAL-Context~\cite{mottaghi14pascalcontext} and PASCAL-Parts~\cite{chen2014detect}, evaluations on the proposed PPP are not directly comparable with those on the aforementioned datasets. 

The PPP dataset preserves the original splitting into 4,998 training and 5,105 validation images. At the scene-level abstraction, PPP contains $\left|\mathcal{L}^\text{Th}\right| = 20$ classes with instance annotations and $\left|\mathcal{L}^\text{St}\right| = 80$ classes without instances. At the part level, it comprises of $\left|\mathfrak{P}\right| = 195$ parts spanning $\left|\mathcal{L}^\text{parts}\right| = 16$ classes, and  $\left|\mathcal{L}^\text{Th} \cap \mathcal{L}^\text{parts}\right| = 16$. For easier comparison with related methods, we provide mappings from PPP  to commonly used subsets in related work: 7 parts for human-part parsing in PASCAL-Person-Parts~\cite{chen2014detect} and 58 parts for the reduced set used in~\cite{michieli2020gmnet, Zhao2019BSANet}. More statistics of PPP can be found in Table~\ref{ch7:tab:related-datasets}.

~

\begin{figure}
	\centering
	\includegraphics[width=0.48\linewidth]{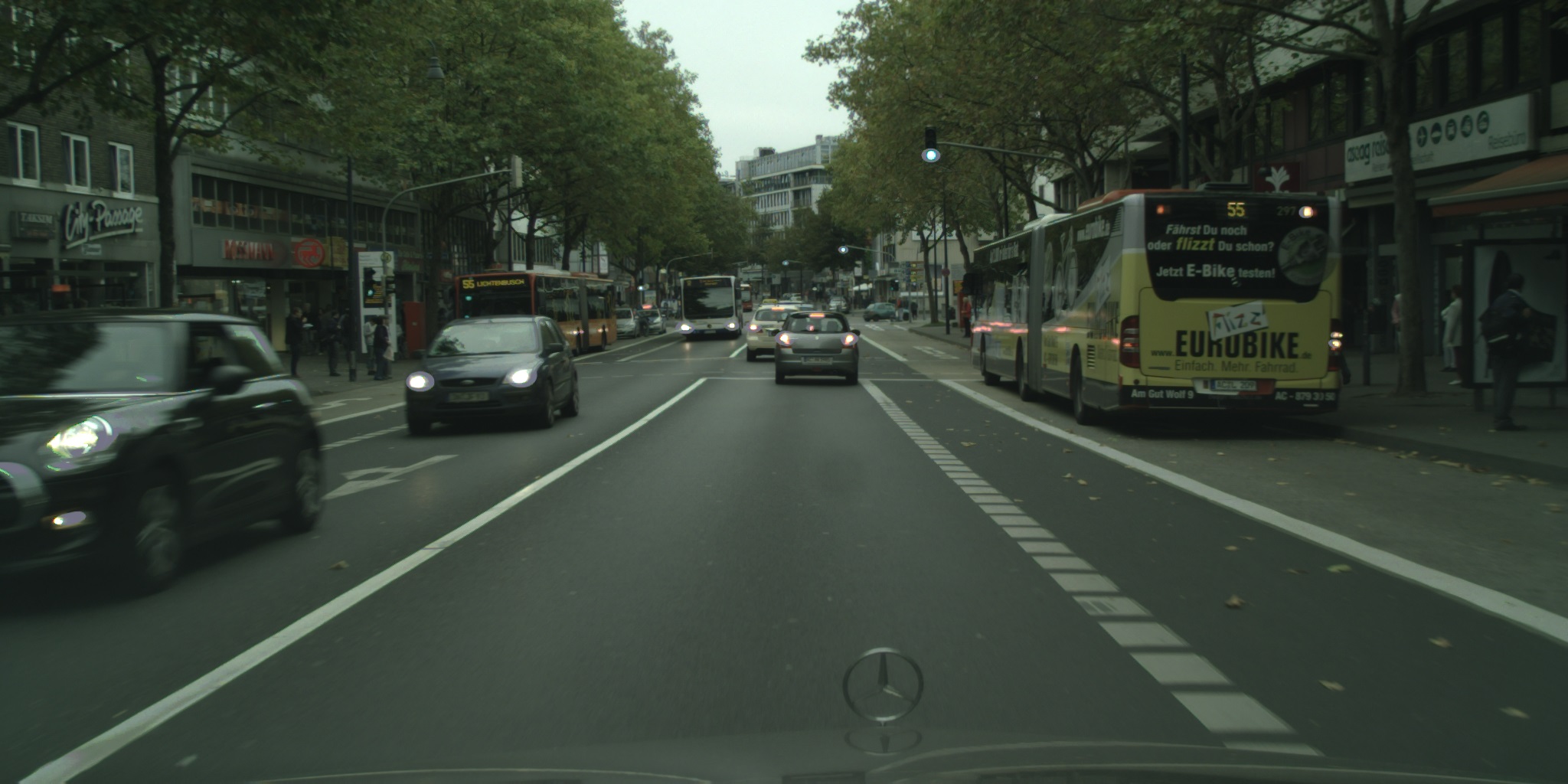}
	~\includegraphics[width=0.48\linewidth]{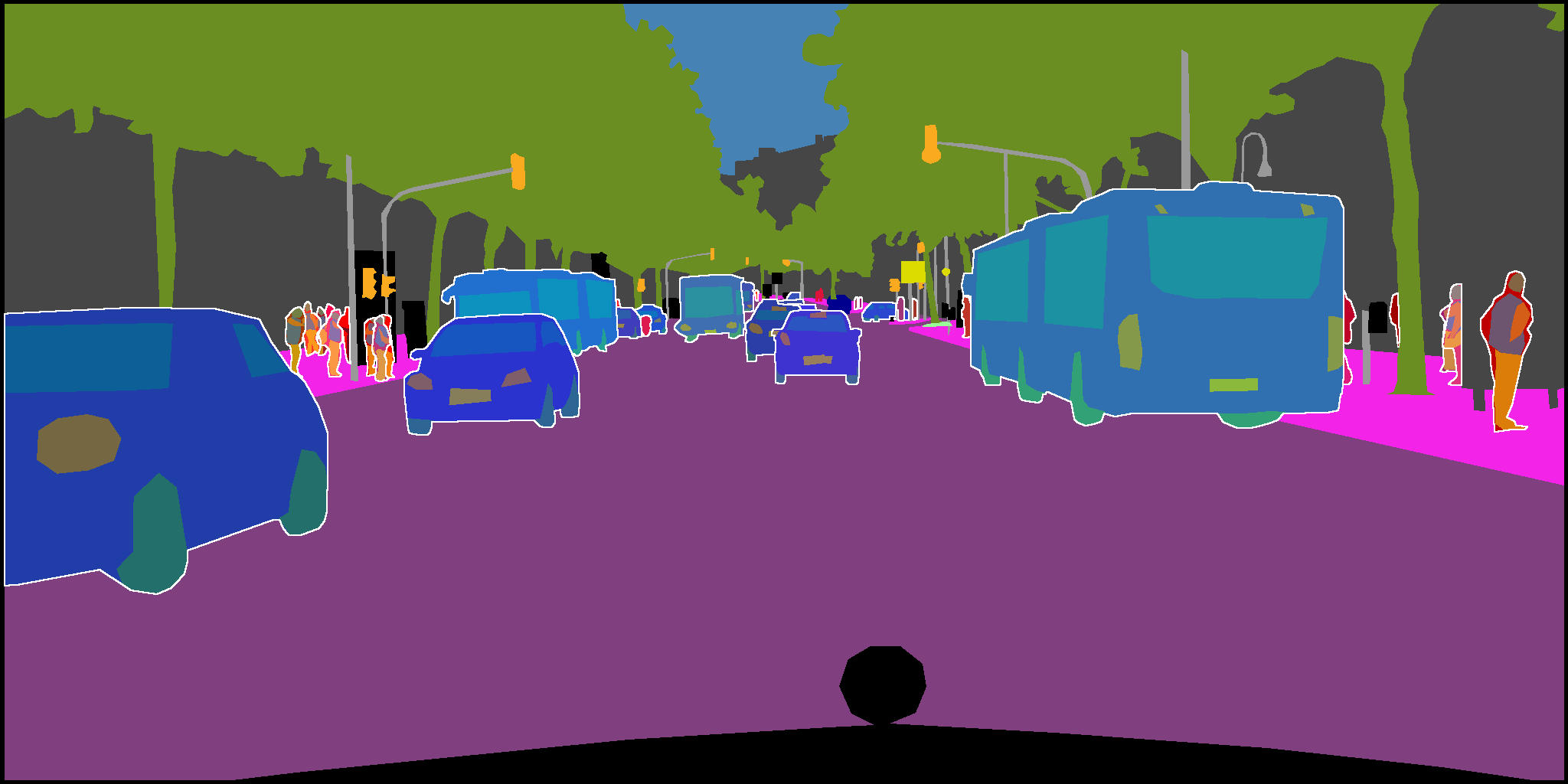}\\
	\includegraphics[width=0.48\linewidth]{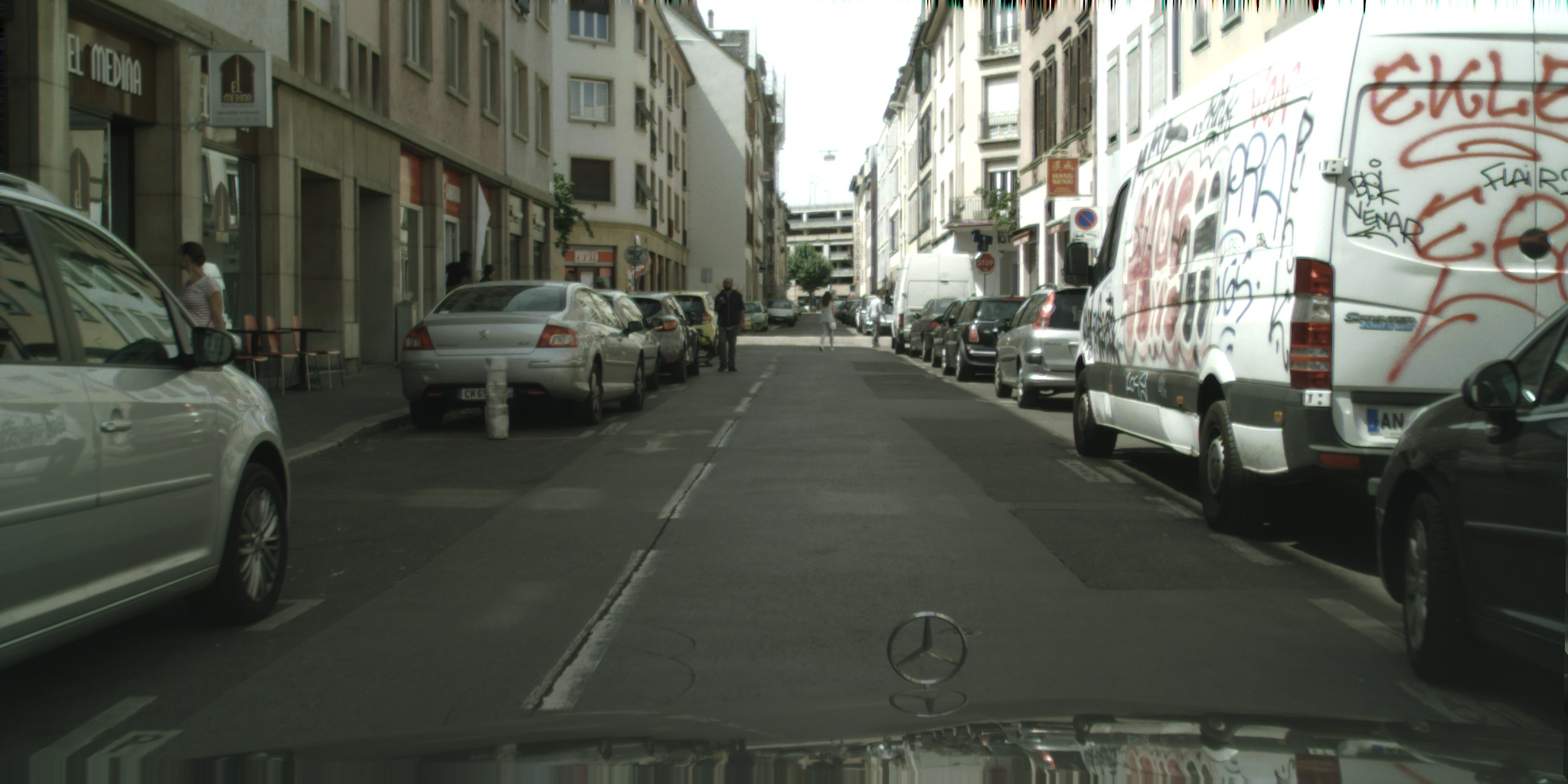}
	~\includegraphics[width=0.48\linewidth]{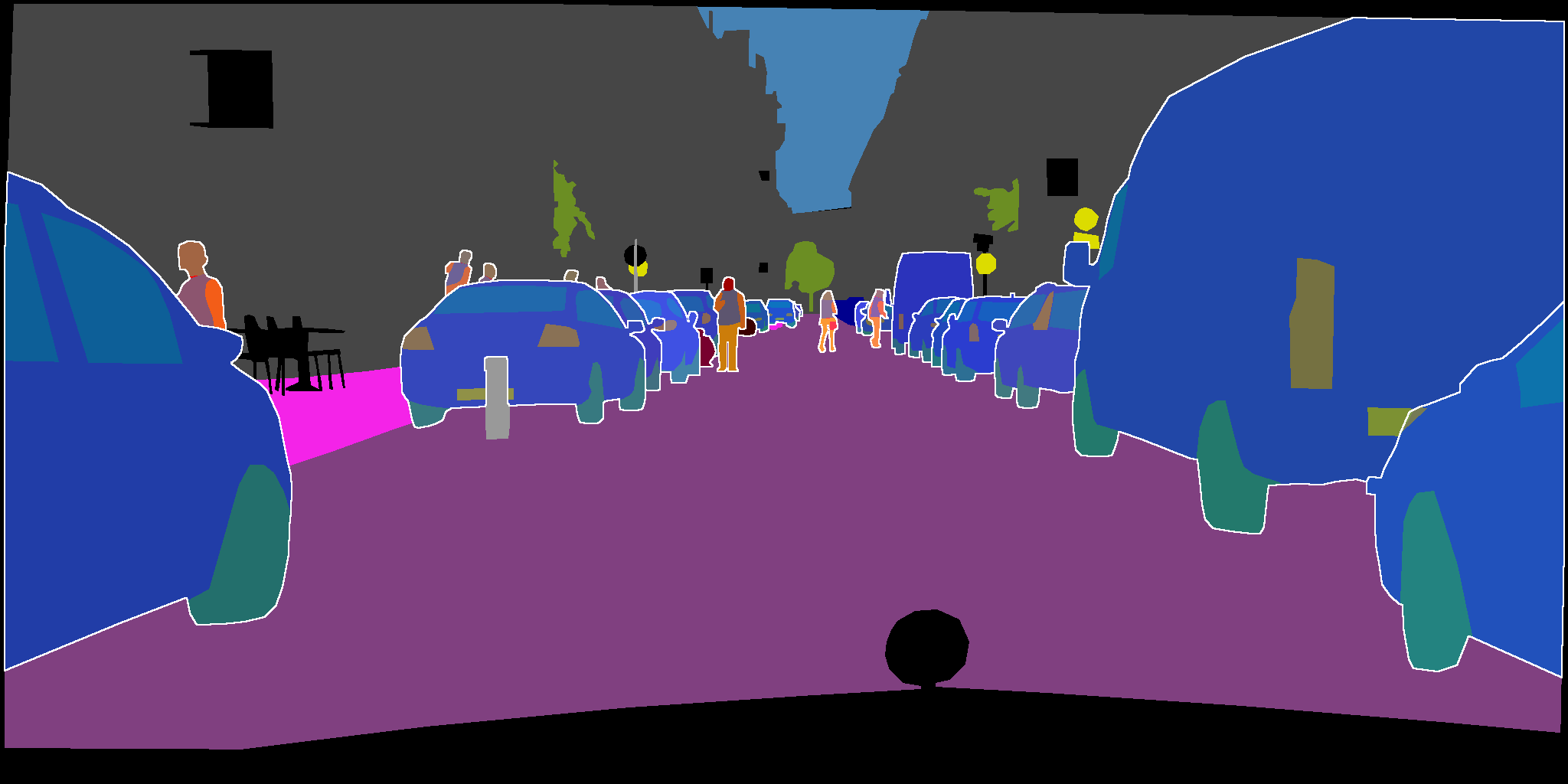}\\
	\includegraphics[width=0.48\linewidth]{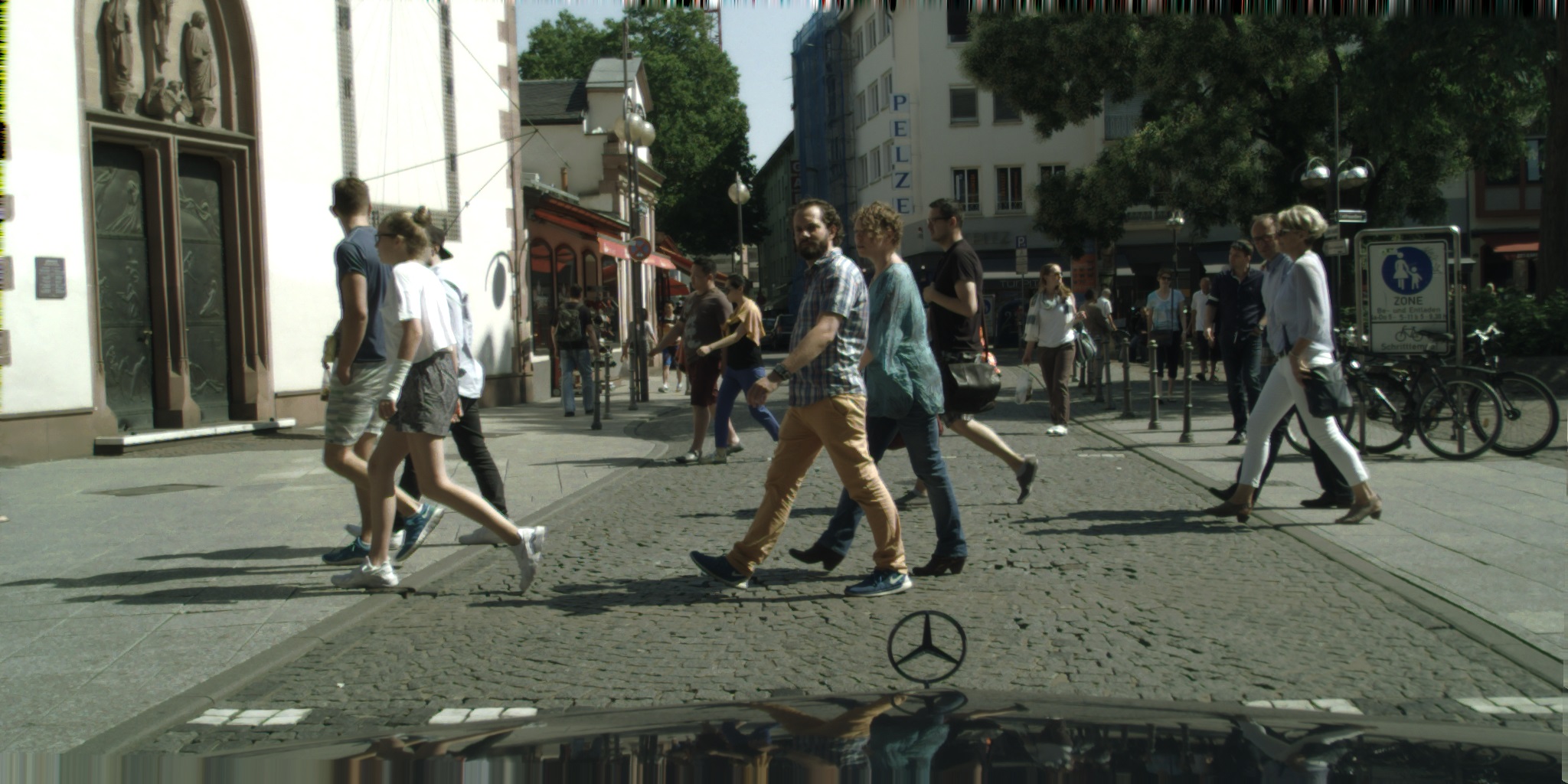}
	~\includegraphics[width=0.48\linewidth]{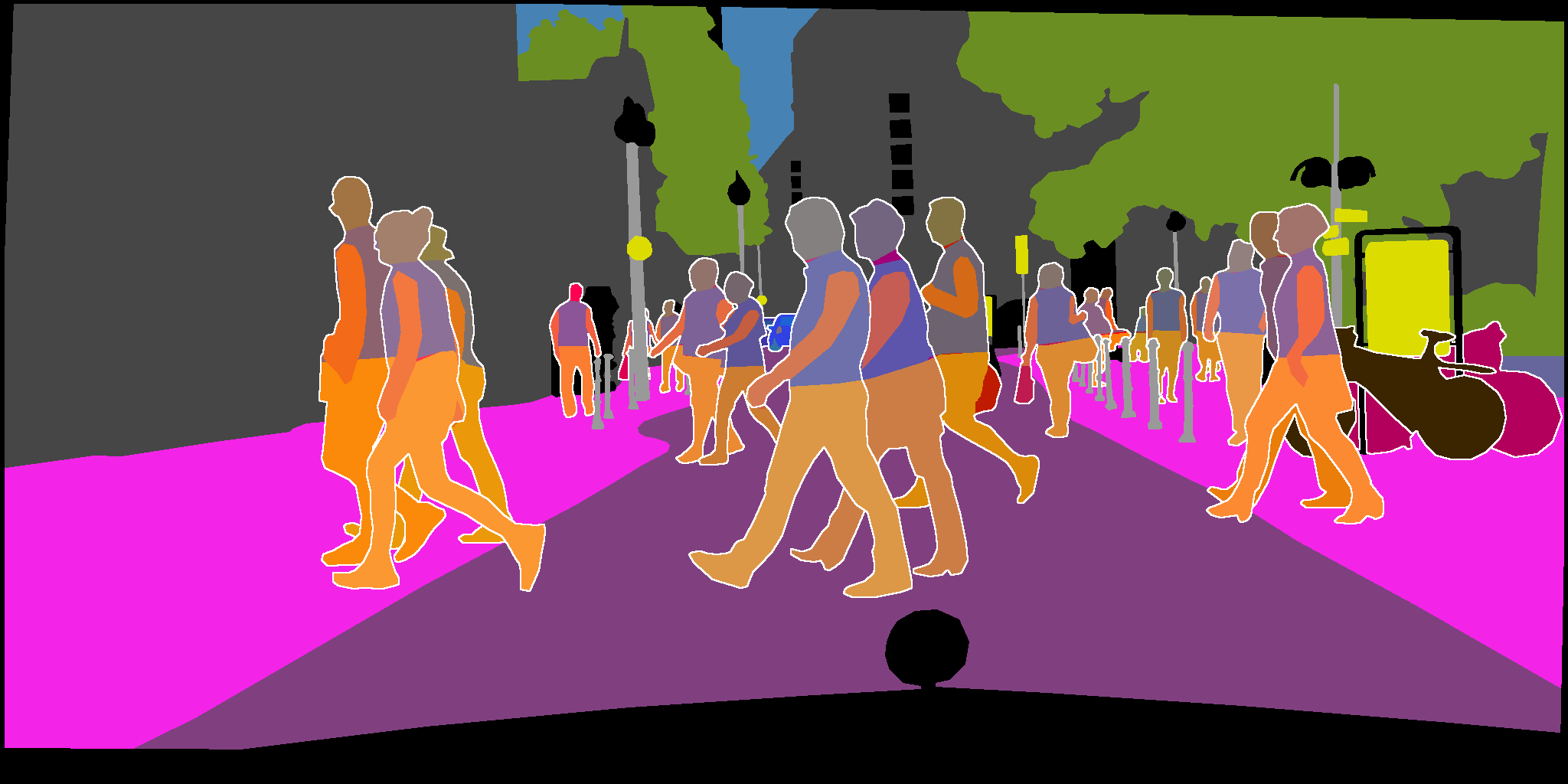}\\
	\includegraphics[width=0.48\linewidth]{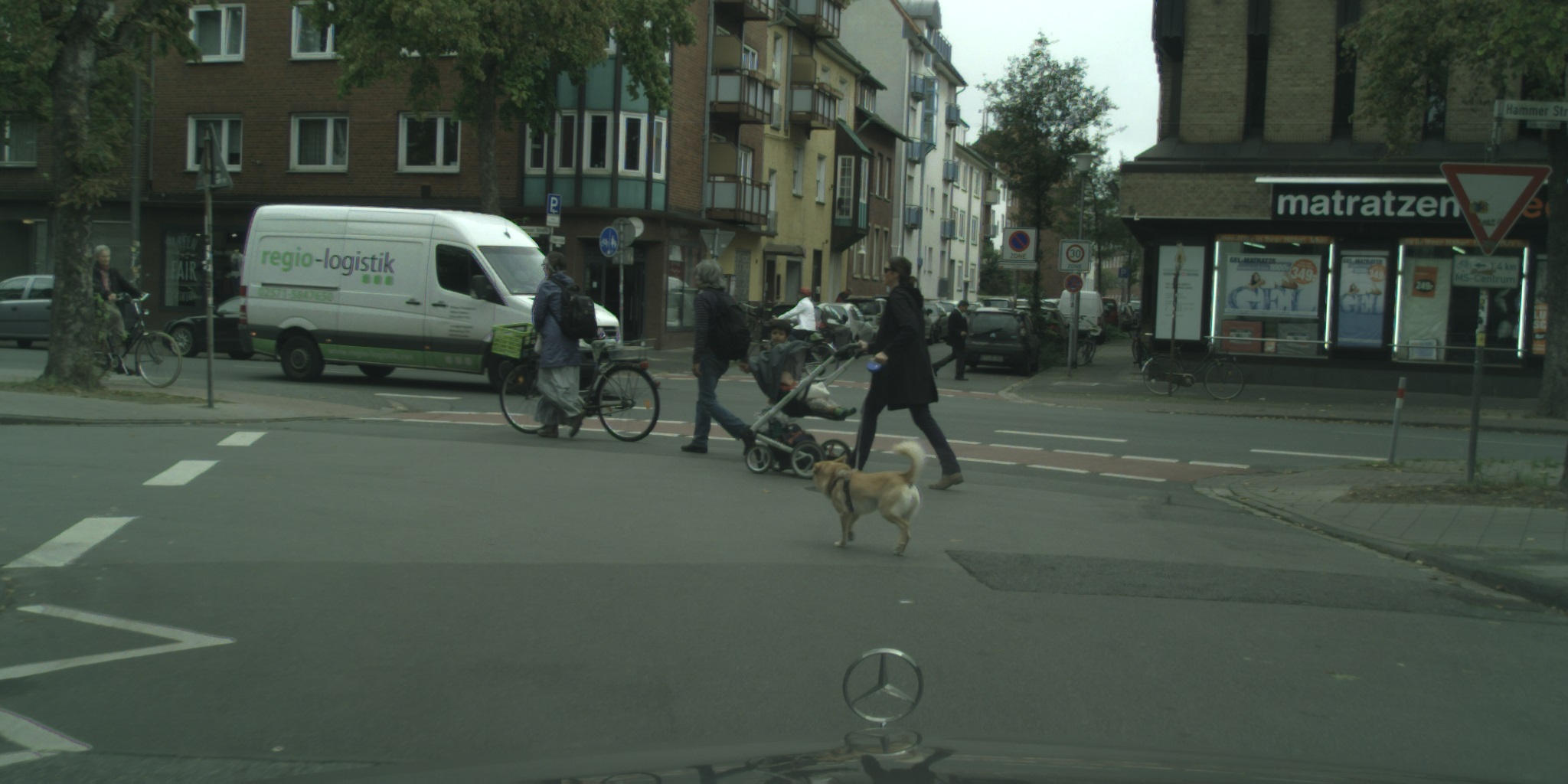}
	~\includegraphics[width=0.48\linewidth]{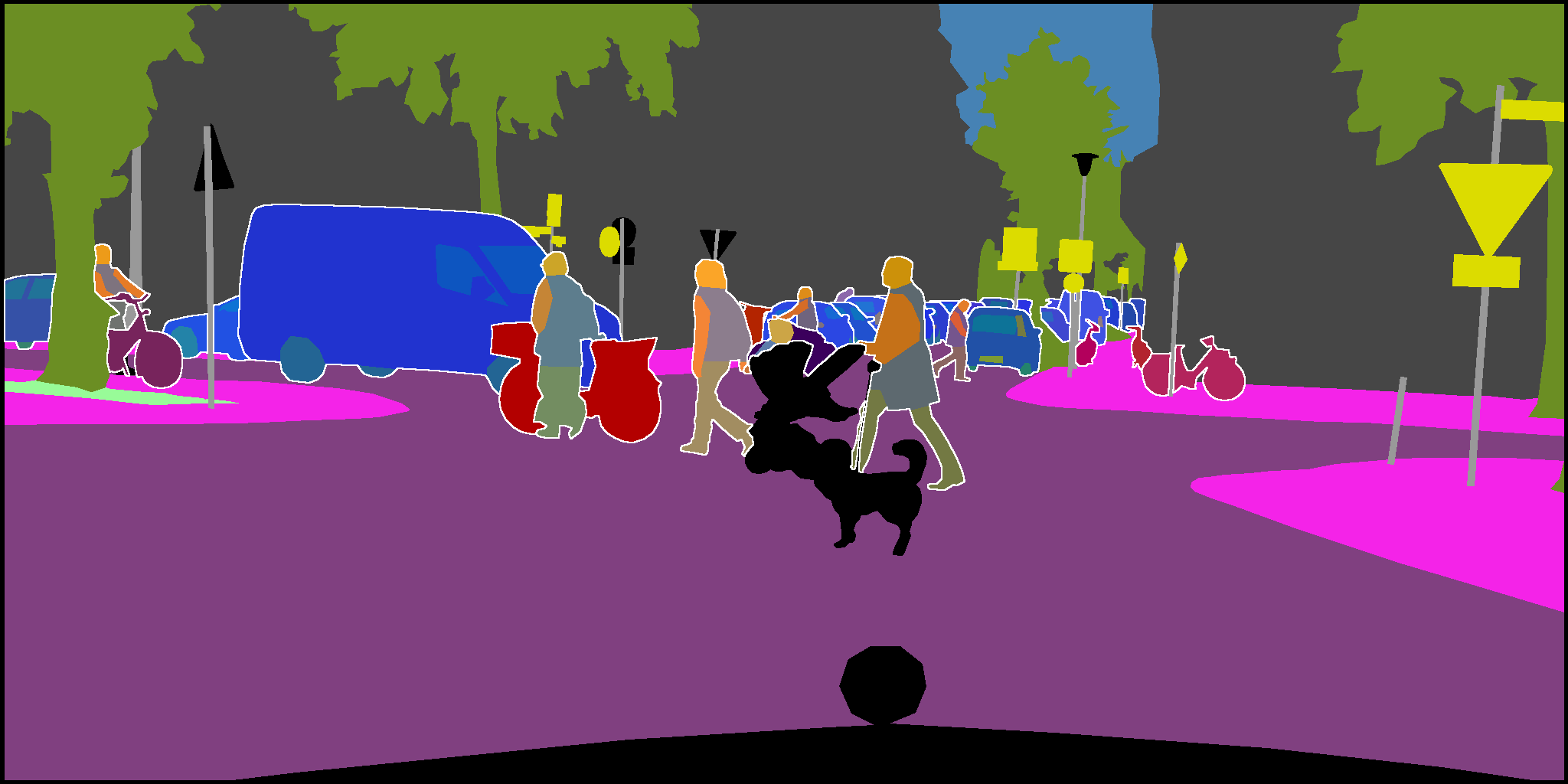}\\
	\vspace{-5pt}
	\caption{Examples from Cityscapes-Panoptic-Parts images and labels. Top two rows: training split. Bottom two rows: validation split.}
	\label{ch7:fig:cpp-examples}
\end{figure}

\begin{figure}
	\centering
	\includegraphics[width=0.40\linewidth]{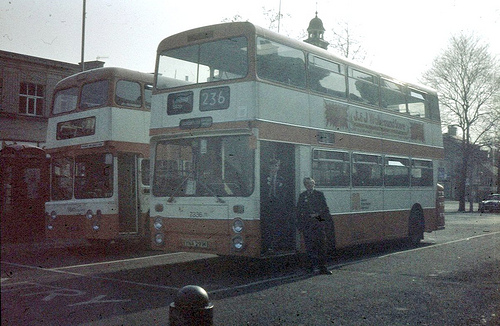}
	~\includegraphics[width=0.40\linewidth]{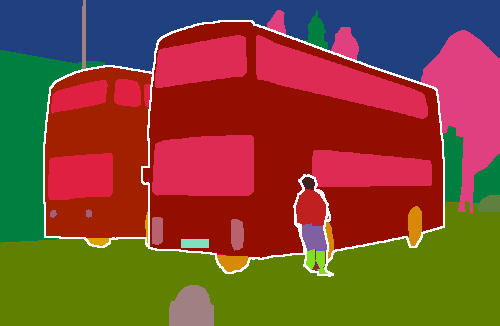}\\
	\includegraphics[width=0.40\linewidth]{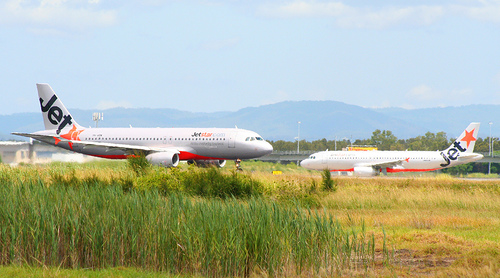}
	~\includegraphics[width=0.40\linewidth]{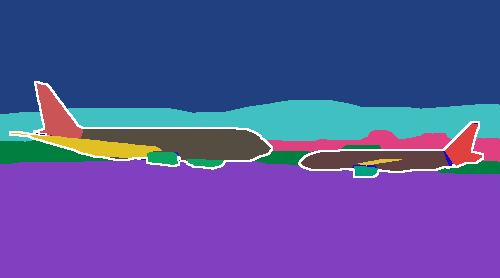}\\
	\includegraphics[width=0.40\linewidth]{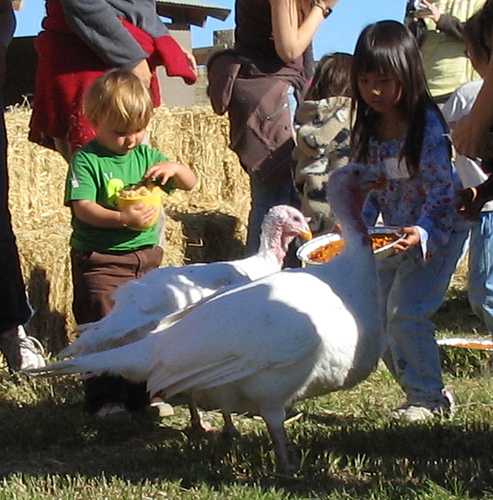}
	~\includegraphics[width=0.40\linewidth]{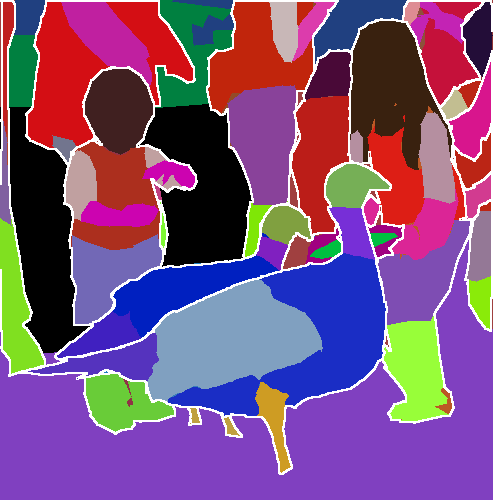}\\
	\includegraphics[width=0.40\linewidth]{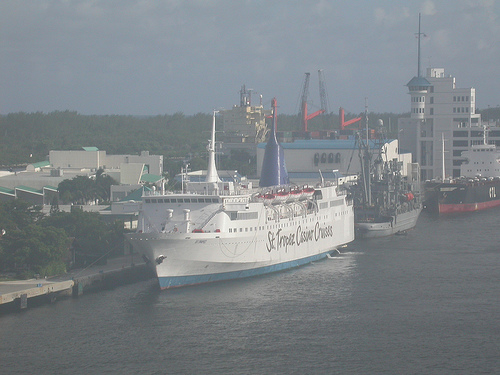}
	~\includegraphics[width=0.40\linewidth]{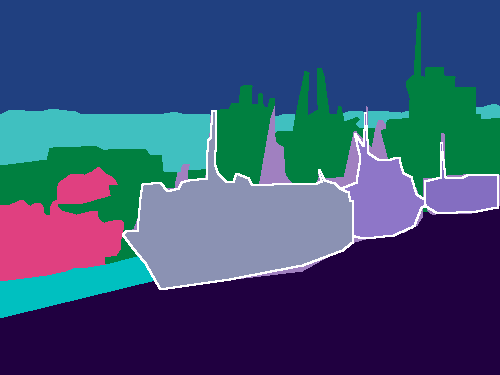}\\
	\includegraphics[width=0.40\linewidth]{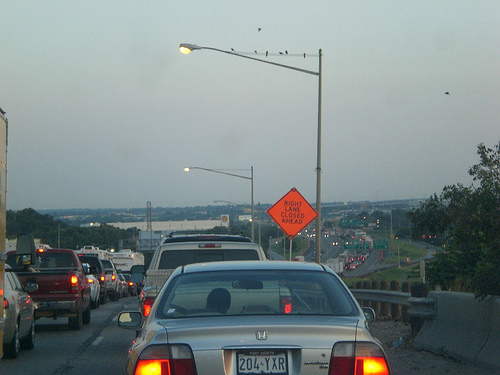}
	~\includegraphics[width=0.40\linewidth]{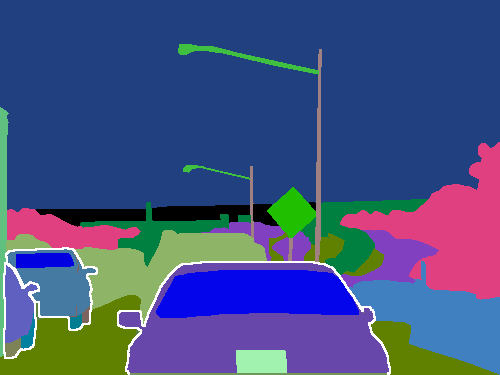}\\
	\vspace{-5pt}
	\caption{Examples of PASCAL-Panoptic-Parts images and labels from the training split. The benefits of our ``best-effort'' merging strategy are clear in the last two images, where the semantic-level labels (from \textit{PASCAL-Context}), \textit{boat} and \textit{car}, provide information for the unlabeled pixels of \textit{PASCAL-Parts}.}
	\label{ch7:fig:ppp-examples}
\end{figure}

\begin{figure}
	\centering
	\includegraphics[width=0.48\linewidth]{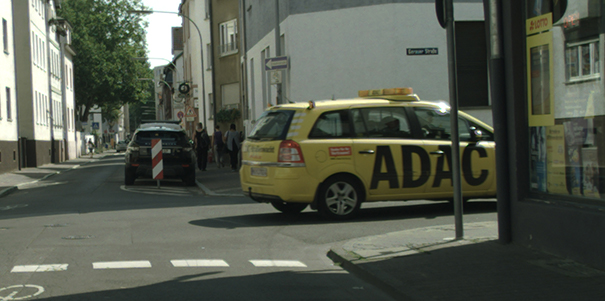}
	\includegraphics[width=0.48\linewidth]{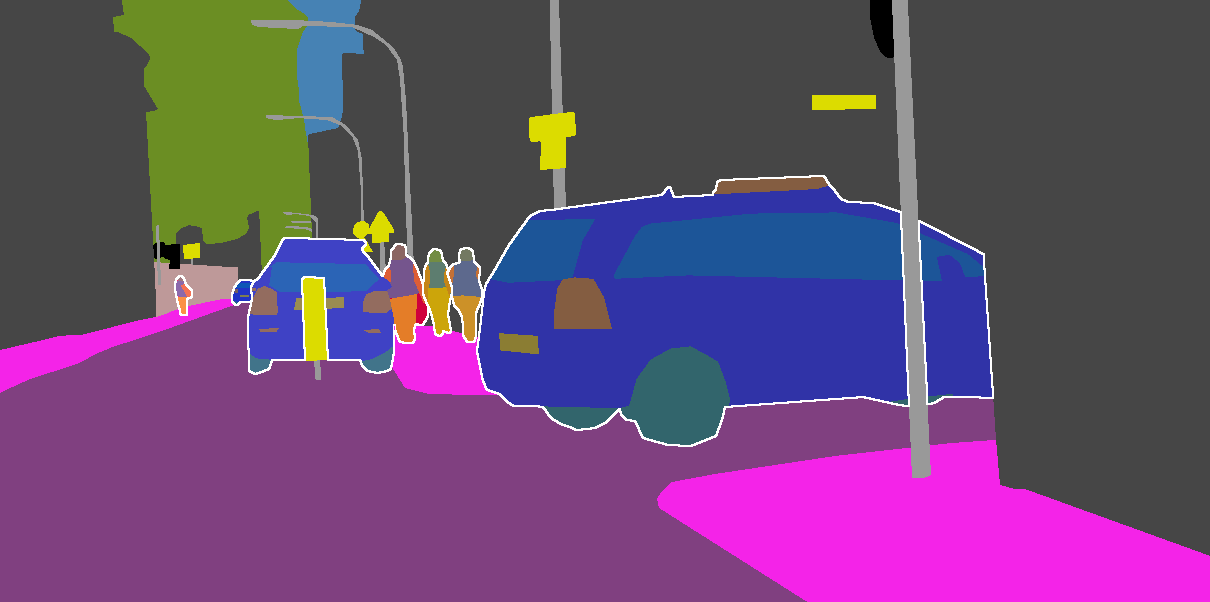}\\ 
	\includegraphics[width=0.48\linewidth]{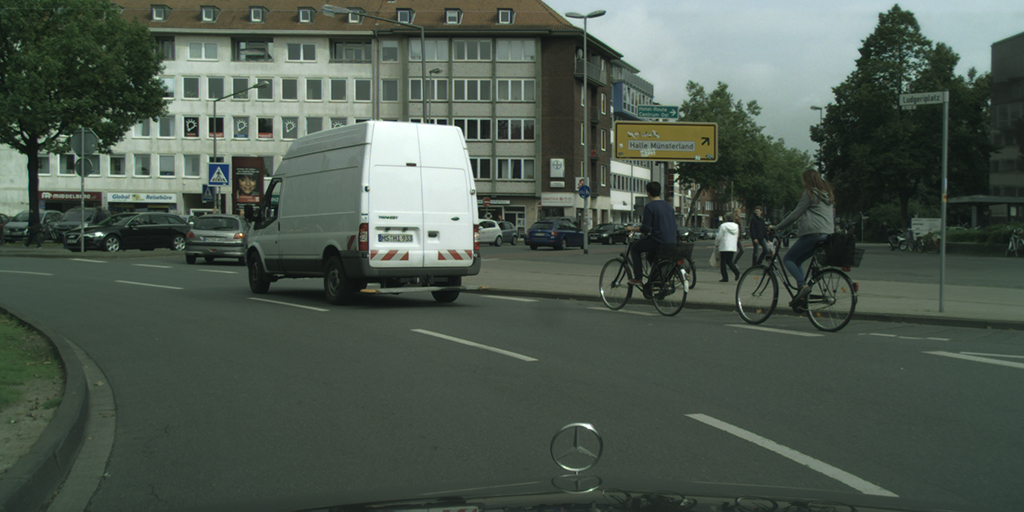}
	\includegraphics[width=0.48\linewidth]{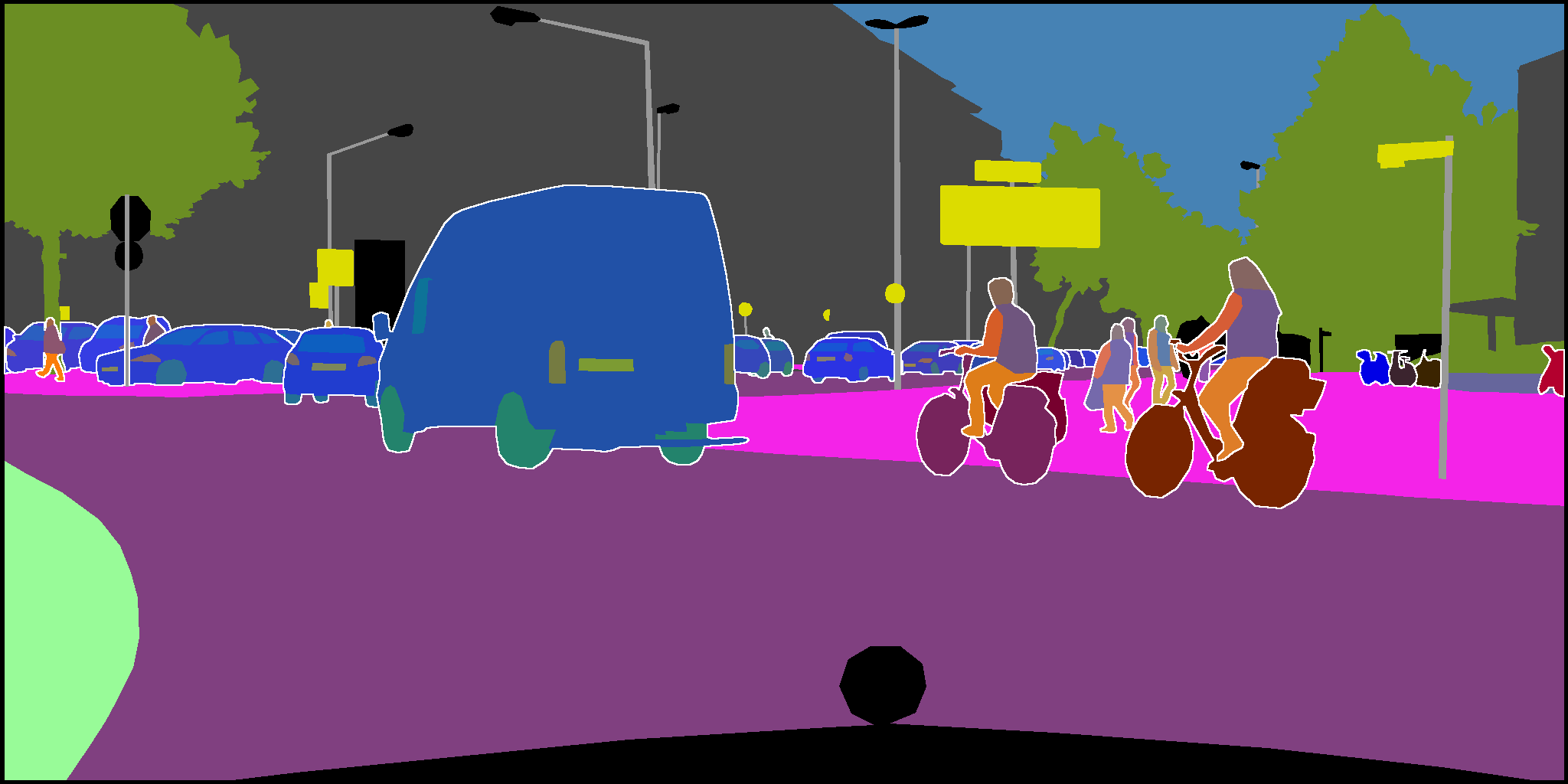}\\ 
	\includegraphics[width=0.48\linewidth]{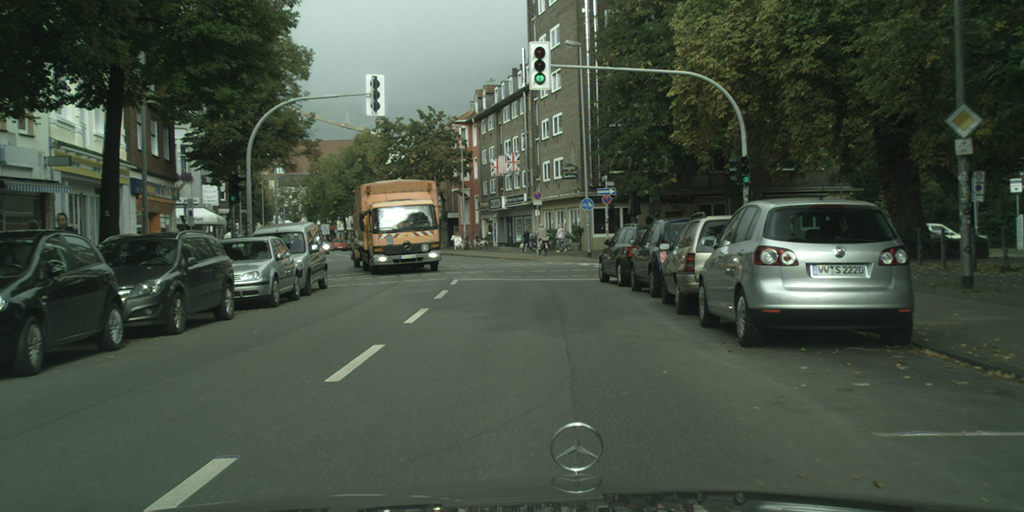}
	\includegraphics[width=0.48\linewidth]{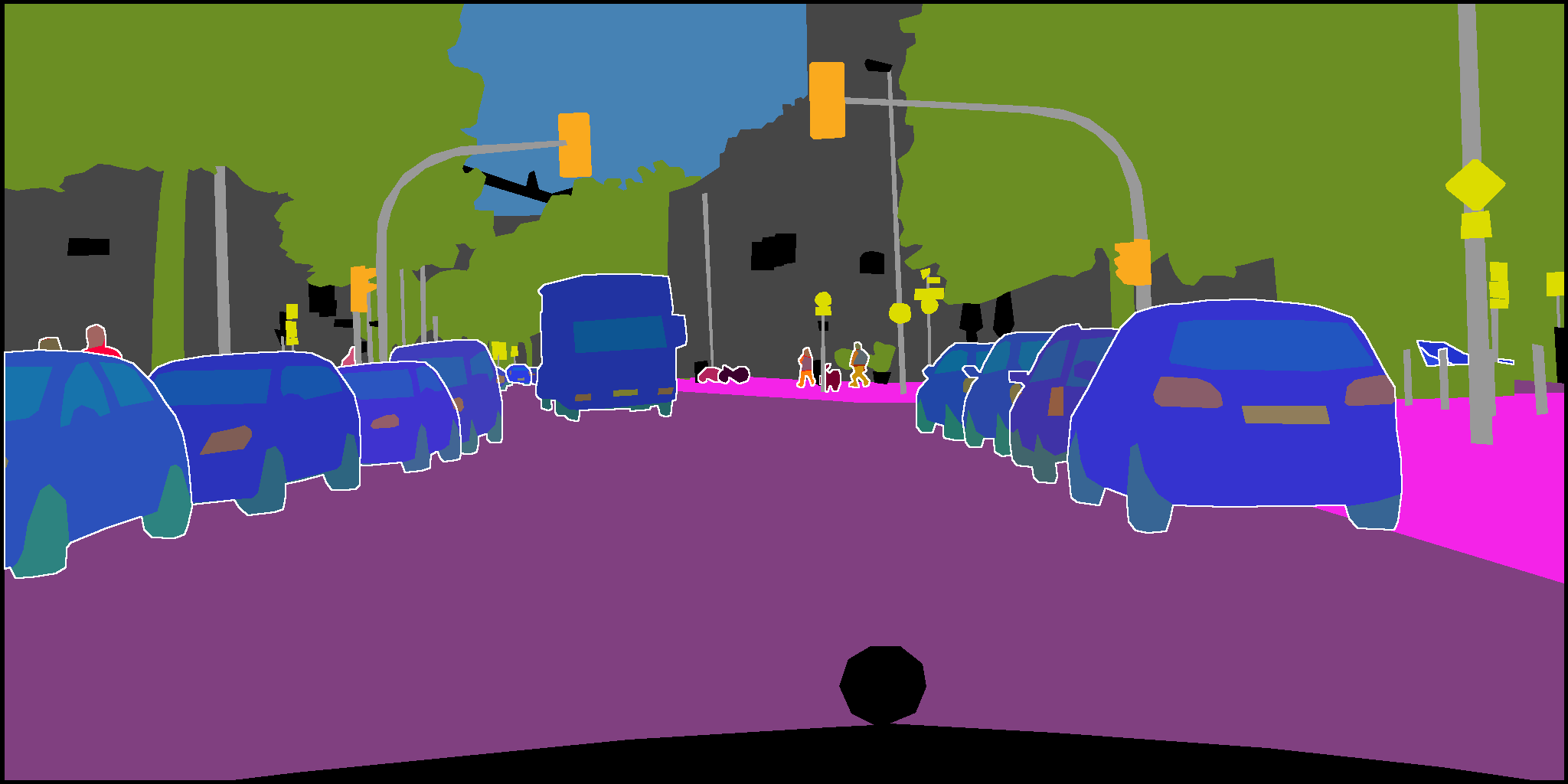}\\ 
	\includegraphics[width=0.48\linewidth]{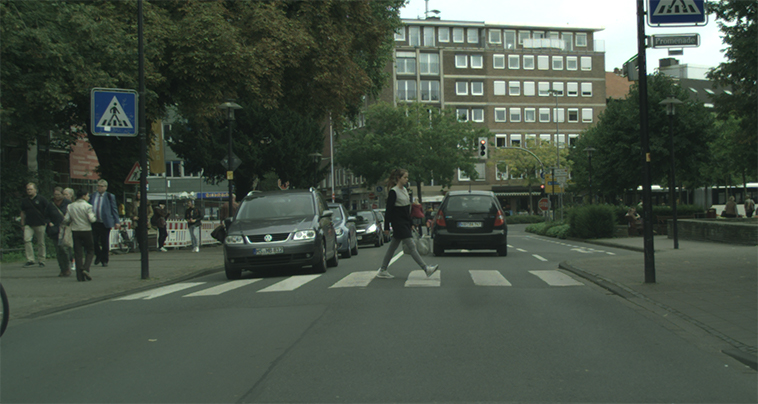}
	\includegraphics[width=0.48\linewidth]{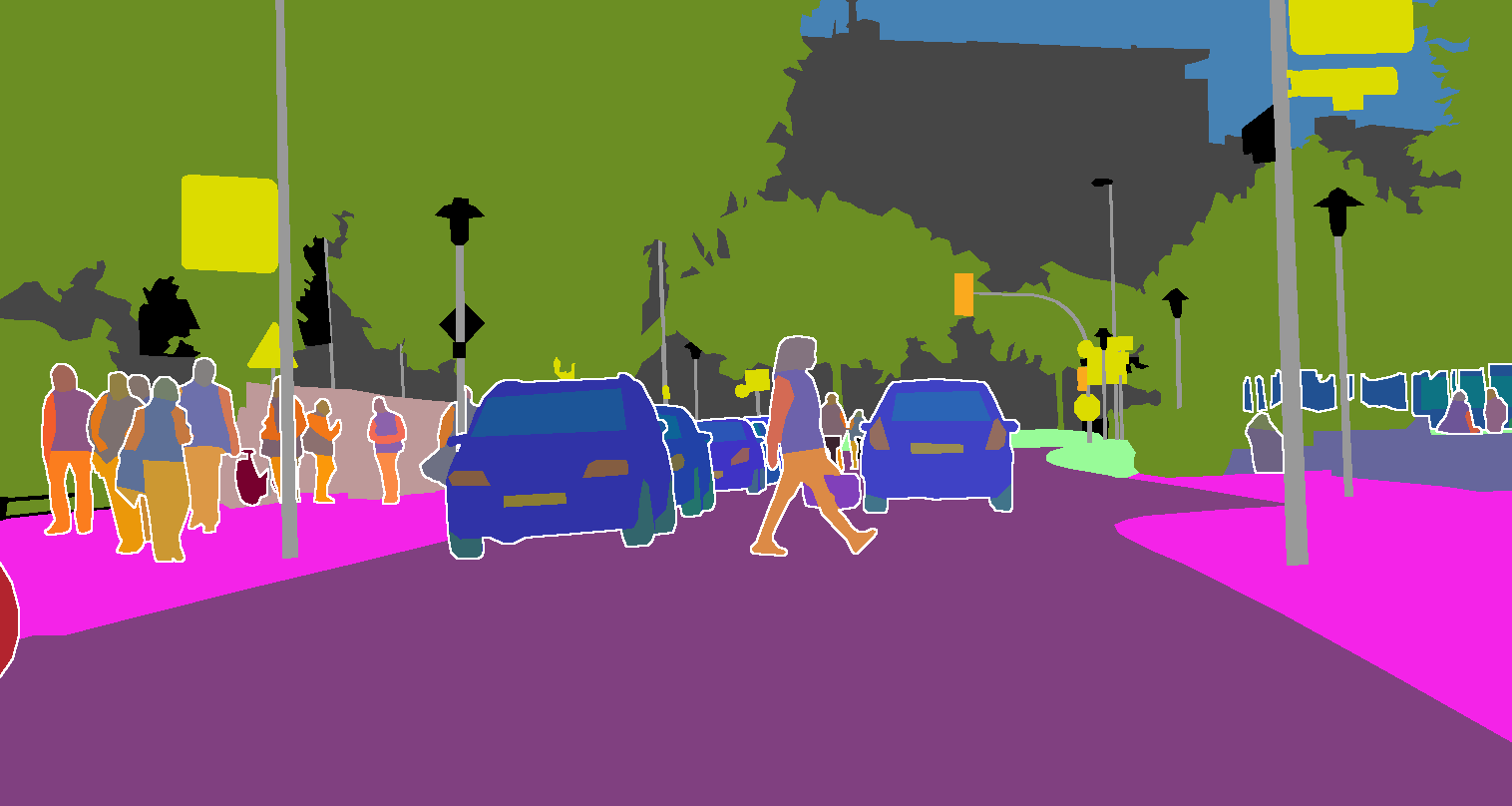}\\ 
	\includegraphics[width=0.48\linewidth]{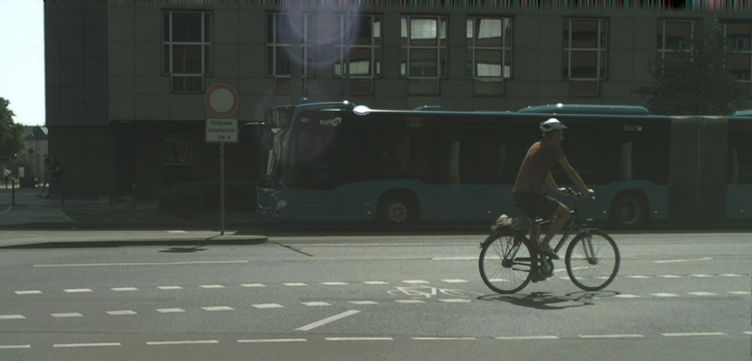}
	\includegraphics[width=0.48\linewidth]{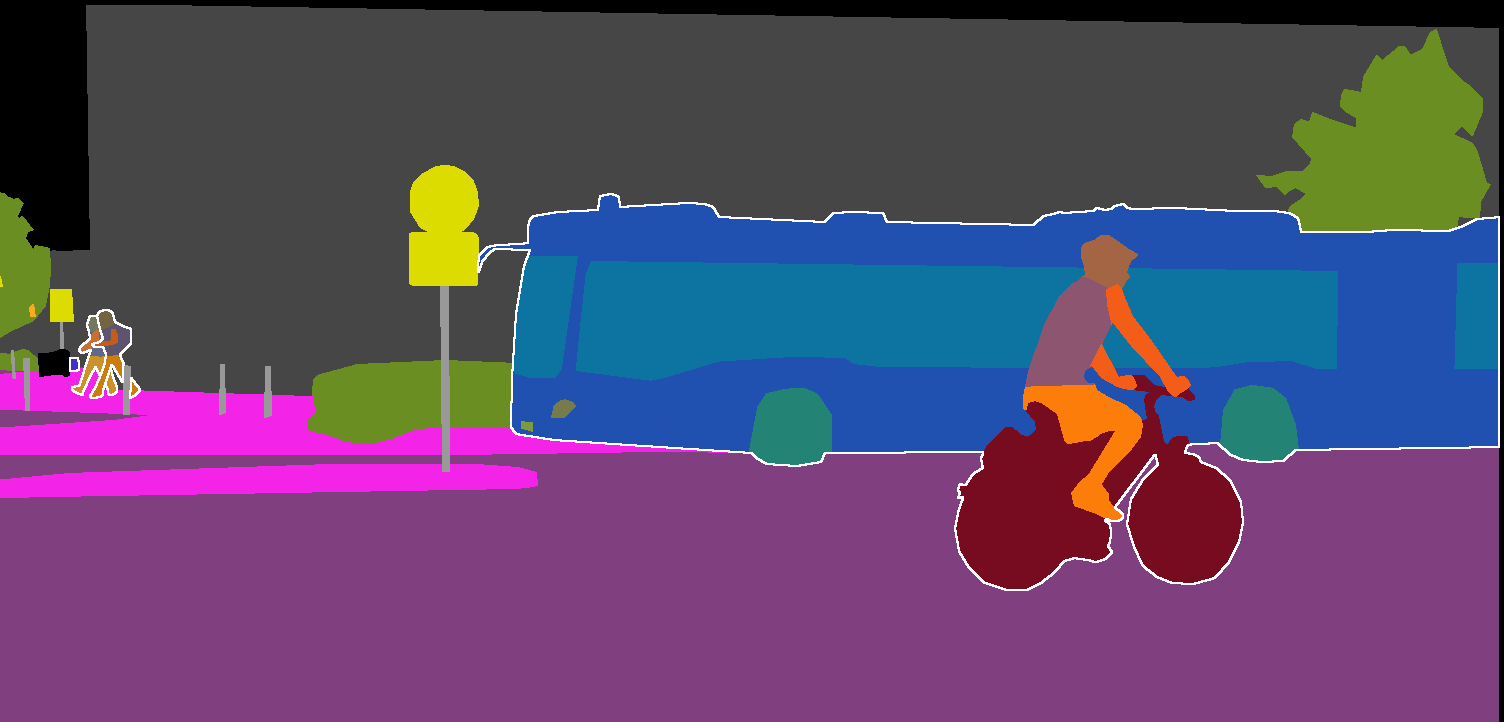}\\
	\includegraphics[width=0.48\linewidth]{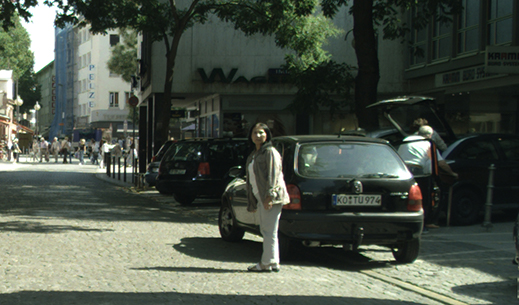}
	\includegraphics[width=0.48\linewidth]{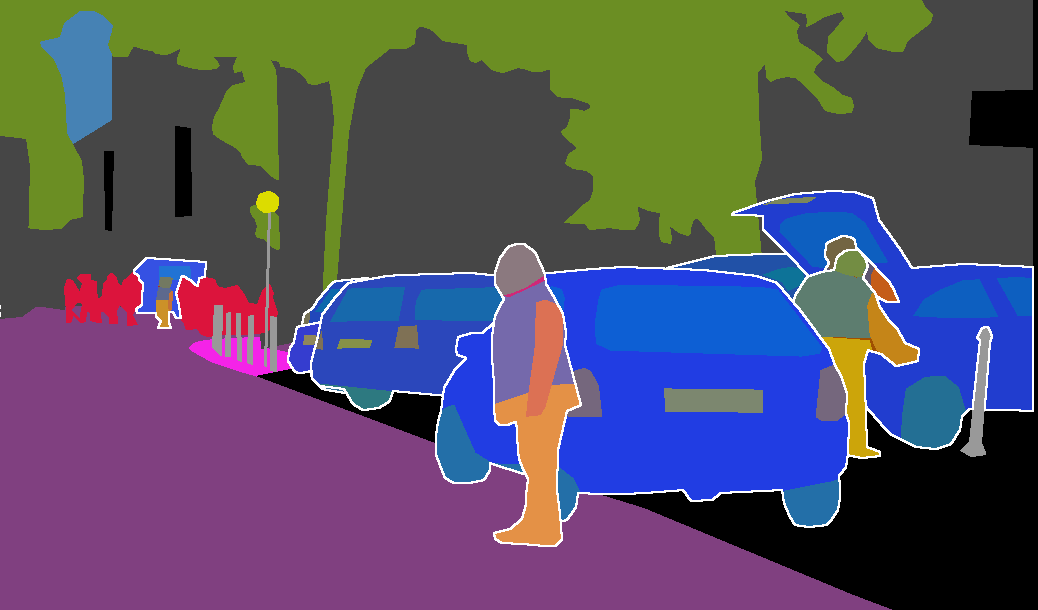}\\
	\caption{More examples from \textit{Cityscapes-Panoptic-Parts}.}
	\label{ch7:fig:cpp-more-examples}
\end{figure}

\begin{figure}
	\centering
	\includegraphics[width=0.35\linewidth]{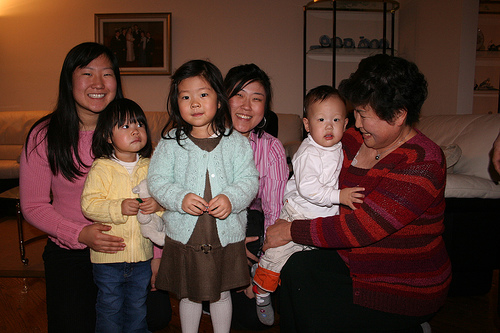}
	\includegraphics[width=0.35\linewidth]{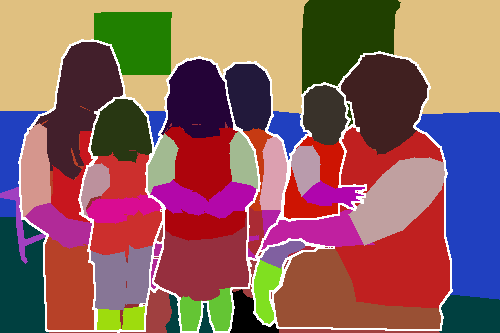}\\
	\includegraphics[width=0.35\linewidth]{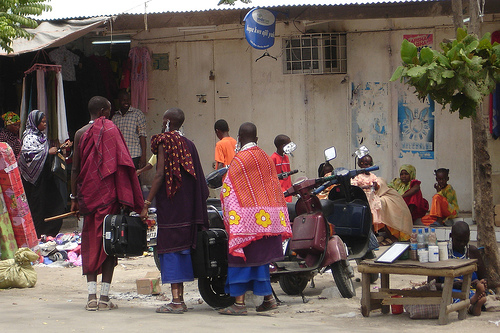}
	\includegraphics[width=0.35\linewidth]{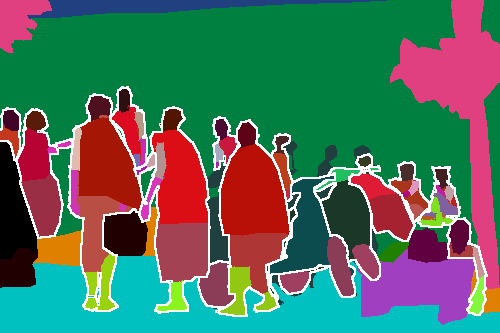}\\
	\includegraphics[width=0.35\linewidth]{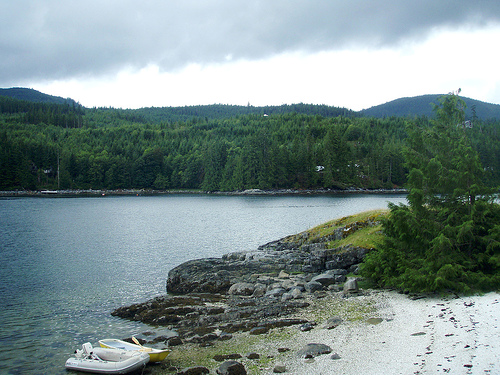}
	\includegraphics[width=0.35\linewidth]{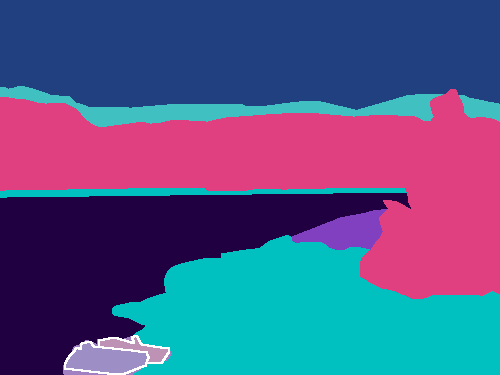}\\
	\includegraphics[width=0.35\linewidth]{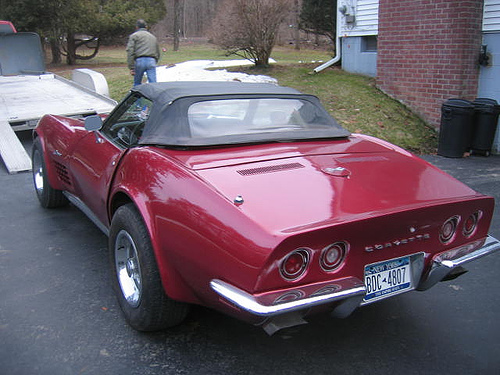}
	\includegraphics[width=0.35\linewidth]{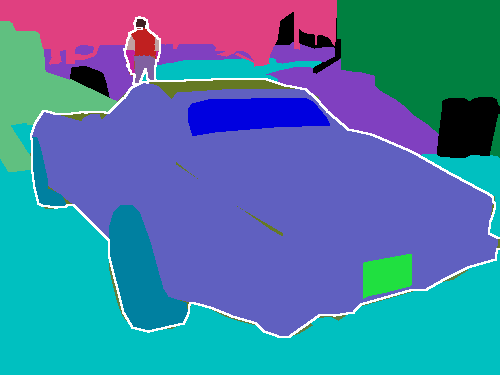}\\
	\includegraphics[width=0.35\linewidth]{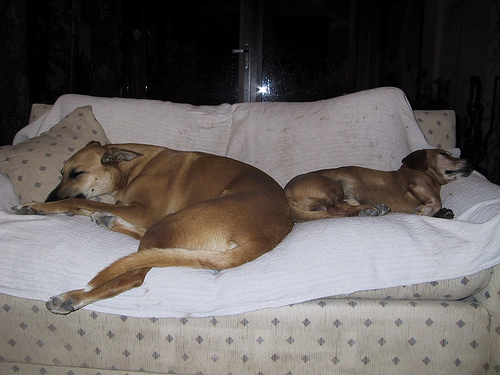}
	\includegraphics[width=0.35\linewidth]{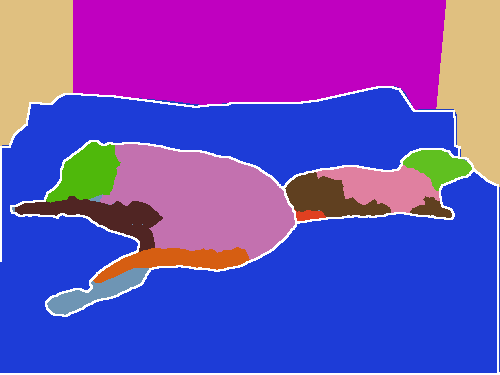}\\
	\includegraphics[width=0.35\linewidth]{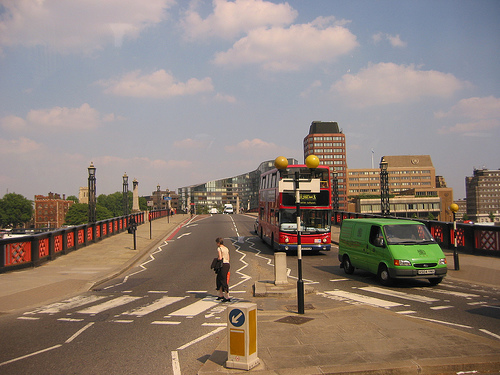}
	\includegraphics[width=0.35\linewidth]{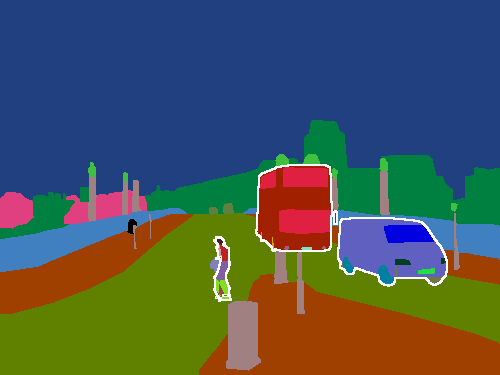}\\
	\caption{More examples from \textit{PASCAL-Panoptic-Parts}.}
	\label{ch7:fig:ppp-more-examples}
\end{figure}

Examples for CPP and PPP datasets are shown in Figures~\ref{ch7:fig:cpp-examples},~\ref{ch7:fig:ppp-examples},~\ref{ch7:fig:cpp-more-examples}, and~\ref{ch7:fig:ppp-more-examples}. For both CPP and PPP, part-level classes are only defined for scene-level things classes. In future work, we anticipate designers of datasets also opt for defining part classes for stuff classes. If so, this is fully compatible with our task definition and metric, since the framework already supports this.

\subsection{Dataset: CPP labeling protocol and part class definitions}
\label{ssec:annot-proc}
A team was assembled containing 12 annotators and three deep learning experts for labeling and constituting reliable data for the experiments. The annotation workload was split such that every city subset in the Cityscapes data was labeled by multiple annotators to ensure that human bias for parts class definitions and labeling errors were minimized. The annotators were asked to start annotating from background to foreground objects, and for each object to start annotating classes in the order of appearance in the above tables. Small objects and indistinguishable parts were not annotated at the part level, and thus maintain the scene-level semantics. Moreover, it was not necessary for objects to have all parts classes. If a class was not visible or behind a foreground object, it was not annotated.

These latter aspects were anticipated in the part-aware panoptic quality metric, which does not evaluate Part-PQ on pixels that do not have part semantics. The procedure for creating the part-level panoptic annotations was therefore defined with the following steps.
\begin{enumerate}[noitemsep, topsep=0pt]
	\item Prior to annotation, an image was masked according to \textit{human} and \textit{vehicle} Cityscapes class labels. Then, these two masked images were provided to the annotators.
	\item The annotators created polygons to label part regions. The polygons at the boundaries of objects and outside of the masked pixels were not asked to be precise, since those areas could be automatically discarded in post-processing.
	\item After annotating, the manual annotated results were supplied to automatic post-processing, which merged original Cityscapes annotations and the new parts annotations into the proposed hierarchical (three-level) format, using the object masks from the first step when required.
\end{enumerate}
The above procedure guarantees that the parts annotations do not overlap with existing Cityscapes annotations not corresponding to objects. Moreover, for regions or instances for which it is difficult to provide part-level annotations due to a) their small size, b) indistinguishable parts, or c) contradictory labels, the original Cityscapes panoptic (two-level) labels are maintained. For example, although some backpack pixels in the original Cityscapes are labeled with the \textit{person} class, part-level annotations are not provided for them and thus maintain the original labels.

\begin{table}
	\centering
	\small
	\begin{tabular}{ll}
		\toprule
		Part class & Definition\\
		\midrule
		Window & Windows, wind shields and other glass surfaces on vehicles. \\
		Wheel & All wheels and tires under vehicles (excluding spare tires). \\
		Light & Light source present on vehicles, including taxi sign. \\
		License plate & License plate on front/back side of vehicles. \\
		Chassis & Part of vehicle body not belonging to above classes. \\
		Unlabeled & Ambiguous or not clearly visible regions. \\
		\bottomrule
	\end{tabular}
	\caption{\textit{Vehicle} part classes for Cityscapes Panoptic Parts.}
	\label{tab:vehicle-parts}
\end{table}

\begin{table}
	\centering
	\small
	\begin{tabular}{ll}
		\toprule
		Part class & Definition\\
		\midrule
		Torso & Core of human body, excluding limbs and head.\\
		Head & Human head. \\
		Arm & Arms, from shoulders to hands. \\
		Leg & Legs, from hips to feet. \\
		Unlabeled & Ambiguous or not clearly visible regions. \\
		\bottomrule
	\end{tabular}
	\caption{\textit{Human} part classes Cityscapes Panoptic Parts.}
	\label{tab:human-parts}
\end{table}

The Cityscapes dataset aims at urban scene understanding and automated driving. Adhering to that direction, we choose to annotate three important \textit{vehicle} classes,~\ie \textit{car}, \textit{truck}, \textit{bus}, and all \textit{human} classes, ~\ie \textit{person}, \textit{rider}. The \textit{vehicle} and \textit{human} categories describe semantic classes with similar parts, thus we define the same semantic parts for each of the classes in these categories. The 23 part classes are defined in Tables~\ref{tab:vehicle-parts},~\ref{tab:human-parts}.

\subsection{Hierarchical label format}
\label{ch7:ssec:label-format}
The amount of information that needs to be encoded in the ground-truth files is much larger than for tasks that involve fewer abstractions and annotation layers. Hence, a compact format needs to be adopted. We have decided to extend the Cityscapes dataset~\cite{Cordts2016Cityscapes} label format, due to its compactness and directness and have included part-level labels in a hierarchical manner. The Cityscapes dataset is labeled pixel-wise with an integer (base 10) \textit{id}, which has up to 5 digits. Every pixel in an image has a \textit{semantic id} (0-99), encoding either \textit{things} or \textit{stuff} semantic classes,~\eg car, person, building, traffic light. If a pixel belongs to a countable object (\textit{thing}), it may also have an \textit{instance id} (0-999), thereby encoding different instances of the same semantic class in an image. The semantic classes are a fixed, predefined set for the whole dataset. The \textit{instance id} is a counter per \textit{things} semantic class and per image.

We have extended this format with a two-digit \textit{part id} (0-99) denoting the part-level semantic classes. The format enables to define up to 100 part classes for every \textit{things} semantic class. Moreover, the parts are bounded to a specific instance, which makes the proposed format compatible with the recently introduced instance-wise object parsing task~\cite{li2017holistic, zhao2018understanding, gong2018instance}.

To summarize the format of the labels each pixel has an:
\begin{itemize}[noitemsep, topsep=0pt]
	\item Up to 2-digit \textit{semantic id}, encoding a \textit{things} or \textit{stuff} semantic class.
\end{itemize}

If in the \textit{things} semantic class, a pixel can optionally have:
\begin{itemize}[noitemsep, topsep=0pt]
	\item Up to 3-digit \textit{instance id}, a counter of instances per image.
\end{itemize}

Finally, if in the \textit{things} semantic class and labeled instance-wise, a pixel can optionally have:
\begin{itemize}[noitemsep, topsep=0pt]
	\item Up to 2-digit \textit{part id}, encoding the parts semantic class per-instance and per-image.
\end{itemize}

The aforementioned format compactly encodes \textit{ids} into an up to 7-digit \textit{id}, for which the first two digits (starting from the left) encode the semantic class, the next 3 digits encode the instance (after zero pre-padding), and the final two digits encode the part class (after zero pre-padding). The following formula is used producing \textit{uids} that can be stored in a single image-like file specified by:

\noindent
{\small
	$
	\textit{uid} = 
	\begin{cases}
		(\textit{semantic id}) & \text{{\footnotesize semantic level}} \\ 
		(\textit{semantic id}) \cdot 10^3 + (\textit{instance id}) & \text{{\footnotesize semantic, instance levels}} \\ 
		(\textit{semantic id}) \cdot 10^5 + (\textit{instance id}) \cdot 10^2 + (\textit{part id}) & \text{{\footnotesize semantic, instance, parts levels}}
	\end{cases}
	$
}

For example, in the Cityscapes Panoptic Parts dataset, a sky (\textit{stuff}) pixel will have $uid = 23$, a car (\textit{things}) pixel that is labeled only on the semantic level will have $uid = 26$, if it is labeled also at instance level it may have $id = 26002$, and a person (\textit{things}) pixel that is labeled at all three levels (of Figure~\ref{fig:tasks}) can have $id = 2401002$.

The unlabeled / void / ``do not care pixels'' are handled in the three levels as follows.
\begin{itemize}[noitemsep,topsep=0pt]
	\item Semantic level: For Cityscapes Panoptic Parts, the original Cityscapes void class is used. For PASCAL Panoptic  Parts the class with $uid = 0$ (first class) is used.
	\item Instance level: For instances the void class is not needed. If a pixel does not belong to an object or cannot be labeled at instance level, then it has only an up to 2-digit \textit{semantic id}.
	\item Parts level: For both datasets, we use the convention that for each semantic class, the part-level class with $uid = 0$ (first class) represents the void pixels. For example, a person pixel, $uid = 2401000$ represents the void parts pixels of instance $10$. The need for a void class arises during the manual annotation process, but in principle it is not needed at the parts level. Thus, we try to minimize void parts level pixels and assign them instead only the semantic and/or instance level labels.
\end{itemize}

\section{\label{ch7:sec:experiments}Setting the baselines}
\begin{figure}
	\centering
	\includegraphics[width=0.9\linewidth]{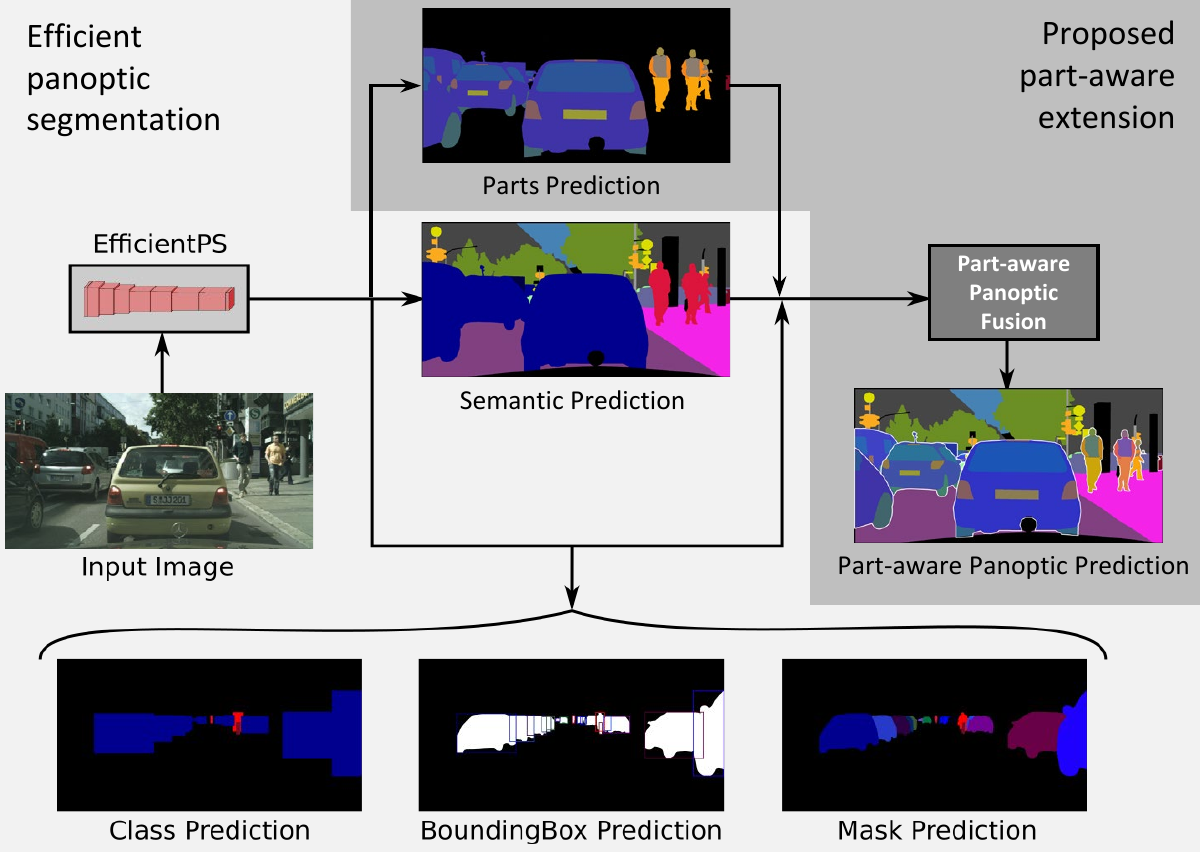}
	\caption{Extensions to Efficient PS network~\cite{mohan2020efficientps} to achieve part-aware panoptic segmentation.}
	\label{ch7:fig:method-citys}
\end{figure}

The two introduced datasets, namely Cityscapes and PASCAL Panoptic Parts, are specifically generated for the novel task of \textit{Part-aware Panoptic Segmentation}. As a consequence, they were not researched as-is in the past. However, the proposed datasets extend the Cityscapes and PASCAL datasets, which have been thoroughly researched in related work under various scene understanding problems, such as part segmentation, semantic segmentation, and panoptic segmentation. This section sets the baselines for part-aware panoptic segmentation, using state-of-the-art panoptic segmentation networks and extends them with part segmentation to achieve the appropriate output format of the predictions. The predictions are then evaluated using the two family of metrics introduced in Section~\ref{ch7:ssec:metrics}.

\subsection{Cityscapes Panoptic Parts}
For the Cityscapes Panoptic Parts (CPP) dataset, we extend the state-of-the-art Efficient Panoptic Segmentation network~\cite{mohan2020efficientps}. This network is design to perform panoptic segmentation, based on a top-down strategy and post-inference merging of semantic and instance segmentation outputs. An extra branch is added aside semantic segmentation, which is responsible for part segmentation of the 23 CPP classes. All outputs,~\ie the original and the parts output, are merged using a part-aware heuristic strategy, to obtain the final part-aware panoptic output. According to the strategy, the original panoptic output has priority over the parts output. Figure~\ref{ch7:fig:method-citys} illustrates this process using a schematic diagram. Results and comparisons with merging methods are provided in Table~\ref{tab:experiments:baselines_cs}.

\begin{landscape}
\begin{table}
\centering
\small
\renewcommand{\arraystretch}{1.4}
\setlength\tabcolsep{5.0pt}
\begin{tabular}{ll||cc|c|ccc||ccc}
	& & \multicolumn{6}{l||}{\textit{Before merging}}  & \multicolumn{3}{l}{\textit{After merging}}  \\
	&  & \multicolumn{2}{c|}{\textbf{mIOU}} & \textbf{AP} & \multicolumn{3}{c||}{\textbf{PQ}} & \multicolumn{3}{c}{\textbf{PartPQ}} \\ 
	& & SemS & PartS & mask & NP+P & NP & P & NP+P & NP & P \\ 
	\midrule
	\multicolumn{3}{l}{\textbf{Merging pre-trained panoptic and part segmentation nets}~\cite{us2021part}} \\
	~~UPSNet \cite{xiong2019upsnet} & DeepLabv3+ \cite{Chen2018deeplabv3plus} &  75.2 & 75.6 & 33.3 & 59.1 & 59.7 & 57.3  & 55.1 & 59.7 & 42.3 \\
	~~DeepLabv3+ \& Mask R-CNN* \cite{Chen2018deeplabv3plus, he2017mask} & DeepLabv3+ \cite{Chen2018deeplabv3plus} & 78.8 & 75.6 & 36.5 & 61.0 & 61.9 & 58.7 &  56.9 & 61.9 & 43.0  \\
	~~EfficientPS \cite{mohan2020efficientps} &  BSANet \cite{Zhao2019BSANet} &  80.3 & 76.0 & 39.7 & 65.0 & 65.2 & 64.2  & 60.2 & 65.2 & 46.1  \\
	~~HRNet-OCR \& PolyTransform* \cite{Yuan2020ocr, Liang2020polytransform} &  BSANet \cite{Zhao2019BSANet} &  81.6 & 76.0 & 44.6  & 66.2 & 67.0 & 64.2 & 61.4 & 67.0 & 45.8 \\
	\multicolumn{2}{l}{\textbf{Integrated trained-from-scratch PPS network} (proposed)} \\
	\multicolumn{2}{l||}{~~EfficientPS with part segmentation} & 80.0 & 75.0 & 39.6 & 64.7 & 65.0 & 64.0 & 59.4 & 64.5 & 45.1
\end{tabular}
\caption{Baseline results for Cityscapes Panoptic Parts as reported in~\cite{us2021part} and results from an integrated solution described in this chapter. The first four rows contain results that are generated by merging predictions from state-of-the-art networks trained on only panoptic segmentation and only part segmentation ground truth. The final row presents results from a single network trained on the CPP dataset. The results are compared before and after the merging of panoptic and part segmentation predictions. mIOU\textsubscript{PartS} indicates the mean IOU for part segmentation on grouped parts (see Subsection~\ref{ch6:ssec:cpp}). Metrics split into \textit{P} and \textit{NP} are evaluated on scene-level classes with and without parts, respectively. Symbol * indicates pretraining on the COCO dataset~\cite{Lin2014COCO}.}
\label{tab:experiments:baselines_cs}
\end{table}
\end{landscape}

\subsection{PASCAL Panoptic Parts}
To the best of our knowledge, there is no panoptic segmentation network publicly available, thus for the PASCAL Panoptic Parts (PPP) dataset, we integrate scene and part-level segmentation and combine the results with an instance segmentation network. This architecture consists of an incremental step over the merging strategy proposed in~\cite{us2021part}, so the results can be compared easier. The scene and part segmentation network is based on DeepLabv3+, augmented with an extra prediction branch that is responsible for part segmentation of the 59 PPP classes. For the merging process, first the scene semantics are merged with instances to obtain a panoptic output. Then, this output is merged with the parts output using a part-aware heuristic strategy, to obtain the final part-aware panoptic output.

The results and comparisons with merging methods are provided in Table~\ref{tab:experiments:baselines_pascal}. As can be noticed from the results, integrating scene and part semantics in a single network is more beneficial for mIOU performance before merging (+0.5\%) than after merging (+0.2\%). The merging strategy prioritizes instances over parts, which is a possible explanation of this degradation. Overall, an increasing trend for all metrics before or after merging is observed, between rows one and three, which have the same base networks. Therefore, apart from the benefit of reducing the computational load by 1/3, solving the semantics within a single network is advantageous and hints that implicit information sharing between the scene and part-level semantics may take place.


We provide a few observations on the obtained results. It is remarked that scene-level things classes that contain parts,~\ie those belonging to $\mathcal{L}^\text{Th} \cap \mathcal{L}^\text{parts}$, have significantly lower PartPQ results, since PartPQ considers in the calculation the segmentation of parts inside every instance (refer to Section~\ref{ch7:ssec:metrics}). PartSQ follows the same trend as it reflects segmentation accuracy, while PartRQ remains the same as it reflects recognition quality. Another observation is that a few classes have a rather low score (\eg \textit{bed}, \textit{bench}, \textit{cloth}, \textit{flower}, \textit{truck}), which is explained by the small number of representative examples (the object size in pixels is at least two orders of magnitude smaller than the most of the classes) of those classes in the dataset. However, the parts inside the objects are still segmented with reasonable scores (\eg \textit{bed} PQ/PartPQ: 3.9\% and SQ/PartSQ: 61.6\%), which confirms the capabilities of the presented concept.

\begin{landscape}
	\begin{table}
		\centering
		\small
		\renewcommand{\arraystretch}{1.4}
		\setlength\tabcolsep{5.0pt}
		\begin{tabular}{ll||cc|c|ccc||ccc}
			& & \multicolumn{6}{l||}{\textit{Before merging}}  & \multicolumn{3}{l}{\textit{After merging}}  \\
			& & \multicolumn{2}{c|}{\textbf{mIOU}} & \textbf{AP} & \multicolumn{3}{c||}{\textbf{PQ}} & \multicolumn{3}{c}{\textbf{PartPQ}} \\ 
			& & SemS & PartS & mask & NP+P & NP & P & NP+P & NP & P \\ 
			\midrule
			\multicolumn{3}{l}{\textbf{Merging pre-trained panoptic and part segmentation nets}~\cite{us2021part}} \\
			~~DeepLabv3+ \& Mask R-CNN \cite{Chen2018deeplabv3plus, he2017mask} & DeepLabv3+ \cite{Chen2018deeplabv3plus} & 47.1 & 53.9 & 38.5 & 35.0 & 26.0  & 61.5 &  31.4 & 26.0 & 47.2  \\
			~~DLv3-ResNeSt269 \& DetectoRS \cite{chen2017deeplabv3, zhang2020resnest, qiao2020detectors} &  BSANet \cite{Zhao2019BSANet} & 55.1 & 58.6 & 44.8 & 42.0 & 33.8 & 66.0 & 38.3 & 33.8 & 51.6\\
			\multicolumn{6}{l}{\textbf{Merging scene/parts sem. segm. and inst. segm. trained-from-scratch nets} (proposed)} \\
			~~DeepLabv3+ & Mask R-CNN & 47.8 & 54.4 & 38.7 & 35.4 & 26.8 & 61.8 &  31.6 & 26.8 & 47.4
		\end{tabular}
		\caption{Baseline results for PASCAL Panoptic Parts as reported in~\cite{us2021part} and results from merging a combined scene and part-level semantics network (semantic and part segmentation) with instance segmentation. The first two rows contain results that are generated by merging predictions from state-of-the-art networks trained on only panoptic segmentation and only part segmentation ground truth. The final row presents results trained on the PPP dataset. The results are compared before and after the merging of panoptic and part segmentation predictions.
		Metrics split into \textit{P} and \textit{NP} are evaluated on scene-level classes with and without parts, respectively.}
		\label{tab:experiments:baselines_pascal}
	\end{table}
\end{landscape}

\section{Conclusions}
\label{ch7:sec:conclusions}
This chapter has introduced the novel task of Part-aware Panoptic Segmentation (PPS). This task belongs to the family of visual scene understanding tasks and unifies the previously separately studied branches of \textit{scene parsing} and \textit{part parsing} within a consistent formulation. This is achieved by building on the panoptic segmentation task and enriching it with part semantics as a second abstraction level, where the part semantics are derived from the semantically meaningful parts of the main objects in the first abstraction level.

Moreover, to support research on this new task we have constructed two datasets, namely Cityscapes Panoptic Parts and PASCAL Panoptic Parts, which extend two established scene understanding datasets to validate part-aware segmentation. Finally, baselines are trained using these datasets and results are compared between networks trained for constituent tasks and the new compound segmentation task. The developed single-network approaches for PPS have shown strong performance, on par with combining state-of-the-art networks for different tasks. This demonstrates that single-network approaches and top-down merging strategies are competent for realizing part-aware panoptic segmentation.

To summarize, this chapter has presented the following contributions.
\begin{itemize}[noitemsep,topsep=0pt]
\item Introduction of the part-aware panoptic segmentation (PPS) task, unifying perception of objects at multiple levels of abstraction.
\item Coherent PPS annotations for two commonly used datasets, which have become publicly available.
\item Baseline solutions exploiting EfficientPS and DeepLabv3+ networks for the PPS task on two datasets.
\end{itemize}

The coherent unification of \textit{scene parsing} and \textit{part parsing} into part-aware panoptic segmentation presents a large step forward towards holistic scene understanding. The results are sufficiently interesting to spark innovations in deep learning that jointly analyze a scene at multiple levels of abstractions,~\ie part level and scene level, and eventually leverage the interaction between these levels.


\chapter{\label{ch:8-conclusions}Conclusions and outlook} 


This thesis addresses 2-D image scene understanding and presents research on semantic segmentation and panoptic segmentation with an emphasis on urban scenes and automated driving applications. This chapter summarizes the conclusions on individual chapters from a thesis-level perspective, discusses the research questions, and gives a future outlook.

\section{Chapter conclusions} 

\noindent \textbf{Chapter 2} identifies the challenges involved in multi-dataset training of CNNs for semantic segmentation and proposes a methodology to solve them based on handling semantic conflicts. These challenges arise from the semantic conflicts within the classes of the label spaces of the employed datasets. To solve these conflicts, a hierarchical organization of the label spaces is proposed, according to their semantics, after which the corresponding architecture of hierarchical classifiers is designed. The experiments are conducted on different combinations of three datasets with conflicting label spaces. The results demonstrate that multi-dataset training improves the segmentation performance and the number of recognizable semantics by a single trained network. The hierarchical structure of classifiers enables simultaneous training and maximum usage of all available supervision resources, regardless of the differences in label spaces. Results from a variant of the proposed methodology have been submitted to the CVPR 2018 Robust Vision Challenge, consisting of 4 datasets, and attained 3\textsuperscript{rd} place overall and 1\textsuperscript{st} - 7\textsuperscript{th} places in individual datasets.

\noindent \textbf{Chapter 3} extends the methodology for multi-dataset training developed in Chapter 2 to include weakly-labeled datasets, complementary to strongly-labeled datasets. Weakly-labeled datasets with bounding-box and image-tag labels are easier to be annotated, hence they contain more images and have higher semantic diversity, compared to pixel-labeled datasets. In order to simultaneously leverage from strong and weak labels, a mixed fully and weakly-supervised training approach is proposed. This approach employs the hierarchical classifiers from Chapter~\ref{ch:3-iv2018} to solve semantic conflicts between datasets and introduces a novel loss function to amend incompatibilities between different supervision types. This loss handles weak and strong supervision in a unified manner, while maintaining the conventional FCN training procedure. The experiments involve two urban-scene datasets and one large-scale generic-scene dataset, and show that weakly-labeled datasets can be successfully leveraged to improve segmentation performance of the semantic classes for which weak supervision is provided. Specifically, per-class IoU gains of up to +13.2\% are achieved for the selected strongly-labeled classes that were complemented with weak supervision. The included bounding-box labels are beneficial for increasing segmentation performance in all cases, while the image-tag labels benefit one of the two datasets in the experiments. Moreover, it is shown that for each order of magnitude of extra bounding-box supervision, the segmentation performance increases proportionally by +0.5\% mIoU.

\noindent \textbf{Chapter 4} improves the computing aspects of the previously proposed multi-dataset training methodology, by devising two (semi-)automated procedures for selecting informative and diverse weakly-labeled data. The increase in segmentation performance and recognized semantics in the previous chapters, achieved by the hierarchical construction of classifiers is acquired at the expense of computing and memory resources. The two proposed selection procedures aim at reducing these required resources by balancing the amount of strong and weak supervision through rejecting or decreasing repeatability of images with low informative value. This is achieved by enhancing the object diversity of selected weakly-labeled images and attaining visual similarity between strongly and weakly-labeled images. The experiments demonstrate that careful selection of image-label pairs from weakly-labeled datasets using the proposed procedures result in similar segmentation performance gains, while using up to 100 times less pairs. The selection procedure reduces the training time by 20 - 90\% and thereby the corresponding energy consumption in a proportional way, while the maximum drop in performance is 0.5\% compared to using all pairs.

\noindent \textbf{Chapter 5} proposes a complete framework for Heterogeneous Training of Semantic Segmentation (HToSS), by re-considering the limitations of the developed methodologies from previous chapters. HToSS achieves multi-dataset training on a variety of strongly-labeled and weakly-labeled datasets and reduces the requirements on compliance for the datasets, in order to be admissible for multi-dataset training. HToSS is more efficient both in memory and computations, compared to hierarchical classification and neither requires manual data relabeling nor the use of external modules as in other literature methods. Extensive experiments on 14 datasets reveal consistent improvements in i) the segmentation performance on test splits of seen (training) datasets, ii) generalization on unseen datasets, and iii) awareness of semantic concepts, expressed by the novel proposed knowledgeability metric. Moreover, the HToSS methodology allows supplementing the conventional pixel-labeled training data with other relevant datasets that otherwise would not be compatible. These properties make HToSS useful for many applications, where training data for semantic segmentation are too scarce to achieve the required performance and generalization. Experiments on combining strongly-labeled datasets by resolving conflicts in their label spaces demonstrate segmentation performance gains up to +20\% mIoU on seen datasets, up to +16.6\% on unseen (generalization) datasets, while obtaining a 3 - 143\% relative increase in the number of recognizable classes. The key results from experiments on combining strongly and weakly-labeled datasets are the increase of +3\% generalization performance and a relative increase in the output classes between 70 - 250\% with maintaining segmentation performance on seen and unseen datasets. Finally, HToSS w.r.t. resources adds only a small number of weights (+10\textsuperscript{3}) and a minor delay (+1.2 ms) in inference compared to various techniques for combining datasets.

\noindent \textbf{Chapter 6} introduces the novel task of part-aware panoptic segmentation and proposes a first baseline network for solving this task. Part-aware Panoptic Segmentation (PPS) unifies the traditionally distinct tasks of part segmentation and panoptic segmentation in an endeavor towards achieving holistic scene understanding. The proposed formulation consistently encompasses part-level and scene-level semantics, together with instance-level object counting, thereby effectively covering all four aspects of scene understanding analyzed in Chapter~\ref{ch:1-intro}. PPS is a super-task of various segmentation tasks (\eg semantic, instance, part, panoptic), and as such, it contains information on multiple abstraction levels for scene understanding. The proposed single-backbone model is based on a state-of-the-art panoptic segmentation model and augments it with part-level semantics. The simultaneous handling of all abstractions within a single model produces coherent and conflict-free pixel-based predictions. The model is trained with two new datasets, which are created to support research on the novel task and extend two established datasets. The trained baseline models are competitive with two combinations of state-of-the-art methods specifically tuned for part-aware panoptic segmentation constituent sub-tasks, however they perform inference at (almost) half the computational cost, as a common feature extraction is used.

\section{Discussion of the research themes}

The objective of this thesis is to improve image segmentation for scene understanding by leveraging multiple heterogeneous datasets and by analyzing a scene at multiple levels of abstraction, as stated in Section~\ref{ch1:sec:challenges-rq}. Chapters~\ref{ch:3-iv2018} and~\ref{ch:6-journal} identify and concisely delineate the challenges that arise from combining multiple datasets. Chapters ~\ref{ch:3-iv2018} and~\ref{ch:4-iv2019} address these challenges by providing a preliminary methodology for multi-dataset training of CNNs to improve segmentation performance and generalization. Chapter~\ref{ch:5-itsc2019-wacv2019} investigates training efficiency and minimizes memory and computation resources of the proposed methodology in the preceding two chapters. Chapter~\ref{ch:6-journal} reconsiders the problem of heterogeneous training for semantic segmentation and devises a generic framework resolving the limitations of the preliminary methodology. The framework improves semantic segmentation performance, generalization, and semantic knowledgeability. Finally, Chapter~\ref{ch:7-panoptic} improves panoptic segmentation at multiple levels of abstraction by extending the task with part-aware semantics, which constitutes an important step towards holistic scene understanding. The remainder of this section discusses the research themes and questions posed in Section~\ref{ch1:sec:challenges-rq}.

\paragraph{RT1}
Leverage heterogeneous datasets from a variety of scene-understanding tasks in order to improve semantic segmentation.

Semantic segmentation is an important aspect of scene understanding for a variety of applications, hence semantic segmentation models should provide high accuracy in a variety of urban and generic environments and understand numerous different semantic concepts appearing in these scenes. Therefore, the training of CNNs for semantic segmentation requires large annotated datasets of rich semantic content. However, due to the required manual effort, the datasets for semantic segmentation are relatively small and have a limited semantic richness, compared to other scene-understanding datasets for related tasks. Moreover, as each dataset is generated to serve a different purpose and is based on different goals and requirements, they have different levels of semantic granularity, which makes them incompatible for training a single network architecture with the combination of all datasets. We advocate the use of the large variety of existing datasets, instead of creating increasingly larger datasets suited for the format of a single task.

\begin{itemize}
\item RQ1a: \textit{Which challenges arise from using heterogeneous datasets to train convolutional networks for semantic segmentation?}

Image scene understanding includes a variety of tasks that require predictions to fulfill different requirements, as outlined in Section~\ref{ch1:sec:scene-und}. Accordingly, datasets created for each of these tasks have different characteristics, number of semantic classes, coverage and localization of annotations. Thus, the combination of multiple datasets for simultaneous training of CNNs for semantic segmentation entails various challenges. The two fundamental challenges are associated with the label spaces and the annotation types of the employed datasets. First, the semantic conflicts between the label spaces do not allow straightforward concatenation/merging to create an unambiguous and conflict-free output label space. Second, the weak annotation types,~\eg image tags and bounding boxes, are incompatible for pixel-based inference. Chapters~\ref{ch:3-iv2018} and~\ref{ch:6-journal} further describe in detail these challenges and formalize the problem of heterogeneous data training.

\item RQ1b: \textit{How can we combine existing datasets for semantic segmentation that are annotated on disjoint label spaces?}

One of the main challenges for combining heterogeneous datasets is represented by the conflicts between the label spaces of separate datasets. Since each dataset is annotated for different purposes, type of scenes, or classes of interest, it is rather common that the label spaces have conflicting semantics. These conflicts occur either due to different semantic granularity of specific classes, or due to different definitions for the same semantic concepts among datasets. In this thesis, two methodologies for solving conflicts are proposed. The first in Chapter~\ref{ch:3-iv2018} aims at creating a hierarchy of the semantic classes from all datasets and then train a matching hierarchy of classifiers. The second methodology, presented in Chapter~\ref{ch:6-journal}, generates a ``flat'' taxonomy of semantic atoms that are fine-level semantic primitives which are extracted from the semantic classes of datasets. The obtained taxonomy is more flexible compared to the hierarchy, can be applied universally to any label space, and is independent from the classifier structure.

\item RQ1c: \textit{Is it possible to use datasets with weak supervision to train convolutional networks for semantic segmentation?}

Semantic segmentation requires pixel-level localization for predictions, so that pixel-labeled ground truth provides the best supervision for training CNNs. In contrast, weak forms of supervision are more common and the number of available datasets is vast. To this end, Chapter~\ref{ch:4-iv2019} proposes a methodology for supervising CNNs with weakly-labeled images. The proposed approach has low complexity, which is achieved by using cues from network predictions and a new loss function. Chapter~\ref{ch:6-journal} improves and generalizes the aforementioned approach to incorporate it within the complete HToSS framework presented in that chapter. The experiments show that weak supervision in the form of bounding boxes or image tags can provide useful learning signals to improve accuracy of predictions. The localization of labels is proportional to the increase in accuracy and inversely proportional to the required amount of weak supervision.
\end{itemize}

\paragraph{RT2}
Efficient training and inference for semantic segmentation with a growing amount of datasets and increasing number of semantic classes.

The benefits of multi-dataset training of CNNs are accompanied by efficiency issues during their deployment. During each training step, representative samples from each dataset and each semantic class should be present in the batch formation. During inference on hardware with limited resources or on integrated platforms, complex architectures or multi-classifier systems have compelling computing requirements. This implies the creation of efficient solutions that decouple or at least reduce dependencies on the considered number of semantic classes and datasets, and those on the mutually dependent network latency, memory, and FLOPS. The following research questions stem from the hierarchical classification scheme proposed in Chapter~\ref{ch:3-iv2018}, but have a generic impact on the overall multi-dataset training scheme.

\begin{itemize}
	\item RQ2a: \textit{Does training for semantic segmentation with multiple datasets scale well with an increasing number of datasets? How is inference influenced when the output label space is large?}

    An increasing number of datasets considered for simultaneous training brings segmentation accuracy benefits, but also induces higher requirements in memory and computation. Since representative examples from each dataset should be present for batch-wise training, this results in a larger output label space requiring more examples from each semantic class to be included. The hierarchical classifier approach described in Chapter~\ref{ch:3-iv2018} effectively introduces a new classifier for each added dataset. Unfortunately, the addition of classifiers increases the parameter count of the network, the required memory, and thereby processing power. As a consequence, this approach does not scale well with an increasing number of datasets. To amend these issues, Chapter~\ref{ch:6-journal} proposes an alternative framework that maintains a single classifier, irrespective of the number of employed datasets, which keeps the network parameters constant. This alternative also grows with the number of classes, albeit with a lower multiplicative factor.

    \item RQ2b: \textit{How can dataset imbalances be mitigated in a multi-dataset training scenario?}
    
    The size of datasets for scene understanding varies in accordance with the tasks involved for those datasets and the level of richness in annotations. In this thesis, the employed datasets range from a couple of thousands to millions of images, which results in imbalanced training for larger datasets or overfitting for smaller datasets. Chapter~\ref{ch:5-itsc2019-wacv2019} attempts to mitigate this challenge by limiting the size of datasets, using smart selection of image-label pairs applied to semantic segmentation. In that chapter, two specific strategies are proposed for selecting pairs from the weakly-labeled datasets, so they do not outnumber the strongly-labeled datasets. The first procedure aims at selecting pairs according to their similarity with the strongly-labeled dataset, while the second procedure pursues to select pairs providing higher object diversity for the training. After the selection is performed, the standard training approach can again be followed, as described in Chapters~\ref{ch:3-iv2018} and~\ref{ch:4-iv2019}.

    \item RQ2c: \textit{Given restricted memory resources, how can we maximize the number of employed datasets and reduce the throughput time for training?}
    
    Chapter~\ref{ch:5-itsc2019-wacv2019} investigates ways of selecting informative image-label pairs and rejecting repetitive or low-informative pairs according to specific criteria for achieving balanced training and reduced training time. Using the proposed selection procedures from the answer to RQ2b, the most visually similar pairs are kept and the ones with high informative value. This approach results in a having a better mixture of samples from the datasets, thereby effectively maximizing the number of employed datasets, while limiting the training time.
\end{itemize}

\paragraph{RT3}
Extend the scene-level panoptic segmentation task with part-level semantics towards holistic scene understanding.

The comprehensive understanding of scenes requires analysis at multiple levels of abstraction and extraction of information that can be utilized by any application, ranging from automated driving to surveillance. Semantic segmentation from the previous chapters provides a description of scene semantics at a very localized level (per-pixel), albeit that it lacks aspects like detecting objects or object parts. Panoptic segmentation augments semantic segmentation with per-pixel instance localization and separation, which provides more informative aspects for applications that depend on the separation between dynamic objects and static elements of a scene. As such, it provides a solid basis for extending it with other levels of abstraction.

\begin{itemize}
\item RQ3a: \textit{How can panoptic segmentation be combined with the concept of part segmentation to enrich the former with semantics of the latter? Can this be incorporated in an unambiguous and consistent manner?}

Panoptic segmentation provides pixel-accurate information on scene-level semantics and object instances. Many real-life applications, such as automated driving, can be enhanced by providing them part-level information, which can be used to anticipate future movements of dynamic objects and understand gestures. Part segmentation provides these cues, however, not at per-object level, since its goal is to segment parts of objects without giving any affinity for these parts. Chapter~\ref{ch:7-panoptic} unifies panoptic segmentation and part segmentation by resolving ambiguities and introducing a part-aware panoptic format. A clear and consistent definition is provided for the task of part-aware panoptic segmentation, which incorporates scene-level and part-level abstractions and information on semantics and instance enumeration.

\item RQ3b: \textit{Is it feasible to train a single network for scene-level and part-level semantics, together with instance-level separation?}

Part-aware panoptic segmentation is a compound task, which requires the simultaneous and unambiguous solution of many sub-tasks, including segmenting an image at the pixel level, separating semantics for things and stuff, detecting objects, identifying parts and associating them with objects. The experiments of Chapter~\ref{ch:7-panoptic} examine the feasibility to solve these sub-tasks using a single feature extraction network having as input a raw image of a scene. The results demonstrate the feasibility of training a single network for part-aware panoptic segmentation, which requires only low overhead over panoptic segmentation and unambiguous inference compared to results taken from networks solving the sub-tasks individually. Besides this, the proposed solution shows a small degradation of the metrics, but almost halves the involved computations.

\item RQ3c: \textit{Are the existing scene-understanding datasets adequate for training and evaluating systems for part-aware panoptic segmentation?}

To the best of our knowledge, there is no publicly available dataset complying to the part-aware panoptic segmentation format. Existing datasets for scene understanding cover information required by sub-tasks of the novel task, hence they can be used for training, but not for evaluation. In order to offer the research community a complete benchmark for part-aware panoptic segmentation, Chapter~\ref{ch:7-panoptic} proposed two new datasets, which are based on existing established benchmarks. The first dataset is extended from the panoptic-compliant Cityscapes dataset with manual part-level annotations. The second dataset is created by combing two sets of annotations on the same set of images, namely instance-aware part-level semantic annotations and scene-level semantic annotations. These two sets of annotations contained conflicts, which are solved by using heuristic rules. The two new datasets comprise an adequate benchmark for part-aware panoptic segmentation, both for training and testing.
\end{itemize}

\section{Discussion and Outlook}
This section provides a general discussion on topics addressed in the thesis and goes beyond its scope by suggesting potential directions for future research and upcoming developments.

The common practice in computer vision for addressing image scene understanding is to split the holistic task of understanding into distinct, well-defined sub-tasks, such as object detection or segmentation. The boundaries of these sub-tasks can be easily defined and specific metrics are devised to evaluate the performance / accuracy of systems. The following question immediately arises: Is it beneficial for a network to solve the holistic problem at once,~\eg by addressing multiple sub-tasks simultaneously, hence providing an unambiguous and comprehensive solution, or address them separately, in which case an extra post-processing algorithm is required to fuse sub-results? In the early years of computer vision, algorithms were limited on solving a specific sub-task,~\eg image classification or face detection, and in a predefined (controlled) environment. The computational capabilities of current systems and the advances in deep learning enable a single network to handle/solve multiple sub-tasks. Observing current trends in industries involving scene understanding,~\eg automated vehicles, it seems that the first approach is still preferred,~\ie solve sub-tasks connected with different autonomous functionalities in an individual way. A justification is that the components can be verified separately, but may be also triggered by legal benefits or standardization for international regulations. In this thesis, we advocate that the second approach is more compelling, since information can be exchanged between sub-tasks to achieve an all-encompassing solution, where the final results are by construction unambiguous and free of conflicts. Although we find that this approach has clear benefits and should be further studied in the future, we refrain from stating that the first is obsolete, and possibly a mixture of both approaches for applied holistic scene understanding is ideal towards achieving general artificial intelligence.
 
Another related discussion revolves around the recent developments in self-supervised learning~\cite{liu2021self,jing2020self,ericsson2021well} and class-agnostic segmentation~\cite{sharma2020class,siam2021video,zhang2019canet}, and their applicability in the topics of this thesis. Two of the most challenging problems faced in the thesis are connected to the training difficulty of networks when the output semantic label spaces are growing and an increasing number of datasets is employed for multi-dataset training. To a large extent, the source of these issues is the current batched gradient-descent-based training algorithms of neural networks. Self-supervised learning and class-agnostic segmentation have the potential to amend these issues. Self-supervised learning has been leveraged in classification and less frequently in other scene understanding tasks as an intermediate step between network initialization and the final network training with task-specific supervision. This self-learning procedure involves the training of a network on a so-called pre-text task, which helps to find a sufficient initialization for the subsequent training on the downstream (practical) task. In the explored cases of multi-dataset training, we have observed that the initialization of networks is of great importance to attain robustness and fast convergence in training, given the current training schemes for neural networks,~\ie gradient-descent algorithms. We hypothesize that self-supervised learning will bring benefits in multi-dataset training and solve the per-dataset and class-candidate issues that arise during batch formation (see Chapters~\ref{ch:5-itsc2019-wacv2019} and~\ref{ch:6-journal}). Regarding class-agnostic segmentation, we have observed that the training difficulty of networks grows with respect to the semantic richness of the output label space. To solve semantic segmentation, the network should intrinsically learn how to localize elements and how to predict semantics. The difficulty of predicting semantics increases when the output label space becomes numerous and more fine-grained. Therefore, we conjecture that a class-agnostic segmentation approach may be advantageous, since it can decouple the segmentation (per-pixel localization), and the prediction of semantics, which can be attributed to segments with a so-called few-shot learning framework. Alternatively, Multi-Class Incremental Learning (MCIL)~\cite{liu2020mnemonics,douillard2021plop} can be employed to address the issue of growing semantics. In MCIL, a model is incrementally updated with new concepts, but this presents the inherent danger of ``catastrophic forgetting'' of the previously learned concepts.

Holistic scene understanding refers to the capability of humans to produce holistic representations of a scene at multiple abstraction levels,~\eg localize elements of the scene and assign semantics to them, then separate them into dynamic or static objects, identify those objects and compute future trajectories for anticipating events. These representations improve modeling of the real world, and subsequently help humans to move around in the scene and safely interact with the elements within their environment. This thesis has focused solely on improving aspects of 2-D scene understanding with input image data specifically captured by visual sensors. The pursued direction for holistic understanding involves extracting a representation from a single image with as many / rich abstractions as possible in a coherent and unambiguous manner between them. Even though 2-D image data provide a thorough view of a scene, they are projections of the 3-D dynamic world produced by camera lenses and sensors. Apart from the introduced aberrations, the projection loses the depth dimension, leading to occlusions and obscures the 3-D structure of the scene. To improve the correct perception of the environment, other visual or non-visual sensory information~\eg Lidars, Radars, GPS, and odometry, should be employed for providing important clues that are not contained within the 2-D image data, like changes the time domain. For further future improvements, it is expected that upcoming systems for holistic scene understanding should smoothly handle multi-modal input information and perform unambiguous, comprehensive 3D temporal scene understanding.

\setlength{\baselineskip}{0pt} 
{\renewcommand*\MakeUppercase[1]{#1}%
\printbibliography[heading=bibintoc,title={Bibliography}]}

\newpage



\chapter*{\label{ch:pubs}Publication List}
\addcontentsline{toc}{chapter}{Publication List}

\thispagestyle{plain}
\pagestyle{plain}

\section*{Journal articles}

\begin{enumerate}[itemsep=0.3em,start=1,label={\bfseries J\arabic*.~~~}]
	\item \textbf{P. Meletis}, and G. Dubbelman. \textit{``Training Semantic Segmentation on Heterogeneous Datasets''}. In: IEEE Transactions on Neural Networks and Learning Systems (under review), 2021.
	\item D. de Geus, \textbf{P. Meletis}, and G. Dubbelman. \textit{``Fast Panoptic Segmentation Network''}. In: IEEE Robotics Automation Letters 5.2, 2020, pp. 1742–1749. \texttt{DOI: 10.1109/LRA.2020.2969919}.
	\item A. Bighashdel, \textbf{P. Meletis}, and G. Dubbelman. \textit{``Diving Into Deep Pedestrian Path Prediction''}. In: (in preparation), 2021.
\end{enumerate}

\section*{International conferences contributions}

\begin{enumerate}[itemsep=0.3em,start=1,label={\bfseries C\arabic*.~~~}]
	\item D. de Geus\footnote[1]{equal contribution},  \textbf{P. Meletis}\footnotemark[1], C. Lu, X. Wen, G. Dubbelman. \textit{``Part-aware Panoptic Segmentation''}. In: IEEE/CVF Conference on Computer Vision and Pattern Recognition. 2021.
	\item C. Sebastian, R. Imbriaco, \textbf{P. Meletis}, G. Dubbelman, E. Bondarev, P. de With. \textit{``TIED: A Cycle Consistent Encoder-Decoder Model for Text-to-Image Retrieval''}. In: IEEE/CVF Conference on Computer Vision and Pattern Recognition Workshops. 2021.
	\item A. Bighashdel, \textbf{P. Meletis}, P. Jancura, and G. Dubbelman. \textit{``Deep Adaptive Multi-Intention Inverse Reinforcement Learning''}. In: European Conference on Machine Learning. 2021.
	\item A. Bighashdel, \textbf{P. Meletis} and G. Dubbelman, \textit{``Towards Equilibrium-based Interaction Modeling for Pedestrian Path Prediction''}, 2020 IEEE 23rd International Conference on Intelligent Transportation Systems (ITSC), Rhodes, Greece, 2020, pp. 1-8. \texttt{DOI: 10.1109/ITSC45102.2020.9294530}.
	\item \textbf{P. Meletis}, R. Romijnders, and G. Dubbelman. \textit{``Data Selection for training Semantic Segmentation CNNs with cross-dataset weak supervision''}. In: IEEE Intelligent Transportation Systems Conference, ITSC 2019, Auckland, New Zealand, October 27-30, 2019. IEEE, 2019, pp. 3682–3688. \texttt{DOI: 10.1109/ITSC.2019.8917069}.
	\item D. de Geus, \textbf{P. Meletis}, and G. Dubbelman. \textit{``Single Network Panoptic Segmentation for Street Scene Understanding''}. In: 2019 IEEE Intelligent Vehicles Symposium, IV 2019, Paris, France, June 9-12, 2019. IEEE, 2019, pp. 709–715. \texttt{DOI: 10.1109/IVS.2019.8813788}.
	\item \textbf{P. Meletis} and G. Dubbelman. \textit{``On Boosting Semantic Street Scene Segmentation with Weak Supervision''}. In: 2019 IEEE Intelligent Vehicles Symposium, IV 2019, Paris, France, June 9-12, 2019. IEEE, 2019, pp. 1334–1339. \texttt{DOI: 10.1109/IVS.2019.8814217}.
	\item R. Romijnders, \textbf{P. Meletis}, and G. Dubbelman. \textit{``A Domain Agnostic Normalization Layer for Unsupervised Adversarial Domain Adaptation''}. In: IEEE Winter Conference on Applications of Computer Vision, WACV 2019, Waikoloa Village, HI, USA, January 7-11, 2019. IEEE, 2019, pp. 1866–1875. \texttt{DOI: 10.1109/WACV.2019.00203}.
	\item \textbf{P. Meletis} and G. Dubbelman. \textit{``Training of Convolutional Networks on Multiple Heterogeneous Datasets for Street Scene Semantic Segmentation''}. In: 2018 IEEE Intelligent Vehicles Symposium, IV 2018, Changshu, Suzhou, China, June 26-30, 2018. IEEE, 2018, pp. 1045–1050. \texttt{DOI: 10.1109/IVS.2018.8500398}.
\end{enumerate}

\section*{Regional conferences contributions and technical reports}

\begin{enumerate}[itemsep=0.3em,start=1,label={\bfseries R\arabic*.~~~}]
	\item \textbf{P. Meletis}, X. Wen, C. Lu, D. de Geus, and G. Dubbelman ``Cityscapes-Panoptic-Parts and PASCAL-Panoptic-Parts datasets for Scene Understanding''. In: CoRR abs/2004.07944, 2021.\\ \texttt{URL: https://arxiv.org/abs/2004.07944}.
	\item R. Romijnders, \textbf{P. Meletis}, and G. Dubbelman. \textit{``An Agnostic Normalization Layer for Unsupervised Adversarial Domain Adaptation''}. In: Netherlands Conference on Computer Vision (NCCV). Eindhoven, the Netherlands, 2019.
	\item \textbf{P. Meletis} and G. Dubbelman. \textit{``Towards holistic scene understanding in autonomous driving''}. Published abstract and oral presentation in: 6th International Conference and Exhibition on Automobile \& Mechanical Engineering. Zurich, Switzerland, 2019. \texttt{ISSN: 2319-9873}.
	\item D. de Geus, \textbf{P. Meletis}, and G. Dubbelman. \textit{``Panoptic Segmentation with a Joint Semantic and Instance Segmentation Network''}. In: CoRR abs/1809.02110, 2018. \texttt{URL: http://arxiv.org/abs/1809.02110}.
	\item \textbf{P. Meletis} and G. Dubbelman. \textit{``Hierarchical Semantic Segmentation for Street Scenes''}. In: IEEE SBE Symposium: Automotive track. Eindhoven University of Technology, Eindhoven, the Netherlands, 2017.
\end{enumerate}

\chapter*{\label{ch:acks}Acknowledgments}
\addcontentsline{toc}{chapter}{Acknowledgments}

\thispagestyle{plain}
\pagestyle{plain}

\begin{spacing}{1.0}
I remember as if it was yesterday when I arrived in Eindhoven for my final interview, where four sunny days in a row were about to come, even if it was March. A PSV match was held the same day and a big party was organized for the win. After meeting my soon-to-be promotors and having felt the Eindhoven vibe, it was enough to accept the offer to pursue a PhD. Luckily, PSV is still winning, but the sunny March was just a luring invitation.

Finalizing my PhD and looking back at my life as a researcher, I have only beautiful memories and experiences to think of. The endeavor would not have been possible without the support and the company of all the colleagues, friends, and family, who assisted me in finding the balance between work and life and maintaining it. I feel gratitude towards each one standing by me, and in these closing words, I would like to thank you personally.

First and foremost, I would like to thank my promotors dr. Gijs Dubbelman and professor Peter de With for showing me their trust and giving me the opportunity to work with them. Gijs you have created an excellent research environment and a healthy group, in which it was a great pleasure to work, experiment, and contemplate new ideas. Your support in every professional and even personal topic was invaluable, shaped me as a person and as a scientist. I want to thank you for the freedom you gave me and your dedication in the days (and nights until the last hours) before the submission deadlines. Peter, I am grateful for the time you spent on this thesis, your patience and persistence in continuously ameliorating it. I enjoyed all of our detailed as well as high-level discussions.

Definitely, a large part of my PhD life was shared with the wonderful colleagues and friends in the VCA lab. I am glad to be part of this group because I made strong friendships that I hope will last after graduating. First, I would like to thank all current and past members of the MPS cluster for spending quality time together and having interesting conversations. Anweshan, Ariyan, Daan, Fabrizio, Chenyang, Pavol, Jos, Narsimlu, Ya, Liang, Dennis, Willem, and Floris, sharing the office with you was fun, I feel very lucky I met you and I am grateful for your support in easy and hard times. Warm thanks to all other VCA members for the coffee discussions, successful collaborations, and the pleasant time we spent together: Marco, Joost, Tim, Luis, Fons, Joy, Cheng, Egor, Rafaelle, and Clint. Also, I cannot forget past members of the VCA lab, Hani, Broni, Ronald, Arash, Farhad, Kostas, Sveta, with who we played sports and had fruitful conversations. Last but not least, I should mention our secretaries Anja, Marieke, and Pascale, who were always there for any assistance that I needed, facilitating bureaucratic procedures, and happily organizing the great summer and winter group events.

Throughout these four years, I have had the honor of supervising several bachelor and master students, Lars, Jake, Floris, Fjord, Vincent, Rob, and Daan, some of whom later became fellow PhDs. Thank you for the excellent collaboration, the outstanding publications we shared, the things you taught me, and the constructive discussions we had.

My thanks also go to my Rotterdam friends and cousins Vladan, and Spyros and Angeliki for their support, advice, and cheerful moments we had. Moreover, I want to thank my friend Tom for the fruitful collaborations and his enthusiasm for future plans. Of course, I cannot omit Dimitris and the whole group of pus* friends for the BBQs and brownies, Manolis for showing me cool snowboard tricks, and George my extreme-sports buddy.

At this point, I would like to thank Cristina, Chara, Eleni, Francesca, Nailia, Antoine, Roger, Arash, Christos, Amir, and Eleonora, members of the famous ``Greek lunch'' group. I enjoyed every bit of time we spent together, through energizing coffee breaks, walks, discussions, and trips. I am proud to be part of this group that knows how to enjoy life.

Finally, I want to mention my dearest friends who accompany me for more than 20 years and they know every strength and weakness of mine: Tereza, Myrsini, Renata, Konstantinos, Kimon, Kostas, Giannis, Nana, and Platon. I am extremely lucky to have you in my life journey and I wish you all the best for your own life journey. Joana, I cannot ask for more, you are supporting me every day, and you have also contributed to this thesis. And of course, I owe my biggest gratitude to my family, cousins, my brother who is my best mate, and my parents for believing in me, supporting me in every choice, instilling me with ethical values, and cultivating the scientific spirit inside me.
\end{spacing}

\chapter*{\label{ch:cv}About the Author}
\addcontentsline{toc}{chapter}{About the Author}

\begin{spacing}{1.0}
	
\begin{floatingfigure}{0.3\textwidth}
	\hspace{-17pt}
	\includegraphics[width=0.3\textwidth]{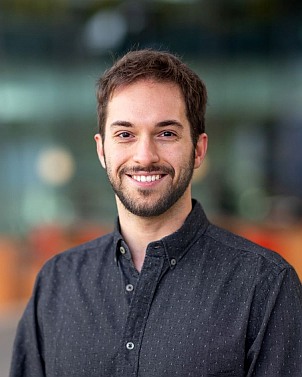}
	\vspace{-5pt}
\end{floatingfigure}

Panagiotis Meletis was born in Athens, Greece, on Saturday, June 8, 1991. After finishing high school at Arsakeio Psychikou, Athens, in 2009, he pursued graduate studies at the National Technical University of Athens in the Department of Electrical and Computer Engineering. In 2015, he received his BSc and MSc degrees with a specialization in Signal Processing and Electronics. For his master thesis, he developed a face detection system using modern deep learning techniques.

In 2016, Panagiotis joined the Video Coding and Architectures group as a PhD researcher, at the Eindhoven University of Technology, the Netherlands. He worked on advancing deep learning techniques for autonomous vehicles within the Mobile Perception Systems lab. His research involved the design of novel algorithms towards holistic visual scene understanding within several European projects. The realization of the algorithms in an autonomous vehicle laboratory platform was used in various live demonstrations for these projects. He competed with a specialized deep network for semantic segmentation at the Robust Vision Challenge of CVPR 2018, where he attained 3\textsuperscript{rd} place overall and 1\textsuperscript{st} place in the WildDash benchmark.

During his PhD study, he has (co-)authored 15 publications, and he has given six presentations at conferences and invited talks. Panagiotis participated in the International Summer School on Deep Learning in Genova, Italy, 2018. Moreover, he served for the period 2017 – 2020 in the TU/e communication ambassadors team.

Aiming to move the field of visual scene understanding forward, he guided a team on generating two novel, publicly available datasets for all-encompassing and holistic scene understanding in 2019. His current research interests include deep learning techniques for machine perception, automated visual reasoning, and explainable and sustainable artificial intelligence.

\end{spacing}

\clearpage
~

\end{document}